\documentclass[10pt,twocolumn,letterpaper]{article}

\usepackage{iccv}
\usepackage{times}
\usepackage{epsfig}
\usepackage{graphicx}
\usepackage{amsmath}
\usepackage{amssymb}

\usepackage{adjustbox}
\usepackage{stackengine}
\usepackage{multirow}
\usepackage{tabularx,booktabs}
\usepackage[table]{xcolor}
\definecolor{lightgray}{gray}{0.9}
\definecolor{lightblue}{rgb}{0.93,0.95,1.0}
\definecolor{darkgreen}{rgb}{0.0,0.6,0.0}
\definecolor{mypink1}{rgb}{0.858, 0.188, 0.478}

\newcommand{\minisection}[1]{\vspace{2mm}\noindent{\textbf{#1}.}}
\newtheorem{theorem}{Theorem}
\newtheorem{definition}{Definition}

\DeclareMathOperator*{\argmin}{argmin}

\usepackage[breaklinks=true,bookmarks=false,colorlinks]{hyperref}

\iccvfinalcopy 

\def\iccvPaperID{6640} 

\begin{document}

\title{Transferability and Hardness of Supervised Classification Tasks}

\author{Anh T.~Tran\thanks{Work at Amazon Web Services, prior to joining current affiliation.}\\
VinAI Research\\
{\tt\small anstar1111@gmail.com}
\and
Cuong V.~Nguyen\\
Amazon Web Services\\
{\tt\small nguycuo@amazon.com}
\and
Tal Hassner$^*$\\
Facebook AI\\
{\tt\small talhassner@gmail.com}
}

\maketitle

\begin{abstract}
We propose a novel approach for estimating the difficulty and transferability of supervised classification tasks. Unlike previous work, our approach is solution agnostic and does not require or assume trained models. Instead, we estimate these values using an information theoretic approach: treating training labels as random variables and exploring their statistics. When transferring from a source to a target task, we consider the conditional entropy between two such variables (i.e., label assignments of the two tasks). We show analytically and empirically that this value is related to the loss of the transferred model. We further show how to use this value to estimate task hardness. We test our claims extensively on three large scale data sets---CelebA (40 tasks), Animals with Attributes~2 (85 tasks), and Caltech-UCSD Birds~200 (312 tasks)---together representing 437 classification tasks. We provide results showing that our hardness and transferability estimates are strongly correlated with empirical hardness and transferability. As a case study, we transfer a learned face recognition model to CelebA attribute classification tasks, showing state of the art accuracy for tasks estimated to be highly transferable.
\end{abstract}

\section{Introduction}
How easy is it to transfer a representation learned for one task to another? How can we tell which of several tasks is hardest to solve? Answers to these questions are vital in planning model transfer and reuse, and can help reveal fundamental properties of tasks and their relationships in the process of developing universal perception engines~\cite{arora2014provable}. The importance of these questions is therefore driving research efforts, with several answers proposed in recent years. 

Some of the answers to these questions established task relationship indices, as in the Taskonomy~\cite{zamir2018taskonomy} and Task2Vec~\cite{achille2019task2vec,achille2019information} projects. Others analyzed task relationships in the context of multi-task learning~\cite{lee2016asymmetric,lu2017fully,veit2017conditional,yang2016deep,zhao2018modulation}. Importantly, however, these and other efforts are computational in nature, and so build on specific machine learning solutions as {\em proxy task representations}.

By relying on such proxy task representations, these approaches are naturally limited in their application: Rather than insights on the tasks themselves, they may reflect relationships between the specific solutions chosen to represent them, as noted by previous work~\cite{zamir2018taskonomy}. Some, moreover, establish task relationships by maintaining model zoos, with existing trained models already available. They may therefore also be computationally expensive~\cite{achille2019task2vec,zamir2018taskonomy}. Finally, in some scenarios, establishing task relationships requires multi-task learning of the models, to measure the influence different tasks have on each other~\cite{lee2016asymmetric,lu2017fully,veit2017conditional,yang2016deep,zhao2018modulation}.

We propose a radically different, {\em solution agnostic} approach: We seek underlying relationships, irrespective of the particular models trained to solve these tasks or whether these models even exist. We begin by noting that supervised learning problems are defined not by the models trained to solve them, but rather by the {\em data sets of labeled examples and a choice of loss functions}. We therefore go to the source and explore tasks directly, by examining their data sets rather than the models they were used to train.

To this end, we consider supervised classification tasks defined over the same input domain. As a loss, we assume the cross entropy function, thereby including most commonly used loss functions. We offer the following surprising result: By assuming an optimal loss on two tasks, the {\em conditional entropy} (CE) between the label sequences of their training sets provides a bound on the {\em transferability} of the two tasks---that is, the log-likelihood on a target task for a trained representation transferred from a source task. We then use this result to obtain a-priori estimates of task transferability and hardness.

Importantly, we obtain effective transferability and hardness estimates by evaluating {\em only training labels}; we do not consider the solutions trained for each task or the input domain. This result is surprising considering that it greatly simplifies estimating task hardness and task relationships, yet, as far as we know, was overlooked by previous work.

We verify our claims with rigorous tests on a total of 437 tasks from the CelebA~\cite{liu2015faceattributes}, Animals with Attributes~2 (AwA2)~\cite{xian2018zero}, and Caltech-UCSD Birds 200 (CUB)~\cite{WelinderEtal2010} sets. We show that our approach reliably predicts task transferability and hardness. As a case study, we evaluate transferability from face recognition to facial attribute classification. On attributes estimated to be highly transferable from recognition, our results outperform the state of the art despite using a simple approach, involving training a linear support vector machine per attribute.

\section{Related work}\label{sec:related}
Our work is related to many fields in machine learning and computer vision, including transfer learning~\cite{ying2018transfer}, meta learning~\cite{rusu2018meta}, domain shifting~\cite{rozantsev2019beyond}, and multi-task learning~\cite{kendall2018multi}. Below we provide only a cursory overview of several methods directly related to us. For more principled surveys on transfer learning, we refer to others~\cite{azizpour2015factors, pan2010survey, weiss2016survey, yosinski2014transferable}.

\minisection{Transfer learning} This paper is related to transfer learning~\cite{pan2010survey, weiss2016survey, ying2018transfer} and our work can be used to select good source tasks and data sets when transferring learned models. Previous theoretical analysis of transfer learning is extensive~\cite{achille2019information, azizzadenesheli2018regularized, ben2010theory, ben2003exploiting, blitzer2008learning, mansour2009domain}. These papers allowed generalization bounds to be proven but they are abstract and hard to compute in practice. Our transferability measure, on the other hand, is easily computed from the training sets and can potentially be useful also for continual learning~\cite{nguyen2019toward, nguyen2018variational, ring1997child}.

\minisection{Task spaces} Tasks in machine learning are often represented as labeled data sets and a loss function. For some applications, qualitative exploration of the training data can reveal relationships between two tasks and, in particular, the biases between them~\cite{torralba2011unbiased}. 

Efforts to obtain more complex task relationships involved trained models. Data sets were compared using fixed dimensional lists of statistics, produced using an autoencoder trained for this purpose~\cite{edwards2016towards}. The successful Taskonomy project~\cite{zamir2018taskonomy}, like us, assumes multiple task labels for the same input images (same input domain). They train one model per-task and then evaluate transfers between tasks thereby creating a task hypergraph---their taxonomy. 

Finally, Task2Vec constructs vector representations for tasks, obtained by mapping partially trained probe networks down to low dimensional task embeddings~\cite{achille2019task2vec,achille2019information}. Unlike these methods, we consider only the labels provided in the training data for each task, without using trained models.

\minisection{Multi-task learning} Training a single model to solve multiple tasks can be mutually beneficial to the individual tasks~\cite{he2017mask,ranjan2017all,wang2015towards}. When two tasks are only weakly related, however, attempting to train a model for them both can produce a model which under-performs compared to models trained for each task separately. Early multi-branch networks and their variants encoded human knowledge on the relationships of tasks in their design, joining related tasks or separating unrelated tasks~\cite{jou2016deep,ranjan2019hyperface,rothe2015dex}.

Others adjusted for related vs.~unrelated tasks during training of a deep multi-task network. Deep cross residual learning does this by introducing cross-residuals for regularization~\cite{jou2016deep}, cross-stitch combines activations from multiple task-specific networks~\cite{misra2016cross}, and UberNet proposed a task-specific branching scheme~\cite{kokkinos2017ubernet}. 

Some sought to discover what and how should be shared across tasks during training by automatic discovery of network designs that would group similar tasks together~\cite{lu2017fully} or by solving tensor factorization problems~\cite{yang2016deep}. Alternatively, parts of the input rather than the network were masked according to the task at hand~\cite{veit2017conditional}. Finally, modulation modules were proposed to seek destructive interferences between unrelated tasks~\cite{zhao2018modulation}.

\section{Transferability via conditional entropy}\label{sec:main}
We seek information on the transferability and hardness of supervised classification tasks. Previous work obtained this information by examining machine learning models developed for these tasks~\cite{achille2019task2vec,achille2019information,zhao2018modulation}. Such models are produced by training on labeled data sets that represent the tasks. These models can therefore be considered views on their training data. In this work we instead use information theory to produce estimates \emph{from the source}: the data itself.

Like others~\cite{zamir2018taskonomy}, we assume our tasks share the same input instances and are different only in the labels they assign to each input. Such settings describe many practical scenarios. A set of face images, for instance, can have multiple labels for each image, representing tasks such as recognition~\cite{Masi:18:learning,masi2019face} and classification of various attributes~\cite{liu2015faceattributes}.

We estimate transferability using the CE between the label sequences of the target and source tasks. Task hardness is similarly estimated: by computing transferability from a trivial task. We next formalize our assumptions and claims.

\subsection{Task transferability}\label{sec:transf}
We assume a single {\em input sequence} of training samples, $X = ( x_1, x_2, \ldots, x_n) \in \mathcal{X}^n$, along with two {\em label sequences} $Y = ( y_1, y_2, \ldots, y_n)\in {\mathcal{Y}}^n$ and $Z = ( z_1, z_2, \ldots, z_n )\in{\mathcal{Z}}^n$, where~$y_i$ and~$z_i$ are labels assigned to~$x_i$ under two separate tasks: source task~$T^Z=(X, Z)$ and target task~$T^Y=(X,Y)$. Here, $\mathcal{X}$ is the domain of the values of $X$, while ${\mathcal Y}=\mathrm{range}(Y)$ and ${\mathcal Z}=\mathrm{range}(Z)$ are the sets of different values in~$Y$ and~$Z$ respectively. Thus, if~$Z$ contains binary labels, then ${\mathcal Z} = \{ 0, 1\}$.

We consider a classification model $M=(w, h)$ on the source task, $T^Z$. The first part, $w : \mathcal{X} \rightarrow \mathbb{R}^D$, is some transformation function, possibly learned, that outputs a $D$-dimensional representation $r= w(x)\in\mathbb{R}^D$ for an input $x \in \mathcal{X}$. The second part, $h : \mathbb{R}^D \rightarrow \mathcal{P}(\mathcal{Z})$, is a classifier that takes a representation $r$ and produces a probability distribution $h(r) \in \mathcal{P}(\mathcal{Z})$, where $\mathcal{P}(\mathcal{Z})$ is the space of all probability distributions over ${\mathcal Z}$.

This description emphasizes the two canonical stages of a machine learning system~\cite{duda2012pattern}: representation followed by classification. As an example, a deep neural network with Softmax output is represented by a learned $w$ which maps the input into some feature space, producing a {\em deep embedding}, $r$, followed by classification layers (one or more), $h$, which maps the embedding into the prediction probability.

Now, assume we train a model $(w_Z, h_Z)$ to solve $T^Z$ by minimizing the cross entropy loss on $Z$:
\begin{equation}
w_Z, h_Z = \argmin_{w, h \in (W, H)} \mathcal{L}_Z(w, h),
\label{eq:source-loss}
\end{equation}
where $W$ and $H$ are our chosen spaces of possible values for $w$ and $h$, and $\mathcal{L}_Z(w, h)$ is the cross entropy loss (equivalently, the negative log-likelihood) of the parameters $(w, h)$:
\begin{equation}
\mathcal{L}_Z(w, h) = - l_Z(w, h) = - \frac{1}{n} \sum_{i=1}^n \log P(z_i | x_i; w, h),
\label{eq:cross-ent}
\end{equation}
where $l_Z(w, h)$ is the log-likelihood of $(w, h)$.

To transfer this model to target task $T^Y$, we fix the function $w_Z$ and retrain only the classifier on the labels of $T^Y$. Denote the new classifier $k_Y$, selected from our chosen space $K$ of target classifiers. Note that $k_Y$ does not necessarily share the same architecture as $h_Z$. We train $k_Y$ by minimizing the cross entropy loss on $Y$ with the fixed~$w_Z$:
\begin{equation}
k_Y = \argmin_{k \in K} \mathcal{L}_Y(w_Z, k),
\label{eq:target-loss}
\end{equation}
where $\mathcal{L}_Y(w_Z, k)$ is defined similarly to Eq.~\eqref{eq:cross-ent} but for the label set $Y$. Under this setup, we define the transferability of task $T^Z$ to task $T^Y$ as follows.

\begin{definition}
The transferability of task $T^Z$ to task $T^Y$ is measured by the expected accuracy of the model $(w_Z, k_Y)$ on a random test example $(x, y)$ of task $T^Y$:
\begin{equation}
\mathrm{Trf}(T^Z \rightarrow T^Y) = \mathbb{E} \left[ \mathrm{acc}(y, x; w_Z, k_Y) \right],
\label{eq:true-transferability}
\end{equation}
which indicates how well a representation $w_Z$ trained on task $T^Z$ performs on task $T^Y$.
\end{definition}

In practice, if the trained model does not overfit, the log-likelihood on the {\em training set}, $l_Y(w_Z, k_Y)$, provides a good indicator of Eq~\eqref{eq:true-transferability}, that is, how well the representation $w_Z$ and the classifier $k_Y$ performs on task $T^Y$. This non-overfitting assumption holds even for large networks that are properly trained and tested on datasets sampled from the \emph{same distribution}~\cite{zhang2016understanding}.
Thus, in the subsequent sections, we instead consider the following log-likelihood as an alternative measure of transferability:
\begin{equation}
\widetilde{\mathrm{Trf}}(T^Z \rightarrow T^Y) = l_Y(w_Z, k_Y).
\label{eq:transferability}
\end{equation}

\begin{figure*}[t!]
\centering
    \includegraphics[clip, trim=0mm 0mm 0mm 0mm,width=.9\linewidth]{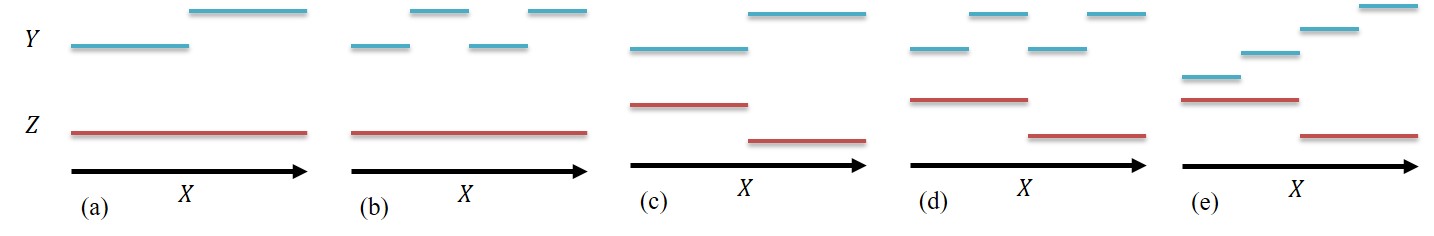}
    \vspace{-2mm}
\caption{{\bf Visualizing toy examples.} The transferability between two tasks, represented as sequences $(X,Y)$ and $(X,Z)$. The horizontal axis represent instances and the values for $Z$ (in red) and $Y$ (cyan). In which of these examples would it be easiest to transfer a model trained for task $T^Z$ to task $T^Y$? See discussion and details in Sec.~\ref{sec:intuition}.}\vspace{-3mm}
\label{fig:visualize}
\end{figure*}

\subsection{The conditional entropy of label sequences}\label{sec:CE}
From the label sequences~$Y$ and~$Z$, we can compute the empirical joint distribution $\hat{P}(y, z)$ for all $(y, z) \in {\mathcal Y} \times {\mathcal Z}$ by counting, as follows:
\begin{equation}
\hat{P}(y, z) = \frac{1}{n}| \{ i : y_i = y \text{ and } z_i = z \}|.\label{eq:empjointdist}
\end{equation}
We now adopt the definition of CE between two random variables~\cite{cover2012elements} to define the CE between our label sequences~$Y$ and~$Z$.
\begin{definition}
The CE of a label sequence~$Y$ given a label sequence~$Z$, $H(Y|Z)$, is the CE of a random variable (or random label) $\bar{y}$ given a random variable (or random label) $\bar{z}$, where $(\bar{y}, \bar{z})$ are drawn from the empirical joint distribution $\hat{P}(y, z)$ of Eq.~\eqref{eq:empjointdist}:
\begin{equation}
H(Y|Z) = -\sum_{y \in {\mathcal Y}} \sum_{z \in {\mathcal Z}} \hat{P}(y, z) \log \frac{\hat{P}(y, z)}{\hat{P}(z)},\label{eq:CE}
\end{equation}
where $\hat{P}(z)$ is the empirical marginal distribution on ${\mathcal Z}$:
\begin{equation}\hat{P}(z) = \sum_{y \in {\mathcal Y}} \hat{P}(y, z) = \frac{1}{n}|\{ i : z_i = z \}|. \label{eq:emp_marg_dst}
\end{equation}
\end{definition}

CE represents a measure of the amount of information provided by the value of one random variable on the value of another. By treating the labels assigned to both tasks as random variables and measuring the CE between them, we are measuring the information required to estimate a label in one task given a (known) label in another task.

We now prove a relationship between the CE of Eq.~\eqref{eq:CE} and the tranferability of Eq.~\eqref{eq:transferability}. In particular, we show that the log-likelihood on the target task~$T^Y$ is lower bounded by log-likelihood on the source task~$T^Z$ minus $H(Y|Z)$, if the optimal input representation $w_Z$ trained on~$T^Z$ is transferred to~$T^Y$.

To prove our theorem, we assume the space $K$ of target classifiers contains a classifier $\bar{k}$ whose log-likelihood lower bounds that of $k_Y$. We construct $\bar{k}$ as follows. For each input $x$, we compute the Softmax output $p_Z = h_Z(w_Z(x))$, which is a probability distribution on $\mathcal{Z}$. We then convert $p_Z$ into a Softmax on $\mathcal{Y}$ by taking the expectation of the empirical conditional probability ${ \hat{P}(y | z) = \hat{P}(y, z)/\hat{P}(z) }$ with respect to $p_Z$. That is, for all $y \in \mathcal{Y}$, we define:
\begin{equation}
p_Y(y) = \mathbb{E}_{z \sim p_Z}[\hat{P}(y | z)] = \sum_{z \in \mathcal{Z}} \hat{P}(y | z) ~ p_Z(z),
\end{equation}
where $p_Z(z)$ is the probability of the label $z$ returned by $p_Z$.
For any input $w_Z(x)$, we let the output of $\bar{k}$ be $p_Y$. That is, $\bar{k}(w_Z(x)) = p_Y$.
We can now prove the following theorem.

\begin{theorem}
Under the training procedure described in Sec.~\ref{sec:transf}, we have:
\begin{align}
\widetilde{\mathrm{Trf}}(T^Z \rightarrow T^Y) \ge l_Z(w_Z, h_Z) - H(Y|Z).
\label{eq:bound}
\end{align}\vspace{-5mm}
\label{thrm:transferability}
\end{theorem}
\minisection{Proof sketch}\footnote{Full derivations provided in the appendix.}
From the definition of $k_Y$ and the assumption that $\bar{k} \in K$, we have $\widetilde{\mathrm{Trf}}(T^Z \rightarrow T^Y) = l_Y(w_Z, k_Y) \ge l_Y(w_Z, \bar{k})$. From the construction of $\bar{k}$, we have:
\begin{align}
l_Y(w_Z, \bar{k}) &= \frac{1}{n} \sum_{i=1}^n \log \left( \sum_{z \in \mathcal{Z}} \hat{P}(y_i | z) P(z | x_i; w_Z, h_Z) \right) \nonumber \\
&{\hskip -46pt} \ge \frac{1}{n} \sum_{i=1}^n \log \left( \hat{P}(y_i | z_i) P(z_i | x_i; w_Z, h_Z) \right) \label{eq:pproofsketch2} \\
&{\hskip -46pt} = \frac{1}{n} \sum_{i=1}^n \log \hat{P}(y_i | z_i) + \frac{1}{n} \sum_{i=1}^n \log P(z_i | x_i; w_Z, h_Z).\label{eq:pproofsketch3}
\end{align}
We can easily show that the first term in Eq.~\eqref{eq:pproofsketch3} equals $-H(Y | Z)$, while the second term is $l_Z(w_Z, h_Z)$.

\minisection{Discussion 1: Generality of our settings} Our settings for spaces $W, H, K$ are general and include a variety of practical use cases. For example, neural networks $W$ will include all possible (vector) values of the network weights until the penultimate layer, while $H$ and $K$ would include all possible (vector) values of the last layer's weights. Alternatively, we can use support vector machines (SVM) for $K$. In this case, $K$ would include all possible values of the SVM parameters~\cite{scholkopf2001learning}. Our result even holds when the features are fixed, as when using tailored representations such as SIFT~\cite{lowe1999object}. In these cases, space $W$ would contain only one transformation function from raw input to the features.

\minisection{Discussion 2: Assumptions} We can easily satisfy the assumption $\bar{k} \in K$ by first choosing a space $K'$ (e.g., the SVMs) which will play the role of $K \setminus \{ \bar{k} \}$. We solve the optimization problem of Eq.~\eqref{eq:target-loss} on $K'$ instead of $K$ to obtain the optimal classifier $k'$. To get the optimal classifier $k_Y$ on $K = K' \cup \{ \bar{k} \}$, we simply compare the losses of $k'$ and $\bar{k}$ and select the best one as $k_Y$.

The optimization problems of Eq.~\eqref{eq:source-loss} and~\eqref{eq:target-loss} are global optimization problems. In practice, for complex deep networks trained with stochastic gradient descent, we often only obtain the local optima of the loss. In this case, we can easily change and prove Theorem~\ref{thrm:transferability} which would include the differences in the losses between the local optimum and the global optimum in the right-hand-side of Eq.~\eqref{eq:bound}. In many practical applications, the difference between local optimum and global optimum is not significant \cite{choromanska2015loss, nguyen2017loss}.

\minisection{Discussion 3: Extending our result to test log-likelihood} In Theorem \ref{thrm:transferability}, we consider the empirical log-likelihood, which is generally unbounded. If we make the (strong) assumption of bounded differences between empirical log-likelihoods, we can apply McDiarmid’s inequality~\cite{mcdiarmid1989method} to get an upper-bound on the left hand side of Eq.~\eqref{eq:bound} by the expected log-likelihood with some probability.

\minisection{Discussion 4: Implications} Theorem~\ref{thrm:transferability} shows that the transferability from task $T^Z$ to task $T^Y$ depends on both the CE $H(Y|Z)$ and the log-likelihood $l_Z(w_Z, h_Z)$. Note that the log-likelihood $l_Z(w_Z, h_Z)$ is optimal for task $T^Z$ and so it represents the hardness (or easiness) of task $T^Z$. Thus, from the theorem, if $l_Z(w_Z, h_Z)$ is small (i.e., the source task is hard), transferability would reduce. Besides, if the CE $H(Y|Z)$ is small, transferability would increase.

Finally, we note that when the source task $T^Z$ is fixed, the log-likelihood $l_Z(w_Z, h_Z)$ is a constant. In this case, the transferability only depends on the CE $H(Y|Z)$. Thus, we can estimate the transferability from one source task to multiple target tasks by considering only the CE.

\subsection{Intuition and toy examples}\label{sec:intuition}
To gain intuition on CE and transferability, consider the toy examples illustrated in Fig.~\ref{fig:visualize}. The (joint) input set is represented by the~$X$ axis. Each input $x_i\in X$ is assigned with two labels, $y_i\in Y$ and $z_i\in Z$, for the two tasks. In Fig.~\ref{fig:visualize}(a,b), task $T^Z$ is the trivial task with a constant label value (red line) and in Fig.~\ref{fig:visualize}(c--e) $T^Z$ is a binary classification task, whereas $T^Y$ is binary in  Fig.~\ref{fig:visualize}(a--d) and multi-label in Fig.~\ref{fig:visualize}(e). In which of these examples would transferring a representation trained for $T^Z$ to $T^Y$ be hardest?

Of the five examples, (c) is the easiest transfer as it provides a 1-1 mapping from $Z$ to $Y$. Appropriately, in this case, $H(Y|Z) = 0$. Next up are (d) and (e) with $H(Y|Z) = \log2$: In both cases each class in $T^Z$ is mapped to two classes in $T^Y$. Note that $T^Y$ being non-binary is naturally handled by the CE. Finally, transfers (a) and (b) have $H(Y|Z)=4\log2$; the highest CE. Because $T^Z$ is trivial, the transfer must account for the greatest difference in the information between the tasks and so the transfer is hardest.

\begin{figure*}[t!]
\centering
\begin{tabular}{c@{~}c@{~}c@{~}c}
    \includegraphics[clip, trim=0mm 0mm 0mm 11mm, width=0.245\textwidth]{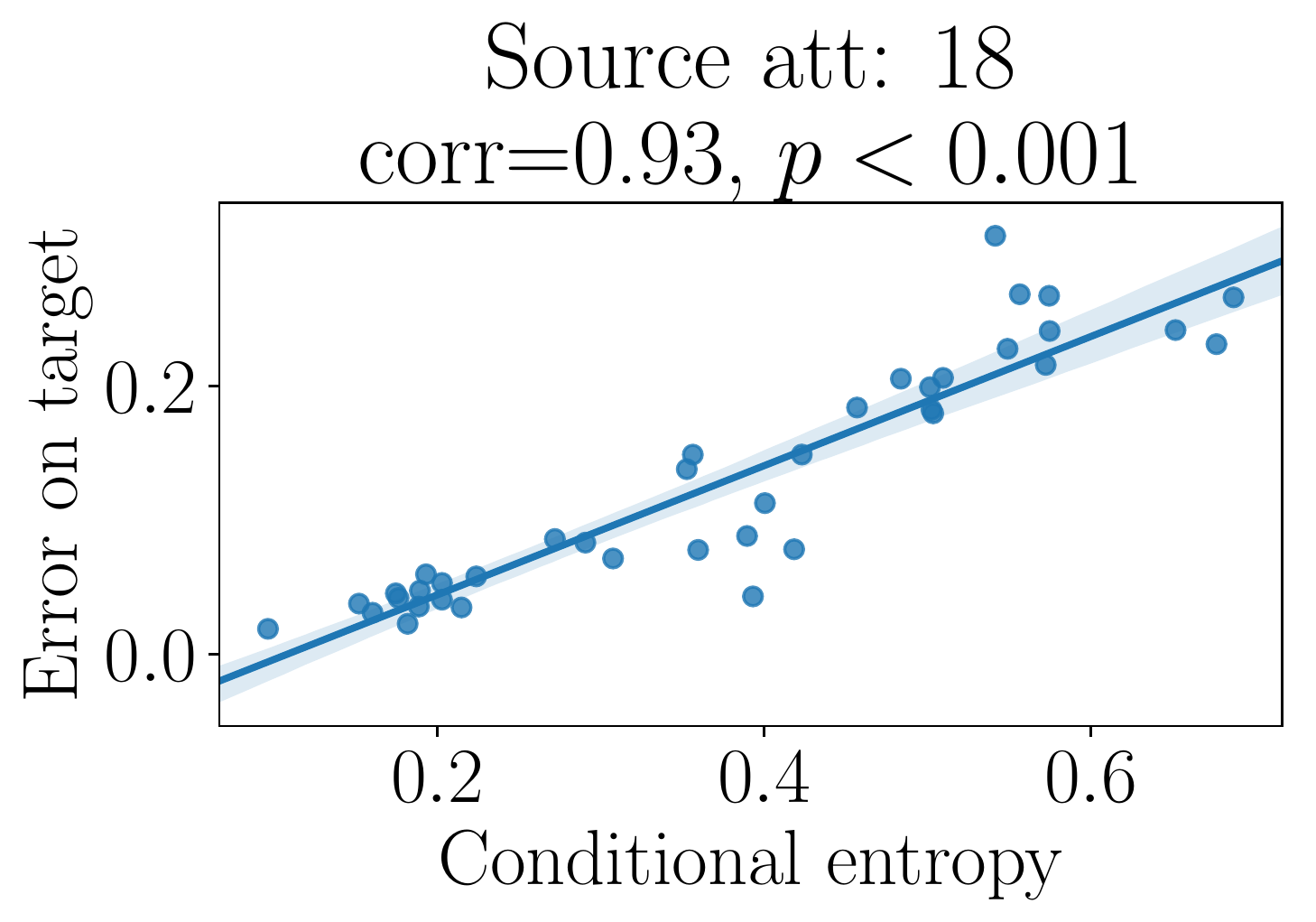}&
    \includegraphics[clip, trim=0mm 0mm 0mm 11mm, width=0.245\textwidth]{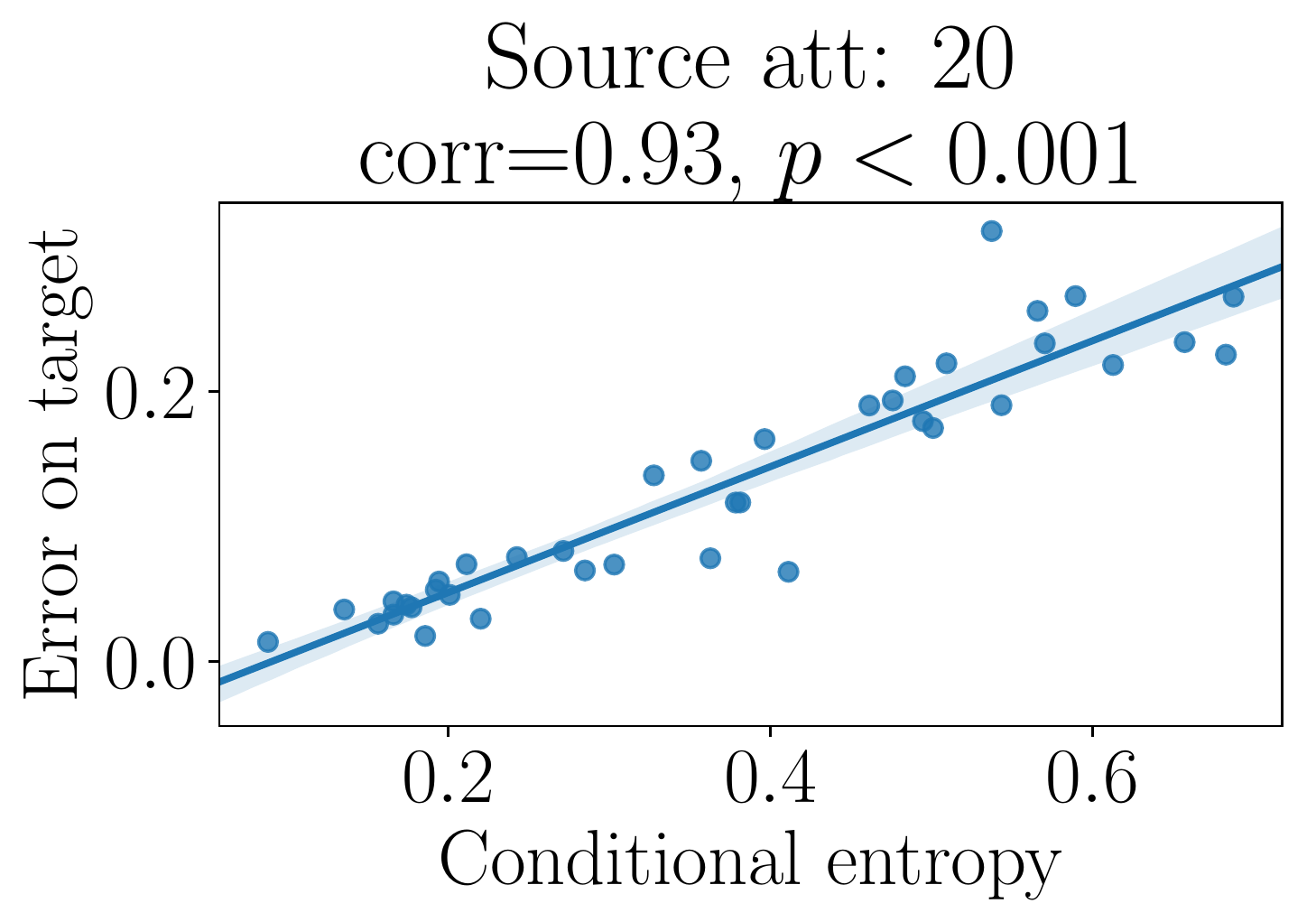}&
    \includegraphics[clip, trim=0mm 0mm 0mm 11mm, width=0.245\textwidth]{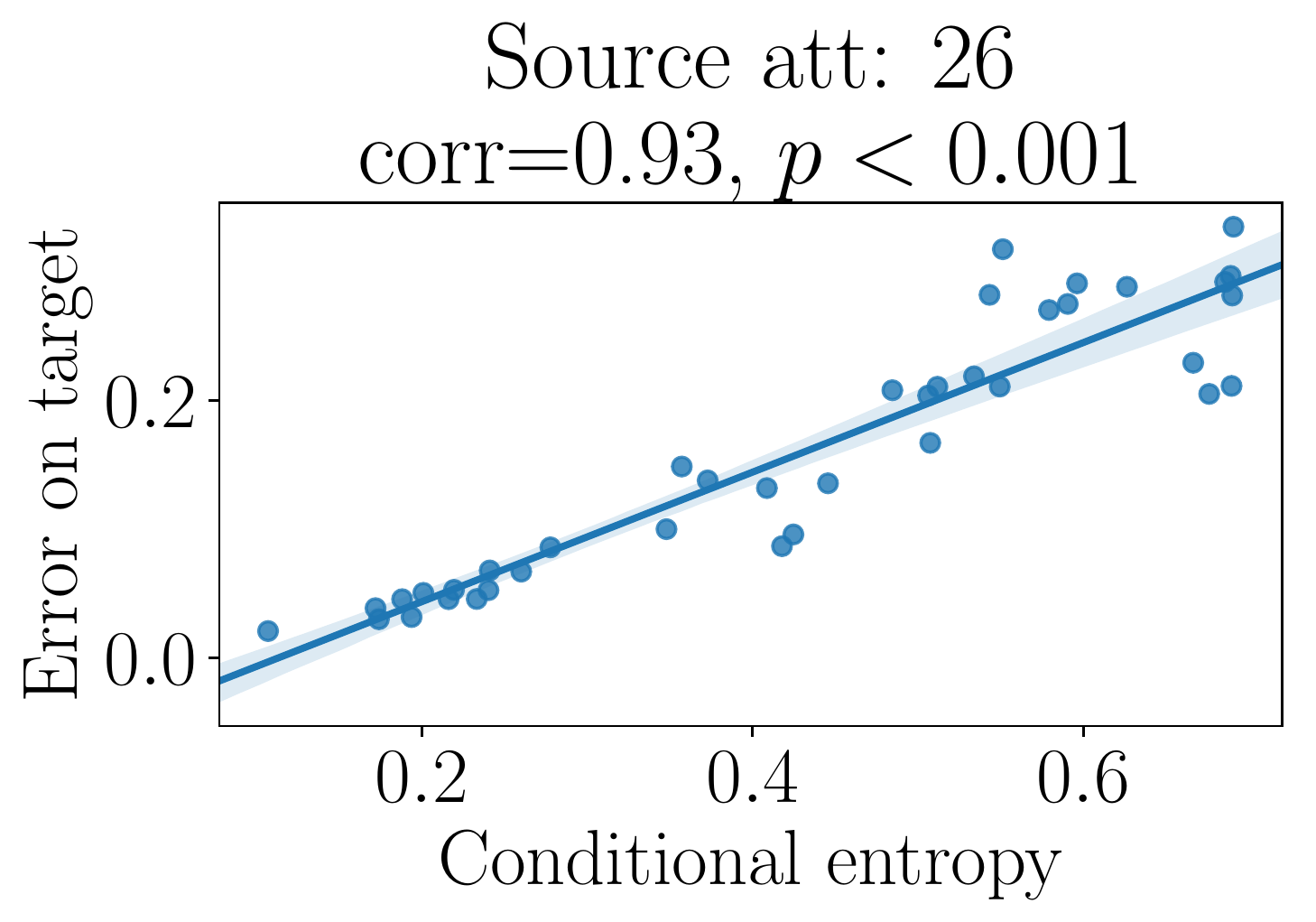}&
    \includegraphics[clip, trim=0mm 0mm 0mm 11mm, width=0.245\textwidth]{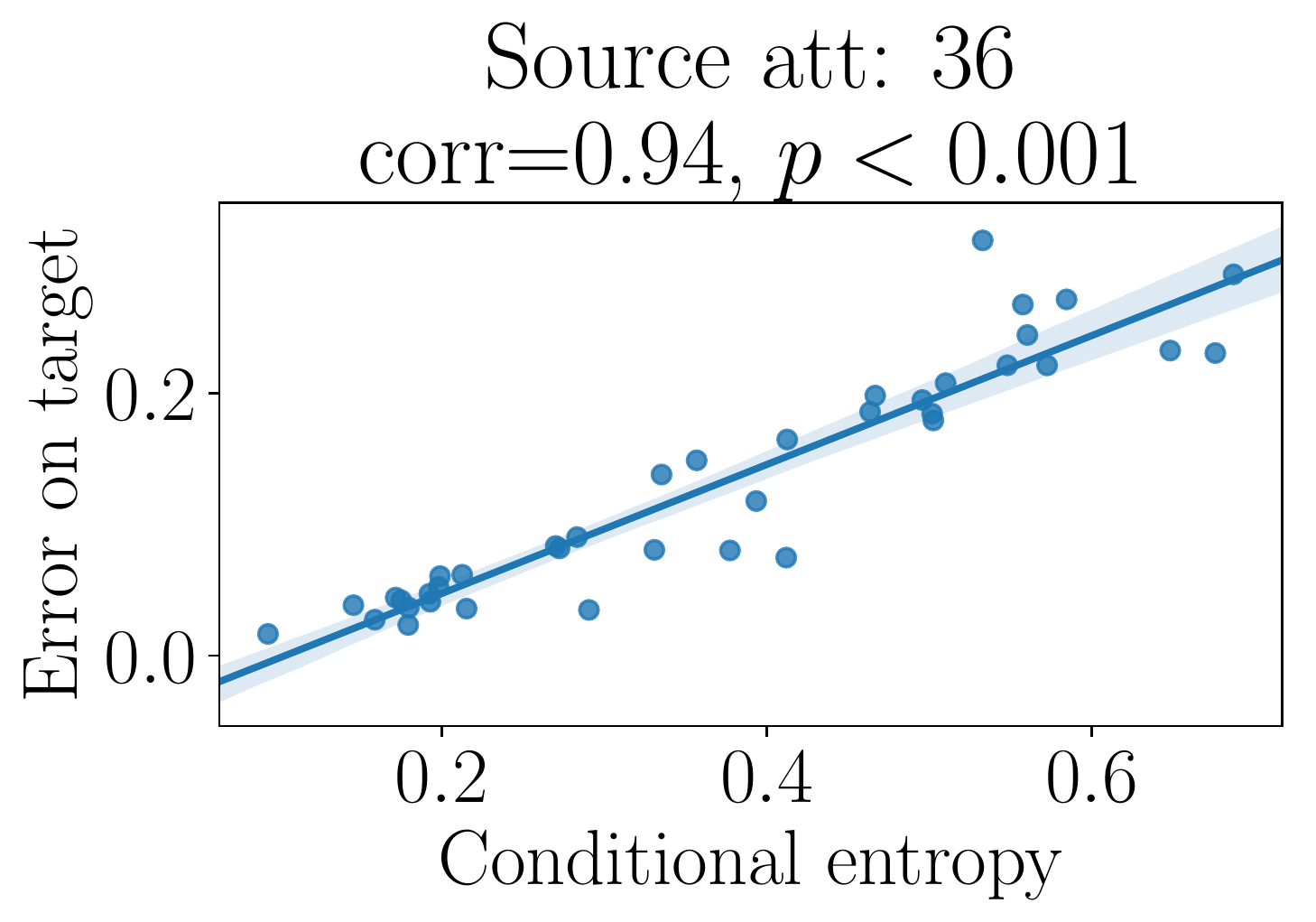}\\
    (a) Heavy makeup & (b) Male & (c) Pale skin & (d) Wearing lipstick\\[5pt]
    \includegraphics[clip, trim=0mm 0mm 0mm 11mm, width=0.245\textwidth]{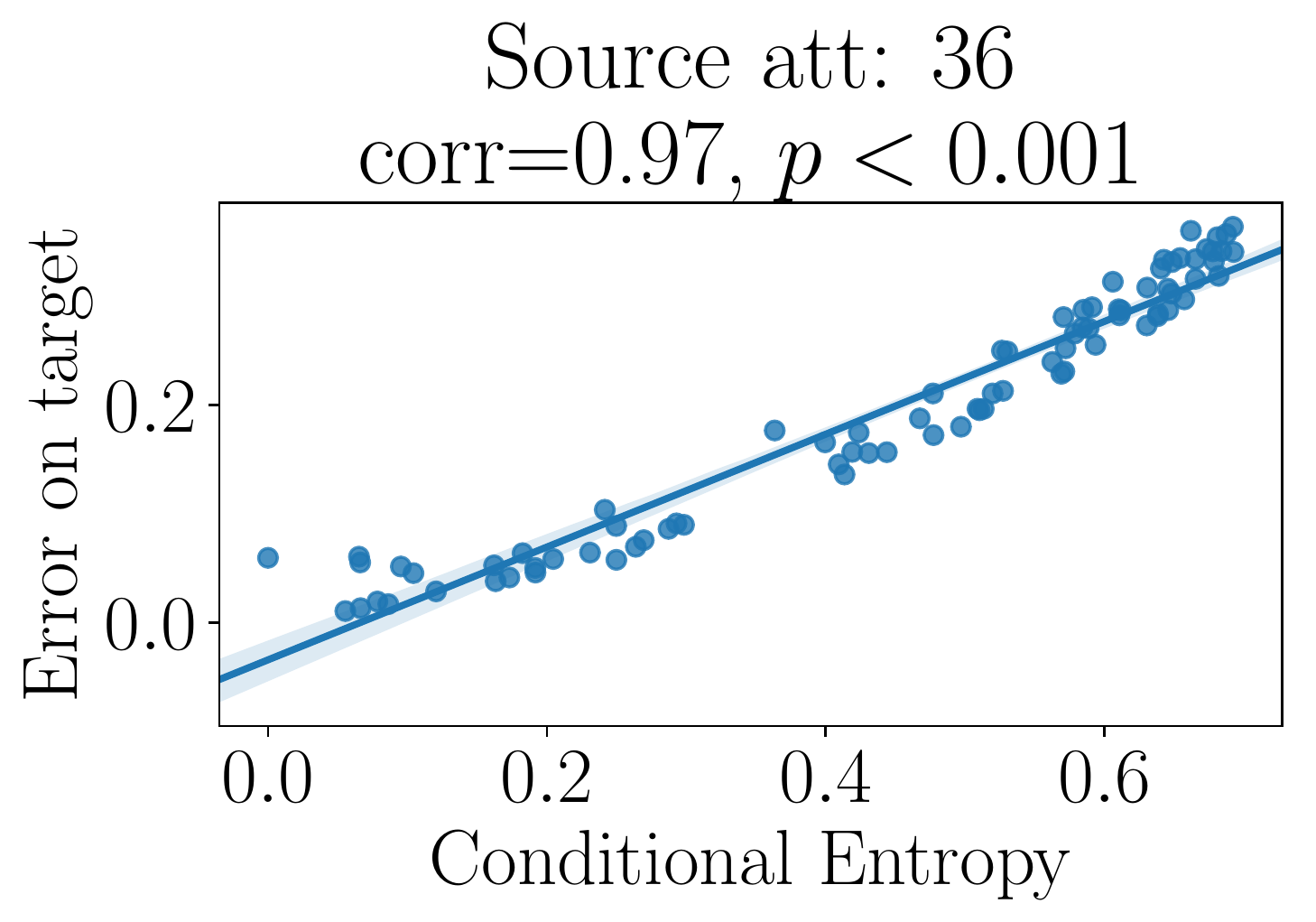}&
    \includegraphics[clip, trim=0mm 0mm 0mm 11mm, width=0.245\textwidth]{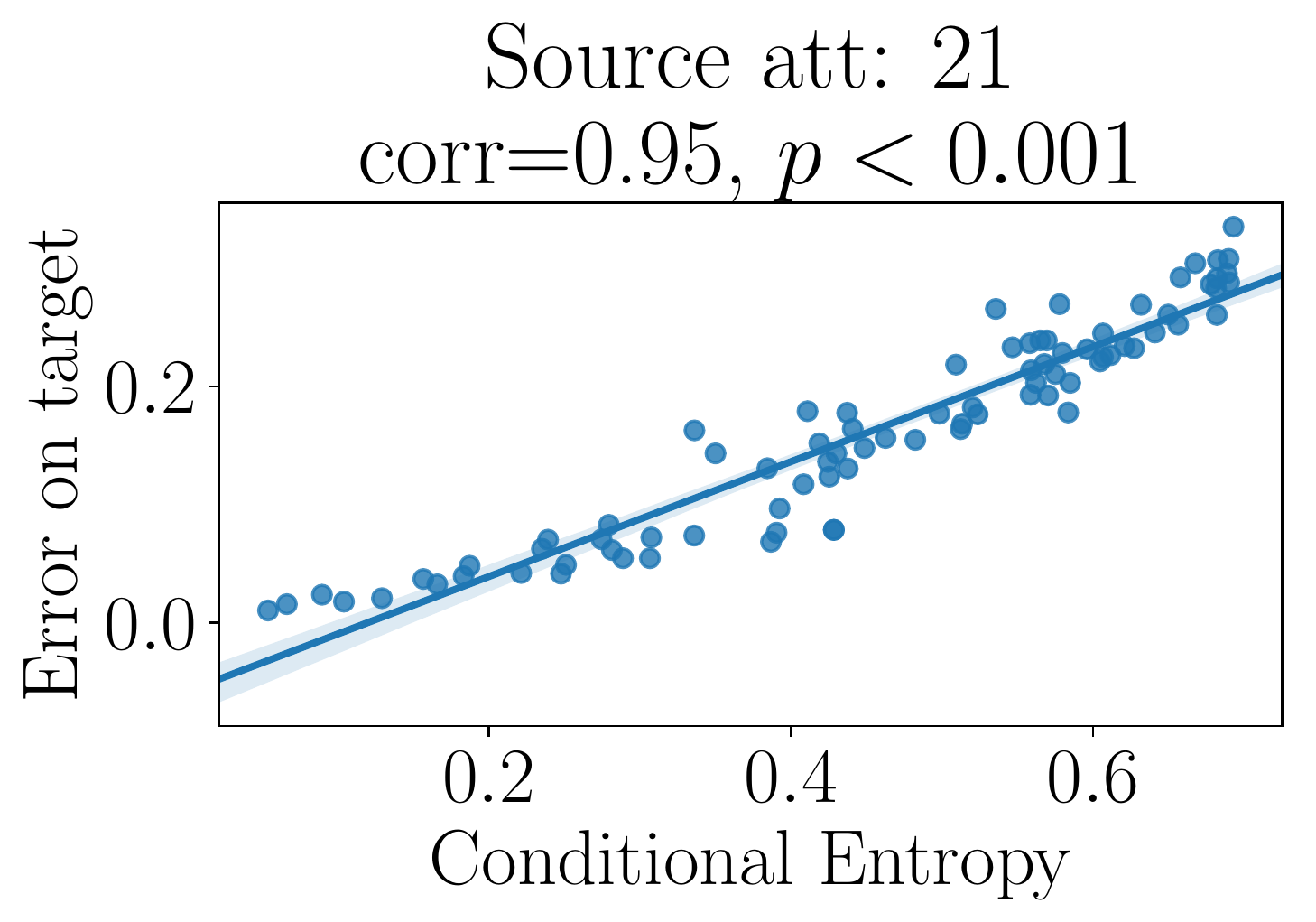}&
    \includegraphics[clip, trim=0mm 0mm 0mm 11mm, width=0.245\textwidth]{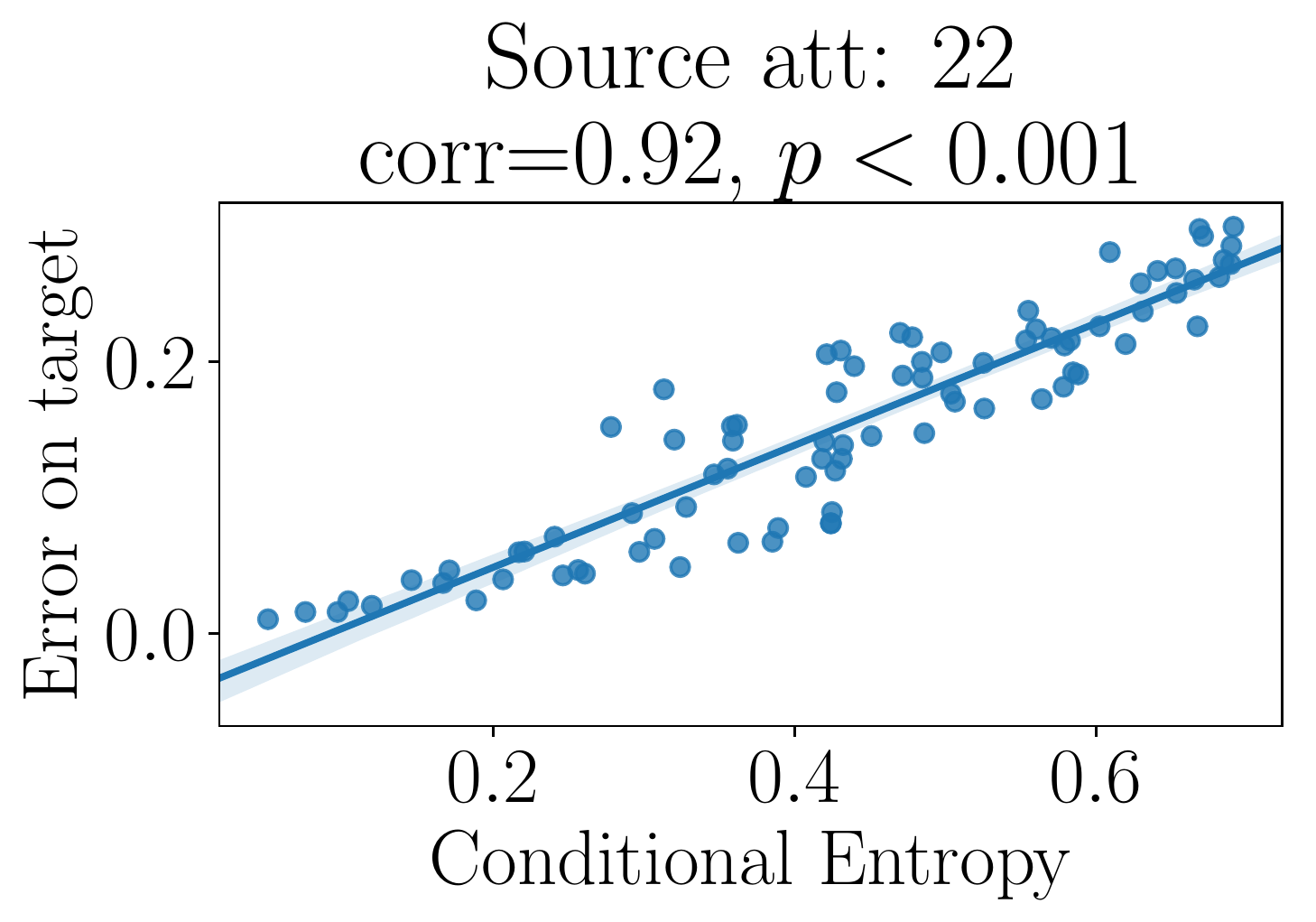}&
    \includegraphics[clip, trim=0mm 0mm 0mm 11mm, width=0.245\textwidth]{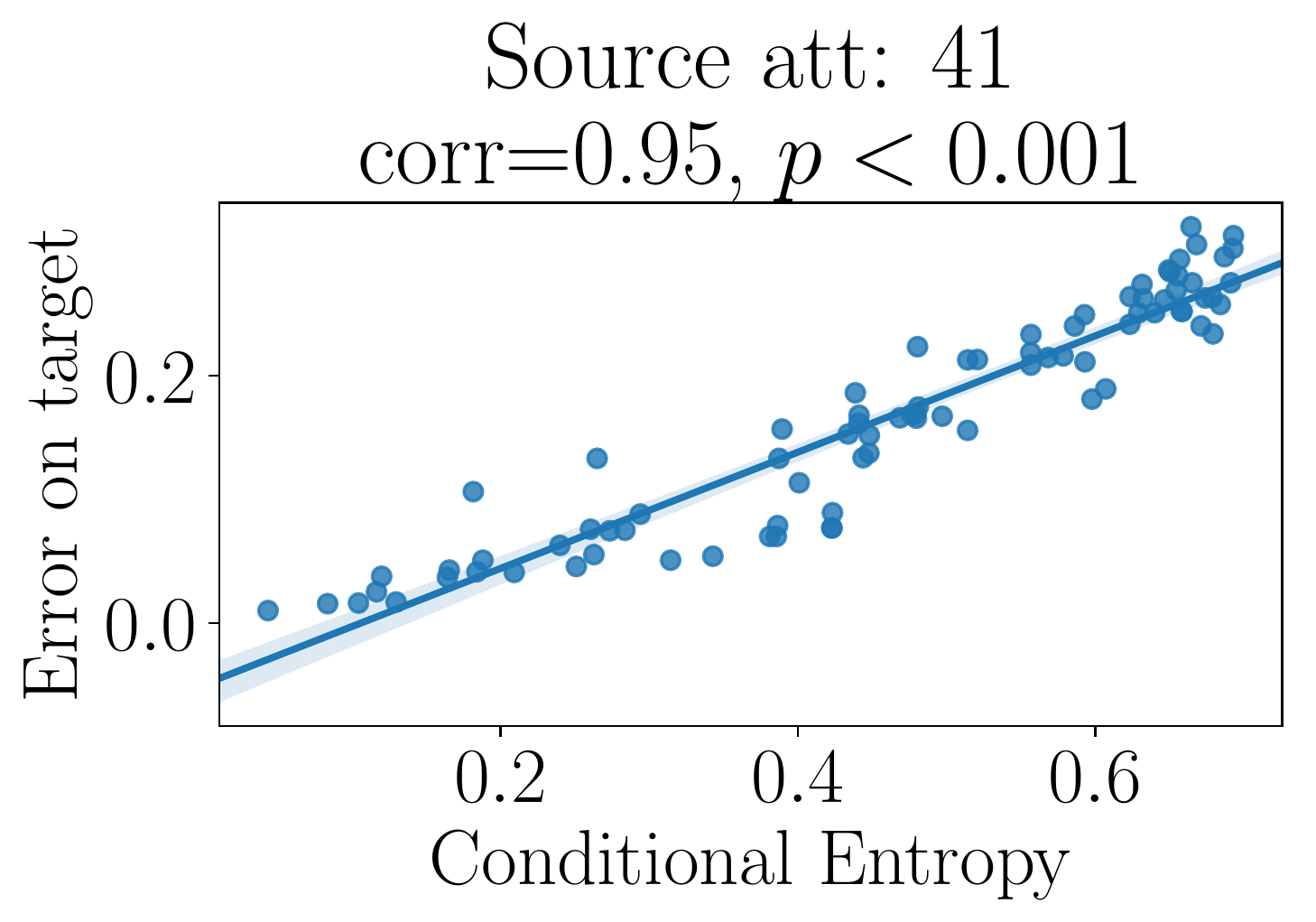}\\
    (e) Swim & (f) Pads & (g) Paws & (h) Strong\\[5pt]
    \includegraphics[clip, trim=0mm 0mm 0mm 11mm, width=0.245\textwidth]{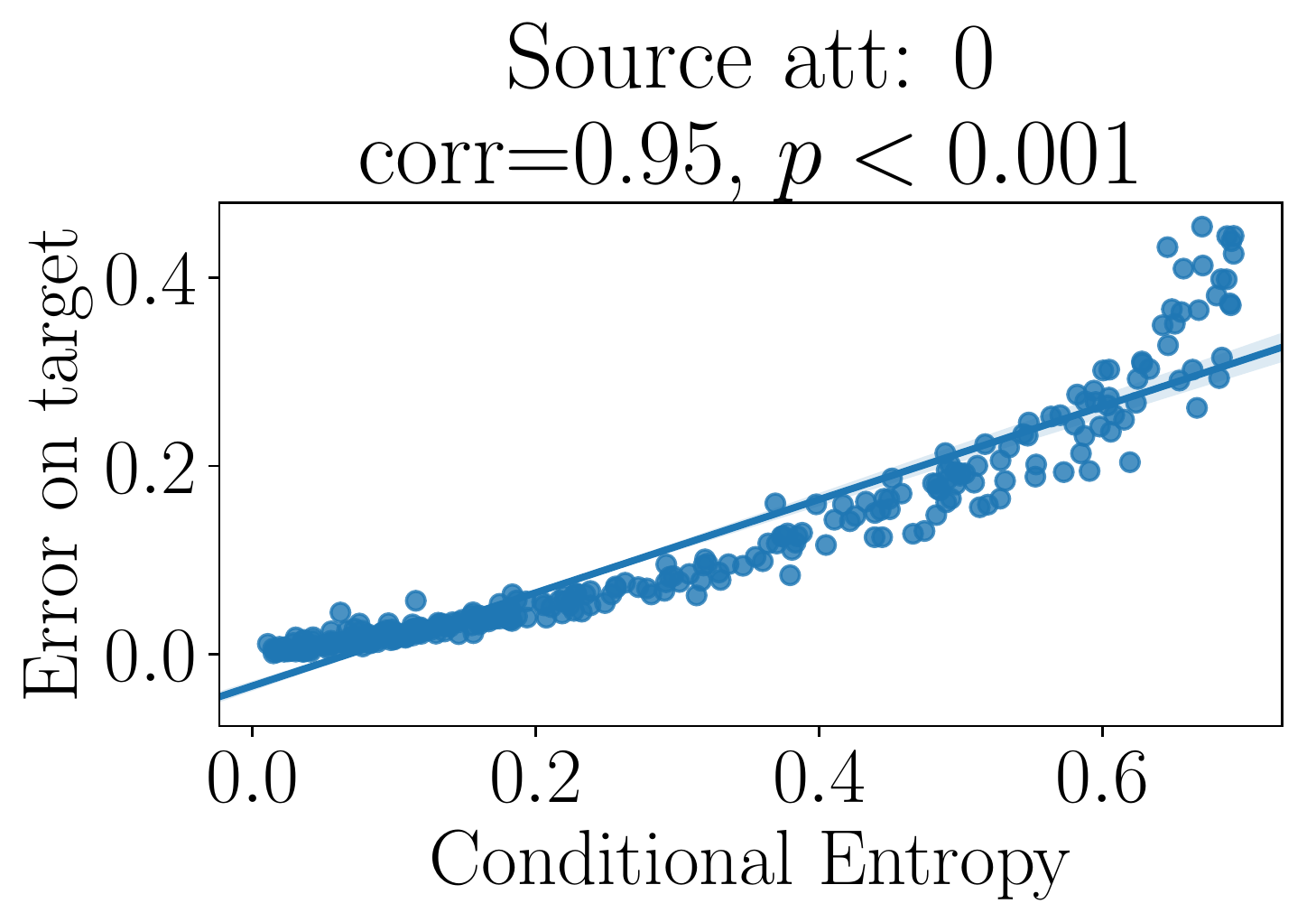}&
    \includegraphics[clip, trim=0mm 0mm 0mm 11mm, width=0.245\textwidth]{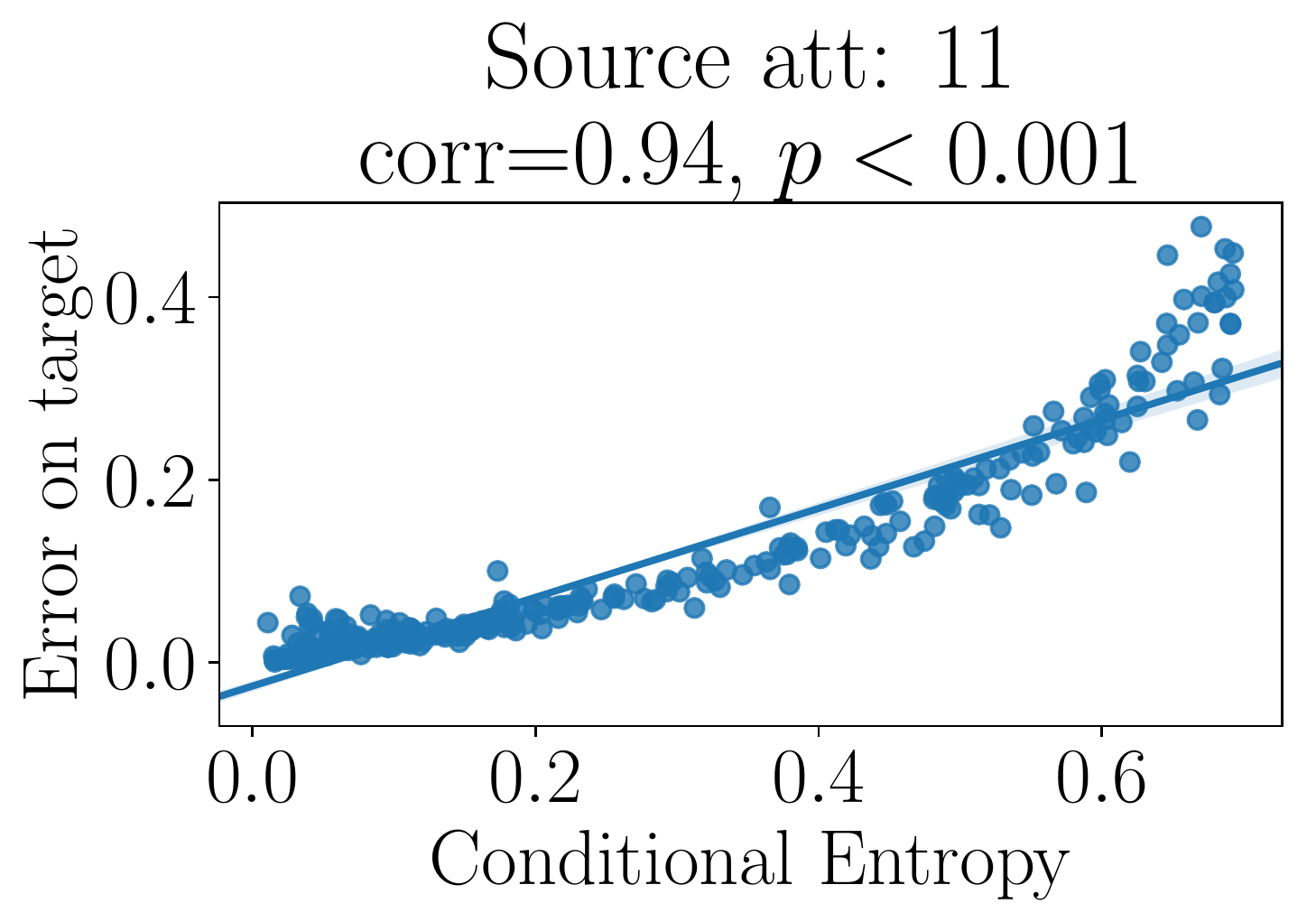}&
    \includegraphics[clip, trim=0mm 0mm 0mm 11mm, width=0.245\textwidth]{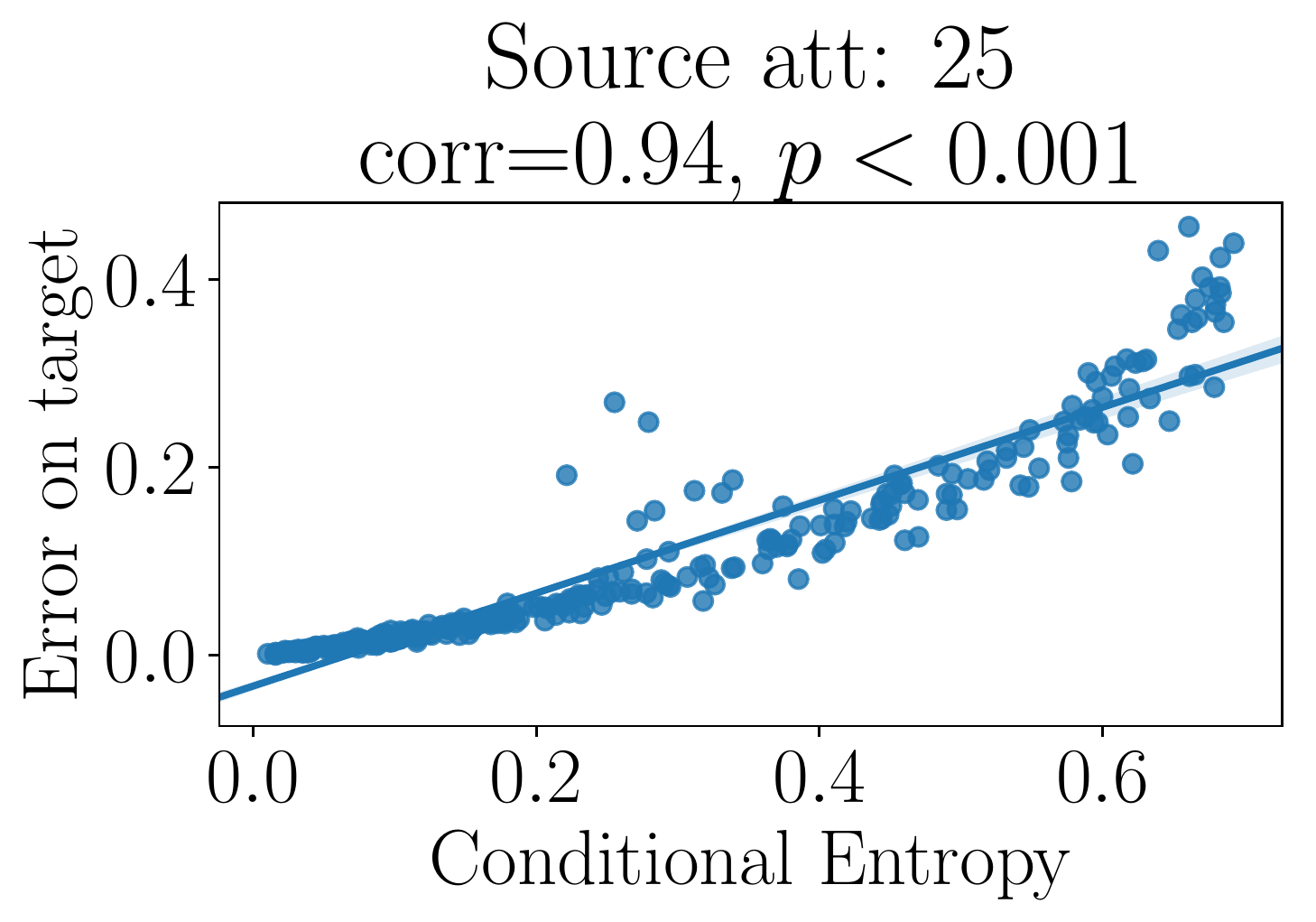}&
    \includegraphics[clip, trim=0mm 0mm 0mm 11mm, width=0.245\textwidth]{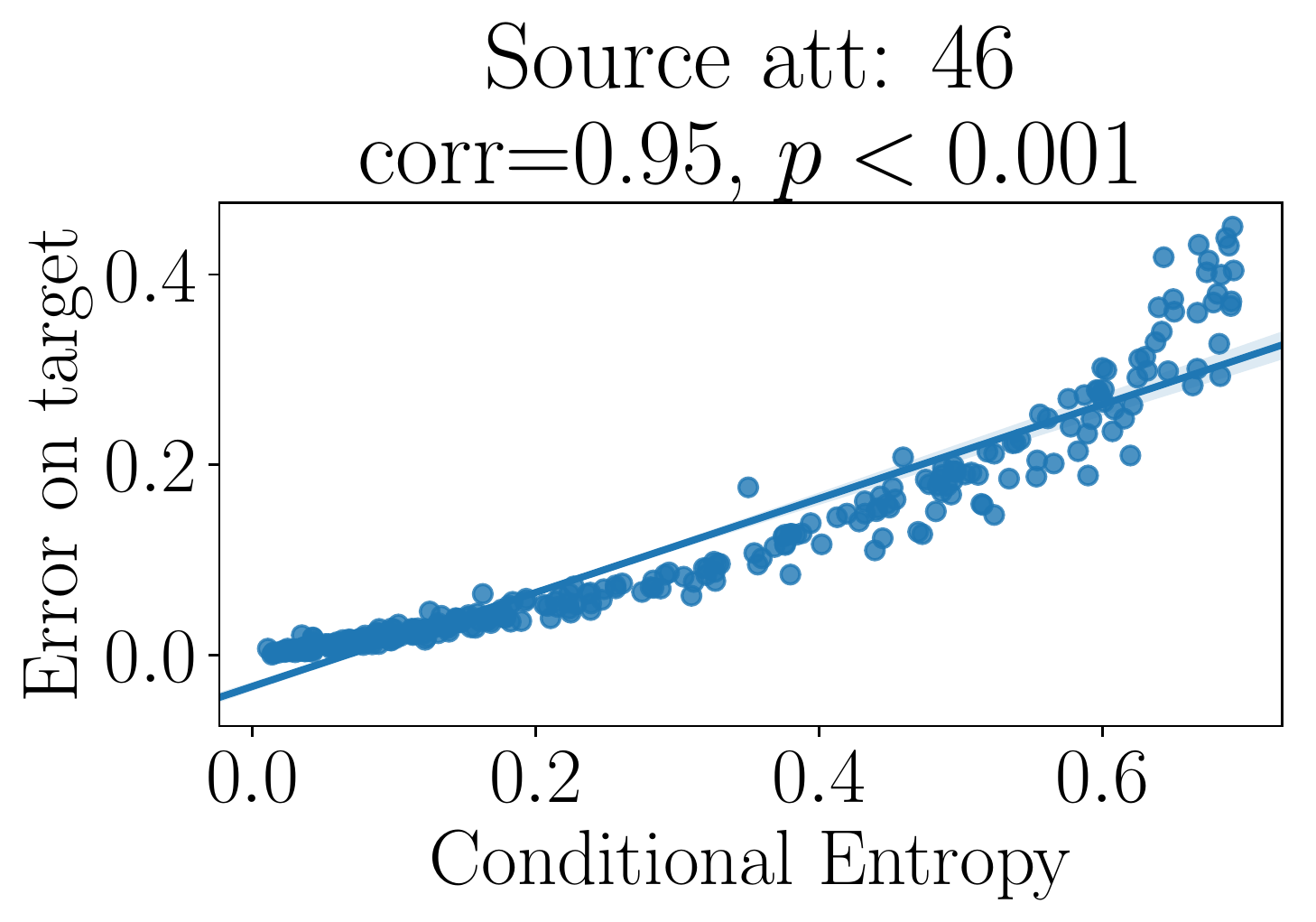}\\
    (i) Curved Bill & (j) Iridescent Wings & (k) Brown Upper Parts & (l) Olive Under Parts
\end{tabular}
\caption{{\bf Attribute prediction; CE vs. test errors on target tasks.} Examples from CelebA (a-d), AwA2 (e-h), and CUB (i-l). Plot titles name the source tasks $T^Z$; points represent different target tasks $T^Y$. Corr is the Pearson correlation coefficient between the two variables and $p$ is the statistical significance of the correlation. In all cases, the correlation is statistically significant. See Sec.~\ref{sec:res:transf} for details.}\vspace{-2mm}
\label{fig:face_att_trans}
\end{figure*}

\begin{table*}[ht]
    \centering{
    \resizebox{1.0\linewidth}{!}{
    \begin{tabular}{c@{~~}lc@{~~~}c@{~~~}c@{~~~}c@{~~~}c@{~~}c@{~~}c@{~~~}c@{~~~}c@{~~~}c@{~~~}c@{~~~}|c}
    \toprule
&Attribute:&Male&Bald&Gray Hair & Mustache & Double Chin &$\hdots$& Attractive& Wavy Hair & High Cheeks & Smiling & Mouth Open&Average (all)\\ \hline
1&LNets+ANet 2015~ \cite{liu2015faceattributes}&0.980&0.980&0.970&0.950&0.920&\multirow{11}{*}{$\hdots$}&0.810&0.800&0.870&0.920&0.920&0.873\\
2&Walk and Learn 2016~\cite{wang2016walk}&0.960&0.920&0.950&0.900&0.930&&0.840&0.850&0.950&0.980&0.970&0.887\\
3&MOON 2016~\cite{rudd2016moon}&0.981&0.988&0.981&0.968&0.963&&0.817&0.825&0.870&0.926&0.935&0.909\\
4&LMLE 2016~\cite{huang2016learning}&0.990&0.900&0.910&0.730&0.740&&0.880&0.830&0.920&0.990&0.960&0.838\\
5&CR-I 2017~\cite{dong2017class}&0.960&0.970&0.950&0.940&0.890&&0.830&0.790&0.890&0.930&0.950&0.866\\
6&MCNN-AUX 2017~\cite{hand2017attributes}&0.982&0.989&0.982&0.969&0.963&&0.831&0.839&0.876&0.927&0.937&0.913\\
7&DMTL 2018~\cite{han2018heterogeneous}&0.980&0.990&0.960&0.970&0.990&&0.850&0.870&0.880&0.940&0.940&0.926\\ 
8&Face-SSD 2019~\cite{jang2019registration}&0.973&0.986&0.976&0.960&0.960&&0.813&0.851&0.868&0.918&0.919&0.903\\ \hline
\rowcolor{lightblue}9&CE$\uparrow$ (decreasing transferability)&0.017&0.026&0.052&0.062&0.083&&0.361&0.381&0.476&0.521&0.551&-\\ \hline
10&Dedicated Res18&0.985&0.990&0.980&0.968&0.959&&0.823&0.842&0.878&0.933&0.943&0.911\\
11&Transfer&0.992&0.991&0.981&0.968&0.963&&0.820&0.800&0.859&0.909&0.901&0.902\\
    \bottomrule
    \end{tabular}
    }
    }
    \caption{{\bf Transferability from face recognition to facial attributes.} Results for CelebA attributes, sorted in ascending order of row~9 (decreasing transferability). Results are shown for the five attributes most and least transferable from recognition. Subject specific attributes, e.g., {\em male} and {\em bald}, are more transferable than expression related attributes such as {\em smiling} and {\em mouth open}. Unsurprisingly, transfer results (row 11) are best on the former than the latter. Rows 1-8 provide published state of the art results. Despite training only an lSVM for attribute, row 11 results are comparable with more elaborate attribute classification systems. For details, see Sec.~\ref{sec:id2attrib}.}\vspace{-3mm}
    \label{tab:id2attribute}
\end{table*}

\section{Task hardness}\label{sec:hardness}
A potential application of transferability is task hardness. 
In Sec.~\ref{sec:CE}, we mentioned that the hardness of a task can be measured from the optimal log-likelihood on that task. Formally, we can measure the hardness of a task $T^Z$ by:
\begin{align}
\mathrm{Hard}(T^Z) =  \min_{w, h \in (W, H)} \mathcal{L}_Z(w, h) = - l_Z(w_Z, h_Z). \label{eq:hardness}
\end{align}
This definition of hardness depends on our choice of $(W, H)$, which may determine various factors such as representation size or network architecture. The intuition behind the definition is that if the task $T^Z$ is hard for all models in $(W, H)$, we should expect higher loss even after training.

Using Theorem~\ref{thrm:transferability}, we can bound $l_Z(w_Z, h_Z)$ in Eq.~\eqref{eq:hardness} by transferring from a trivial task $T^C$ to $T^Z$. We define a {\em trivial task} as the task for which all input values are assigned the same, constant label. Let $C$ be the (constant) label sequences of the trivial task $T^C$. From Theorem~\ref{thrm:transferability} and Eq.~\eqref{eq:hardness}, we can easily show that:
\begin{equation}
\mathrm{Hard}(T^Z) = - l_Z(w_Z, h_Z) \le H(Z|C). \label{eq:est-hardness}
\end{equation}
Thus, we can approximate the hardness of task $T^Z$ by looking at the CE $H(Z|C)$. We note that the CE $H(Z|C)$ is also used to estimate the transferability ${\widetilde{\mathrm{Trf}}(T^C \rightarrow T^Z)}$. So, $\mathrm{Hard}(T^Z)$ is closely related to ${\widetilde{\mathrm{Trf}}(T^C \rightarrow T^Z)}$. Particularly, if task $T^Z$ is hard, we expect it is more difficult to transfer from a trivial task to $T^Z$.

This relationship between hardness and transferability from a trivial task is similar to the one proposed by Task2Vec~\cite{achille2019task2vec}. They too indexed task hardness as the distance from a trivial task. To compute task hardness, however, they required training deep models, whereas we obtain this measure by simply computing $H(Z|C)$ using Eq.~\eqref{eq:CE}.

Of course, estimating the hardness by $H(Z|C)$ ignores the input and is hence only an approximation. In particular, one could possibly design scenarios where this measure would not accurately reflect the hardness of a given task. Our results in Sec.~\ref{sec:res:taskhard} show, however, that these label statistics provide a strong cue for task hardness. 

\begin{figure}[t!]
\centering
\includegraphics[clip, trim=0mm 0mm 0mm 11mm, width=.7\linewidth]{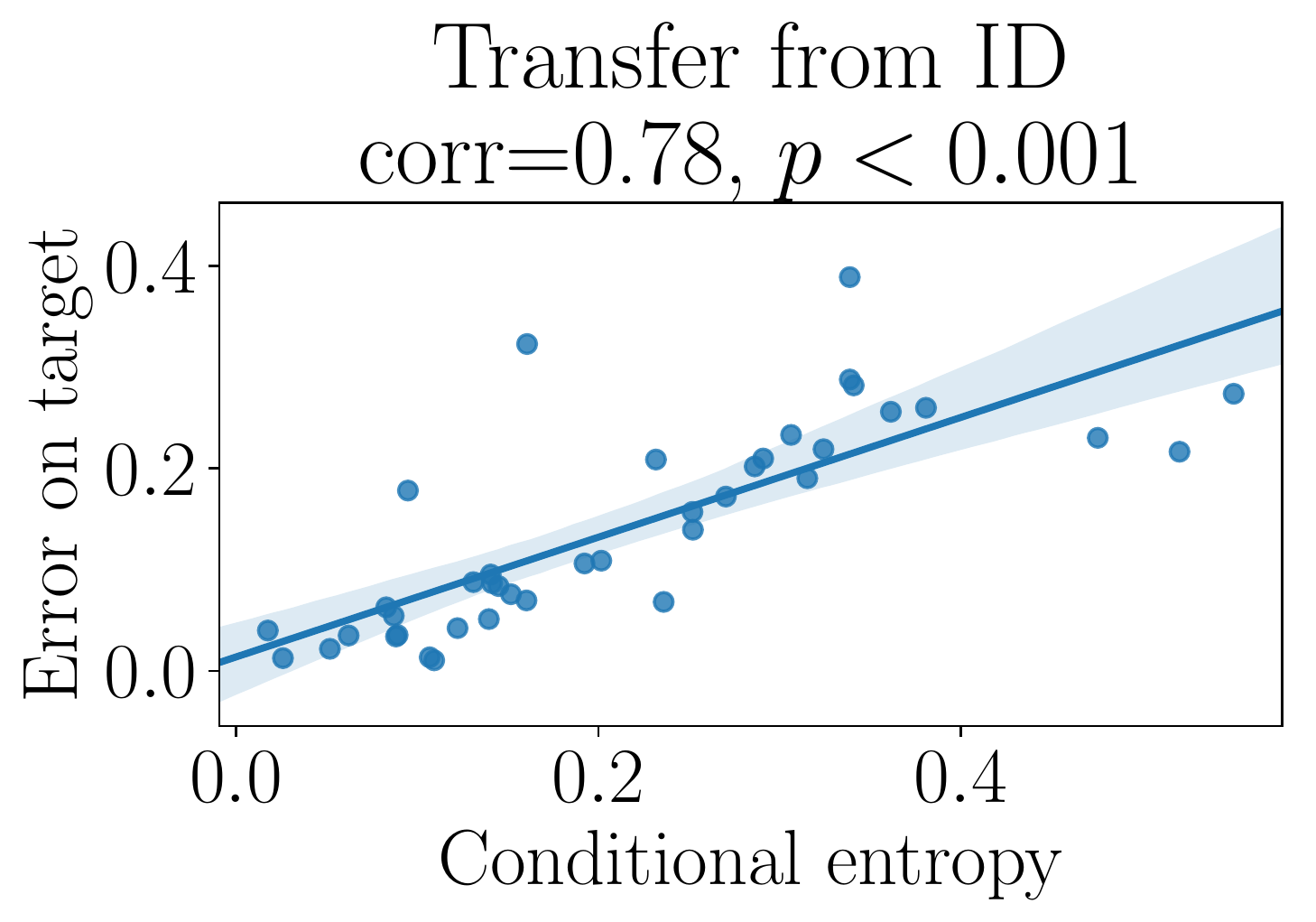}
\caption{{\bf Identity to attribute; CE vs. test errors on target tasks.} Predicting 40 CelebA attributes using a face recognition network. Corr is the Pearson correlation coefficient between the two variables, and $p$ is the statistical significance of the correlation.}
\label{fig:id2att}
\end{figure}

\begin{figure}[t!]
\centering
\includegraphics[clip, trim=0mm 2mm 0mm 0mm, width=.85\linewidth]{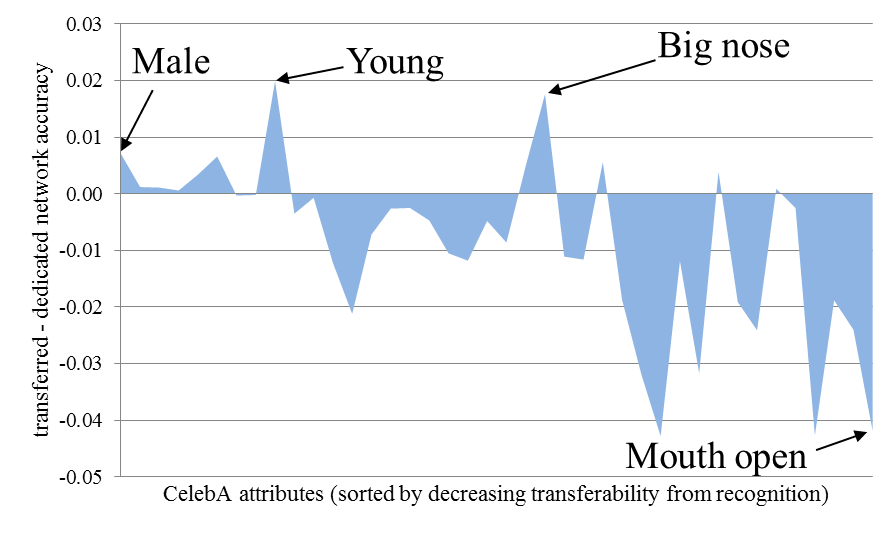}
\caption{{\bf Identity to attribute; transferred $-$ dedicated accuracy.} Differences between CelebA accuracy of transferred recognition model and models trained for each attribute. Results are sorted by decreasing transferability (same as Table~\ref{tab:id2attribute}).
}\vspace{-2mm}
\label{fig:id2att:comparewithdedicated}
\end{figure}

\section{Experiments}\label{experiments}
We rigorously evaluate our claims using three large scale, widely used data sets representing 437 classification tasks. Although Sec.~\ref{sec:CE} provides a bound on the training loss, test accuracy is generally more important. We thus report results on test images not included in the training data.

\minisection{Benchmarks} The {\em Celeb Faces Attributes} (CelebA) set~\cite{liu2015faceattributes} was extensively used to evaluate transfer learning~\cite{dupont2018learning,liu2018exploring,liu2018detach,shankar2016refining}. CelebA contains over 202k face images of 10,177 subjects. Each image is labeled with subject identity as well as 40 binary attributes. We used the standard train / test splits (182,626 / 19,961 images, respectively). To our knowledge, of the three sets, it is the only one that provides baseline results for attribute classification.

{\em Animals with Attributes~2} (AwA2)~\cite{xian2018zero} includes over 37k images labeled as belonging to one of 50 animals classes. Images are labeled based on their class association with 85 different attributes. Models were trained on 33,568 training images and tested on 3,754 separate test images.

Finally, {\em Caltech-UCSD Birds 200} (CUB)~\cite{WelinderEtal2010} offers 11,788 images of 200 bird species, labeled with 312 attributes as well as {\em Turker Confidence} attributes. Labels were averaged across multiple Turkers using confidences. Finally, we kept only reliable labels, using a threshold of 0.5 on the average confidence value. We used 5,994 images for training and 5,794 images for testing. 

We note that the {\em Task Bank} set with its 26 tasks was also used for evaluating task relationships~\cite{zamir2018taskonomy}. We did not use it here as it mostly contains regression tasks rather than the classification problems we are concerned with.

\subsection{Evaluating task transferability}\label{sec:res:transf}
We compared our transferability estimates from the CE to the actual transferability of Eq.~\eqref{eq:true-transferability}. To this end, for each attribute~$T^Z$ in a data set, we measure the actual transferability, $\mathrm{Trf}(T^Z \rightarrow T^Y)$, to all other attributes~$T^Y$ in that set using the test split. We then compare these transferability scores to the corresponding CE estimates of Eq.~\eqref{eq:CE} using an existing correlation analysis~\cite{nguyen2019toward}.

We note again that when the source task is fixed, as in this case, the transferability estimates can be obtained by considering only the CE. Furthermore, since $\mathrm{Trf}(T^Z \rightarrow T^Y)$ and the CE $H(Y|Z)$ are negative correlated, we compare the correlation between the test error rate, $1 - \mathrm{Trf}(T^Z \rightarrow T^Y)$, and the CE $H(Y|Z)$ instead.

\minisection{Transferring representations} We keep the learned representation, $w_Z$, and produce a new classifier $k_Y$ by training on the target task (Sec.~\ref{sec:transf}). We used ResNet18~\cite{He_2016_CVPR}, trained with standard cross entropy loss, on each source task~$T^Z$ (source attribute). These networks were selected as they were deep enough to obtain good accuracy on our benchmarks, but not too deep to overfit~\cite{zhang2016understanding}. The penultimate layer of these networks produce embeddings $r\in \mathbb{R}^{2048}$ which the networks classified using $h_Z$---their last, fully connected (FC) layers---to binary attribute values.

We transferred from source to target task by freezing the networks, only replacing their FC layers with linear SVM (lSVM). These lSVM were trained to predict the binary labels of {\em target} tasks given the embeddings produced for the source tasks by~$w_Z$ as their input. The test errors of the lSVM, which are measures of ${1 - \mathrm{Trf}(T^Z \rightarrow T^Y)}$, were then compared with the CE, $H(Y|Z)$.

We use lSVM as it allows us to focus on the information passed from $T^Z$ to $T^Y$. A more complex classifier could potentially mask this information by being powerful enough to offset any loss of information due to the transfer. In practical use cases, when transferring a deep network from one task to another, it may be preferable to fine tune the last layers of the network or its entirety, provided that the training data on the target task is large enough.\vspace{-1mm}

\begin{figure}[t!]
\centering
\includegraphics[clip, trim=0mm 2mm 0mm 2mm, width=.72\linewidth]{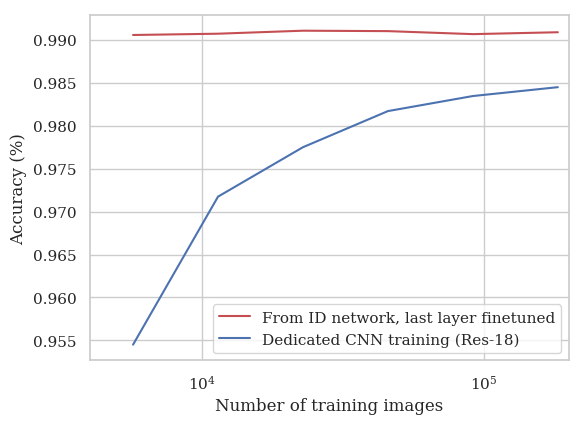}\\
(a) Male\\
\includegraphics[clip, trim=0mm 2mm 0mm 2mm, width=.72\linewidth]{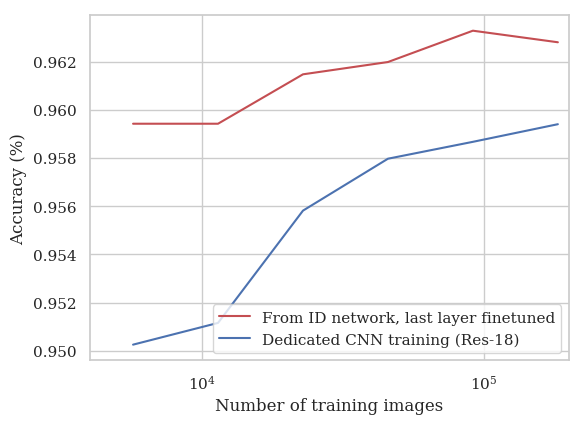}\\
(b) Double chin
\caption{{\bf Classification accuracy for varying training set sizes.} Top: {\em male}; bottom: {\em double chin}. Dedicated classification networks trained from scratch (blue) vs. face recognition network transferred to the attributes with an lSVM (red). Because recognition transfers well to these attributes, we obtain accurate classification with a fraction of the training data and effort.}\vspace{-3mm}
\label{fig:id2att4trainsize}
\end{figure}

\minisection{Transferability results} Fig.~\ref{fig:face_att_trans} reports selected quantitative transferability results on the three sets.\footnote{For full results see the appendix.} Each point in these graphs represents the CE, $H(Y|Z)$, vs. the target test error, ${1 - \mathrm{Trf}(T^Z \rightarrow T^Y)}$. The graphs also provide the linear regression model fit with 95\% confidence interval, the Pearson correlation coefficients between the two values, and the statistical significance of the correlation,~$p$.

In all cases, the CE and target test error are highly positively correlated with statistical significance. These results testify that the CE of Eq.~\eqref{eq:CE} is indeed a good predictor for the actual transferability of Eq.~\eqref{eq:true-transferability}. This is remarkable especially since the relationship between tasks is evaluated without considering the input domain or the machine learning models trained to solve these tasks.

\begin{figure*}[t!]
\centering
\begin{tabular}{ccc}
     \includegraphics[clip,trim=0mm 2mm 0mm 2mm, width=.30\linewidth]{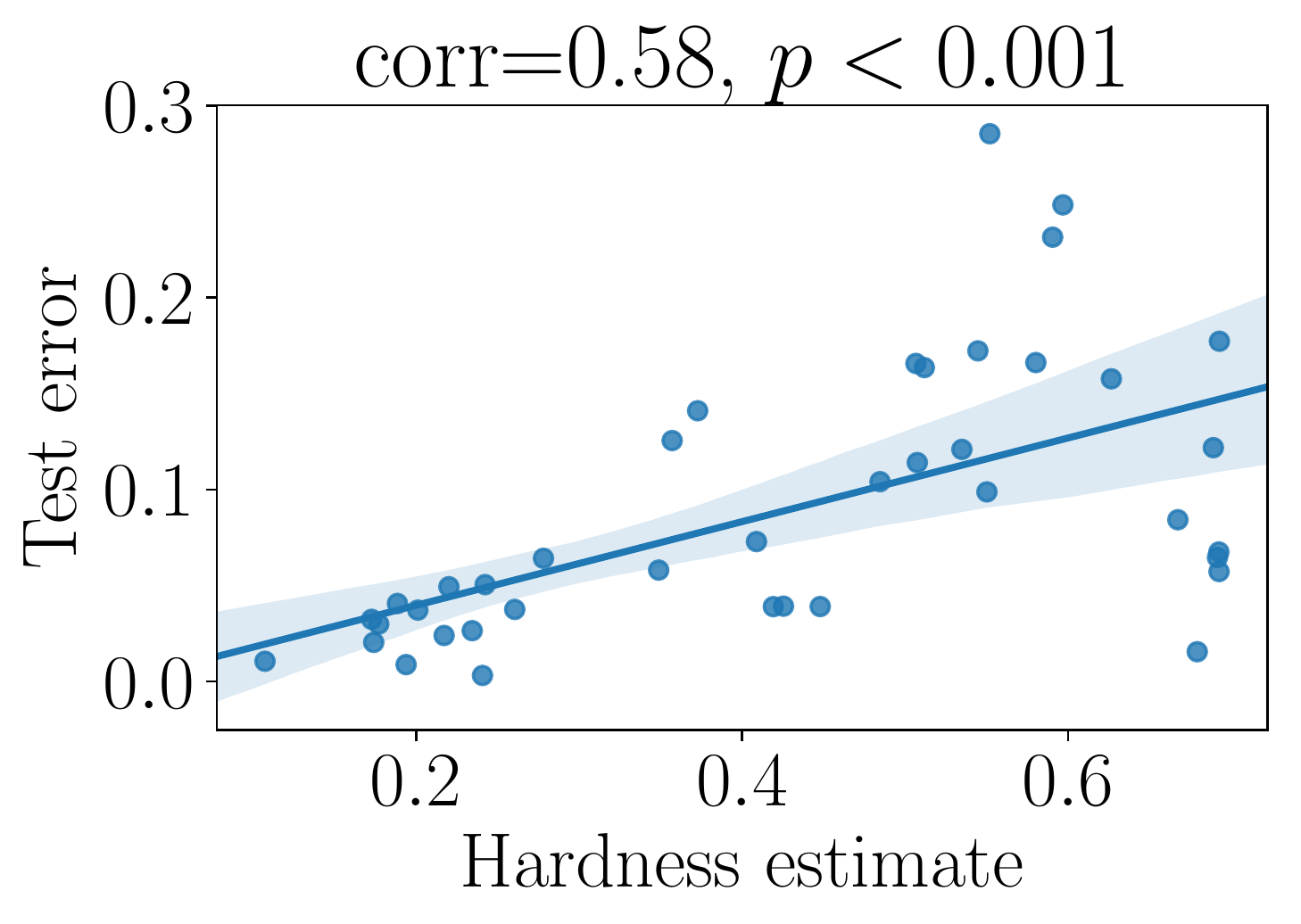}&
     \includegraphics[clip,trim=0mm 2mm 0mm 2mm, width=.30\linewidth]{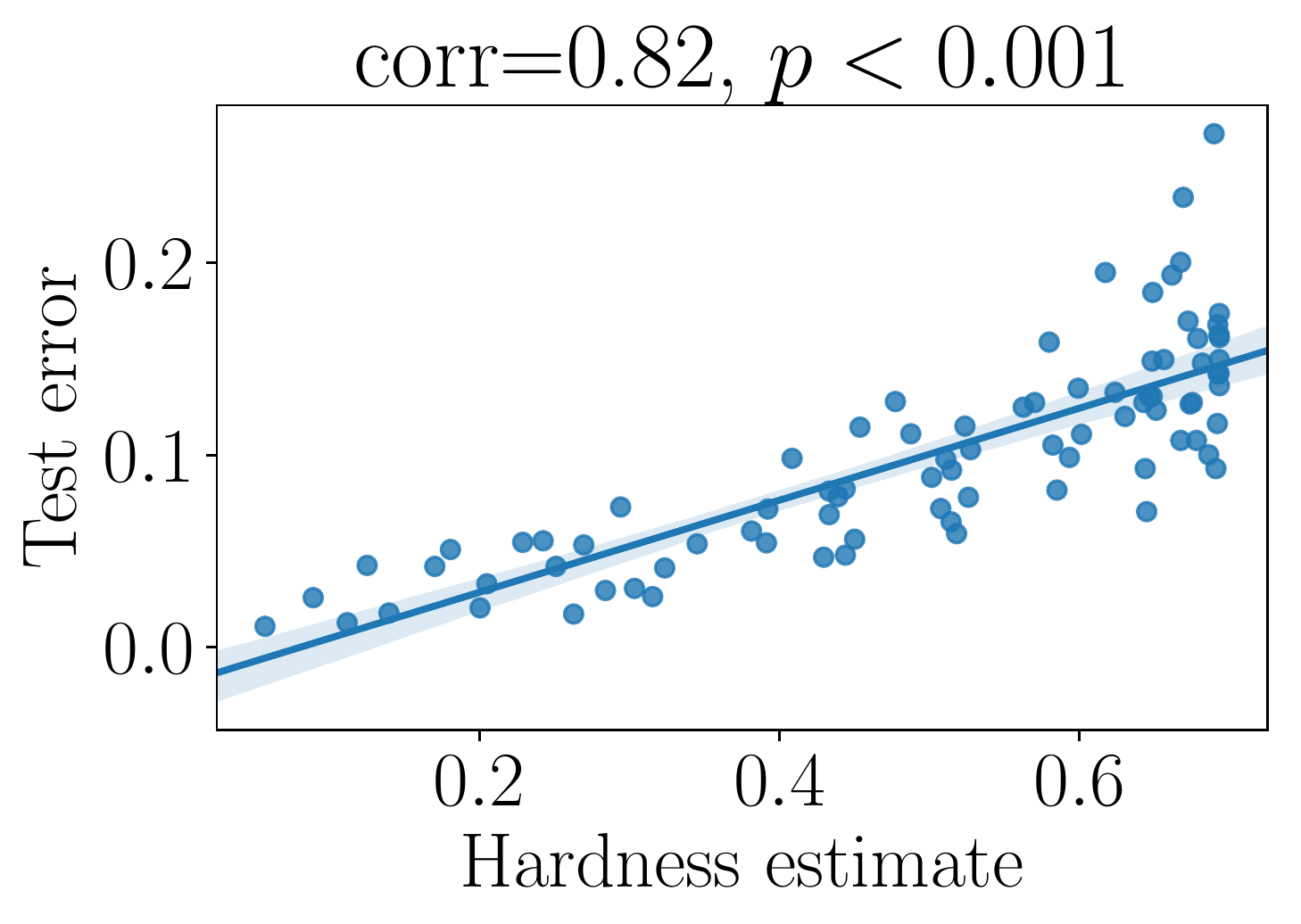} & 
    \includegraphics[clip,trim=0mm 2mm 0mm 2mm, width=.30\linewidth]{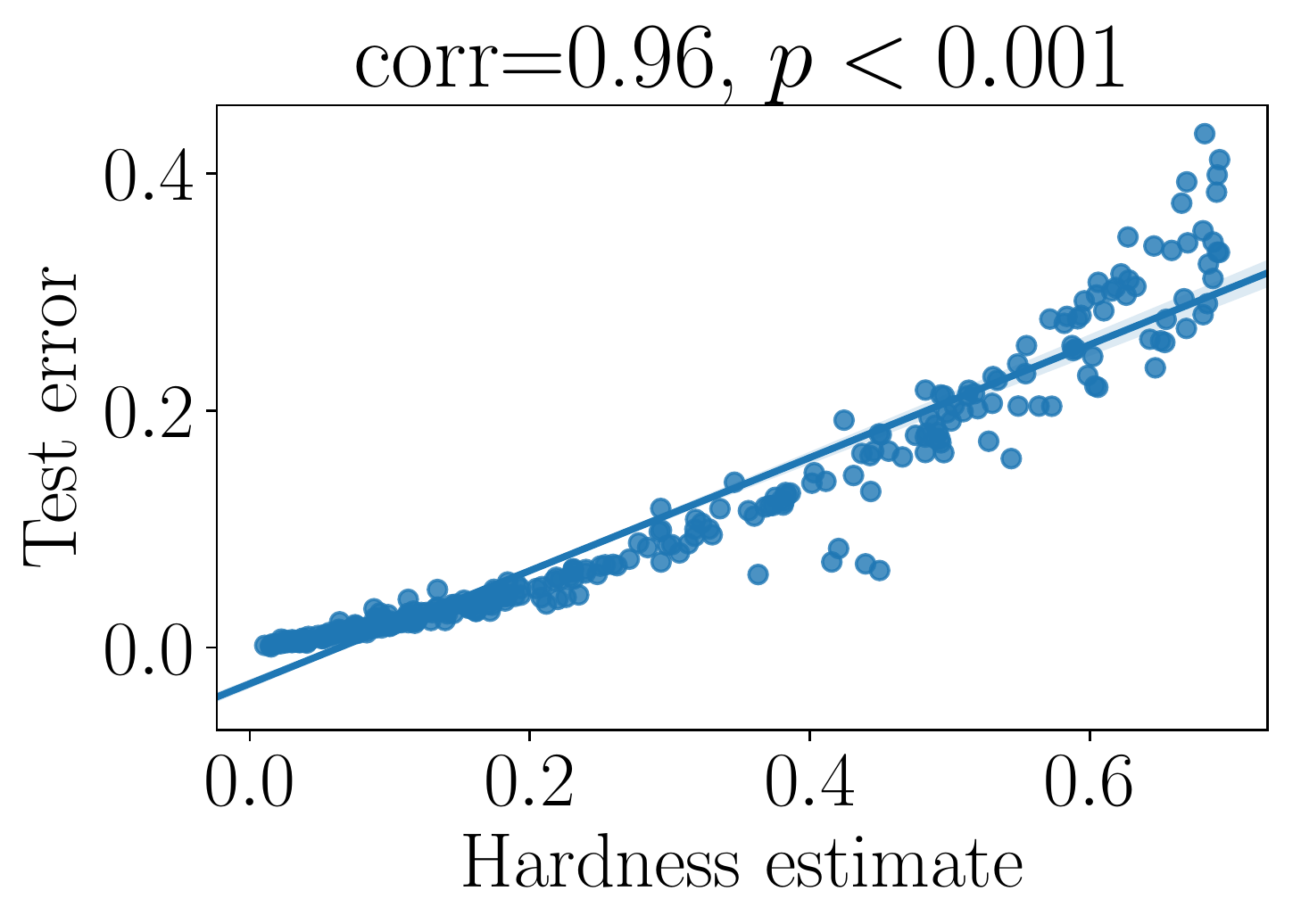}\\
    (a) CelebA & (b) AwA2 & (c) CUB
\end{tabular}
\caption{{\bf Estimated task hardness vs. empirical errors} on the three benchmarks. Estimated hardness is well correlated with empirical hardness with significance $p<0.001$.}\vspace{-3mm}
\label{fig:hardness}
\end{figure*}

\subsection{Case study: Identity to facial attributes}\label{sec:id2attrib}
A key challenge when training effective attribute classifiers is the difficulty of obtaining labeled attribute training data. Whereas face images are often uploaded to the Internet along with subject names~\cite{Cao18,guo2016msceleb}, it is far less common to find images labeled with attributes such as {\em high cheek bones}, {\em bald}, or even {\em male}~\cite{LH:CVPRw15:age}. It is consequently harder to assemble training sets for attribute classification at the same scale and diversity as those used to train other tasks.

To reduce the burden of collecting attribute data, we therefore explore transferring a representation learned for face recognition. In this setting, we can also compute estimated transferability scores (via the CE) between the subject labels provided by CelebA and the labels of each attribute. We note that unlike the previous examples, the source labels are not binary and include over 10k values.

\minisection{Face recognition network} We compare our estimated transferability vs. actual transferability using a deep face recognition network. To this end, we use a ResNet101 architecture trained for face recognition on the union of the MS-Celeb-1M~\cite{guo2016msceleb} and VGGFace2~\cite{Cao18} training sets (following removal of subjects included in CelebA), with a cosine margin loss ($m=0.4$)~\cite{wang2018cosface}. This network achieves accuracy comparable to the state of the art reported by others, with different systems, on standard benchmarks~\cite{dhar2019measuring}.

\minisection{Transferability results: recognition to attributes} Table~\ref{tab:id2attribute} reports results for the five attributes most transferable from recognition (smallest CE; Eq.~\eqref{eq:CE}) and the five least transferable (largest CE). Columns are sorted by increasing CE values (decreasing transferability), listed in row 9. Row 11 reports accuracy of the transferred network with the lSVM trained on the target task. Estimated vs. actual transferability is further visualized in Fig.~\ref{fig:id2att}. Evidently, correlation between the two is statistically significant, testifying that Eq.~\eqref{eq:CE} is a good predictor of actual transferability, here demonstrated on a source task with multiple labels.

For reference, Table~\ref{tab:id2attribute} provides in Row 10 the accuracy of the dedicated ResNet18 networks trained for each attribute. Finally, rows 1 through 8 provide results for published state of the art on the same tasks.

\minisection{Analysis of results} Subject specific attributes such as {\em male} and {\em bald} are evidently more transferable from recognition (left columns of Table~\ref{tab:id2attribute}) than attributes that are related to expressions (e.g., {\em smiling} and {\em mouth open}, right columns). Although this relationship has been noted by others, previous work used domain knowledge to determine which attributes are more transferable from identity~\cite{liu2015faceattributes}, as others have done in other domains~\cite{ghadiyaram2019large,mahajan2018exploring}. By comparison, our work shows how these relationships emerge from our estimation of transferability. 

Also, notice that for the transferable attributes, our results are comparable to dedicated networks trained for each attribute, although they gradually drop off for the less transferable attributes in the last columns. This effect is visualized in Fig.~\ref{fig:id2att:comparewithdedicated} which shows the growing differences in attribute classification accuracy for a transferred face recognition model and models trained for each attribute. Results are sorted by decreasing transferability (same as in Table~\ref{tab:id2attribute}).

Results in Fig.~\ref{fig:id2att:comparewithdedicated} show a few notable exceptions where transfer performs substantially better than dedicated models (e.g., the two positive {\em peaks} representing attributes {\em young} and {\em big nose}). These and other occasional discrepancies in our results can be explained in the difference between the true transferability of Eq.~\eqref{eq:true-transferability}, which we measure on the test sets, and Eq.~\eqref{eq:transferability}, defined on the training sets and shown in Sec.~\ref{sec:CE} to be bounded by the CE.

Finally, we note that our goal is not to develop a state of the art facial attribute classification scheme. Nevertheless, results obtained by training an lSVM on embeddings transferred from a face recognition network are only 2.4\% lower than the best scores reported by DMTL 2018~\cite{han2018heterogeneous} (last column of Table~\ref{tab:id2attribute}). The effort involved in developing a state of the art face recognition network can be substantial. By transferring this network to attributes these efforts are amortized in training multiple facial attribute classifiers.

To emphasize this last point, consider Fig.~\ref{fig:id2att4trainsize} which reports classification accuracy on {\em male} and {\em double chin} for growing training set sizes. These attributes were selected as they are highly transferable from recognition (see Table~\ref{tab:id2attribute}). The figure compares the accuracy obtained by training a dedicated network (in blue) to a network transferred from recognition (red). Evidently, on these attributes, transferred accuracy is much higher with far less training data.

\subsection{Evaluating task hardness}\label{sec:res:taskhard}
\vspace{-1mm}
We evaluate our hardness estimates for all attribute classification tasks in the three data sets, using the CE $H(Z|C)$ in Eq.~\eqref{eq:est-hardness}. Fig.~\ref{fig:hardness} compares the hardness estimates for each task vs. the errors of our dedicated networks, trained from scratch to classify each attribute. Results are provided for CelebA, AwA2, and CUB.

The correlation between estimated hardness and classification errors is statistically significant with $p<0.001$, suggesting that the CE $H(Z|C)$ in Eq.~\eqref{eq:est-hardness} indeed captures the hardness of these tasks. That is, in the three data sets, test error rates strongly correlate with our estimated hardness: the harder a task is estimated to be, the higher the errors produced by the model trained for the task. Of course, this result does not imply that the input domain has no impact on task hardness; only that the distribution of training labels already provides a strong predictor for task hardness.

\vspace{-2mm}
\section{Conclusions}
\vspace{-1.5mm}
We present a practical method for estimating the hardness and transferability of supervised classification tasks. We show that, in both cases, we produce reliable estimates by exploring training label statistics, particularly the conditional entropy between the sequences of labels assigned to the training data of each task. This approach is simpler than existing work, which obtains similar estimates by assuming the existence of trained models or by careful inspection of the training process. In our approach, computing conditional entropy is cheaper than training deep models, required by others for the same purpose.

We assume that different tasks share the same input domain (the same input images). It would be useful to extend our work to settings where the two tasks are defined over different domains (e.g., face vs. animal images). Our work further assumes discrete labels. Conditional entropy was originally defined over distributions. It is therefore reasonable that CE could be extended to non-discrete labeled tasks, such as, for faces, 3D reconstruction~\cite{tran2017extreme}, pose estimation~\cite{chang17fpn,chang2019deep} or segmentation~\cite{nirkin2018face}.

\vspace{1mm}
\noindent\textbf{Acknowledgements.} We thank Alessandro Achille, Pietro Perona, and the reviewers for their helpful discussions.

{\small
\bibliographystyle{ieee_fullname}
\bibliography{transf}

\begin{thebibliography}{10}\itemsep=-1pt

\bibitem{achille2019task2vec}
Alessandro Achille, Michael Lam, Rahul Tewari, Avinash Ravichandran, Subhransu
  Maji, Charless Fowlkes, Stefano Soatto, and Pietro Perona.
\newblock {Task2Vec}: Task embedding for meta-learning.
\newblock {\em arXiv preprint arXiv:1902.03545}, 2019.

\bibitem{achille2019information}
Alessandro Achille, Giovanni Paolini, Glen Mbeng, and Stefano Soatto.
\newblock The information complexity of learning tasks, their structure and
  their distance.
\newblock {\em arXiv preprint arXiv:1904.03292}, 2019.

\bibitem{arora2014provable}
Sanjeev Arora, Aditya Bhaskara, Rong Ge, and Tengyu Ma.
\newblock Provable bounds for learning some deep representations.
\newblock In {\em Int. Conf. Mach. Learning}, pages 584--592, 2014.

\bibitem{azizpour2015factors}
Hossein Azizpour, Ali~Sharif Razavian, Josephine Sullivan, Atsuto Maki, and
  Stefan Carlsson.
\newblock Factors of transferability for a generic convnet representation.
\newblock {\em Trans. Pattern Anal. Mach. Intell.}, 38(9):1790--1802, 2015.

\bibitem{azizzadenesheli2018regularized}
Kamyar Azizzadenesheli, Anqi Liu, Fanny Yang, and Animashree Anandkumar.
\newblock Regularized learning for domain adaptation under label shifts.
\newblock In {\em Int. Conf. on Learning Representations}, 2019.

\bibitem{ben2010theory}
Shai Ben-David, John Blitzer, Koby Crammer, Alex Kulesza, Fernando Pereira, and
  Jennifer~Wortman Vaughan.
\newblock A theory of learning from different domains.
\newblock {\em Mach. Learn.}, 79(1-2):151--175, 2010.

\bibitem{ben2003exploiting}
Shai Ben-David and Reba Schuller.
\newblock Exploiting task relatedness for multiple task learning.
\newblock In {\em Learning Theory and Kernel Machines}, pages 567--580.
  Springer, 2003.

\bibitem{blitzer2008learning}
John Blitzer, Koby Crammer, Alex Kulesza, Fernando Pereira, and Jennifer
  Wortman.
\newblock Learning bounds for domain adaptation.
\newblock In {\em Neural Inform. Process. Syst.}, pages 129--136, 2008.

\bibitem{Cao18}
Q. Cao, L. Shen, W. Xie, O.~M. Parkhi, and A. Zisserman.
\newblock {VGGFace2}: A dataset for recognising faces across pose and age.
\newblock In {\em Automatic Face and Gesture Recognition}, 2018.

\bibitem{chang17fpn}
Feng-Ju Chang, Anh Tran, Tal Hassner, Iacopo Masi, Ram Nevatia, and G\'{e}rard
  Medioni.
\newblock Faceposenet: Making a case for landmark-free face alignment.
\newblock In {\em Proc. Int. Conf. Comput. Vision Workshops}, 2017.

\bibitem{chang2019deep}
Feng-Ju Chang, Anh~Tuan Tran, Tal Hassner, Iacopo Masi, Ram Nevatia, and
  G{\'e}rard Medioni.
\newblock Deep, landmark-free fame: Face alignment, modeling, and expression
  estimation.
\newblock {\em Int. J. Comput. Vision}, 127(6-7):930--956, 2019.

\bibitem{chen2015mxnet}
Tianqi Chen, Mu Li, Yutian Li, Min Lin, Naiyan Wang, Minjie Wang, Tianjun Xiao,
  Bing Xu, Chiyuan Zhang, and Zheng Zhang.
\newblock Mxnet: A flexible and efficient machine learning library for
  heterogeneous distributed systems.
\newblock {\em arXiv preprint arXiv:1512.01274}, 2015.

\bibitem{choromanska2015loss}
Anna Choromanska, Mikael Henaff, Michael Mathieu, G{\'e}rard~Ben Arous, and
  Yann LeCun.
\newblock The loss surfaces of multilayer networks.
\newblock In {\em Artificial Intelligence and Statistics}, pages 192--204,
  2015.

\bibitem{cover2012elements}
Thomas~M Cover and Joy~A Thomas.
\newblock {\em Elements of information theory}.
\newblock John Wiley \& Sons, 2012.

\bibitem{dhar2019measuring}
Prithviraj Dhar, Carlos Castillo, and Rama Chellappa.
\newblock On measuring the iconicity of a face.
\newblock In {\em Winter Conf. on App. of Comput. Vision}, pages 2137--2145.
  IEEE, 2019.

\bibitem{dong2017class}
Qi Dong, Shaogang Gong, and Xiatian Zhu.
\newblock Class rectification hard mining for imbalanced deep learning.
\newblock In {\em Proc. Conf. Comput. Vision Pattern Recognition}, pages
  1851--1860, 2017.

\bibitem{duda2012pattern}
Richard~O Duda, Peter~E Hart, and David~G Stork.
\newblock {\em Pattern classification}.
\newblock John Wiley \& Sons, 2012.

\bibitem{dupont2018learning}
Emilien Dupont.
\newblock Learning disentangled joint continuous and discrete representations.
\newblock In {\em Neural Inform. Process. Syst.}, pages 708--718, 2018.

\bibitem{edwards2016towards}
Harrison Edwards and Amos Storkey.
\newblock Towards a neural statistician.
\newblock {\em arXiv preprint arXiv:1606.02185}, 2016.

\bibitem{ghadiyaram2019large}
Deepti Ghadiyaram, Du Tran, and Dhruv Mahajan.
\newblock Large-scale weakly-supervised pre-training for video action
  recognition.
\newblock In {\em Proc. Conf. Comput. Vision Pattern Recognition}, pages
  12046--12055, 2019.

\bibitem{guo2016msceleb}
Yandong Guo, Lei Zhang, Yuxiao Hu, Xiaodong He, and Jianfeng Gao.
\newblock M{S}-{C}eleb-1{M}: A dataset and benchmark for large scale face
  recognition.
\newblock In {\em European Conf. Comput. Vision}. Springer, 2016.

\bibitem{han2018heterogeneous}
Hu Han, Anil~K Jain, Fang Wang, Shiguang Shan, and Xilin Chen.
\newblock Heterogeneous face attribute estimation: A deep multi-task learning
  approach.
\newblock {\em Trans. Pattern Anal. Mach. Intell.}, 40(11):2597--2609, 2018.

\bibitem{hand2017attributes}
Emily~M Hand and Rama Chellappa.
\newblock Attributes for improved attributes: A multi-task network utilizing
  implicit and explicit relationships for facial attribute classification.
\newblock In {\em AAAI Conf. on Artificial Intelligence}, 2017.

\bibitem{he2017mask}
Kaiming He, Georgia Gkioxari, Piotr Doll{\'a}r, and Ross Girshick.
\newblock Mask r-cnn.
\newblock In {\em Proc. Int. Conf. Comput. Vision}, pages 2961--2969, 2017.

\bibitem{He_2016_CVPR}
Kaiming He, Xiangyu Zhang, Shaoqing Ren, and Jian Sun.
\newblock Deep residual learning for image recognition.
\newblock In {\em Proc. Conf. Comput. Vision Pattern Recognition}, June 2016.

\bibitem{huang2016learning}
Chen Huang, Yining Li, Chen Change~Loy, and Xiaoou Tang.
\newblock Learning deep representation for imbalanced classification.
\newblock In {\em Proc. Conf. Comput. Vision Pattern Recognition}, pages
  5375--5384, 2016.

\bibitem{jang2019registration}
Youngkyoon Jang, Hatice Gunes, and Ioannis Patras.
\newblock Registration-free face-ssd: Single shot analysis of smiles, facial
  attributes, and affect in the wild.
\newblock {\em Comput. Vision Image Understanding}, 2019.

\bibitem{jou2016deep}
Brendan Jou and Shih-Fu Chang.
\newblock Deep cross residual learning for multitask visual recognition.
\newblock In {\em Int. Conf. Multimedia}, pages 998--1007. ACM, 2016.

\bibitem{kendall2018multi}
Alex Kendall, Yarin Gal, and Roberto Cipolla.
\newblock Multi-task learning using uncertainty to weigh losses for scene
  geometry and semantics.
\newblock In {\em Proceedings of the IEEE Conference on Computer Vision and
  Pattern Recognition}, pages 7482--7491, 2018.

\bibitem{kokkinos2017ubernet}
Iasonas Kokkinos.
\newblock Ubernet: Training a universal convolutional neural network for low-,
  mid-, and high-level vision using diverse datasets and limited memory.
\newblock In {\em Proc. Conf. Comput. Vision Pattern Recognition}, pages
  6129--6138, 2017.

\bibitem{lee2016asymmetric}
Giwoong Lee, Eunho Yang, and Sung Hwang.
\newblock Asymmetric multi-task learning based on task relatedness and loss.
\newblock In {\em Int. Conf. Mach. Learning}, pages 230--238, 2016.

\bibitem{LH:CVPRw15:age}
Gil Levi and Tal Hassner.
\newblock Age and gender classification using convolutional neural networks.
\newblock In {\em Proc. Conf. Comput. Vision Pattern Recognition Workshops},
  June 2015.

\bibitem{liu2018exploring}
Yu Liu, Fangyin Wei, Jing Shao, Lu Sheng, Junjie Yan, and Xiaogang Wang.
\newblock Exploring disentangled feature representation beyond face
  identification.
\newblock In {\em Proc. Conf. Comput. Vision Pattern Recognition}, pages
  2080--2089, 2018.

\bibitem{liu2018detach}
Yen-Cheng Liu, Yu-Ying Yeh, Tzu-Chien Fu, Sheng-De Wang, Wei-Chen Chiu, and
  Yu-Chiang Frank~Wang.
\newblock Detach and adapt: Learning cross-domain disentangled deep
  representation.
\newblock In {\em Proc. Conf. Comput. Vision Pattern Recognition}, pages
  8867--8876, 2018.

\bibitem{liu2015faceattributes}
Ziwei Liu, Ping Luo, Xiaogang Wang, and Xiaoou Tang.
\newblock Deep learning face attributes in the wild.
\newblock In {\em Proc. Int. Conf. Comput. Vision}, 2015.

\bibitem{lowe1999object}
David~G Lowe.
\newblock Object recognition from local scale-invariant features.
\newblock In {\em Proc. Int. Conf. Comput. Vision}, page 1150, 1999.

\bibitem{lu2017fully}
Yongxi Lu, Abhishek Kumar, Shuangfei Zhai, Yu Cheng, Tara Javidi, and Rogerio
  Feris.
\newblock Fully-adaptive feature sharing in multi-task networks with
  applications in person attribute classification.
\newblock In {\em Proc. Conf. Comput. Vision Pattern Recognition}, pages
  5334--5343, 2017.

\bibitem{mahajan2018exploring}
Dhruv Mahajan, Ross Girshick, Vignesh Ramanathan, Kaiming He, Manohar Paluri,
  Yixuan Li, Ashwin Bharambe, and Laurens van~der Maaten.
\newblock Exploring the limits of weakly supervised pretraining.
\newblock In {\em European Conf. Comput. Vision}, pages 181--196, 2018.

\bibitem{mansour2009domain}
Yishay Mansour, Mehryar Mohri, and Afshin Rostamizadeh.
\newblock Domain adaptation: Learning bounds and algorithms.
\newblock In {\em Conference on Learning Theory}, 2009.

\bibitem{Masi:18:learning}
I. Masi, F.~J. Chang, J. Choi, S. Harel, J. Kim, K. Kim, J. Leksut, S. Rawls,
  Y. Wu, T. Hassner, W. AbdAlmageed, G. Medioni, L.~P. Morency, P. Natarajan,
  and R. Nevatia.
\newblock Learning pose-aware models for pose-invariant face recognition in the
  wild.
\newblock {\em Trans. Pattern Anal. Mach. Intell.}, 2018.

\bibitem{masi2019face}
Iacopo Masi, Anh~Tuan Tran, Tal Hassner, Gozde Sahin, and G{\'e}rard Medioni.
\newblock Face-specific data augmentation for unconstrained face recognition.
\newblock {\em Int. J. Comput. Vision}, 127(6-7):642--667, 2019.

\bibitem{mcdiarmid1989method}
Colin McDiarmid.
\newblock On the method of bounded differences.
\newblock {\em Surveys in combinatorics}, 141(1):148--188, 1989.

\bibitem{misra2016cross}
Ishan Misra, Abhinav Shrivastava, Abhinav Gupta, and Martial Hebert.
\newblock Cross-stitch networks for multi-task learning.
\newblock In {\em Proc. Conf. Comput. Vision Pattern Recognition}, pages
  3994--4003, 2016.

\bibitem{nguyen2019toward}
Cuong~V Nguyen, Alessandro Achille, Michael Lam, Tal Hassner, Vijay Mahadevan,
  and Stefano Soatto.
\newblock Toward understanding catastrophic forgetting in continual learning.
\newblock {\em arXiv:1908.01091}, 2019.

\bibitem{nguyen2018variational}
Cuong~V Nguyen, Yingzhen Li, Thang~D Bui, and Richard~E Turner.
\newblock Variational continual learning.
\newblock In {\em Int. Conf. on Learning Representations}, 2018.

\bibitem{nguyen2017loss}
Quynh Nguyen and Matthias Hein.
\newblock The loss surface of deep and wide neural networks.
\newblock In {\em Int. Conf. Mach. Learning}, pages 2603--2612, 2017.

\bibitem{nirkin2018face}
Yuval Nirkin, Iacopo Masi, Anh~Tran Tuan, Tal Hassner, and Gerard Medioni.
\newblock On face segmentation, face swapping, and face perception.
\newblock In {\em Int. Conf. on Automatic Face and Gesture Recognition}, pages
  98--105. IEEE, 2018.

\bibitem{pan2010survey}
Sinno~Jialin Pan and Qiang Yang.
\newblock A survey on transfer learning.
\newblock {\em Trans. Knowledge and Data Eng.}, 22(10):1345--1359, 2010.

\bibitem{scikit-learn}
F. Pedregosa, G. Varoquaux, A. Gramfort, V. Michel, B. Thirion, O. Grisel, M.
  Blondel, P. Prettenhofer, R. Weiss, V. Dubourg, J. Vanderplas, A. Passos, D.
  Cournapeau, M. Brucher, M. Perrot, and E. Duchesnay.
\newblock Scikit-learn: Machine learning in {P}ython.
\newblock {\em J. Mach. Learning Research}, 12:2825--2830, 2011.

\bibitem{ranjan2019hyperface}
Rajeev Ranjan, Vishal~M Patel, and Rama Chellappa.
\newblock Hyperface: A deep multi-task learning framework for face detection,
  landmark localization, pose estimation, and gender recognition.
\newblock {\em Trans. Pattern Anal. Mach. Intell.}, 41(1):121--135, 2019.

\bibitem{ranjan2017all}
Rajeev Ranjan, Swami Sankaranarayanan, Carlos~D Castillo, and Rama Chellappa.
\newblock An all-in-one convolutional neural network for face analysis.
\newblock In {\em Int. Conf. on Automatic Face and Gesture Recognition}, pages
  17--24. IEEE, 2017.

\bibitem{ring1997child}
Mark~B Ring.
\newblock {CHILD}: A first step towards continual learning.
\newblock {\em Mach. Learn.}, 28(1):77--104, 1997.

\bibitem{rothe2015dex}
Rasmus Rothe, Radu Timofte, and Luc Van~Gool.
\newblock Dex: Deep expectation of apparent age from a single image.
\newblock In {\em Proc. Int. Conf. Comput. Vision Workshops}, pages 10--15,
  2015.

\bibitem{rozantsev2019beyond}
Artem Rozantsev, Mathieu Salzmann, and Pascal Fua.
\newblock Beyond sharing weights for deep domain adaptation.
\newblock {\em Trans. Pattern Anal. Mach. Intell.}, 41(4):801--814, 2019.

\bibitem{rudd2016moon}
Ethan~M Rudd, Manuel G{\"u}nther, and Terrance~E Boult.
\newblock Moon: A mixed objective optimization network for the recognition of
  facial attributes.
\newblock In {\em European Conf. Comput. Vision}, pages 19--35. Springer, 2016.

\bibitem{rusu2018meta}
Andrei~A Rusu, Dushyant Rao, Jakub Sygnowski, Oriol Vinyals, Razvan Pascanu,
  Simon Osindero, and Raia Hadsell.
\newblock Meta-learning with latent embedding optimization.
\newblock In {\em Int. Conf. on Learning Representations}, 2019.

\bibitem{scholkopf2001learning}
Bernhard Scholkopf and Alexander~J Smola.
\newblock {\em Learning with kernels: support vector machines, regularization,
  optimization, and beyond}.
\newblock MIT press, 2001.

\bibitem{shankar2016refining}
Sukrit Shankar, Duncan Robertson, Yani Ioannou, Antonio Criminisi, and Roberto
  Cipolla.
\newblock Refining architectures of deep convolutional neural networks.
\newblock In {\em Proc. Conf. Comput. Vision Pattern Recognition}, pages
  2212--2220, 2016.

\bibitem{torralba2011unbiased}
A Torralba and AA Efros.
\newblock Unbiased look at dataset bias.
\newblock In {\em Proc. Conf. Comput. Vision Pattern Recognition}, pages
  1521--1528. IEEE Computer Society, 2011.

\bibitem{tran2017extreme}
Anh~Tuan Tran, Tal Hassner, Iacopo Masi, Eran Paz, Yuval Nirkin, and G\'{e}rard
  Medioni.
\newblock Extreme {3D} face reconstruction: Looking past occlusions.
\newblock In {\em Proc. Conf. Comput. Vision Pattern Recognition}, 2018.

\bibitem{veit2017conditional}
Andreas Veit, Serge Belongie, and Theofanis Karaletsos.
\newblock Conditional similarity networks.
\newblock In {\em Proc. Conf. Comput. Vision Pattern Recognition}, pages
  830--838, 2017.

\bibitem{wang2018cosface}
Hao Wang, Yitong Wang, Zheng Zhou, Xing Ji, Dihong Gong, Jingchao Zhou, Zhifeng
  Li, and Wei Liu.
\newblock {Cosface}: Large margin cosine loss for deep face recognition.
\newblock In {\em Proc. Conf. Comput. Vision Pattern Recognition}, pages
  5265--5274, 2018.

\bibitem{wang2016walk}
Jing Wang, Yu Cheng, and Rogerio Schmidt~Feris.
\newblock Walk and learn: Facial attribute representation learning from
  egocentric video and contextual data.
\newblock In {\em Proc. Conf. Comput. Vision Pattern Recognition}, pages
  2295--2304, 2016.

\bibitem{wang2015towards}
Peng Wang, Xiaohui Shen, Zhe Lin, Scott Cohen, Brian Price, and Alan~L Yuille.
\newblock Towards unified depth and semantic prediction from a single image.
\newblock In {\em Proc. Conf. Comput. Vision Pattern Recognition}, pages
  2800--2809, 2015.

\bibitem{weiss2016survey}
Karl Weiss, Taghi~M Khoshgoftaar, and DingDing Wang.
\newblock A survey of transfer learning.
\newblock {\em Journal of Big Data}, 3(1):9, 2016.

\bibitem{WelinderEtal2010}
P. Welinder, S. Branson, T. Mita, C. Wah, F. Schroff, S. Belongie, and P.
  Perona.
\newblock {Caltech-UCSD Birds 200}.
\newblock Technical Report CNS-TR-2010-001, California Institute of Technology,
  2010.

\bibitem{xian2018zero}
Yongqin Xian, Christoph~H Lampert, Bernt Schiele, and Zeynep Akata.
\newblock Zero-shot learning-a comprehensive evaluation of the good, the bad
  and the ugly.
\newblock {\em Trans. Pattern Anal. Mach. Intell.}, 2018.

\bibitem{yang2016deep}
Yongxin Yang and Timothy Hospedales.
\newblock Deep multi-task representation learning: A tensor factorisation
  approach.
\newblock In {\em Int. Conf. on Learning Representations}, 2017.

\bibitem{ying2018transfer}
Wei Ying, Yu Zhang, Junzhou Huang, and Qiang Yang.
\newblock Transfer learning via learning to transfer.
\newblock In {\em Int. Conf. Mach. Learning}, pages 5072--5081, 2018.

\bibitem{yosinski2014transferable}
Jason Yosinski, Jeff Clune, Yoshua Bengio, and Hod Lipson.
\newblock How transferable are features in deep neural networks?
\newblock In {\em Neural Inform. Process. Syst.}, pages 3320--3328, 2014.

\bibitem{zamir2018taskonomy}
Amir~R Zamir, Alexander Sax, William Shen, Leonidas~J Guibas, Jitendra Malik,
  and Silvio Savarese.
\newblock Taskonomy: Disentangling task transfer learning.
\newblock In {\em Proc. Conf. Comput. Vision Pattern Recognition}, pages
  3712--3722, 2018.

\bibitem{zhang2016understanding}
Chiyuan Zhang, Samy Bengio, Moritz Hardt, Benjamin Recht, and Oriol Vinyals.
\newblock Understanding deep learning requires rethinking generalization.
\newblock In {\em Int. Conf. on Learning Representations}, 2017.

\bibitem{zhao2018modulation}
Xiangyun Zhao, Haoxiang Li, Xiaohui Shen, Xiaodan Liang, and Ying Wu.
\newblock A modulation module for multi-task learning with applications in
  image retrieval.
\newblock In {\em European Conf. Comput. Vision}, pages 401--416, 2018.

\end{thebibliography}
}

\newpage
\appendix

\section{Proof of theorem 1}
From the definition of $\widetilde{\mathrm{Trf}}(T^Z \rightarrow T^Y)$, we have:
\begin{align}
&\widetilde{\mathrm{Trf}}(T^Z \rightarrow T^Y) \notag \\
&= l_Y(w_Z, k_Y) \tag{definition of $\widetilde{\mathrm{Trf}}$}\\
&\ge l_Y(w_Z, \bar{k}) \tag{definition of $k_Y$ and $\bar{k} \in K$}\\
&= \frac{1}{n} \sum_{i=1}^n \log \left( \sum_{z \in \mathcal{Z}} \hat{P}(y_i | z) P(z | x_i; w_Z, h_Z) \right) \tag{construction of $\bar{k}$} \\
&\ge \frac{1}{n} \sum_{i=1}^n \log \left( \hat{P}(y_i | z_i) P(z_i | x_i; w_Z, h_Z) \right) \tag{replacing the sum by one of its elements}\\
&= \frac{1}{n} \sum_{i=1}^n \log \hat{P}(y_i | z_i) + \frac{1}{n} \sum_{i=1}^n \log P(z_i | x_i; w_Z, h_Z).\label{eq:pproof3}
\end{align}

Note that the second term in Eq.~\eqref{eq:pproof3} is:
\begin{align}
\frac{1}{n} \sum_{i=1}^n \log P(z_i | x_i; w_Z, h_Z) = l_Z(w_Z, h_Z). \label{eq:term2}
\end{align}

Furthermore, the first term in Eq.~\eqref{eq:pproof3} is:
\begin{align}
&\frac{1}{n} \sum_{i=1}^n \log \hat{P}(y_i | z_i) \notag \\
&= \frac{1}{n} \sum_{y \in \mathcal{Y}} \sum_{z \in \mathcal{Z}} \left( \sum_{i~:~y_i=y \text{ and } z_i=z} \log \hat{P}(y | z) \right) \tag{group the summands by values of $y_i$ and $z_i$}\\
&= \frac{1}{n} \sum_{y \in \mathcal{Y}} \sum_{z \in \mathcal{Z}} \left( \left| \left\{ i : y_i=y \text{ and } z_i=z \right\} \right| ~ \log \hat{P}(y | z) \right) \tag{by counting}\\
&= \sum_{y \in \mathcal{Y}} \sum_{z \in \mathcal{Z}} \left( \frac{\left| \left\{ i : y_i=y \text{ and } z_i=z \right\} \right|}{n} ~ \log \hat{P}(y | z) \right) \nonumber\\
&= \sum_{y \in \mathcal{Y}} \sum_{z \in \mathcal{Z}} \left( \hat{P}(y, z) ~ \log \frac{\hat{P}(y, z)}{\hat{P}(z)} \right) \tag{definitions of $\hat{P}(y, z)$ and $\hat{P}(y | z)$}\\
&= -H(Y | Z). \label{eq:term1}
\end{align}

From Eq.~\eqref{eq:pproof3},~\eqref{eq:term2}, and~\eqref{eq:term1}, we have ${ \widetilde{\mathrm{Trf}}(T^Z \rightarrow T^Y)} \ge l_Z(w_Z, h_Z) - H(Y | Z)$. Hence, the theorem holds.

\section{More details on task hardness}

\minisection{On the definition of task hardness}
In our paper, we assume non-overfitting of trained models. When train and test sets are sampled from the \emph{same distribution}, this assumption typically holds for appropriately trained models~\cite{zhang2016understanding}. This property also shows that our definition of hardness, Eq.~\eqref{eq:hardness}, does not conflict with the results of Zhang et al.~\cite{zhang2016understanding}: In such cases, the training loss of Eq.~\eqref{eq:hardness} correlates with the test error, and thus this definition indeed reflects task hardness, explaining the relationships between train and test errors observed in our hardness results.

\minisection{On the representation for trivial tasks}
\emph{Any} representation for a trivial source task can fit the constant label perfectly (zero training loss). In theory, if we choose the optimal $w_Z$ in Eq.~\eqref{eq:source-loss} as our representation, we can show Eq.~\eqref{eq:est-hardness}. In practice, of course we cannot infer the optimal $w_Z$ from the trivial source task, but Eq.~\eqref{eq:est-hardness} shows that we can still connect it to~$H(Z|C)$.

\section{Technical implementation details}

\minisection{Computing the CE}
Computing the CE is straightforward and involves the following steps:
\begin{enumerate}
\item Loop through the training labels of both tasks $T^Z$ and $T^Y$ and compute the empirical joint distribution $\hat{P}(y, z)$ by counting (Eq.~\eqref{eq:empjointdist} in the paper).
\item Loop through the training labels again and compute the CE using Eq.~\eqref{eq:term1} above. That is, 
$$H(Y | Z) = -\frac{1}{n} \sum_{i=1}^n \log \hat{P}(y_i | z_i).$$
\end{enumerate}
Thus, computing the CE only requires running two loops through the training labels. This process is computationally efficient. In the most extreme case, computing the transferability of face recognition ($|\mathcal{Z}|>10k$) to a facial attribute, with $|\mathcal{Y}|=2$, required {\em less than a second} on a standard CPU. 

This run time should be compared with the {\em hours (or days)} required to train deep models in order to empirically measure transferability following the process described by previous work. In particular, Taskonomy~\cite{zamir2018taskonomy} reported {\em over 47 thousand hours of GPU runtime} in order to establish relationships between their 26 tasks.

\minisection{Dedicated attribute training}
Given a source task $T^Z$, we train a dedicated CNN for this task with standard ResNet-18 V2 implemented in the MXNet deep learning library~\cite{chen2015mxnet}.\footnote{Model available from:~\url{https://mxnet.apache.org/api/python/gluon/model_zoo.html}.} We set the initial learning rate to 0.01. Learning rate was then divided by 10 after each 12 epochs. Training converged in less than 40 epochs in all 437 tasks.

\minisection{Task transfer with linear SVM}
After training a deep representation for a source task $T^Z$, we transfer it to a target task $T^Y$ using linear support vector machines (lSVM). 

First, we use the trained CNN, denoted in the paper as $w_Z$, to extract deep embeddings for the entire training data (one embedding per input image. Each embedding is a vector $r\in \mathbb{R}^{2048}$, which we obtain from the penultimate layer of the network. We then use these embeddings, along with the corresponding labels for target task, $T^Y$, to train a standard lSVM classifier, implemented by SK-Learn~\cite{scikit-learn}. The lSVM parameters were kept unchanged from their default values. 

Given unseen testing data, we first extract their embeddings with $w_Z$. We then apply the trained lSVM classifier on these features to predict labels for target task, $T^Y$.

\section{Additional results: Generalization to multi-class} 
Transferability generalizes well to multi-class, as evident in our face recognition (10k labels)-to-attribute tests in Sec.~\ref{sec:id2attrib}. Table~\ref{tab:id2mulattr} below reports hardness tests with multi-class, CelebA, attribute aggregates. Generally speaking, the harder the task, the lower the accuracy obtained.

\begin{table}[h!]
    \centering
    \resizebox{\linewidth}{!}{
    \begin{tabular}{l|c|c|c}
    \toprule
Multi-class& Straight/Wavy/Other & Black/Blonde/Other & Arched/Bushy/Other  \\
\rowcolor{lightblue} Hardness $\downarrow$& 1.040 & 0.925 & 0.867\\
Dedicated Res18 & 0.713 & 0.859 & 0.797\\ \hline
Multi-class&Bangs/Receding/Other & Gray/Blonde/Other & Goatee/Beard/None \\
\rowcolor{lightblue} Hardness $\downarrow$ & 0.690 & 0.575 & 0.557 \\
Dedicated Res18 & 0.900 & 0.943 & 0.937 \\
    \bottomrule
    \end{tabular}
    }
    \vspace{1mm}
    \caption{Multi-class hardness examples on CelebA data.}
    \label{tab:id2mulattr}
\end{table}

\section{Full transferability results}
\begin{figure*}[ht]
\centering
\footnotesize
\begin{tabular}{c@{~}c@{~}c@{~}c@{~}c}
    \includegraphics[clip, trim=0mm 0mm 0mm 13mm, width=0.19\textwidth]{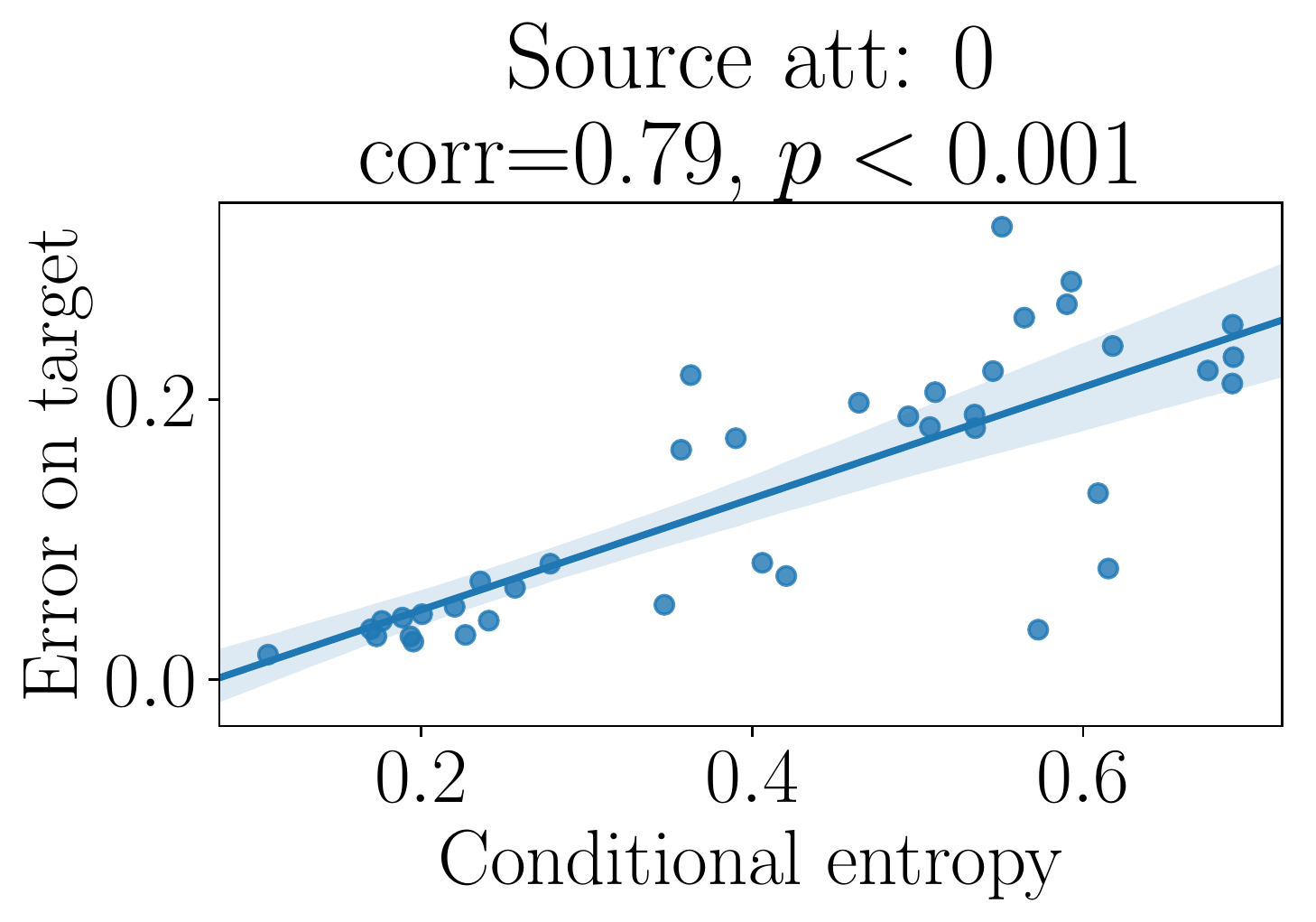}&
    \includegraphics[clip, trim=0mm 0mm 0mm 13mm, width=0.19\textwidth]{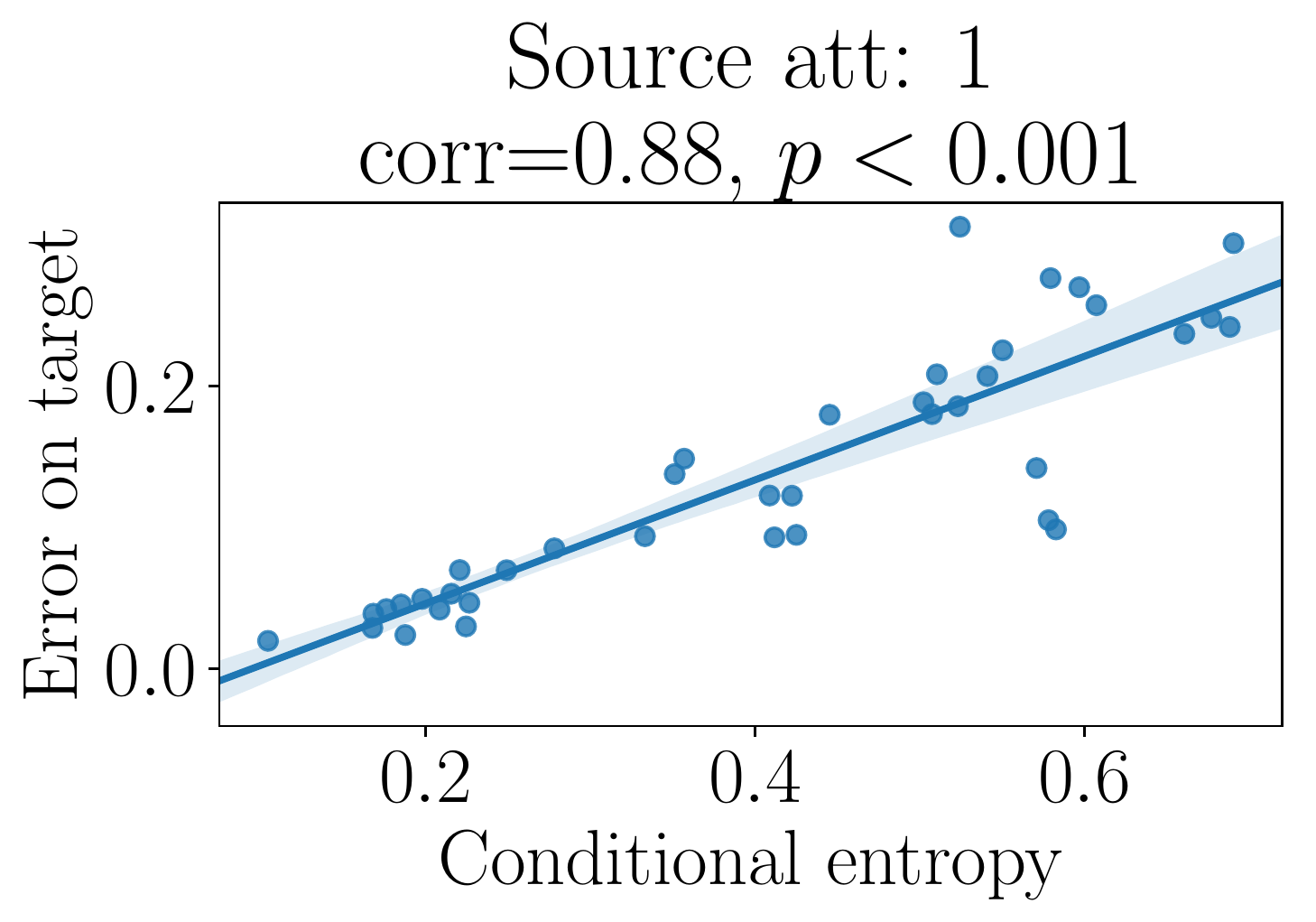}&
    \includegraphics[clip, trim=0mm 0mm 0mm 13mm, width=0.19\textwidth]{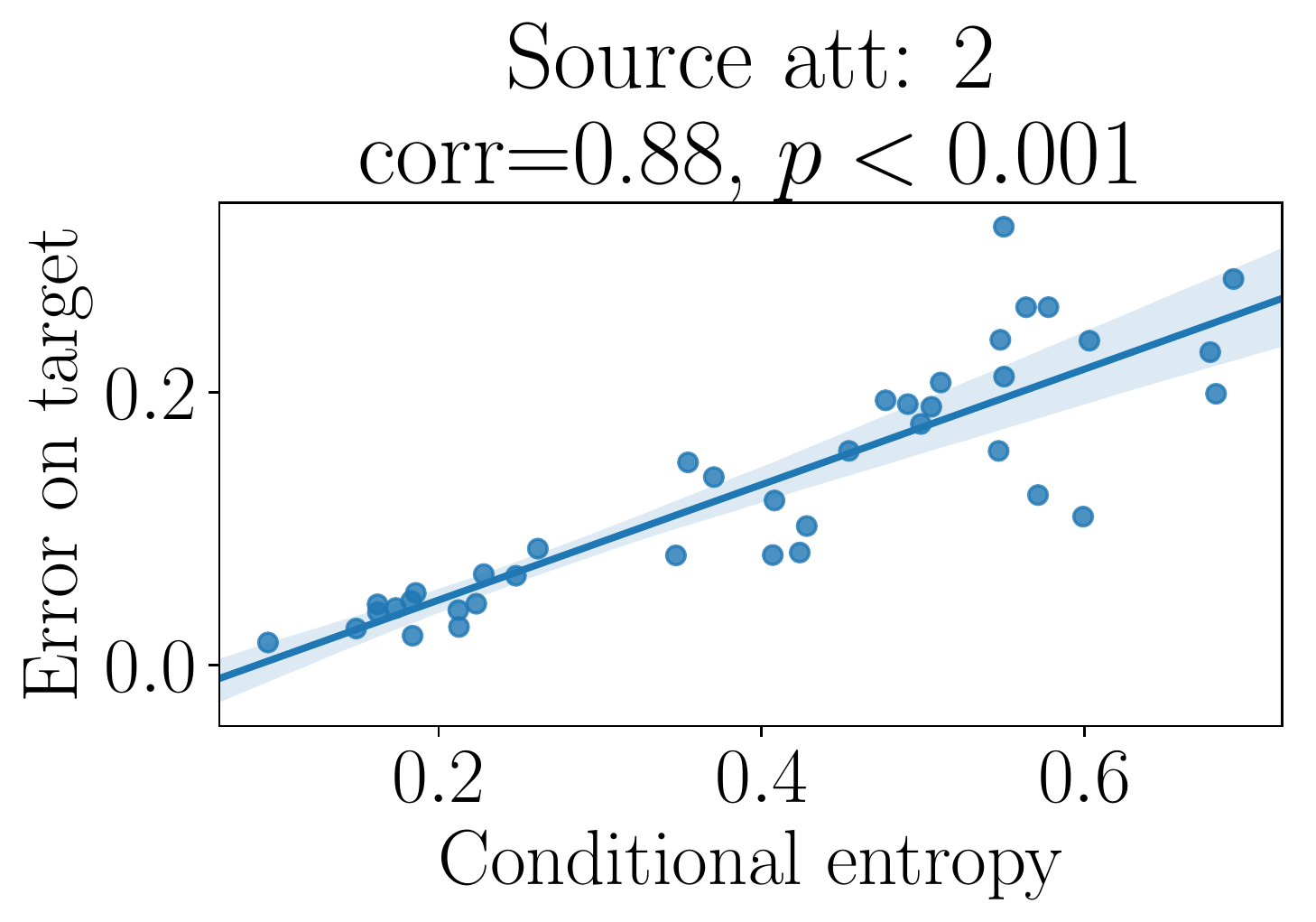}&
    \includegraphics[clip, trim=0mm 0mm 0mm 13mm, width=0.19\textwidth]{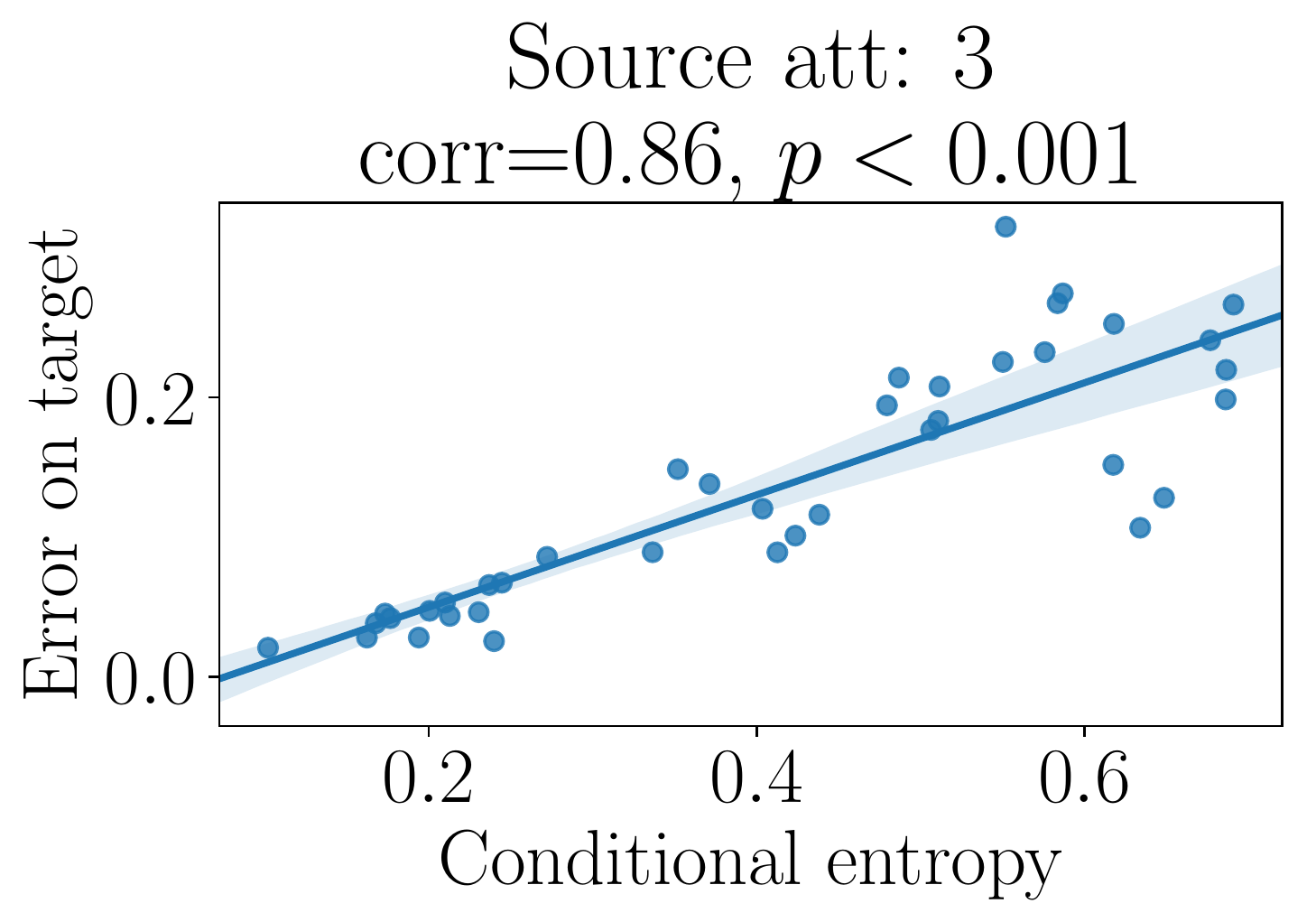}&
    \includegraphics[clip, trim=0mm 0mm 0mm 13mm, width=0.19\textwidth]{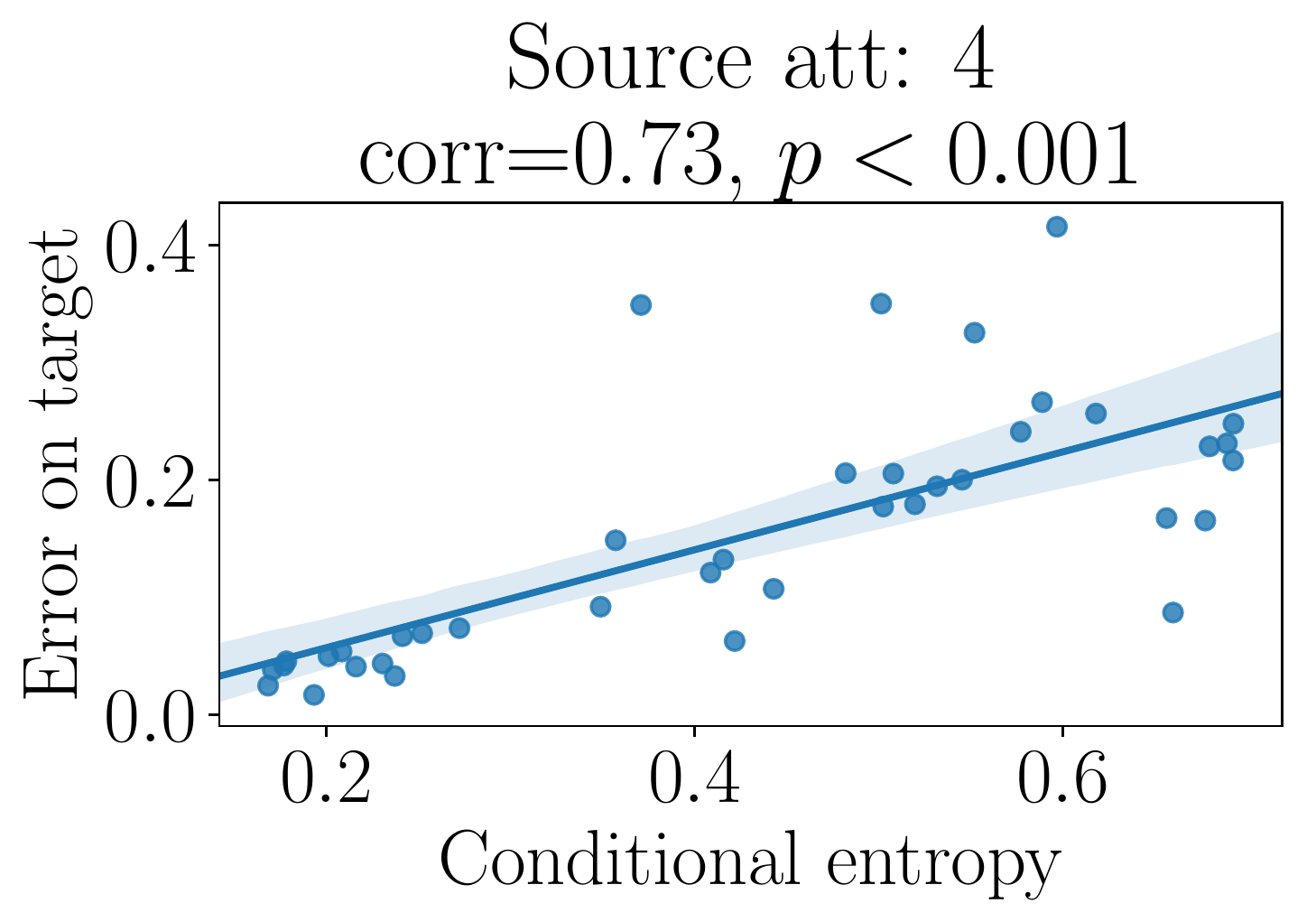}\\[-2pt]
    (0) 5 o'clock shadow & (1) Arched eyebrows & (2) Attractive & (3) Bags under eyes & (4) Bald\\[6pt]
    \includegraphics[clip, trim=0mm 0mm 0mm 11mm, width=0.19\textwidth]{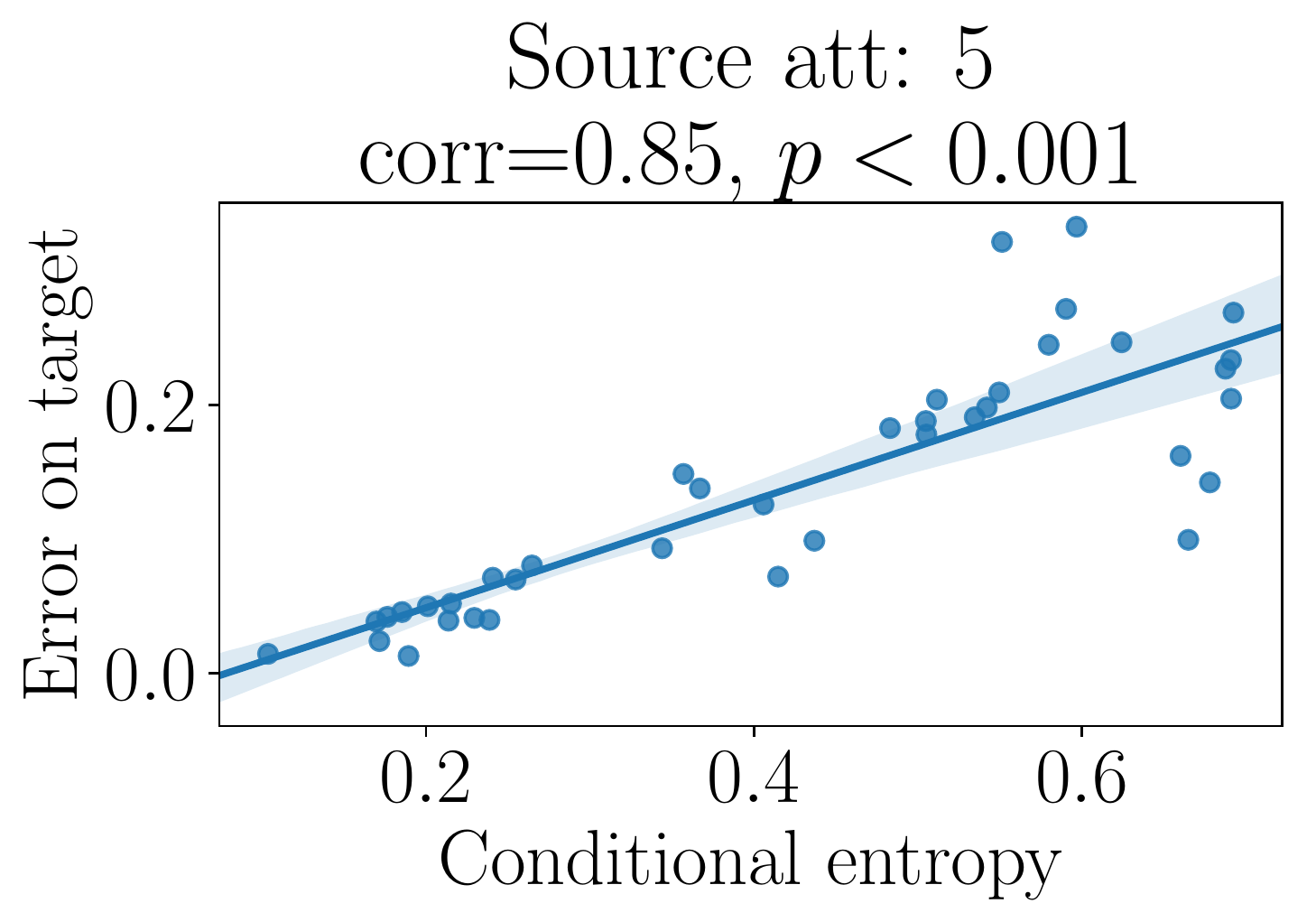}&
    \includegraphics[clip, trim=0mm 0mm 0mm 11mm, width=0.19\textwidth]{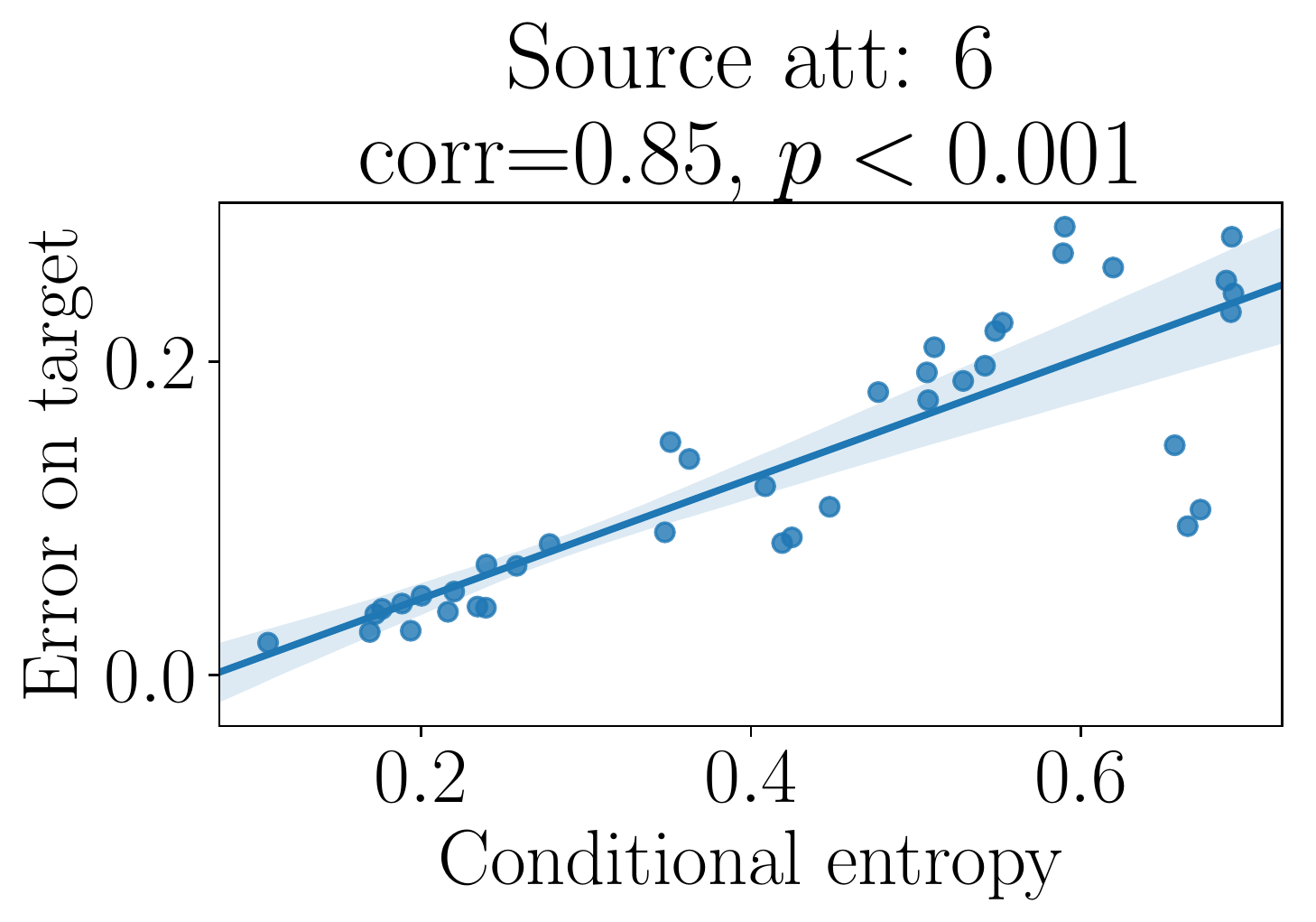}&
    \includegraphics[clip, trim=0mm 0mm 0mm 11mm, width=0.19\textwidth]{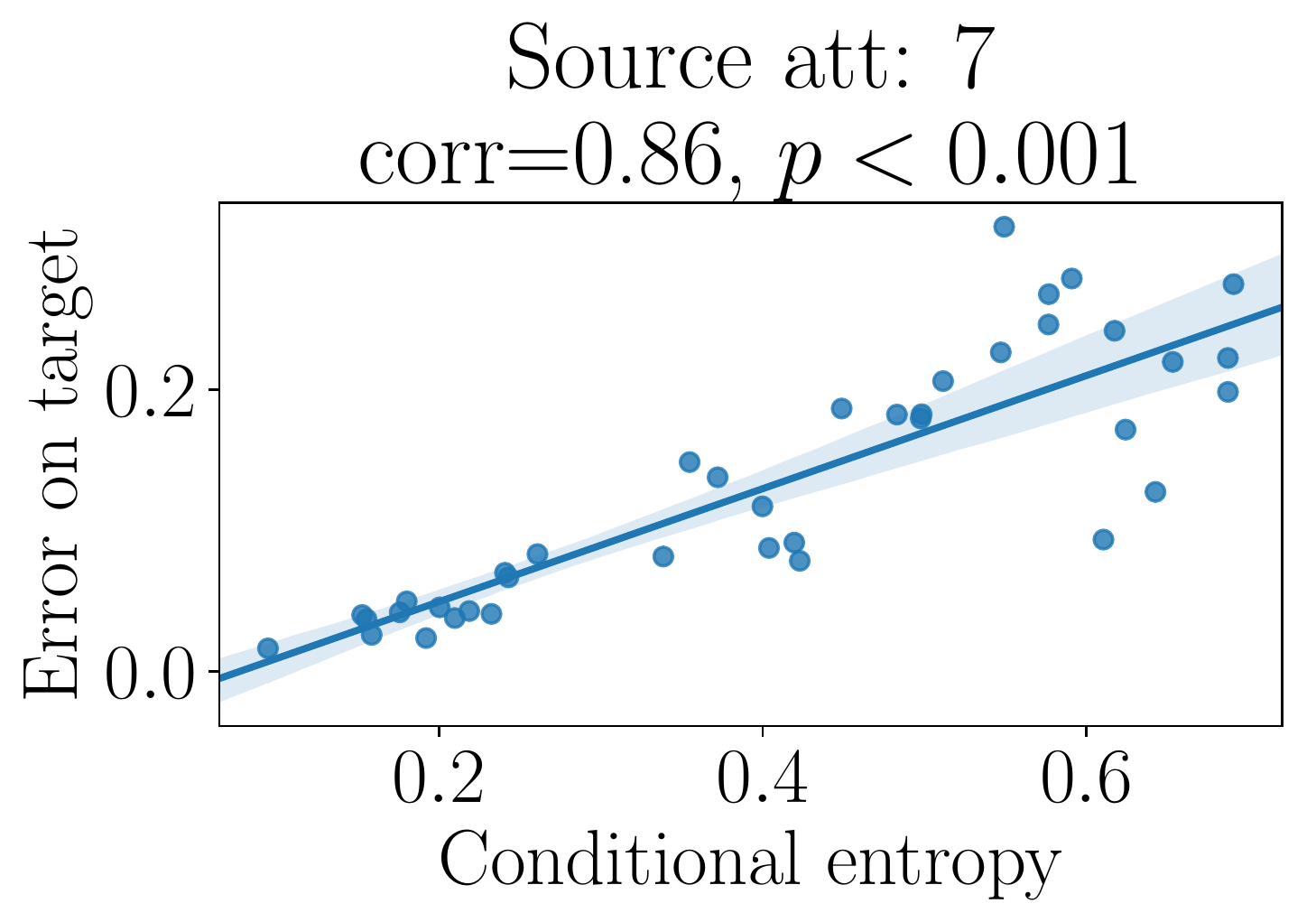}&
    \includegraphics[clip, trim=0mm 0mm 0mm 11mm, width=0.19\textwidth]{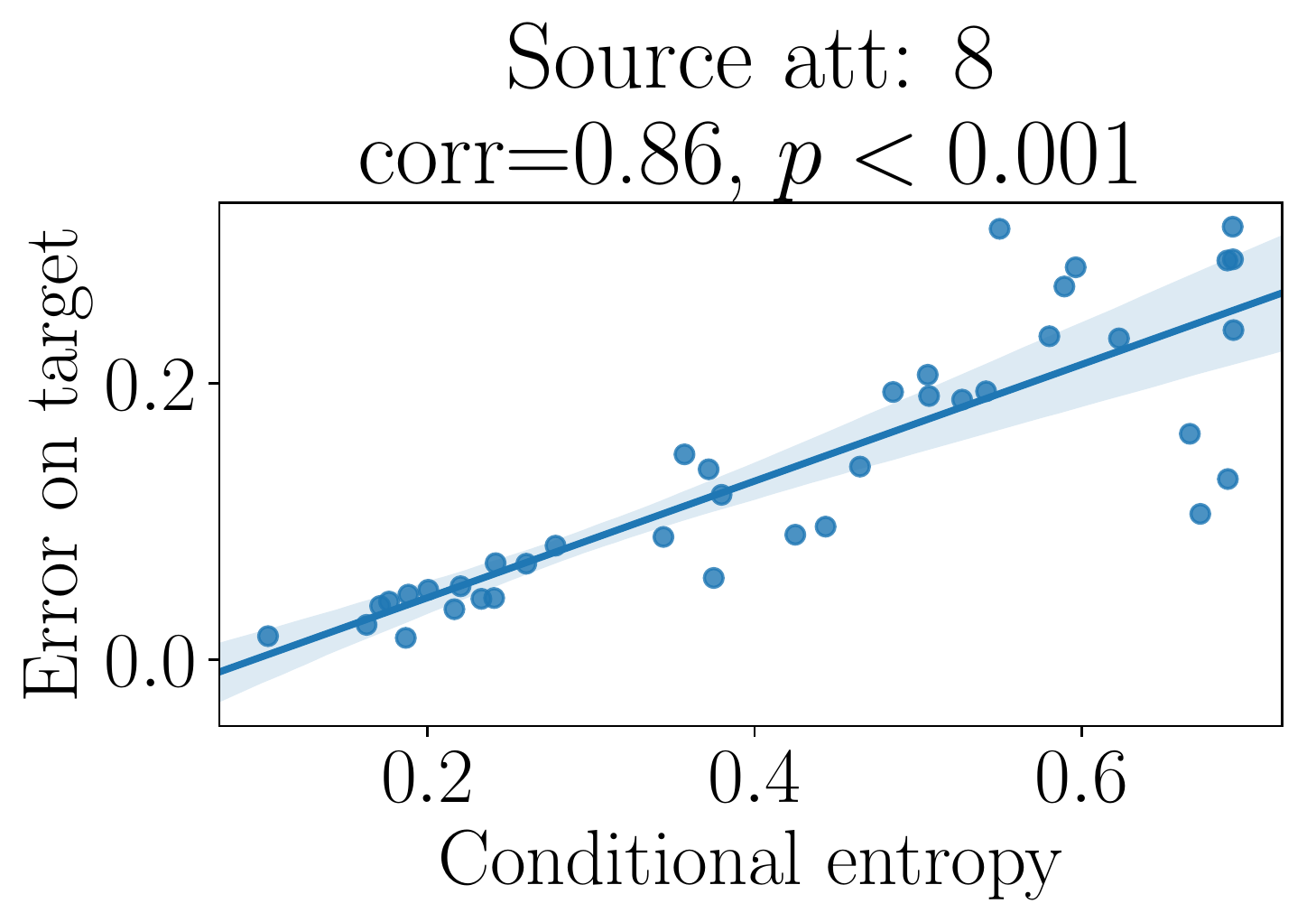}&
    \includegraphics[clip, trim=0mm 0mm 0mm 11mm, width=0.19\textwidth]{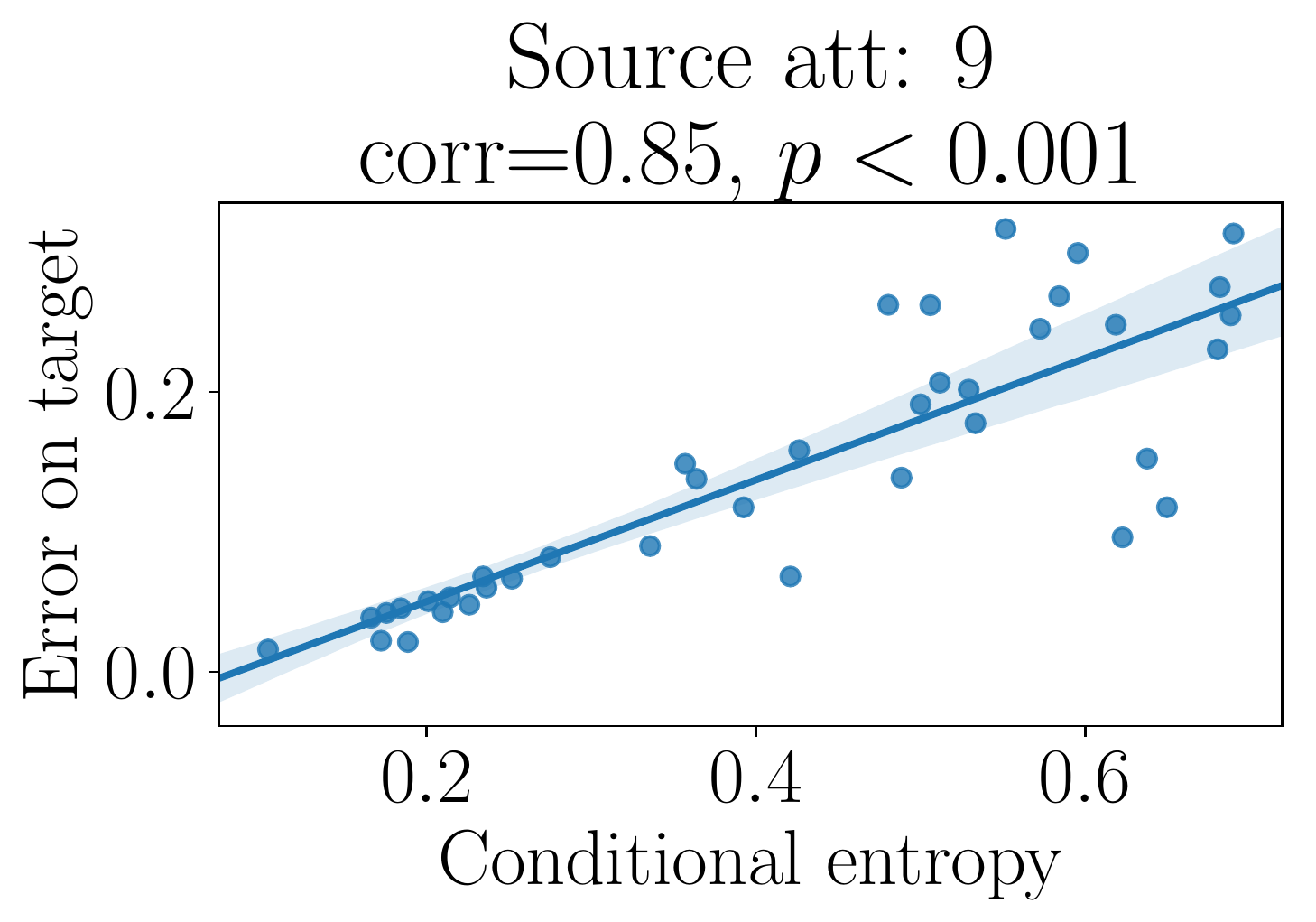}\\[-2pt]
    (5) Bangs & (6) Big lips & (7) Big nose & (8) Black hair  & (9) Blond hair\\[6pt]
    \includegraphics[clip, trim=0mm 0mm 0mm 11mm, width=0.19\textwidth]{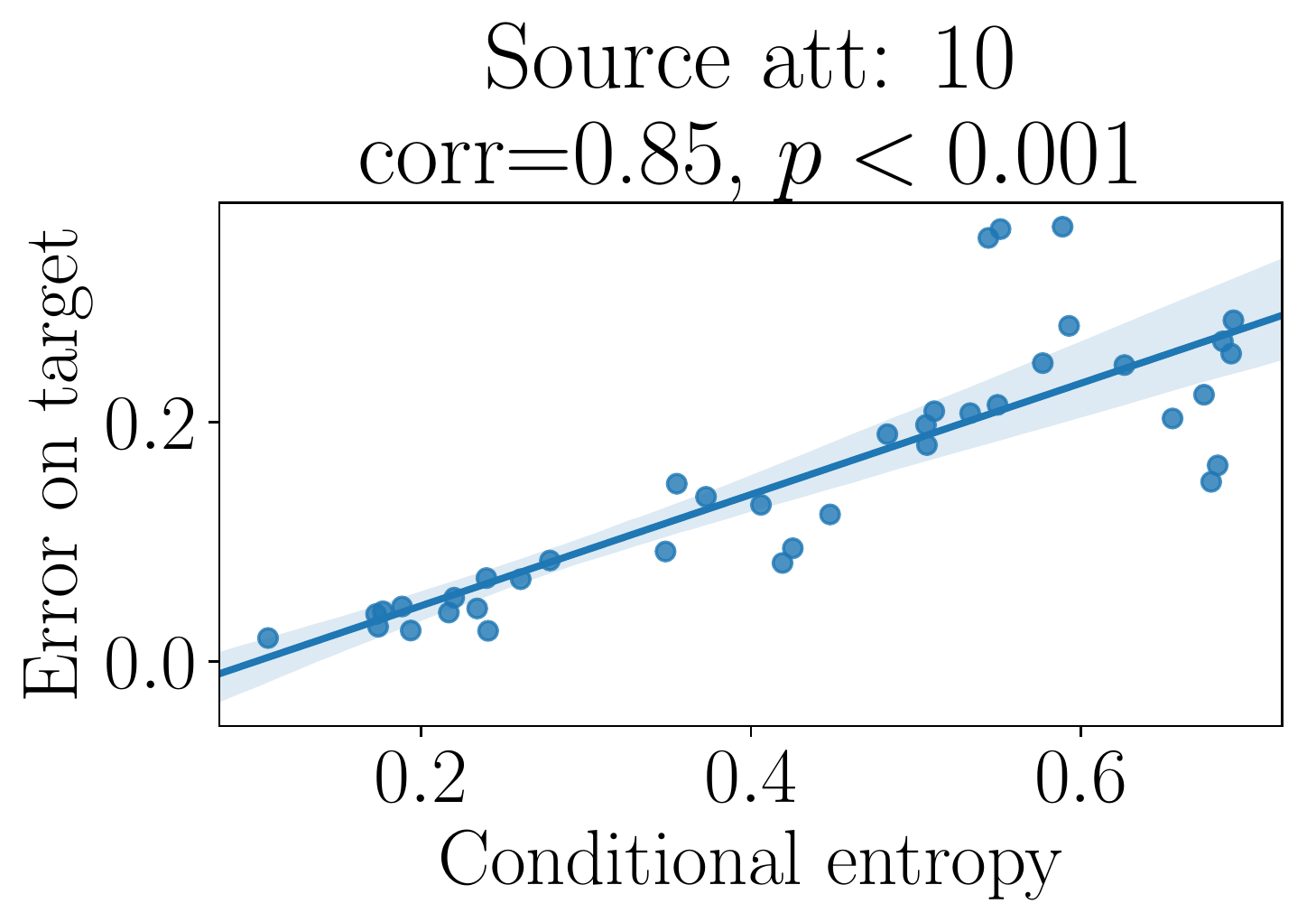}&
    \includegraphics[clip, trim=0mm 0mm 0mm 11mm, width=0.19\textwidth]{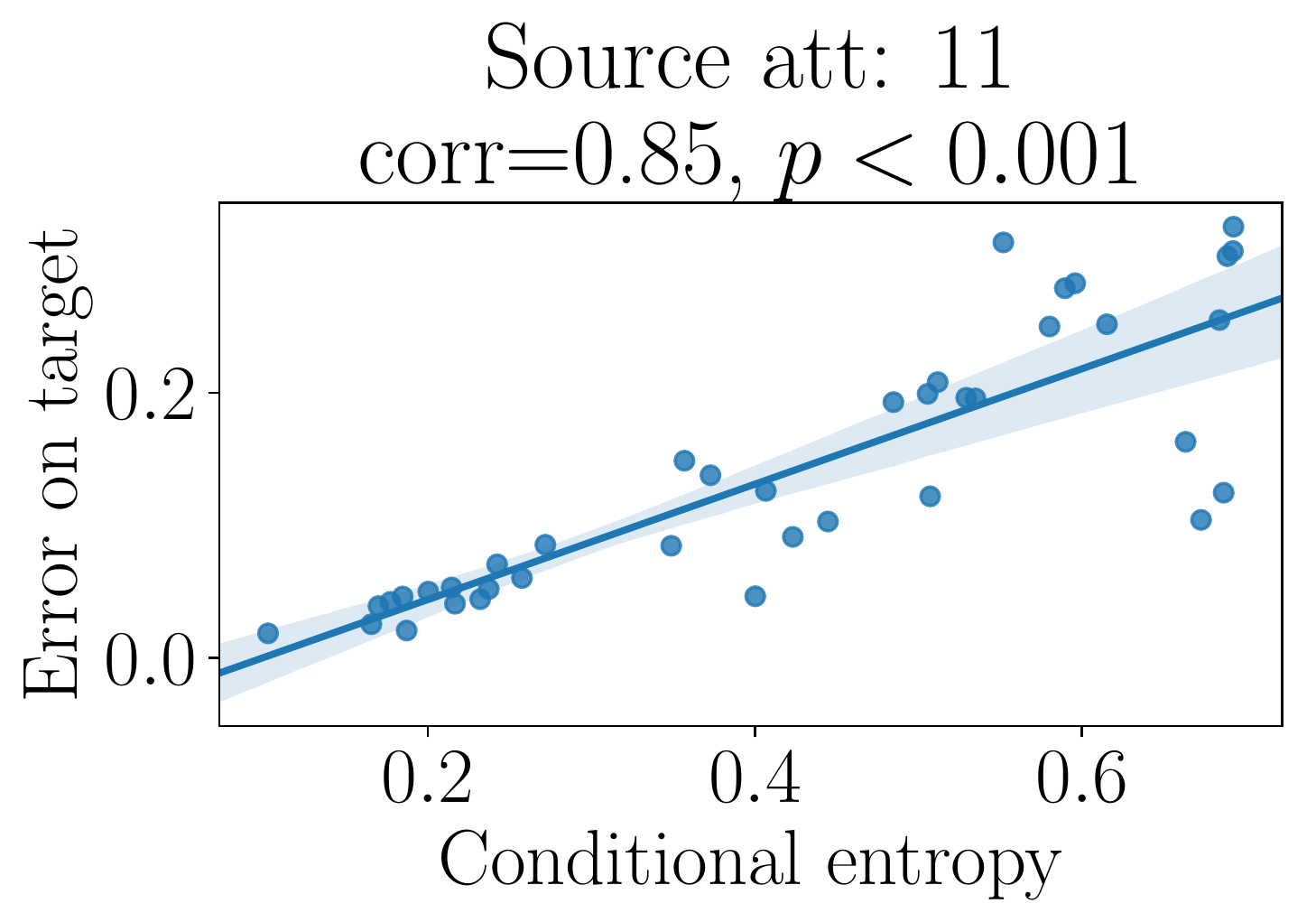}&
    \includegraphics[clip, trim=0mm 0mm 0mm 11mm, width=0.19\textwidth]{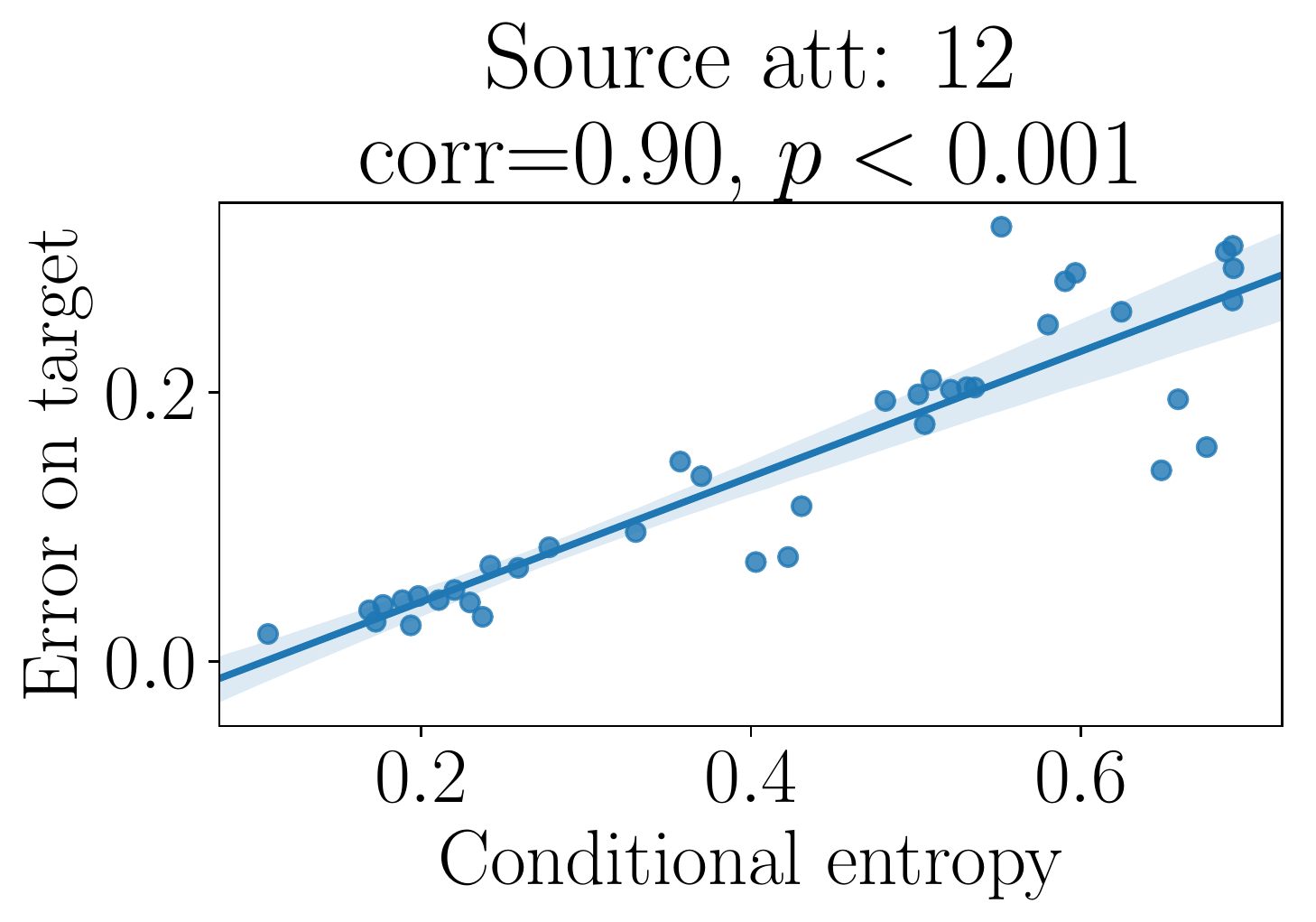}&
    \includegraphics[clip, trim=0mm 0mm 0mm 11mm, width=0.19\textwidth]{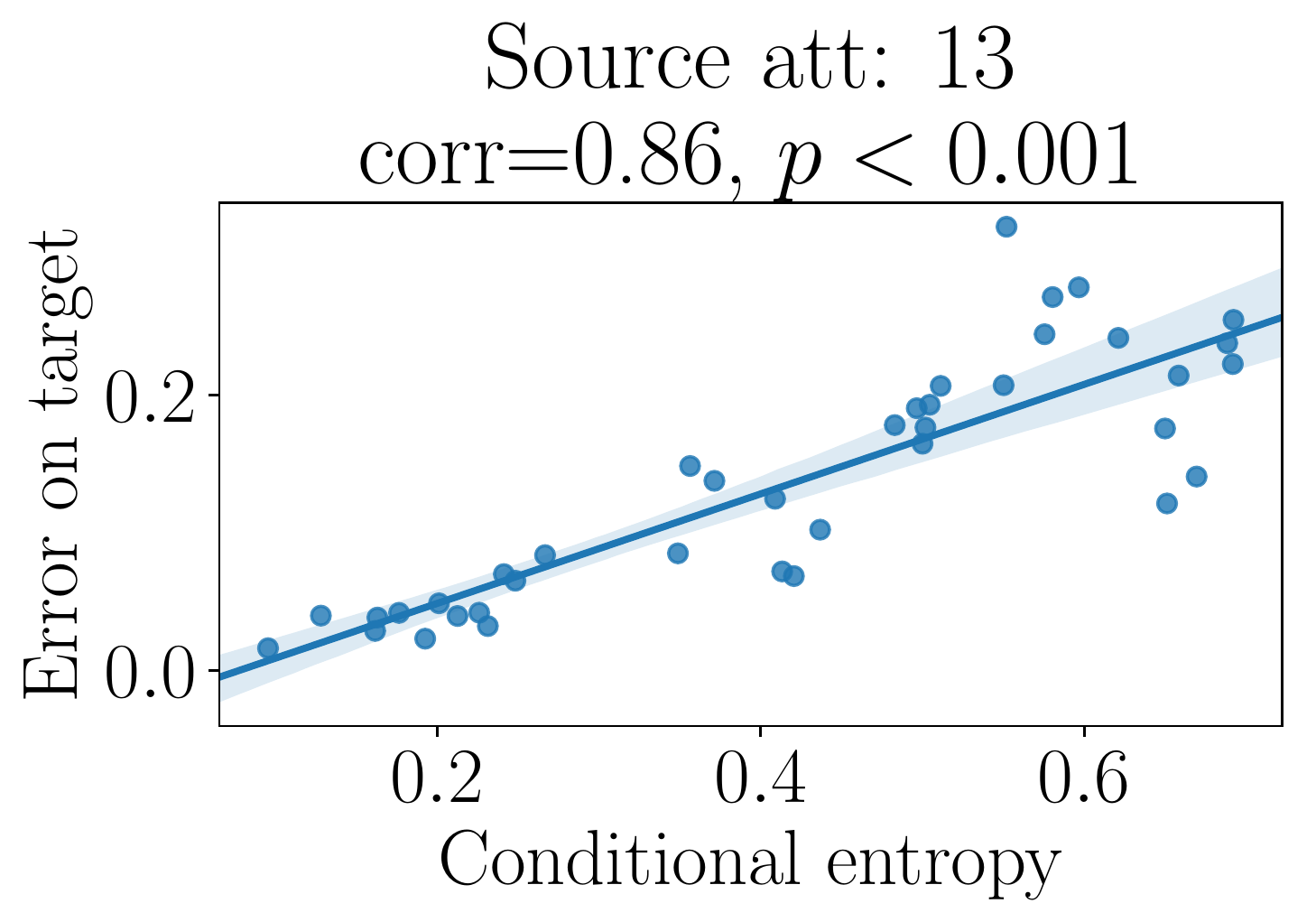}&
    \includegraphics[clip, trim=0mm 0mm 0mm 11mm, width=0.19\textwidth]{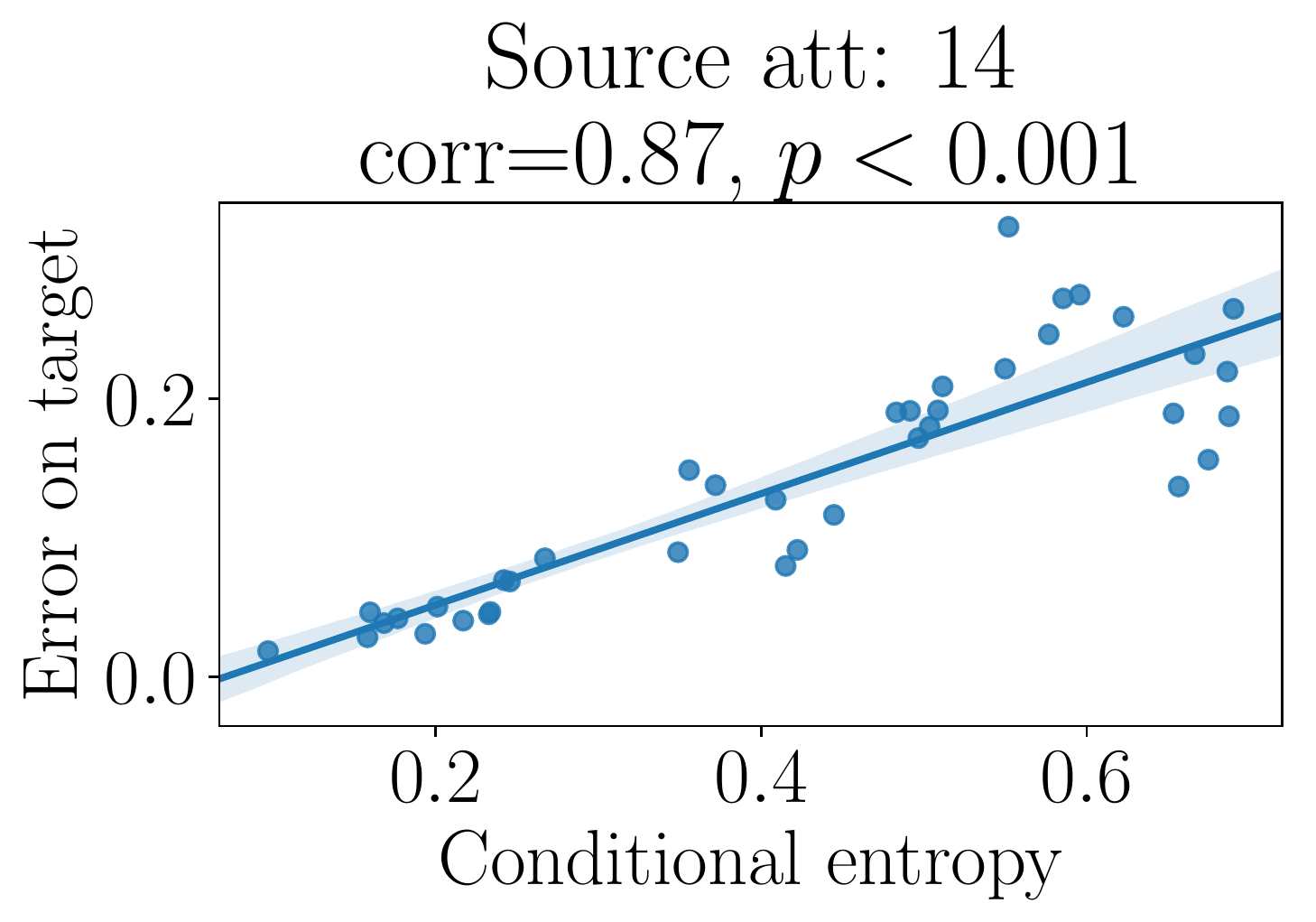}\\[-2pt]
    (10) Blurry & (11) Brown hair & (12) Bushy eyebrows & (13) Chubby  & (14) Double chin\\[6pt]
    \includegraphics[clip, trim=0mm 0mm 0mm 11mm, width=0.19\textwidth]{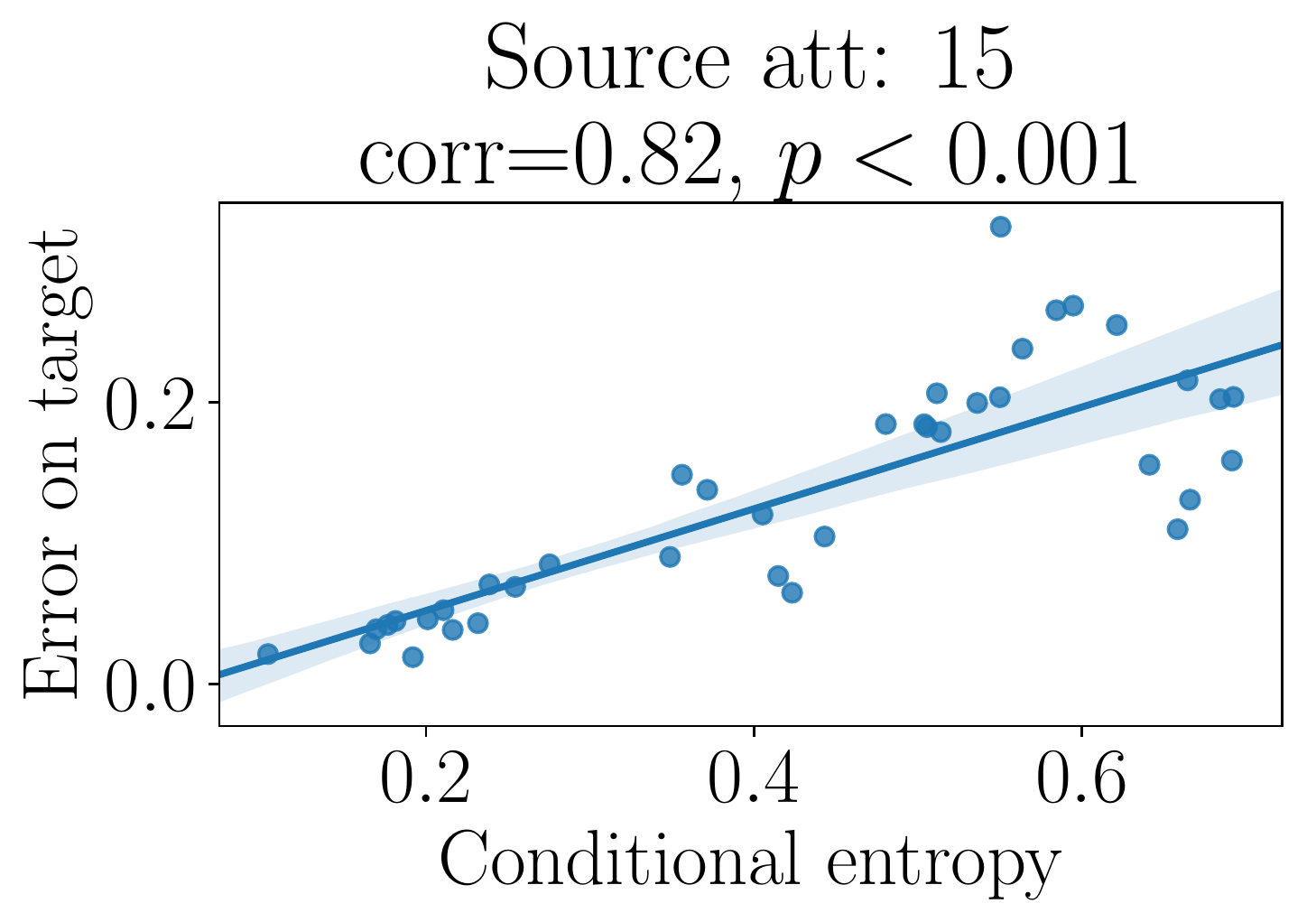}&
    \includegraphics[clip, trim=0mm 0mm 0mm 11mm, width=0.19\textwidth]{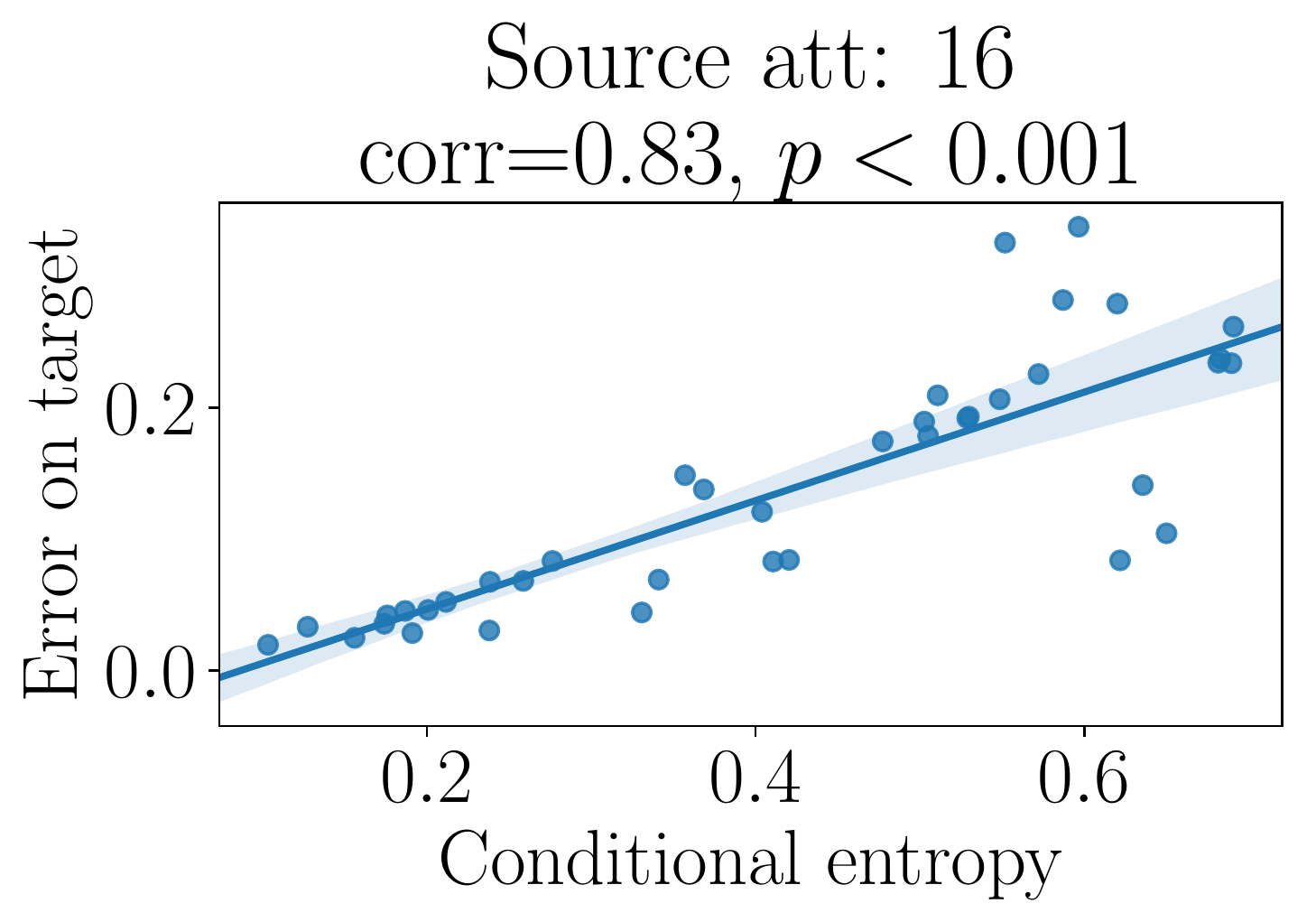}&
    \includegraphics[clip, trim=0mm 0mm 0mm 11mm, width=0.19\textwidth]{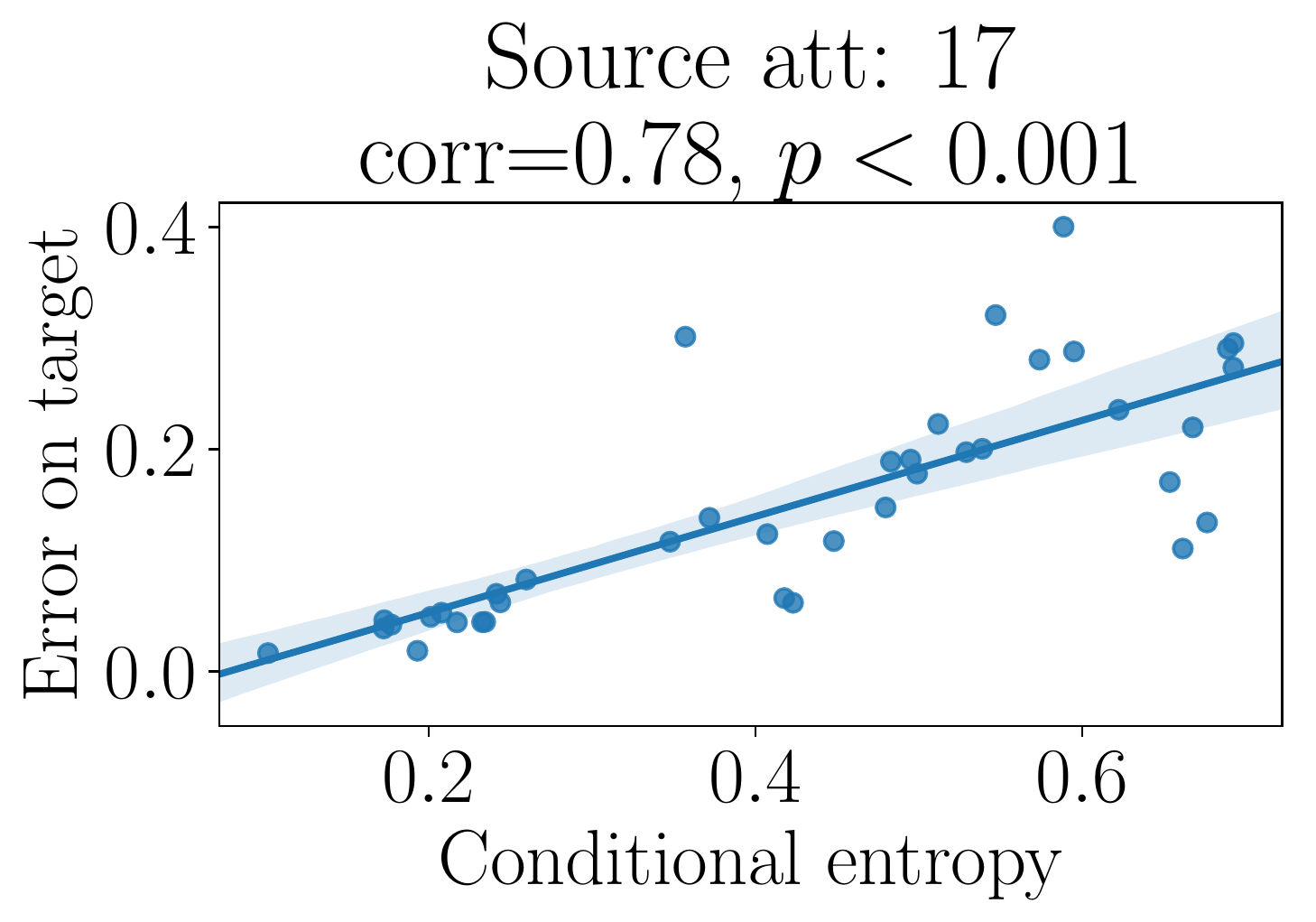}&
    \includegraphics[clip, trim=0mm 0mm 0mm 11mm, width=0.19\textwidth]{figures/tran_from_18.pdf}&
    \includegraphics[clip, trim=0mm 0mm 0mm 11mm, width=0.19\textwidth]{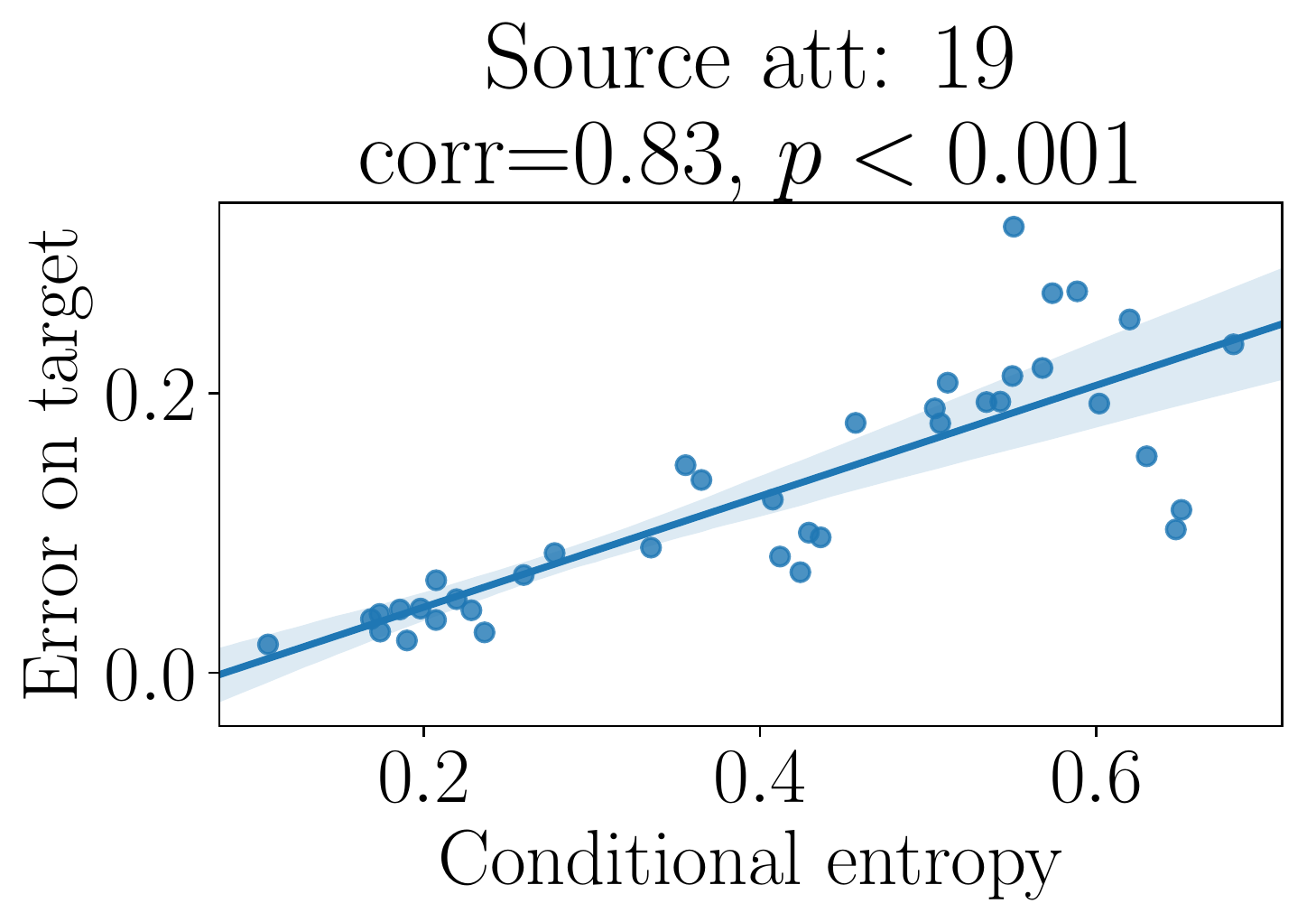}\\[-2pt]
    (15) Eyeglasses & (16) Goatee & (17) Gray hair & (18) Heavy makeup  & (19) High cheekbones\\[6pt]
    \includegraphics[clip, trim=0mm 0mm 0mm 11mm, width=0.19\textwidth]{figures/tran_from_20.pdf}&
    \includegraphics[clip, trim=0mm 0mm 0mm 11mm, width=0.19\textwidth]{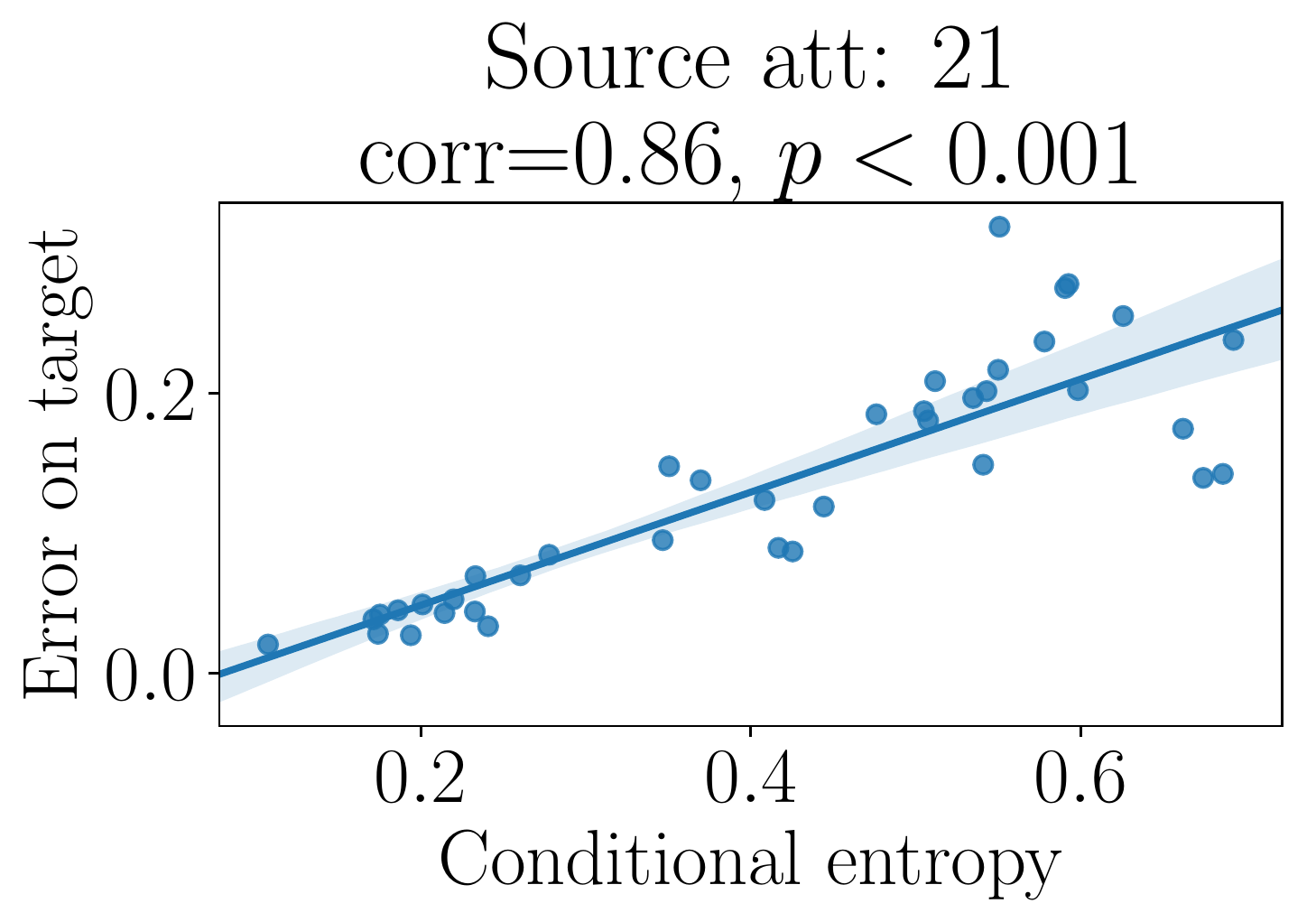}&
    \includegraphics[clip, trim=0mm 0mm 0mm 11mm, width=0.19\textwidth]{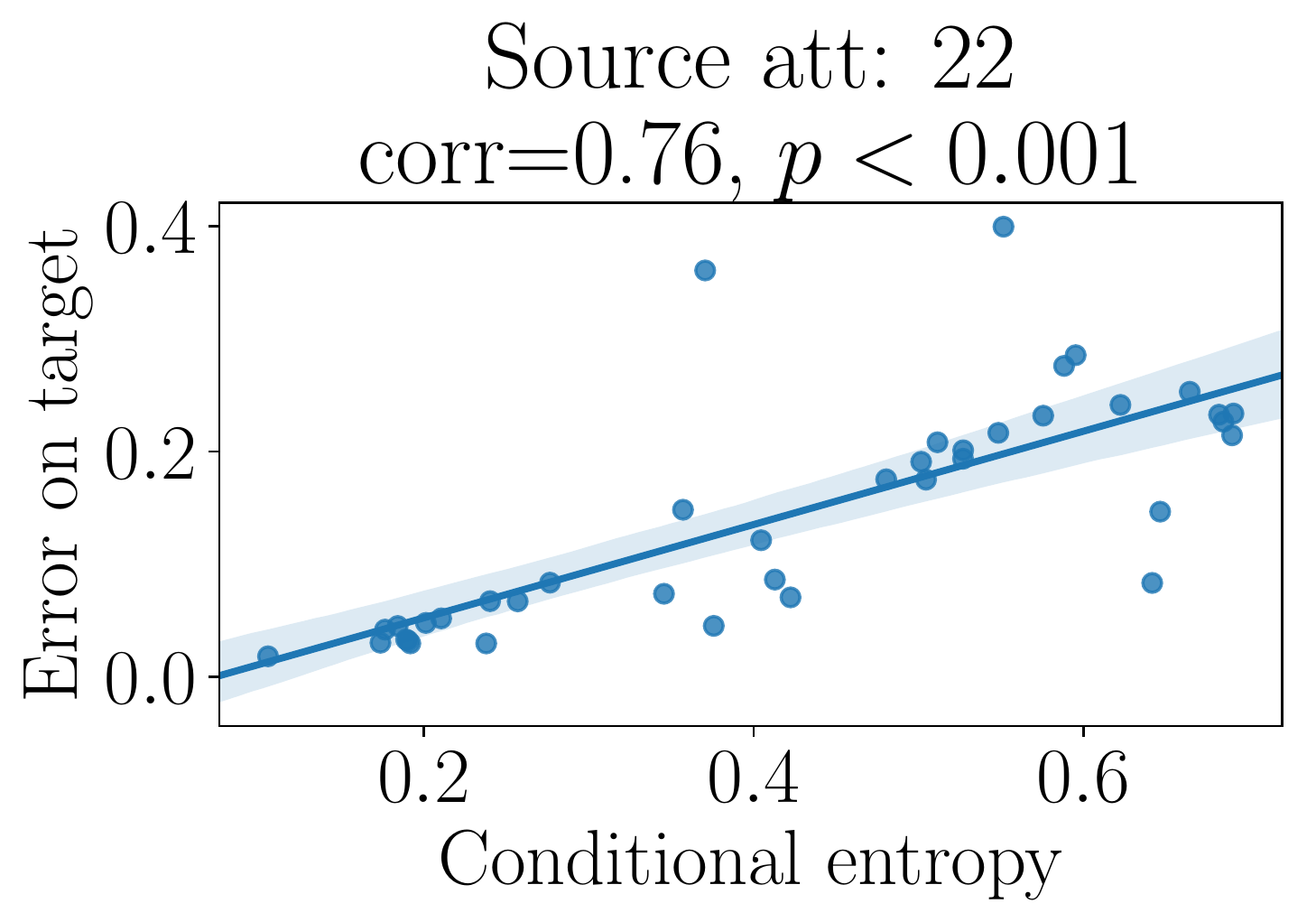}&
    \includegraphics[clip, trim=0mm 0mm 0mm 11mm, width=0.19\textwidth]{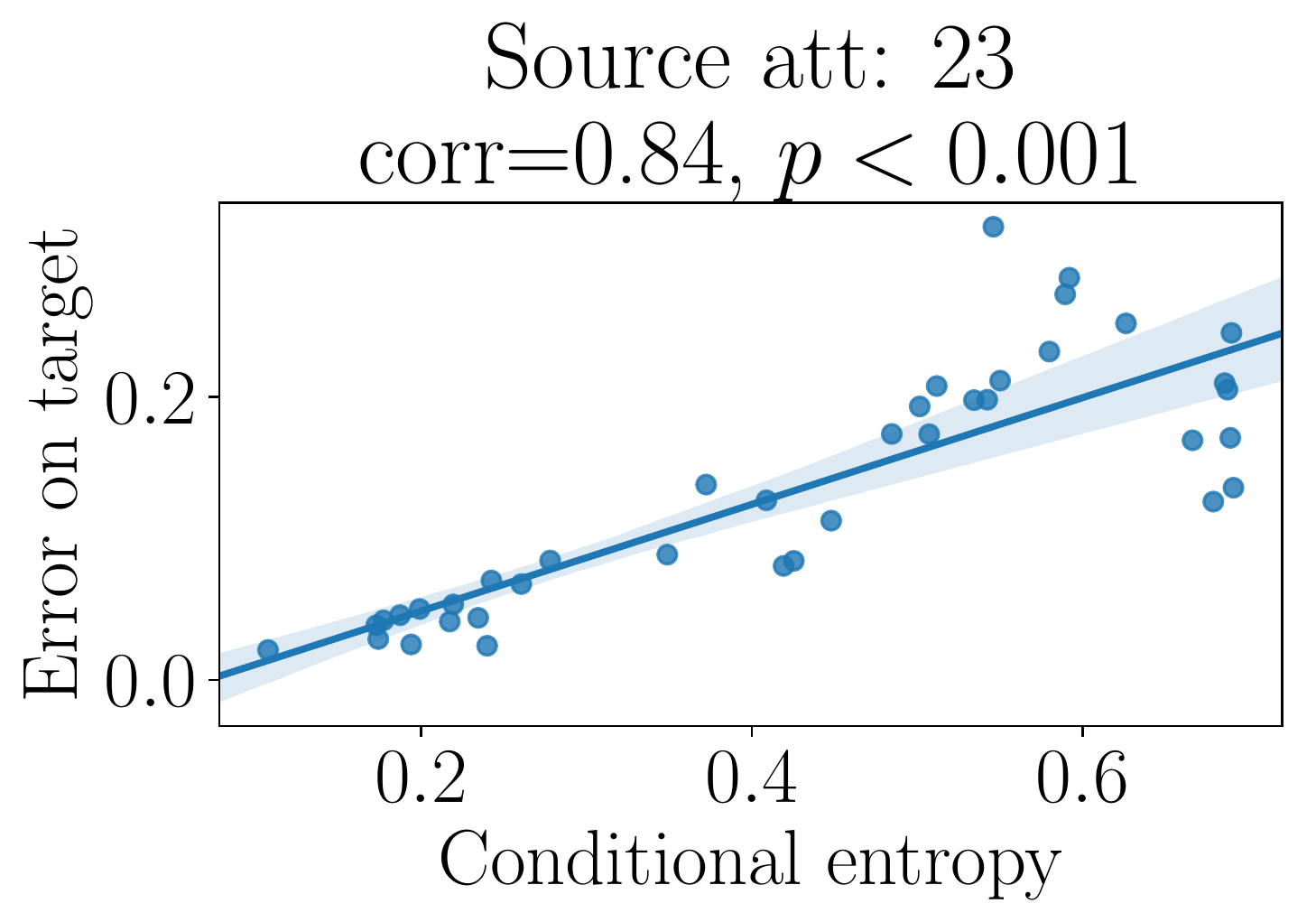}&
    \includegraphics[clip, trim=0mm 0mm 0mm 11mm, width=0.19\textwidth]{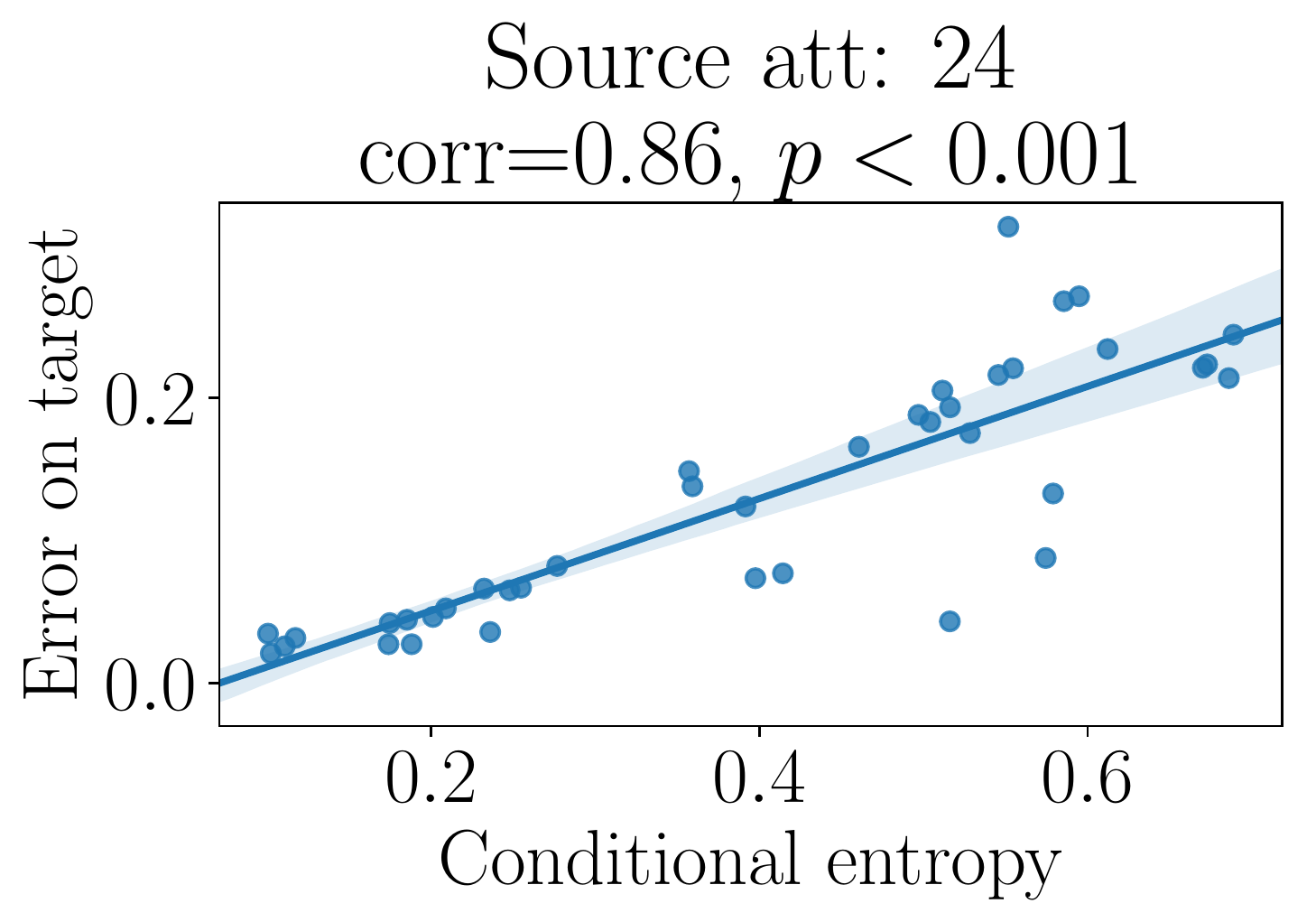}\\[-2pt]
    (20) Male & (21) Mouth slightly open & (22) Mustache & (23) Narrow eyes  & (24) No Beard\\[6pt]
    \includegraphics[clip, trim=0mm 0mm 0mm 11mm, width=0.19\textwidth]{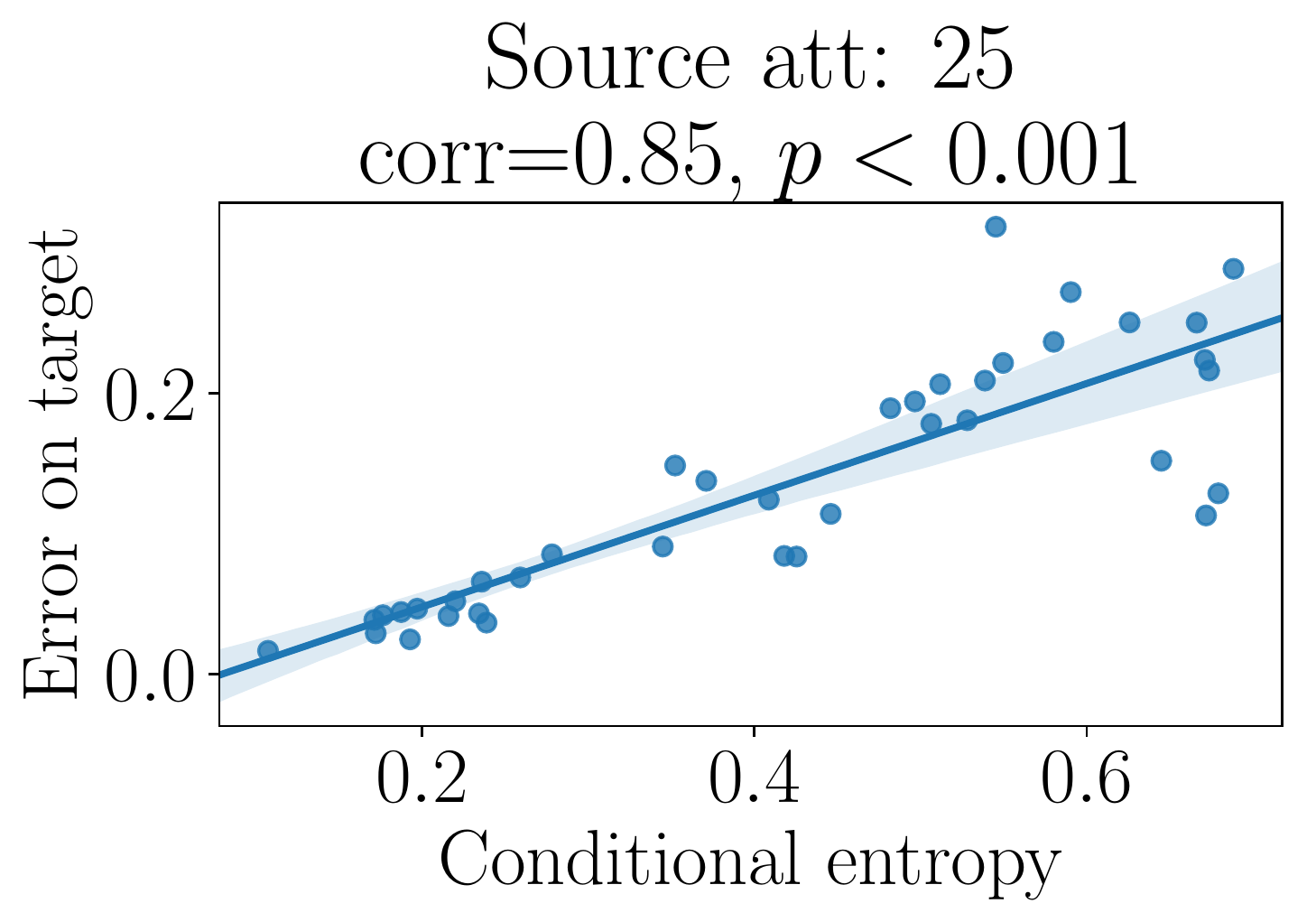}&
    \includegraphics[clip, trim=0mm 0mm 0mm 11mm, width=0.19\textwidth]{figures/tran_from_26.pdf}&
    \includegraphics[clip, trim=0mm 0mm 0mm 11mm, width=0.19\textwidth]{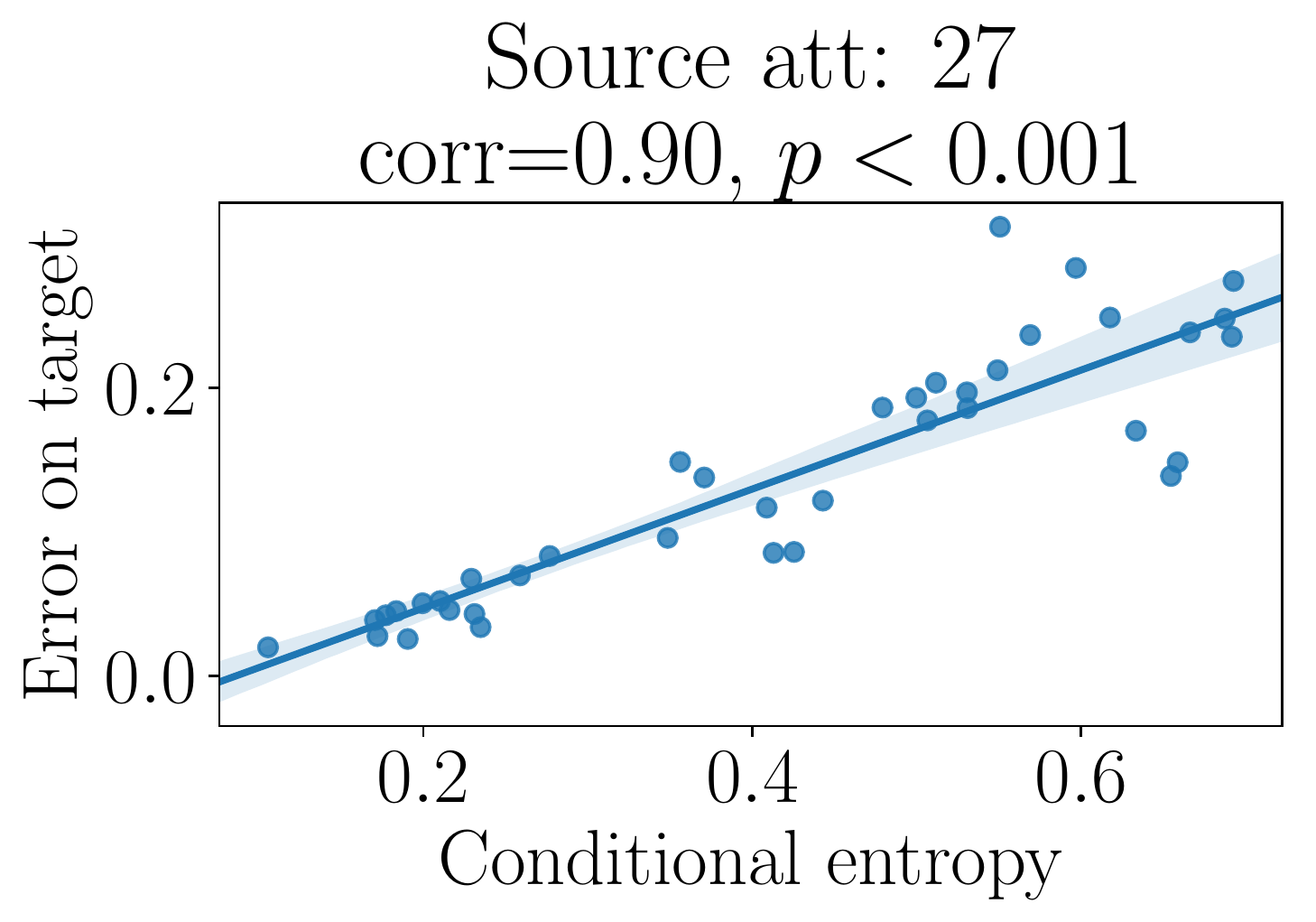}&
    \includegraphics[clip, trim=0mm 0mm 0mm 11mm, width=0.19\textwidth]{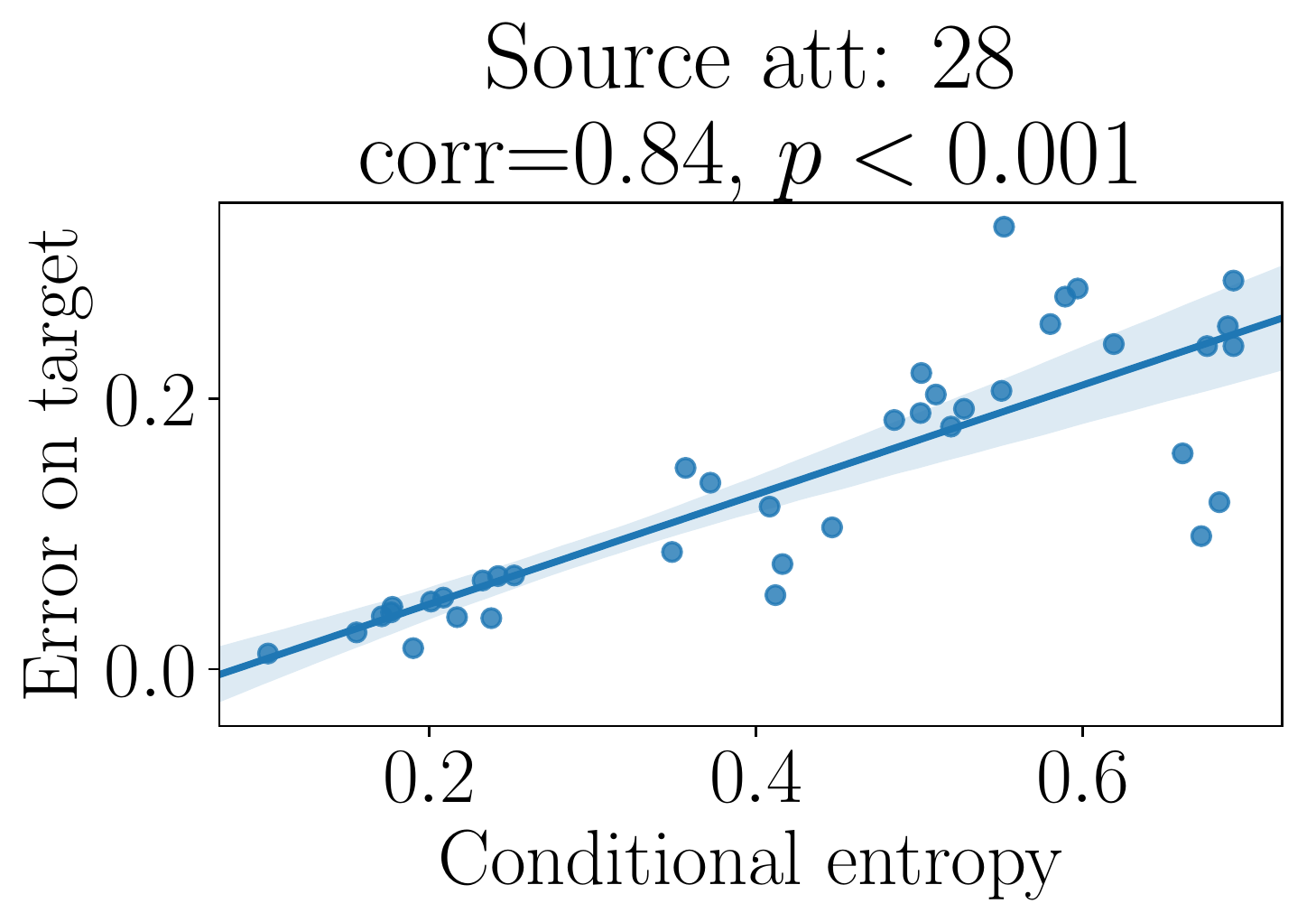}&
    \includegraphics[clip, trim=0mm 0mm 0mm 11mm, width=0.19\textwidth]{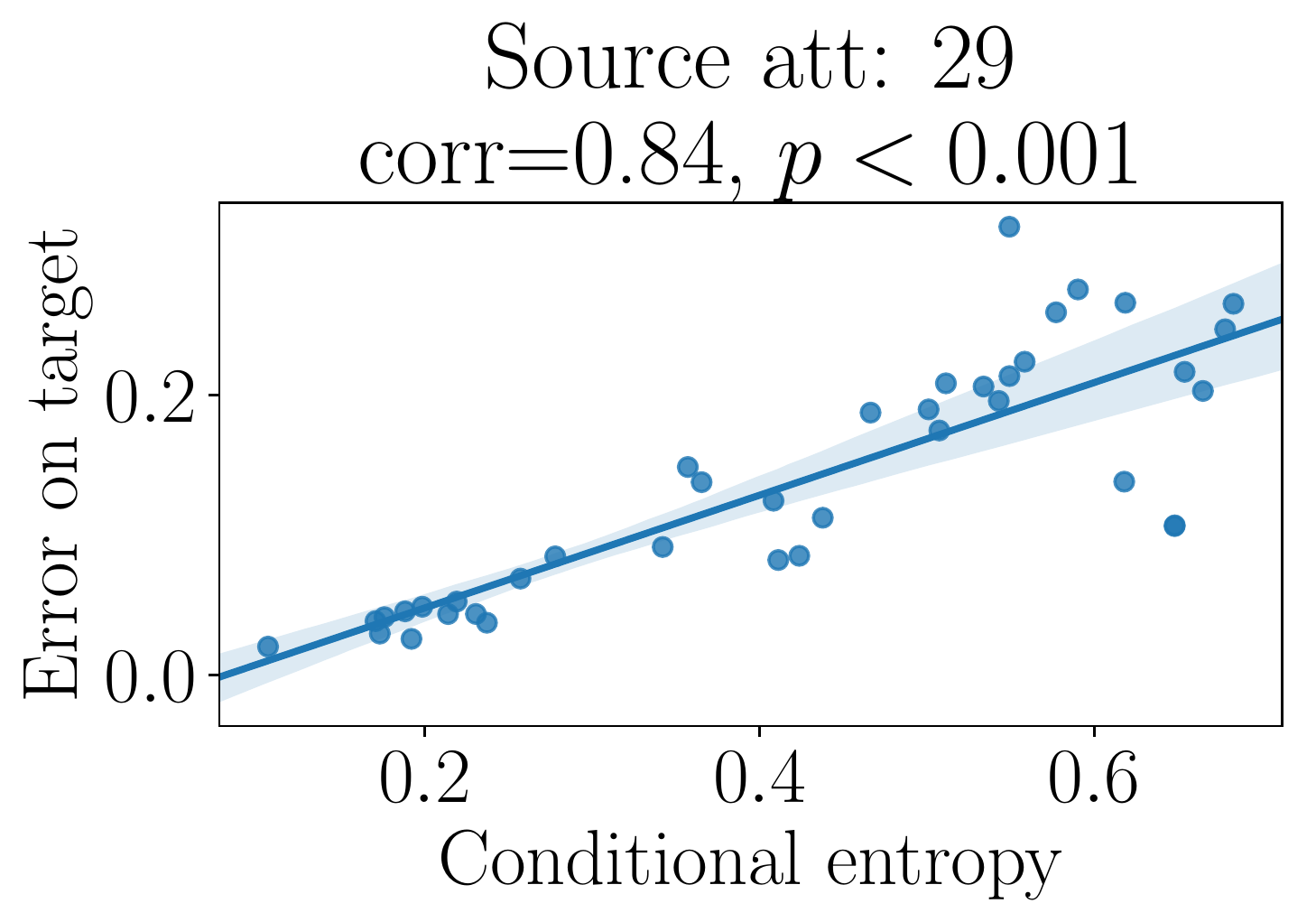}\\[-2pt]
    (25) Oval face & (26) Pale skin & (27) Pointy nose & (28) Receding hairline  & (29) Rosy cheeks\\[6pt]
    \includegraphics[clip, trim=0mm 0mm 0mm 11mm, width=0.19\textwidth]{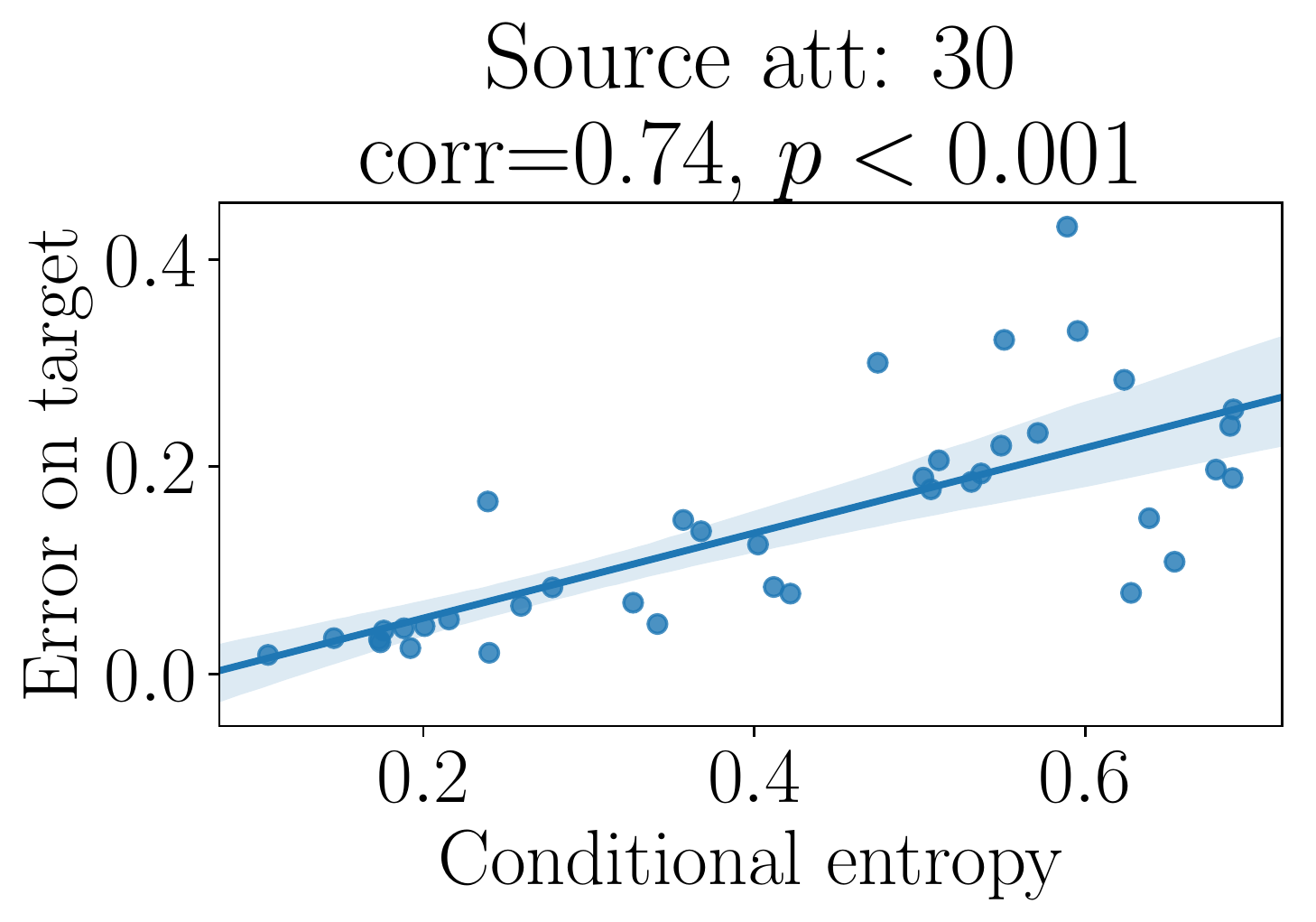}&
    \includegraphics[clip, trim=0mm 0mm 0mm 11mm, width=0.19\textwidth]{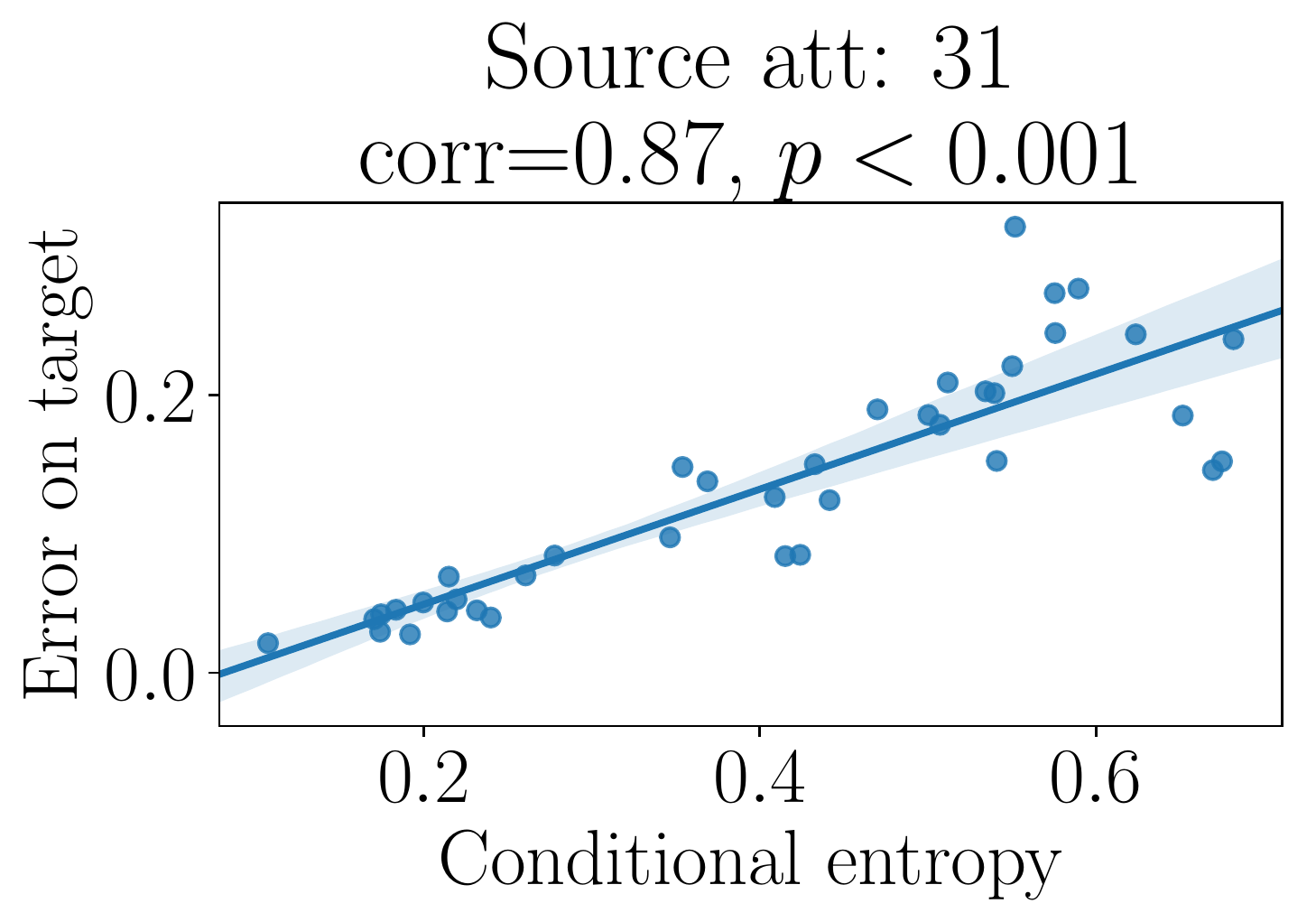}&
    \includegraphics[clip, trim=0mm 0mm 0mm 11mm, width=0.19\textwidth]{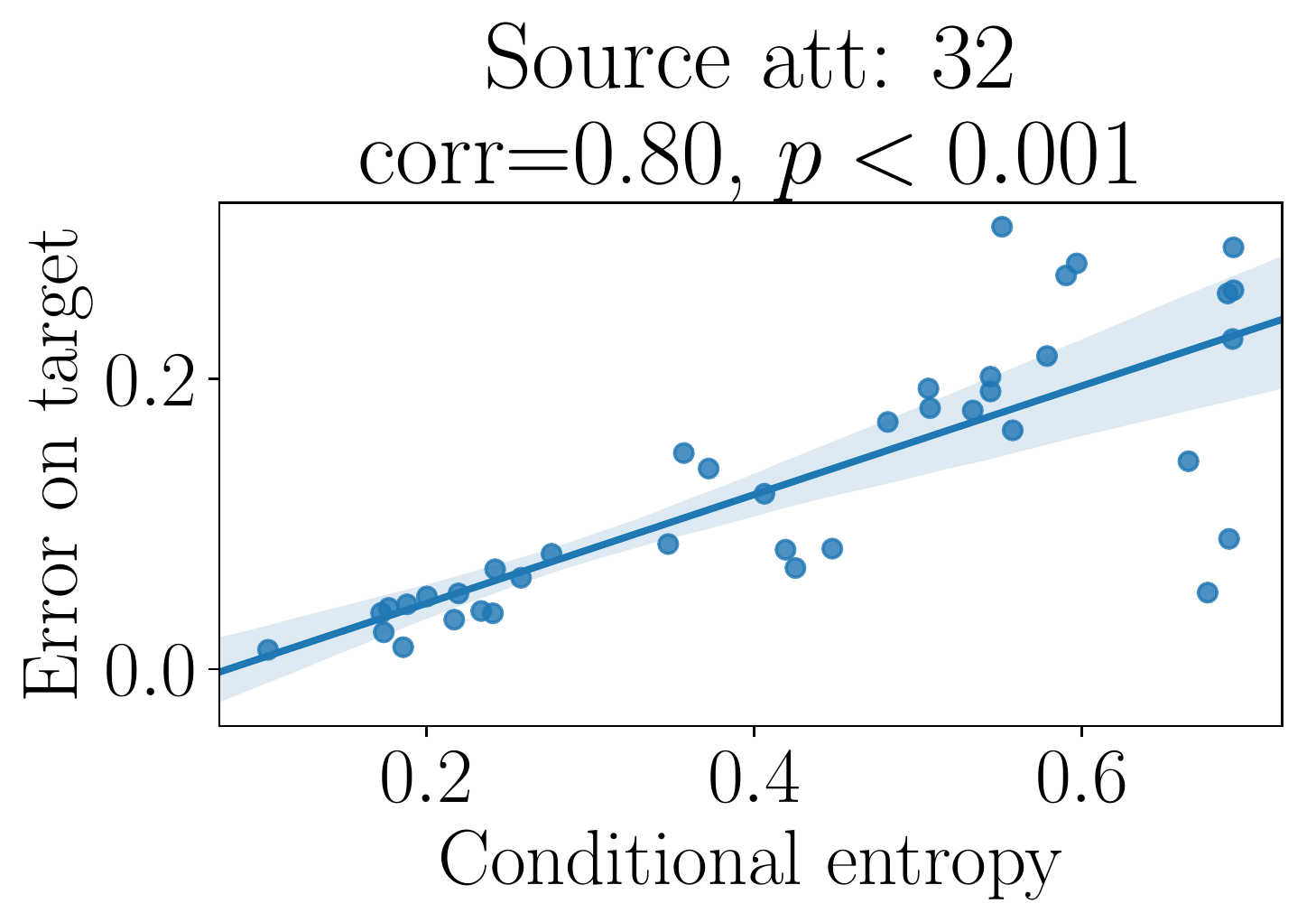}&
    \includegraphics[clip, trim=0mm 0mm 0mm 11mm, width=0.19\textwidth]{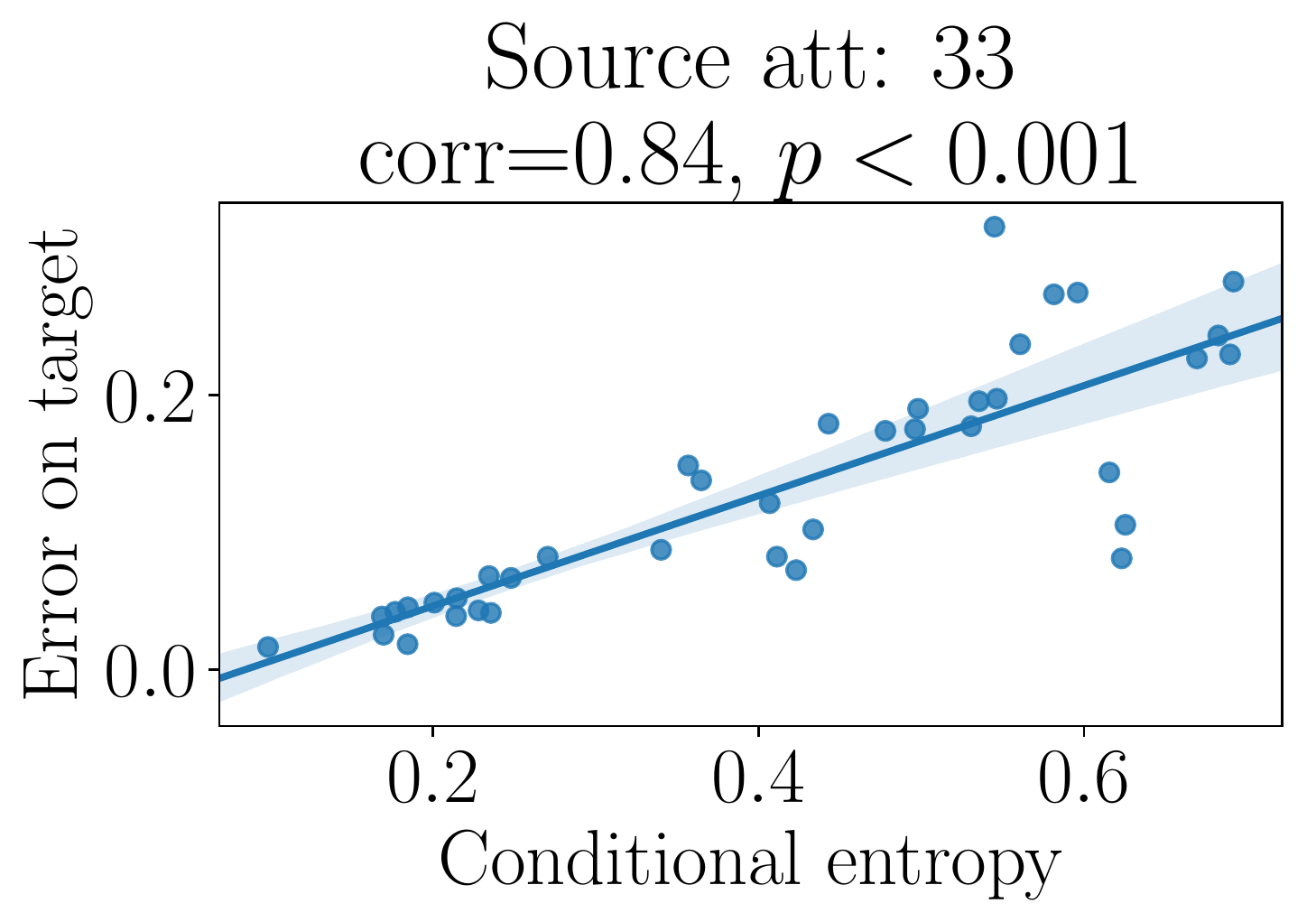}&
    \includegraphics[clip, trim=0mm 0mm 0mm 11mm, width=0.19\textwidth]{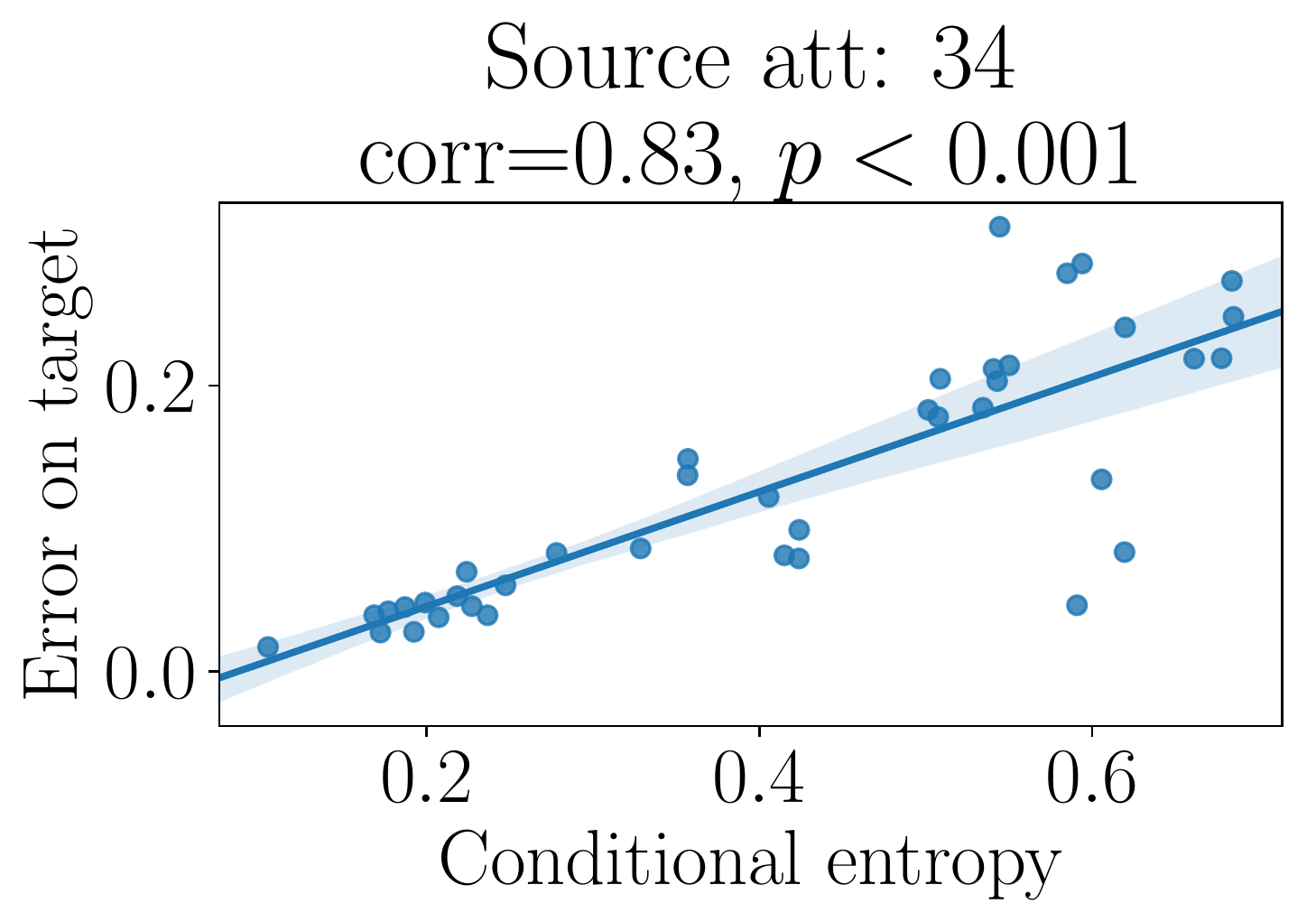}\\[-2pt]
    (30) Sideburns & (31) Smiling & (32) Straight hair & (33) Wavy hair & (34) Wearing earrings\\[6pt]
    \includegraphics[clip, trim=0mm 0mm 0mm 11mm, width=0.19\textwidth]{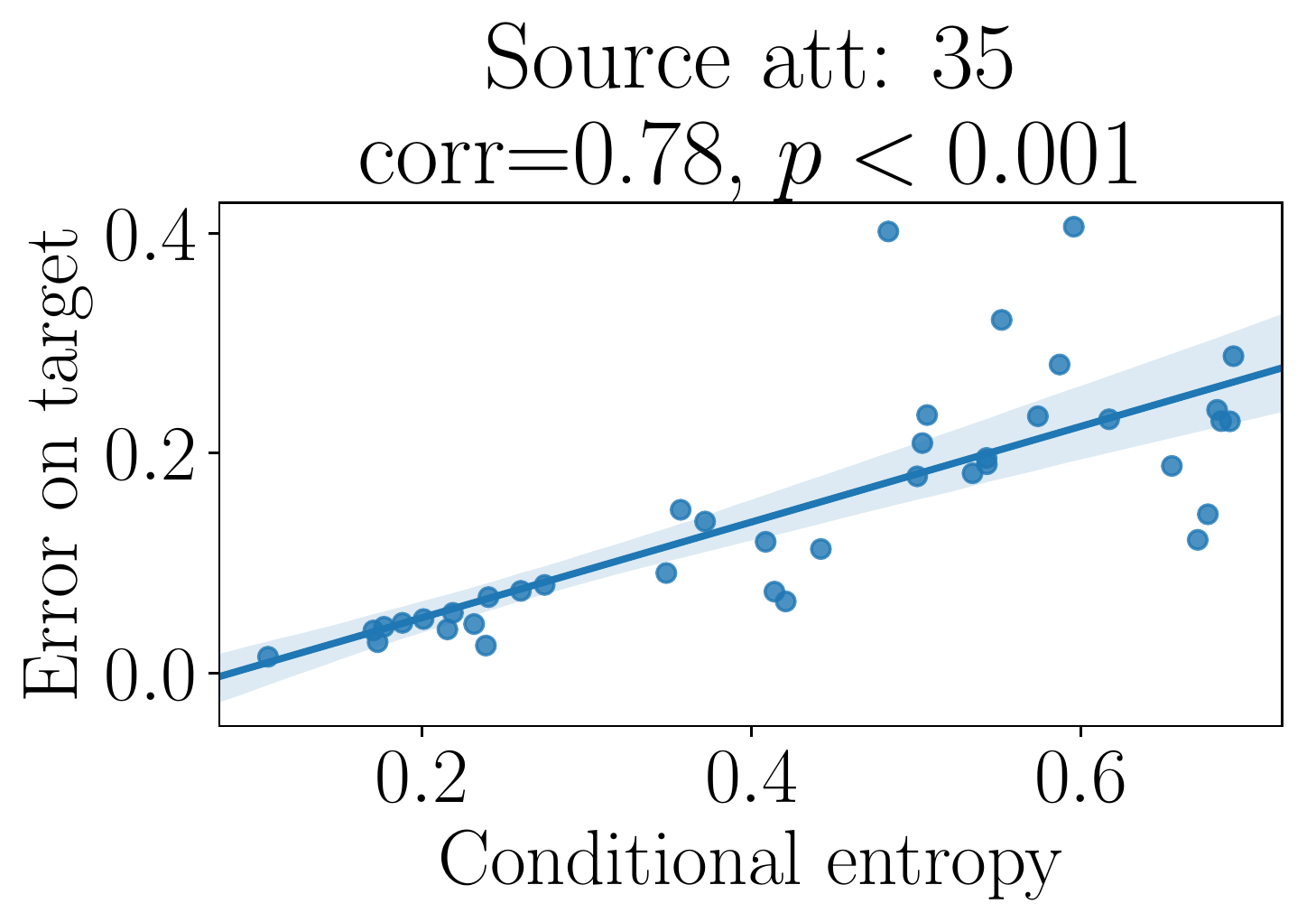}&
    \includegraphics[clip, trim=0mm 0mm 0mm 11mm, width=0.19\textwidth]{figures/tran_from_36.pdf}&
    \includegraphics[clip, trim=0mm 0mm 0mm 11mm, width=0.19\textwidth]{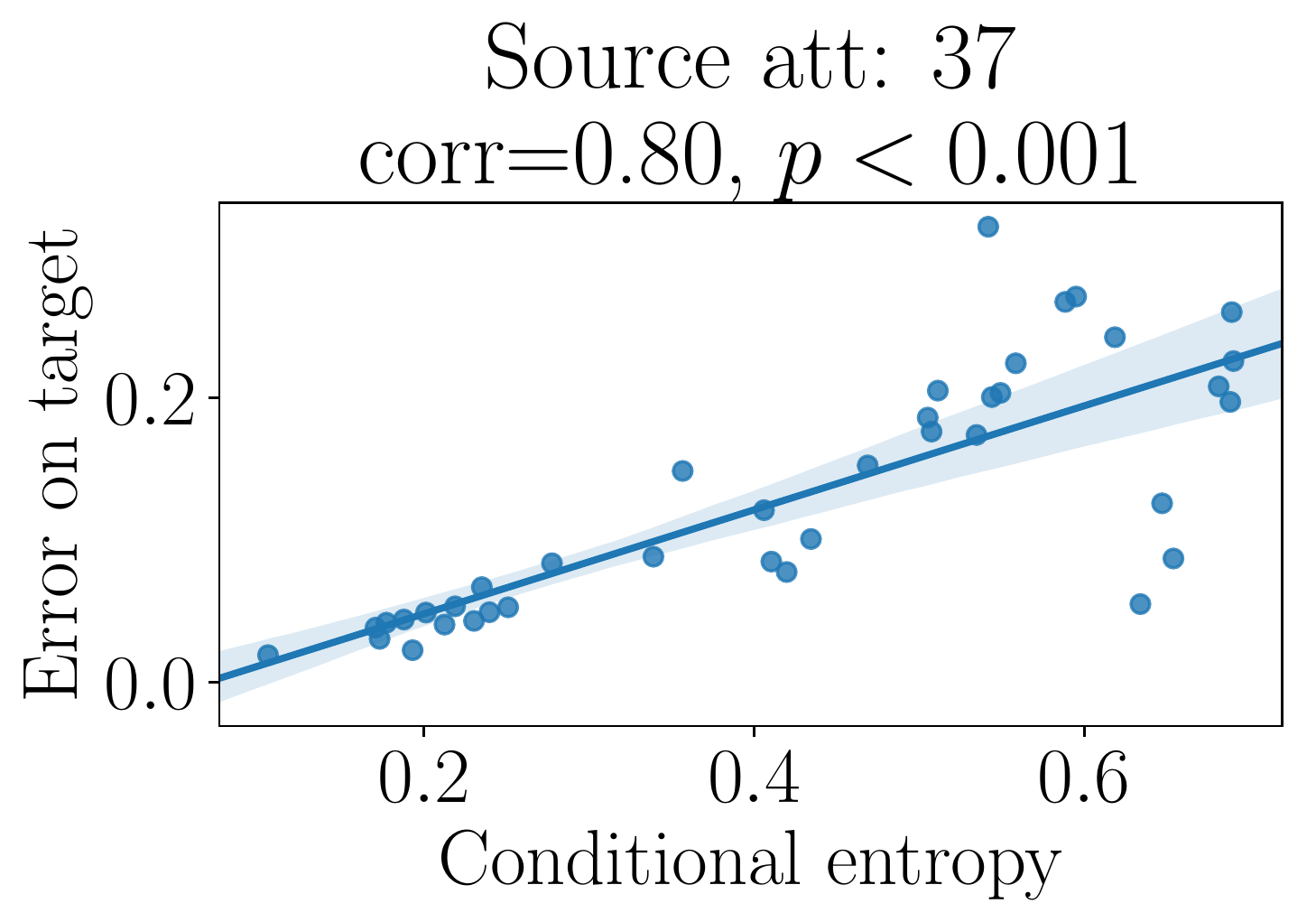}&
    \includegraphics[clip, trim=0mm 0mm 0mm 11mm, width=0.19\textwidth]{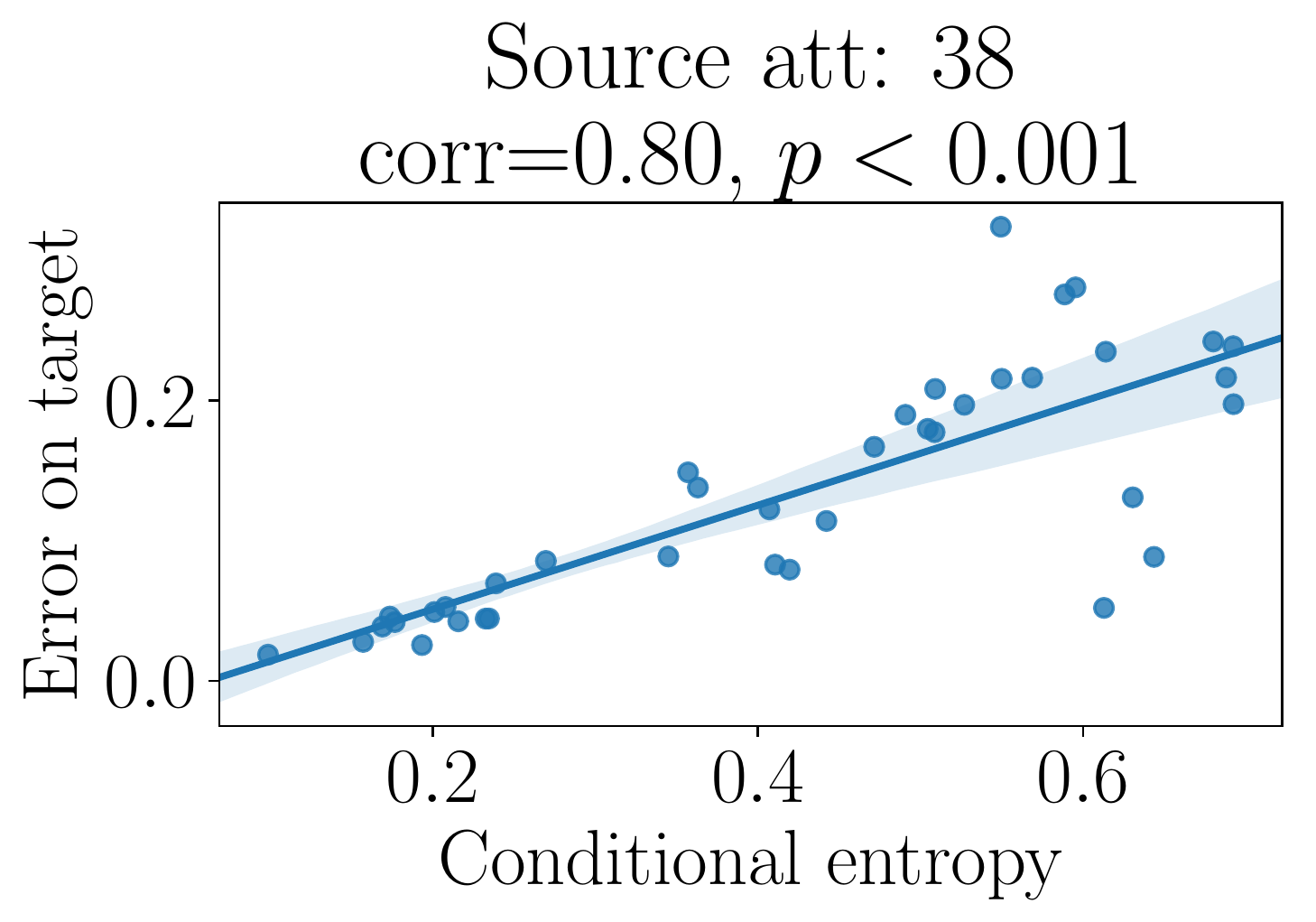}&
    \includegraphics[clip, trim=0mm 0mm 0mm 11mm, width=0.19\textwidth]{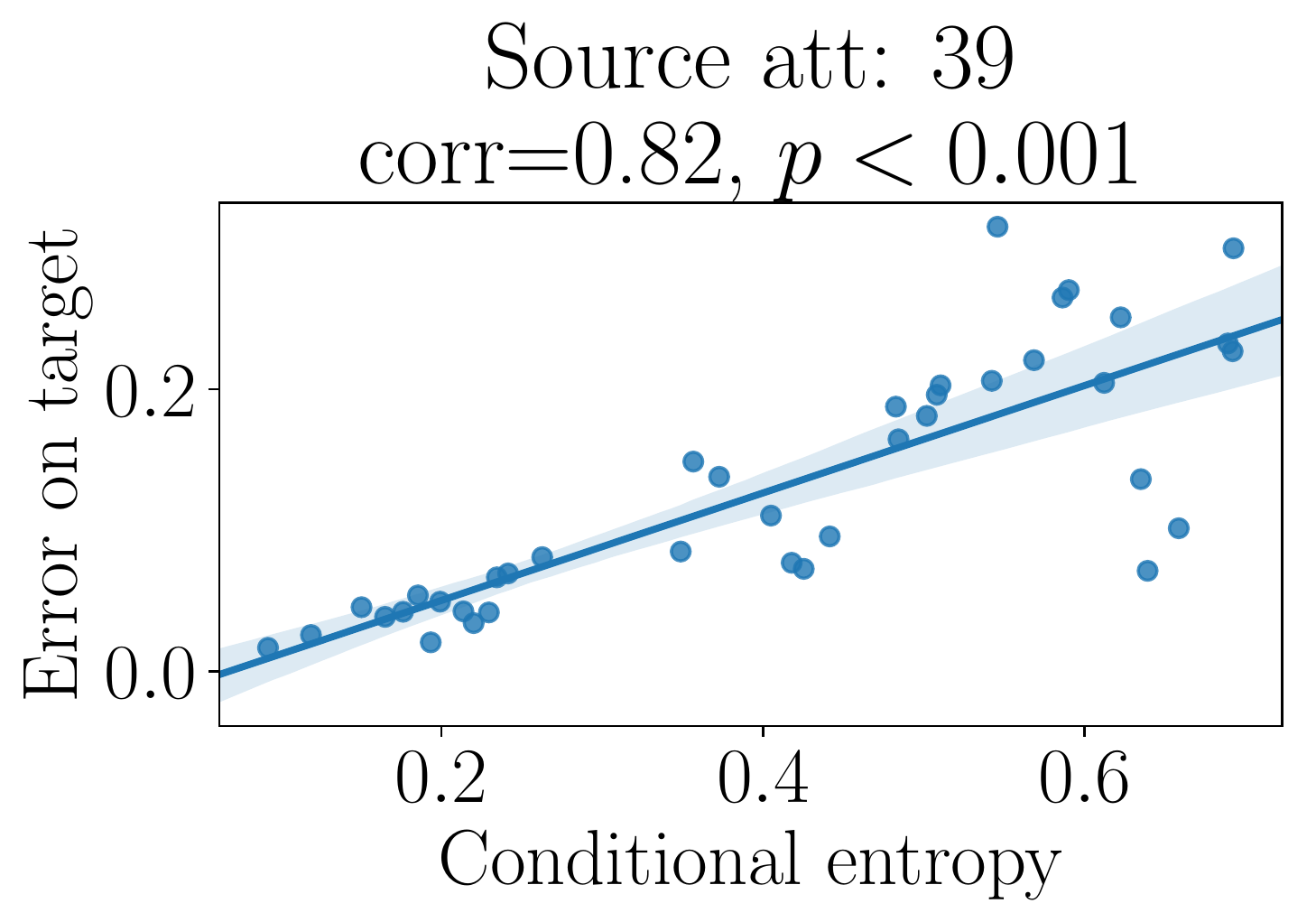}\\[-2pt]
    (35) Wearing hat & (36) Wearing lipstick & (37) Wearing necklace & (38) Wearing necktie & (39) Young\\[6pt]
\end{tabular}
\caption{{\bf Attribute prediction; CE vs. test errors on CelebA (Extended from Fig.~\ref{fig:face_att_trans}(a-d) in the paper).} The source attribute, $T^Z$, in each plot is named in the plot title. Points represent different target tasks $T^Y$. Corr is the Pearson correlation coefficient between the two variables and $p$ is the statistical significance of the correlation. In all cases, the correlation is statistically significant.}
\label{fig:trans_celebA}
\end{figure*}

\begin{figure*}[ht]
\centering
\footnotesize
\begin{tabular}{c@{~}c@{~}c@{~}c@{~}c}
\includegraphics[clip, trim=0mm 0mm 0mm 13mm, width=0.19\textwidth]{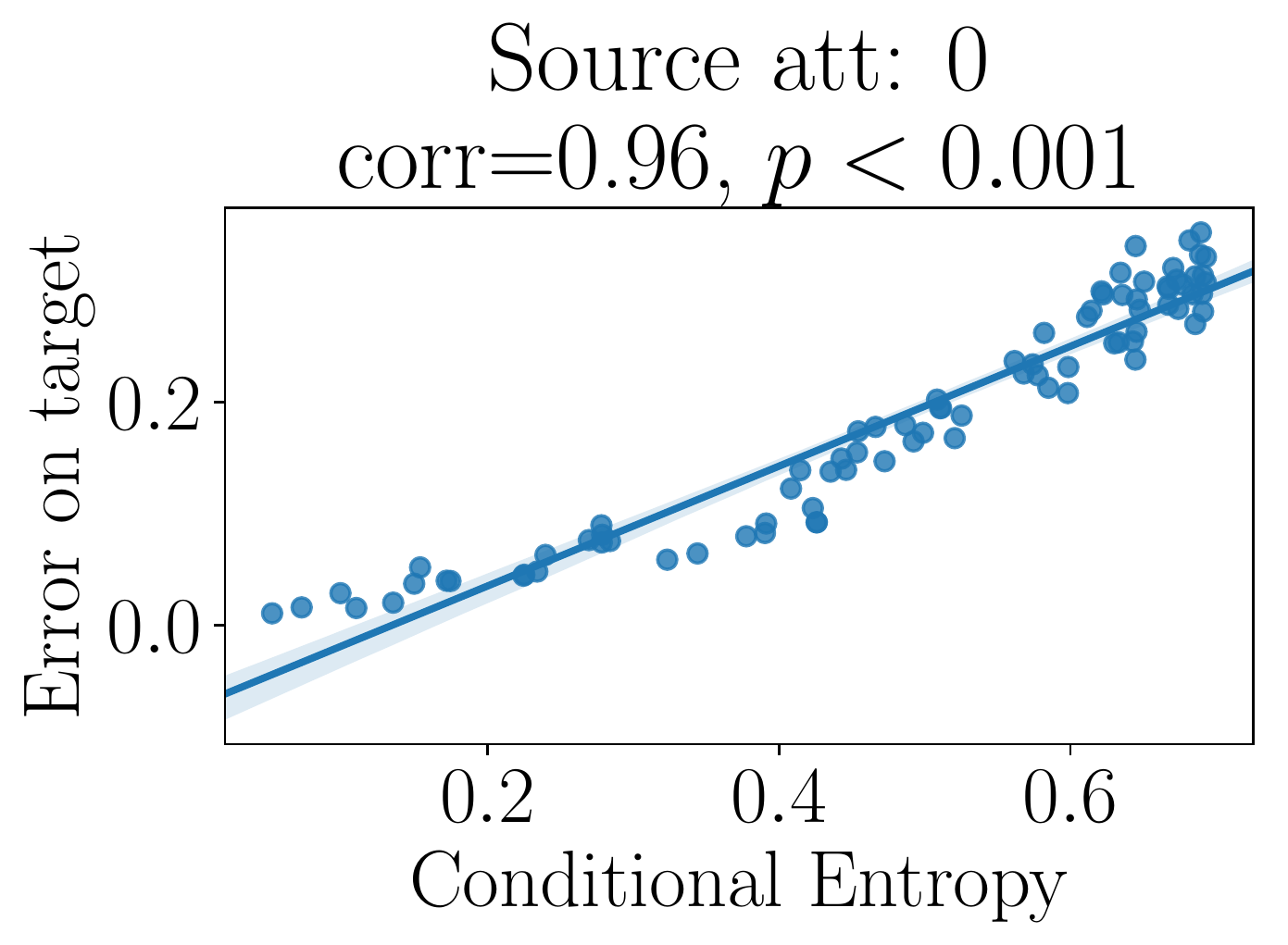}&

\includegraphics[clip, trim=0mm 0mm 0mm 13mm, width=0.19\textwidth]{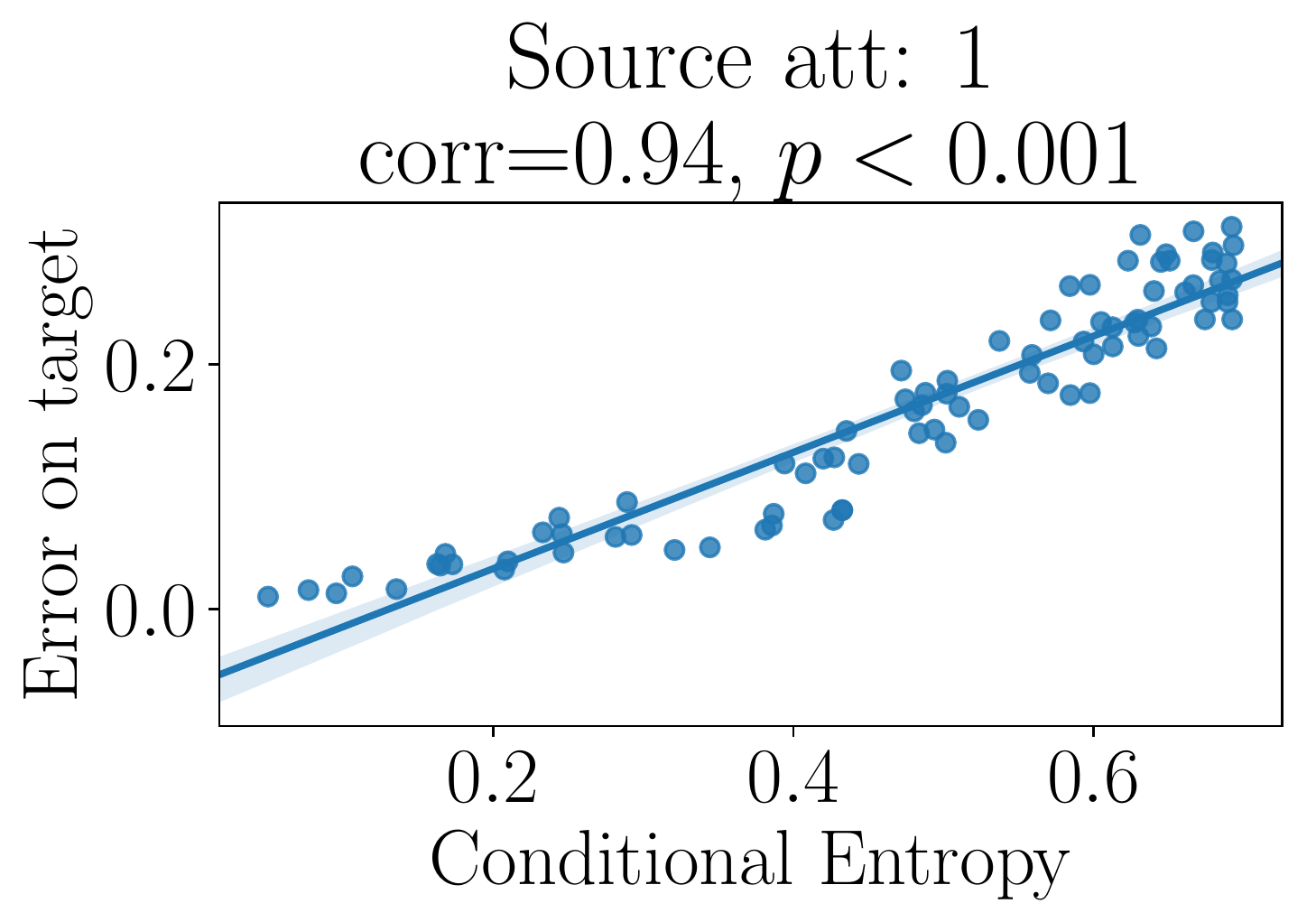}&

\includegraphics[clip, trim=0mm 0mm 0mm 13mm, width=0.19\textwidth]{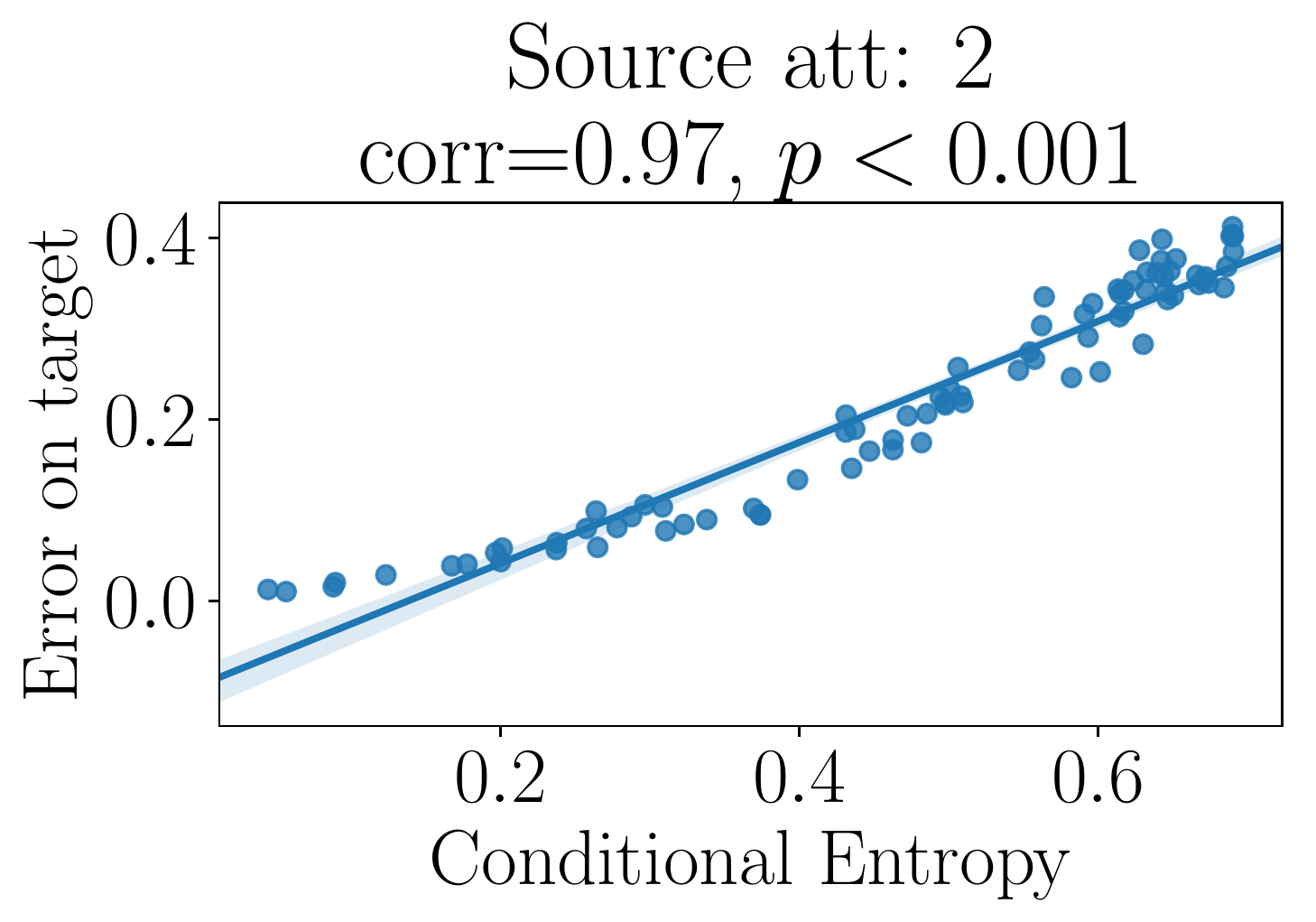}&

\includegraphics[clip, trim=0mm 0mm 0mm 13mm, width=0.19\textwidth]{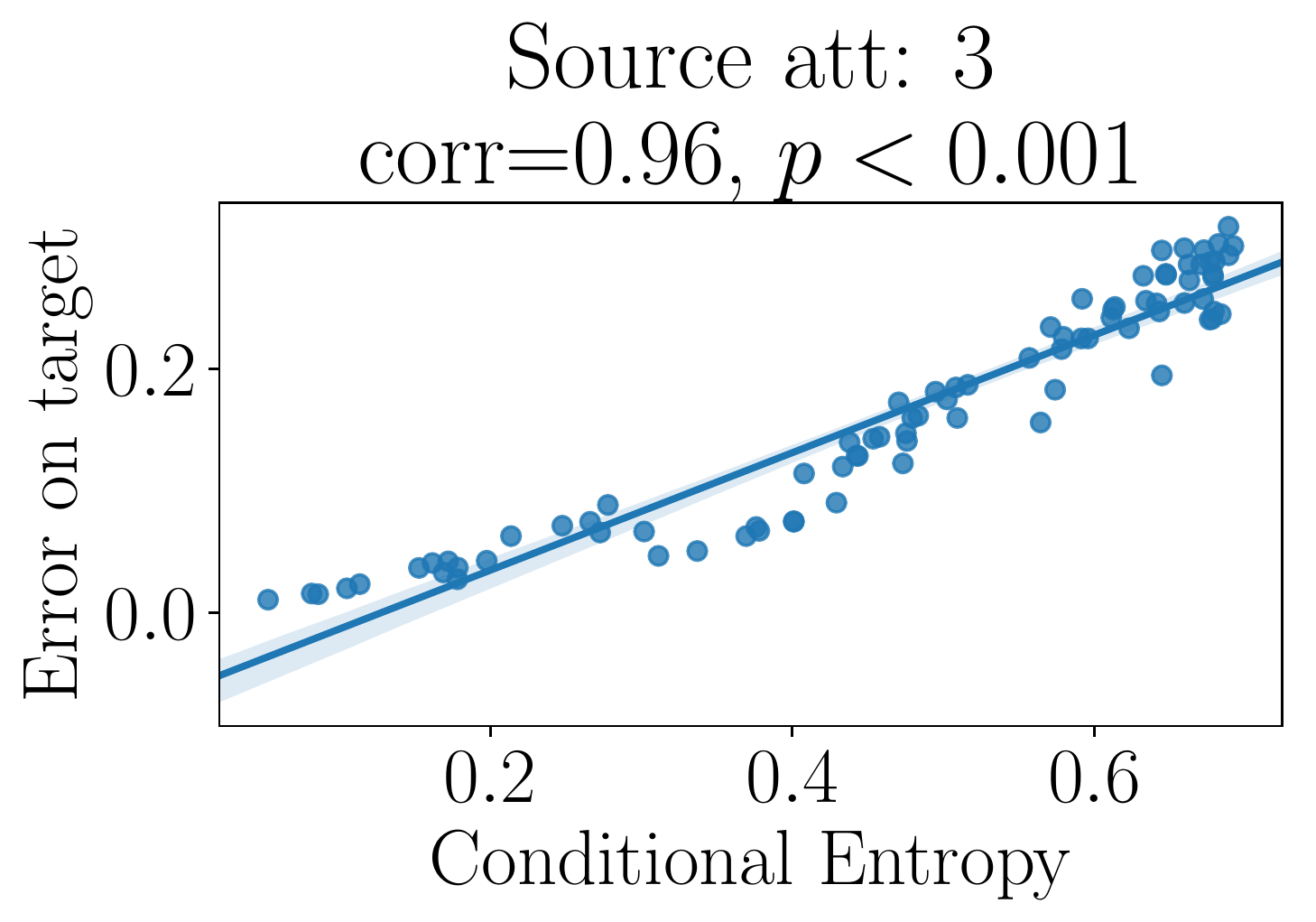}&

\includegraphics[clip, trim=0mm 0mm 0mm 13mm, width=0.19\textwidth]{figures/AWA2_att3.pdf}\\[-2pt]

(0) Black & (1) White & (2) Blue & (3) Brown & (4) Gray\\[6pt]
\includegraphics[clip, trim=0mm 0mm 0mm 13mm, width=0.19\textwidth]{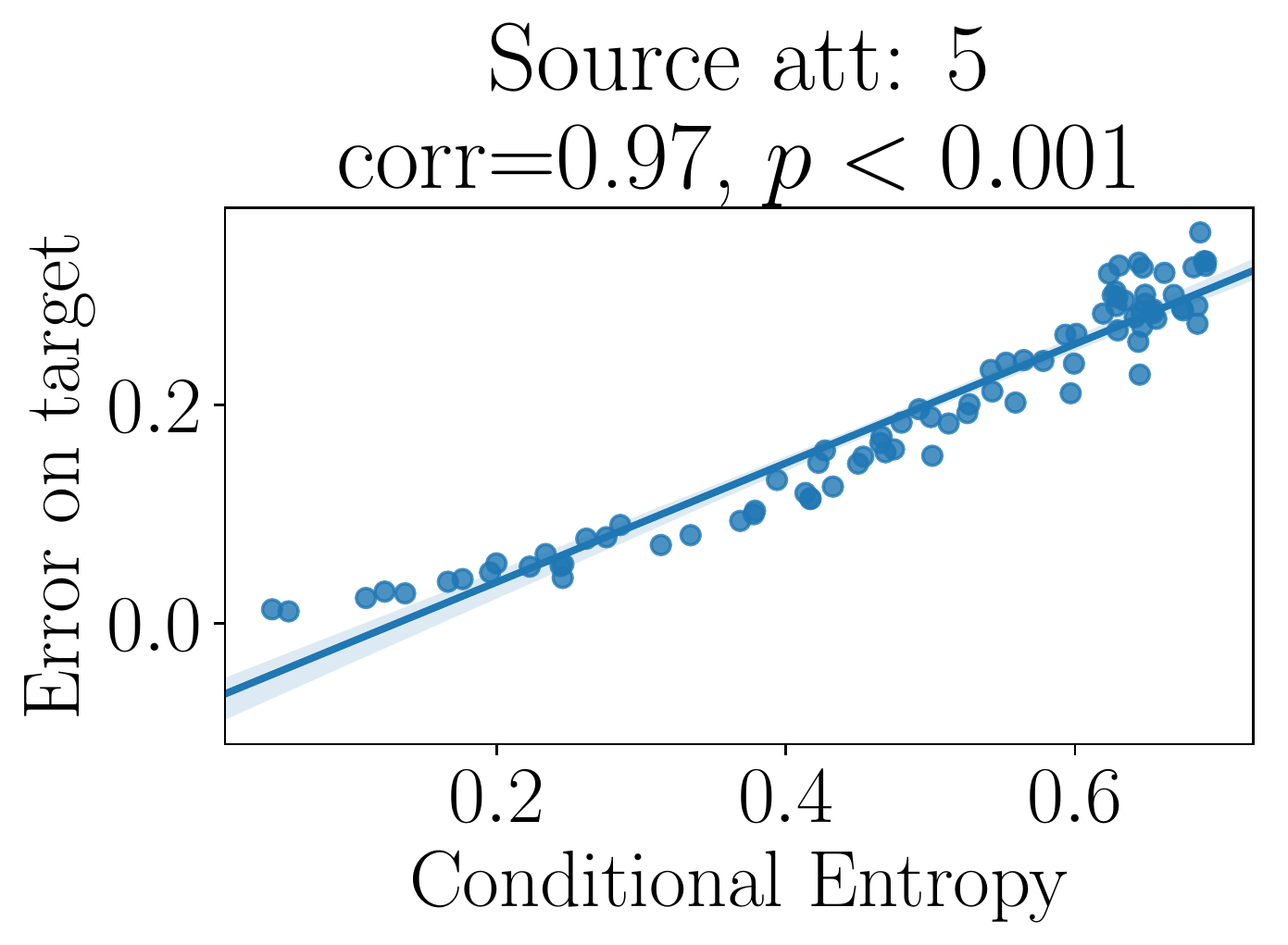}&

\includegraphics[clip, trim=0mm 0mm 0mm 13mm, width=0.19\textwidth]{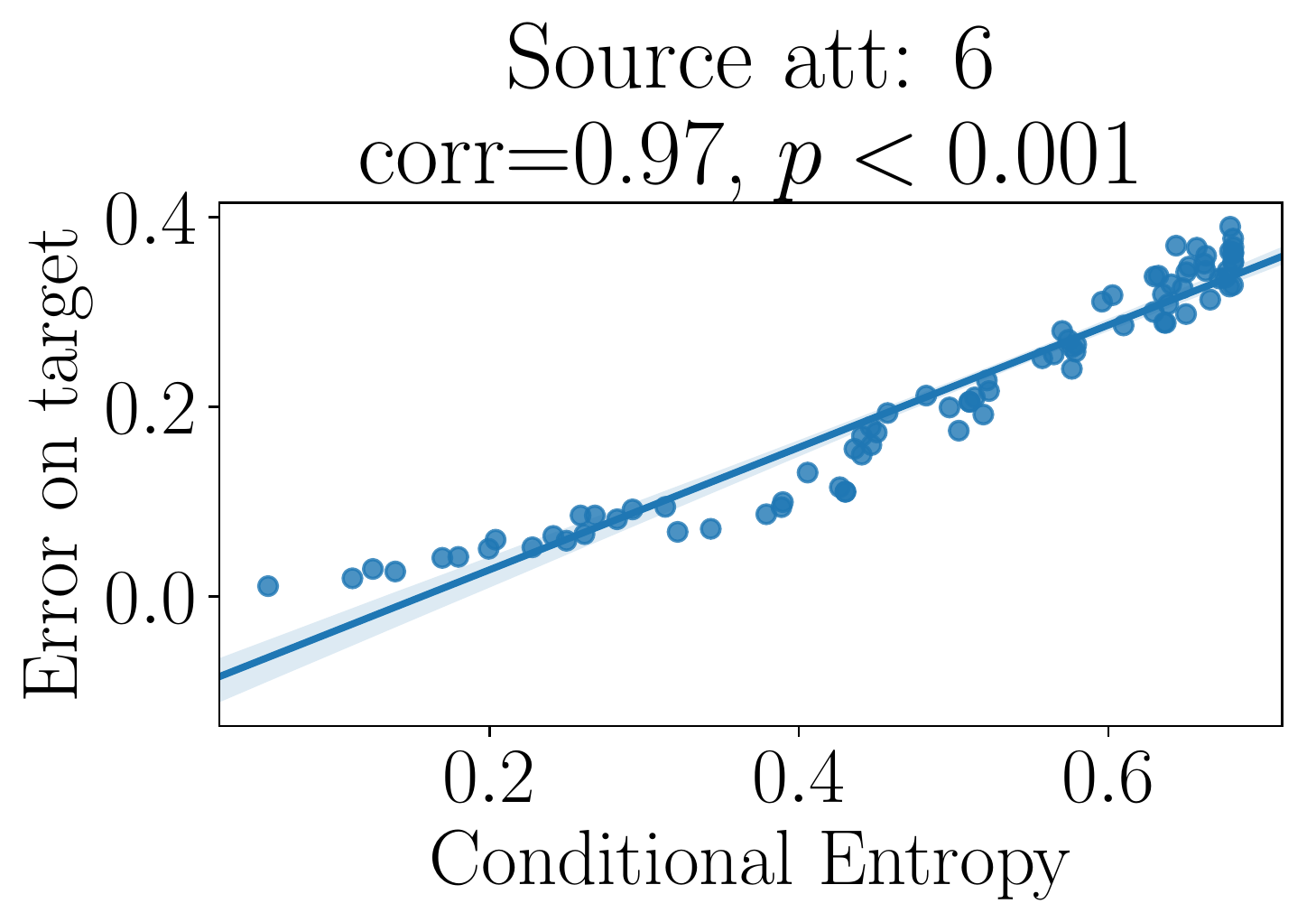}&

\includegraphics[clip, trim=0mm 0mm 0mm 13mm, width=0.19\textwidth]{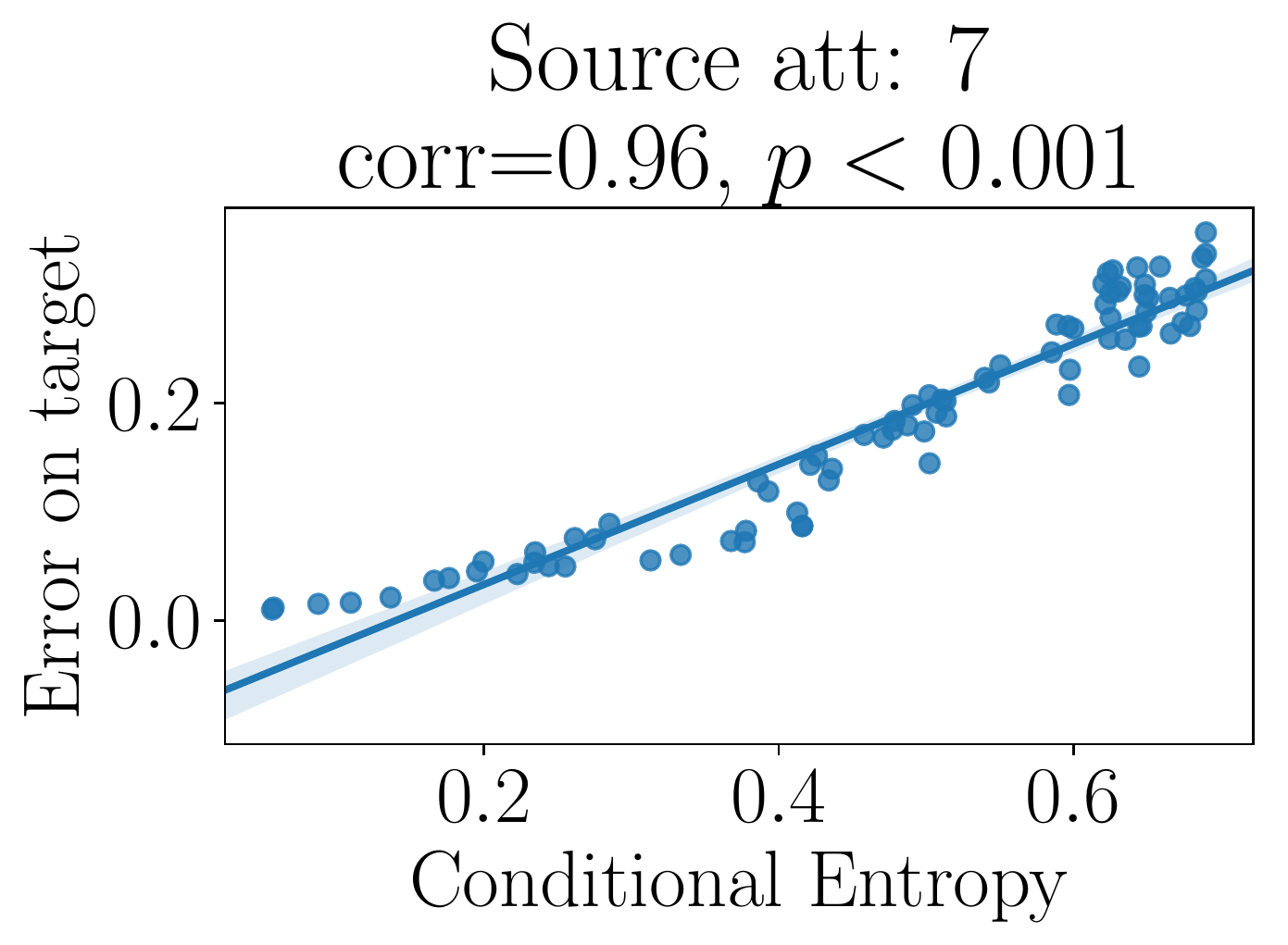}&

\includegraphics[clip, trim=0mm 0mm 0mm 13mm, width=0.19\textwidth]{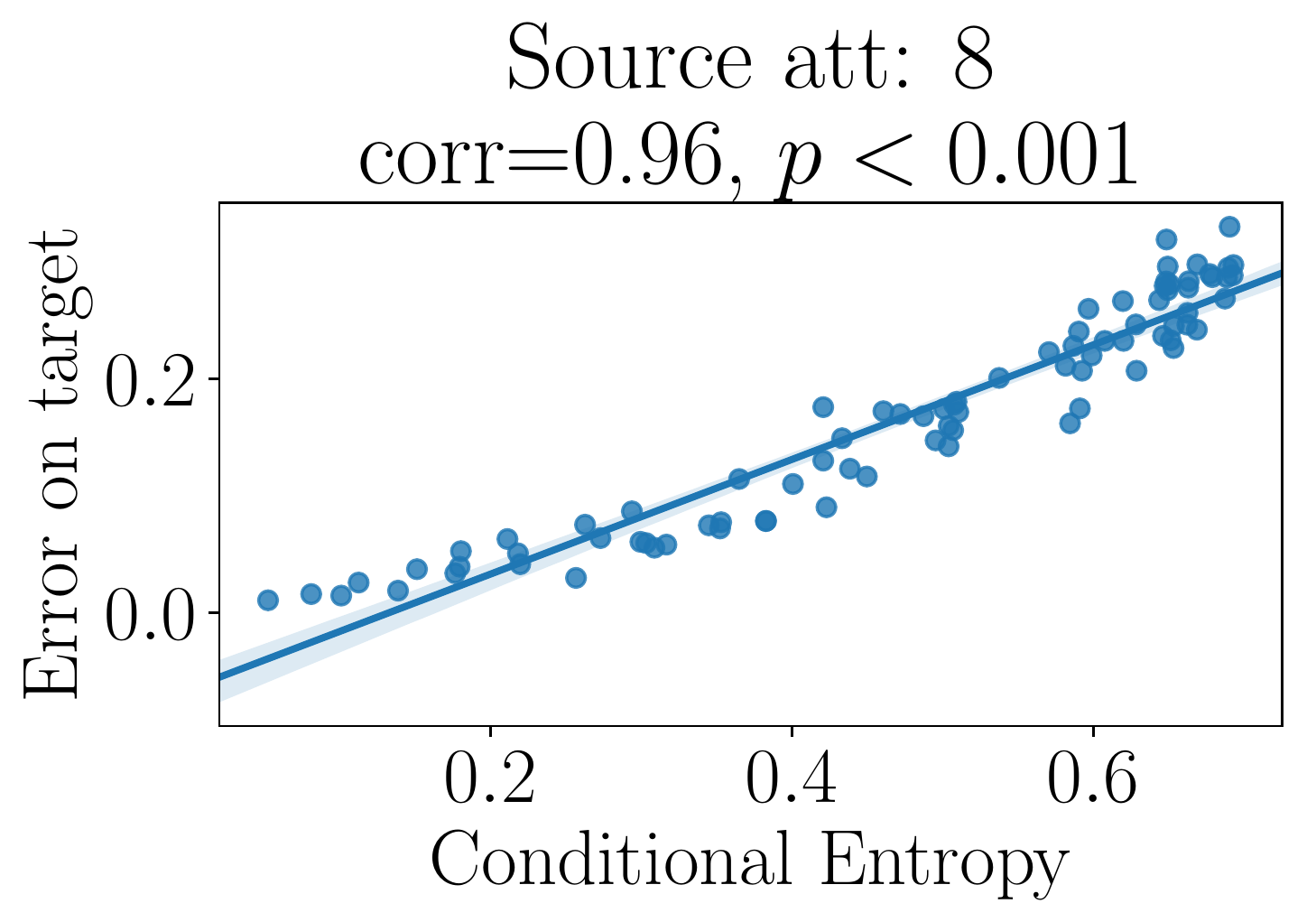}&

\includegraphics[clip, trim=0mm 0mm 0mm 13mm, width=0.19\textwidth]{figures/AWA2_att8.pdf}\\[-2pt]

(5) Orange & (6) Red & (7) Yellow & (8) Patches & (9) Spots\\[6pt]
\includegraphics[clip, trim=0mm 0mm 0mm 13mm, width=0.19\textwidth]{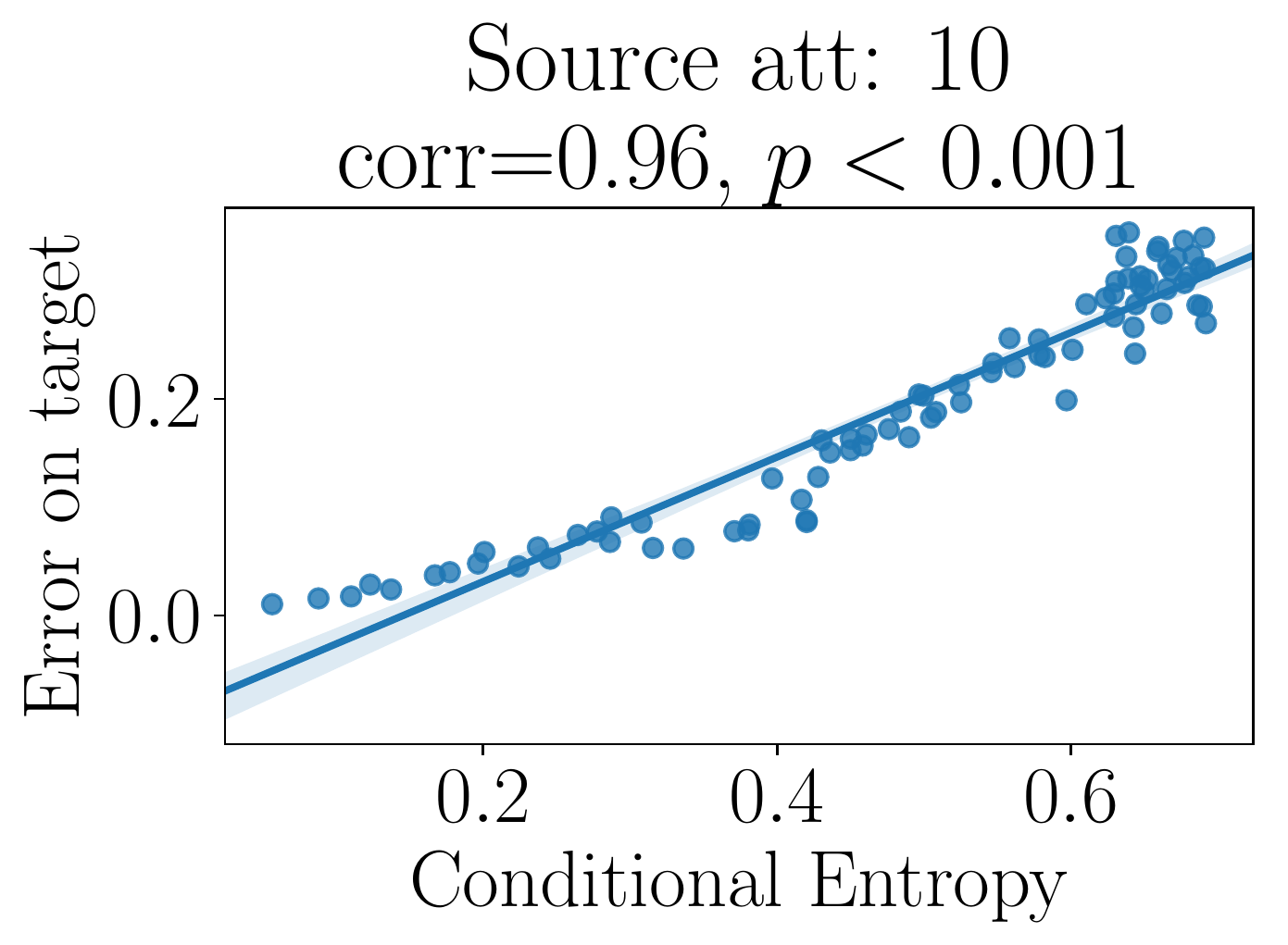}&

\includegraphics[clip, trim=0mm 0mm 0mm 13mm, width=0.19\textwidth]{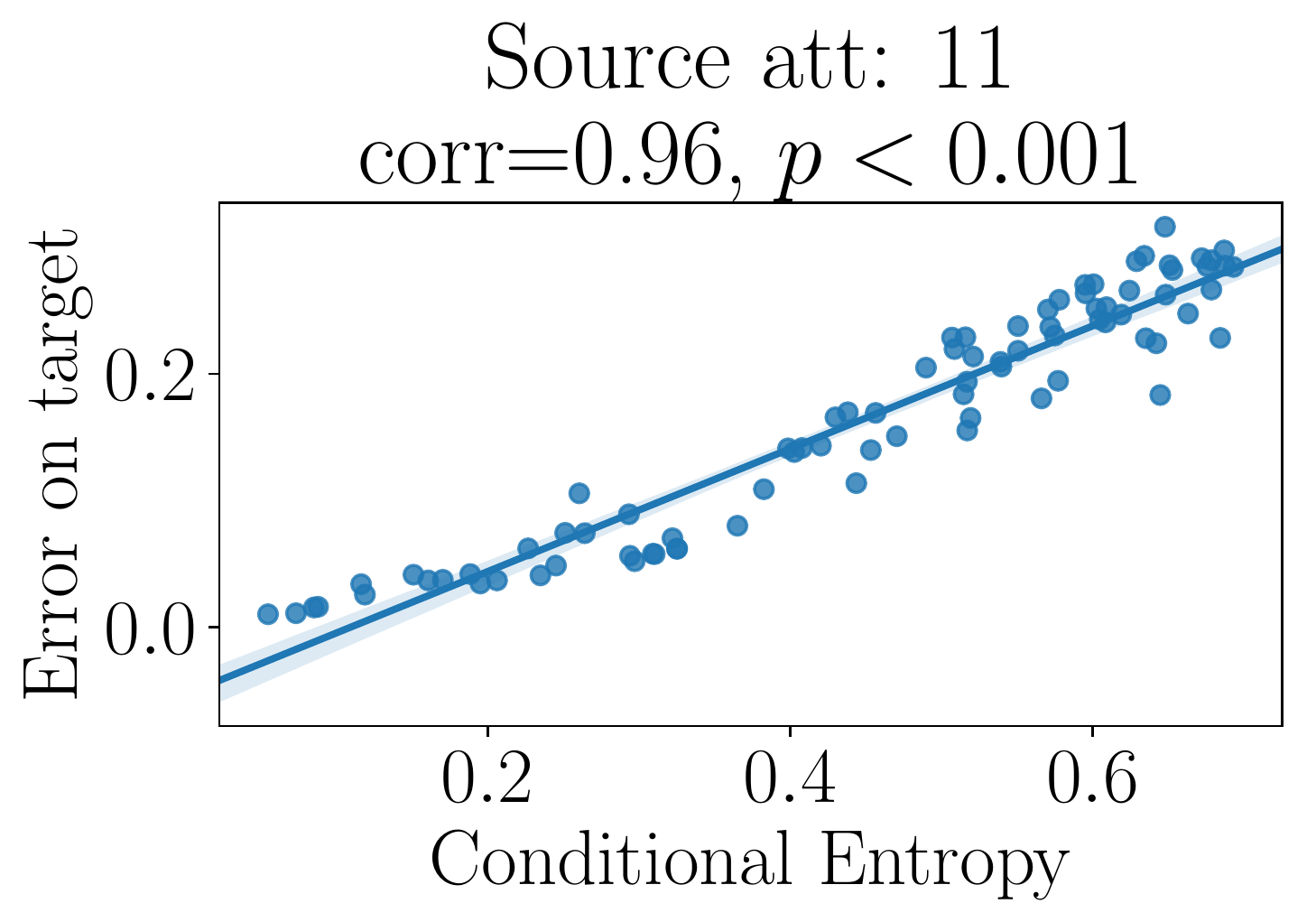}&

\includegraphics[clip, trim=0mm 0mm 0mm 13mm, width=0.19\textwidth]{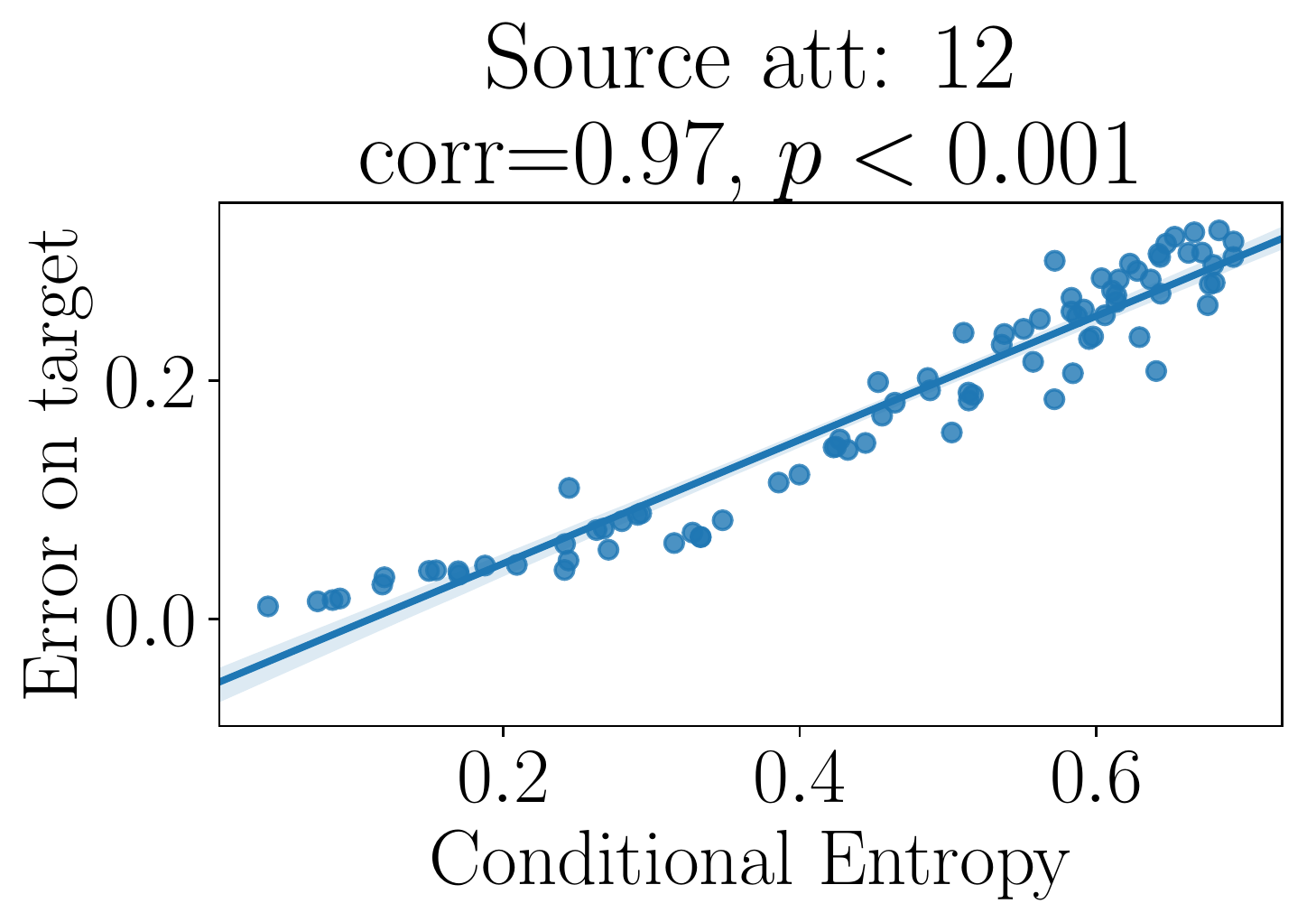}&

\includegraphics[clip, trim=0mm 0mm 0mm 13mm, width=0.19\textwidth]{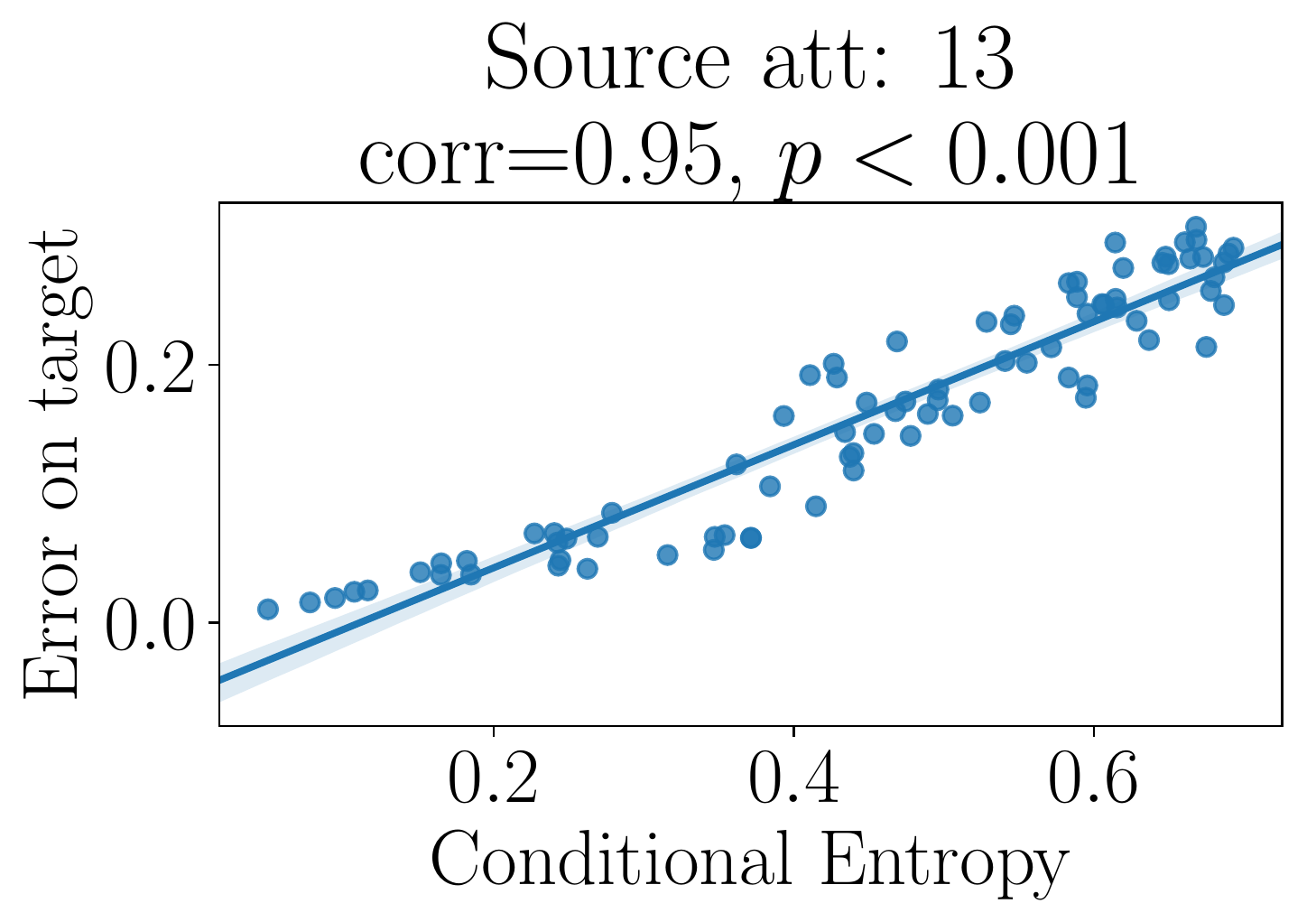}&

\includegraphics[clip, trim=0mm 0mm 0mm 13mm, width=0.19\textwidth]{figures/AWA2_att13.pdf}\\[-2pt]

(10) Stripes & (11) Furry & (12) Hairless & (13) Toughskin & (14) Big\\[6pt]
\includegraphics[clip, trim=0mm 0mm 0mm 13mm, width=0.19\textwidth]{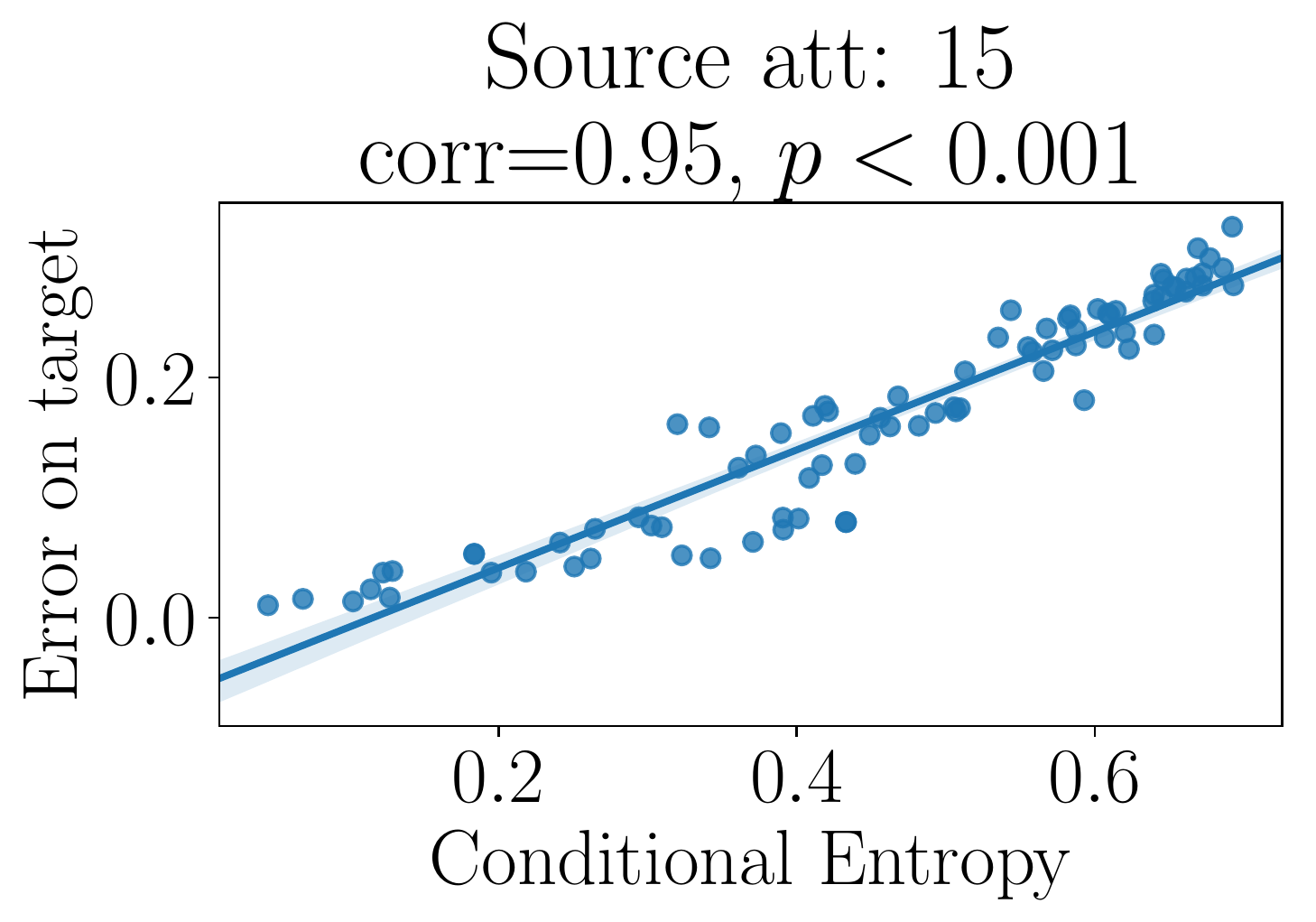}&

\includegraphics[clip, trim=0mm 0mm 0mm 13mm, width=0.19\textwidth]{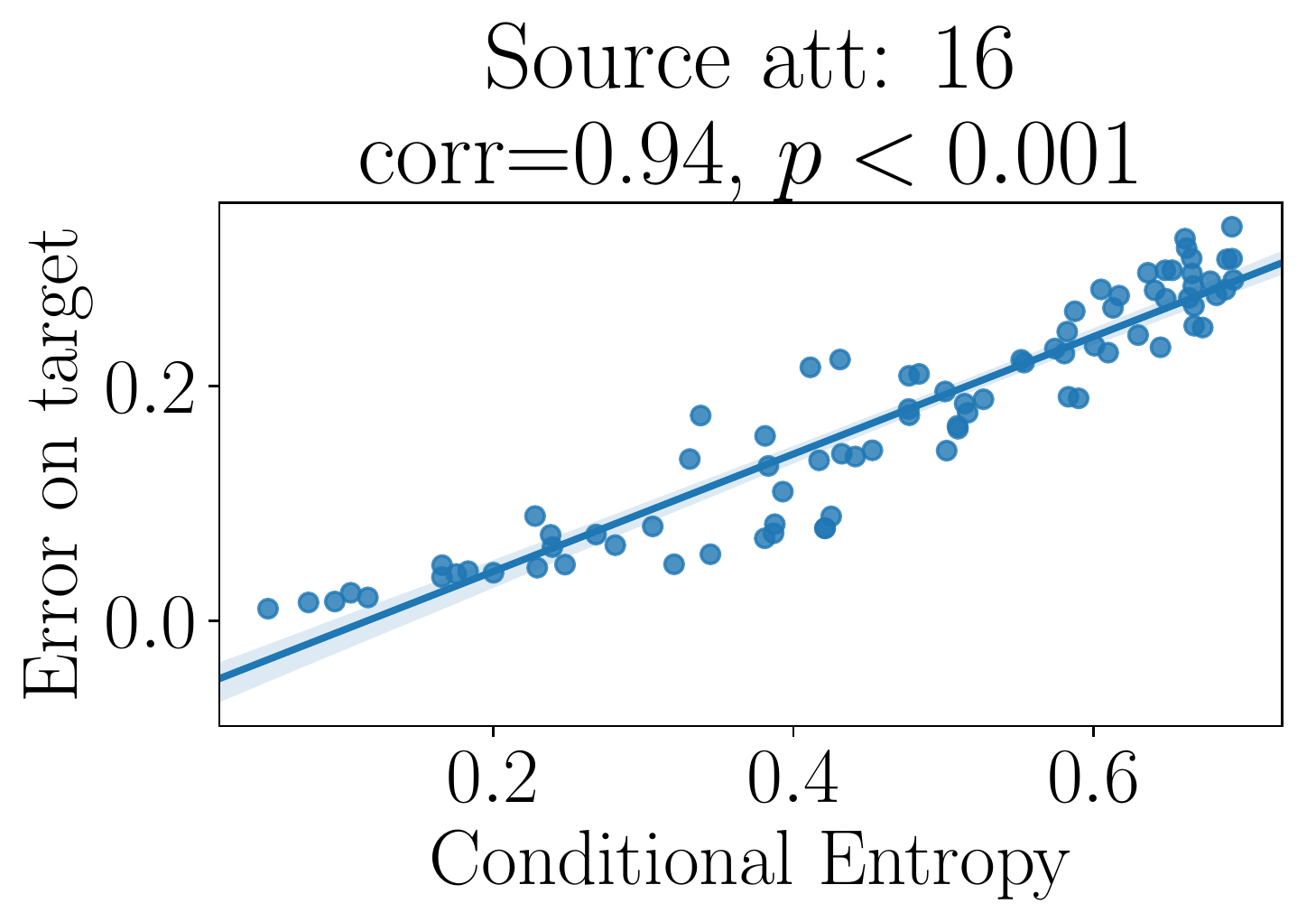}&

\includegraphics[clip, trim=0mm 0mm 0mm 13mm, width=0.19\textwidth]{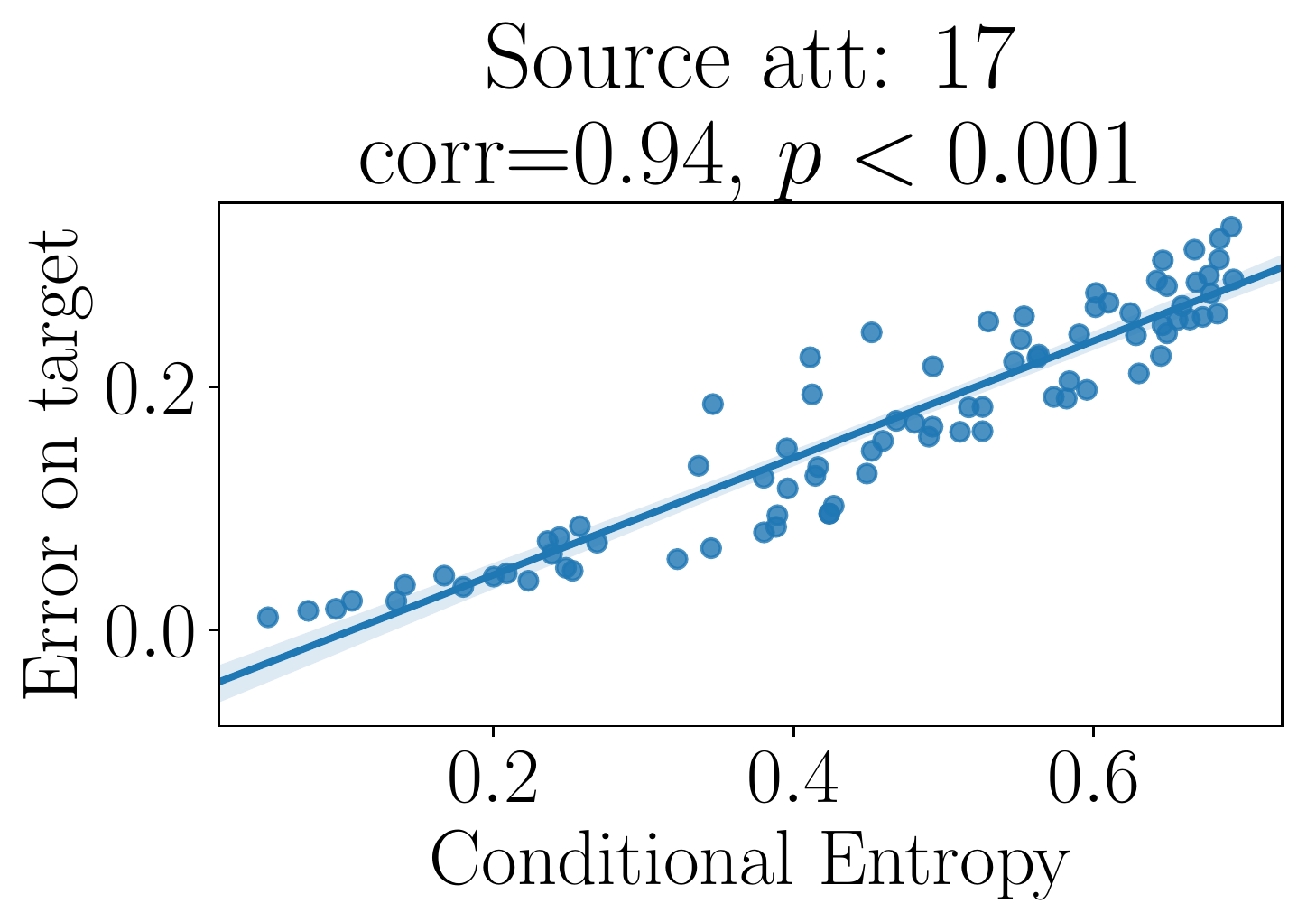}&

\includegraphics[clip, trim=0mm 0mm 0mm 13mm, width=0.19\textwidth]{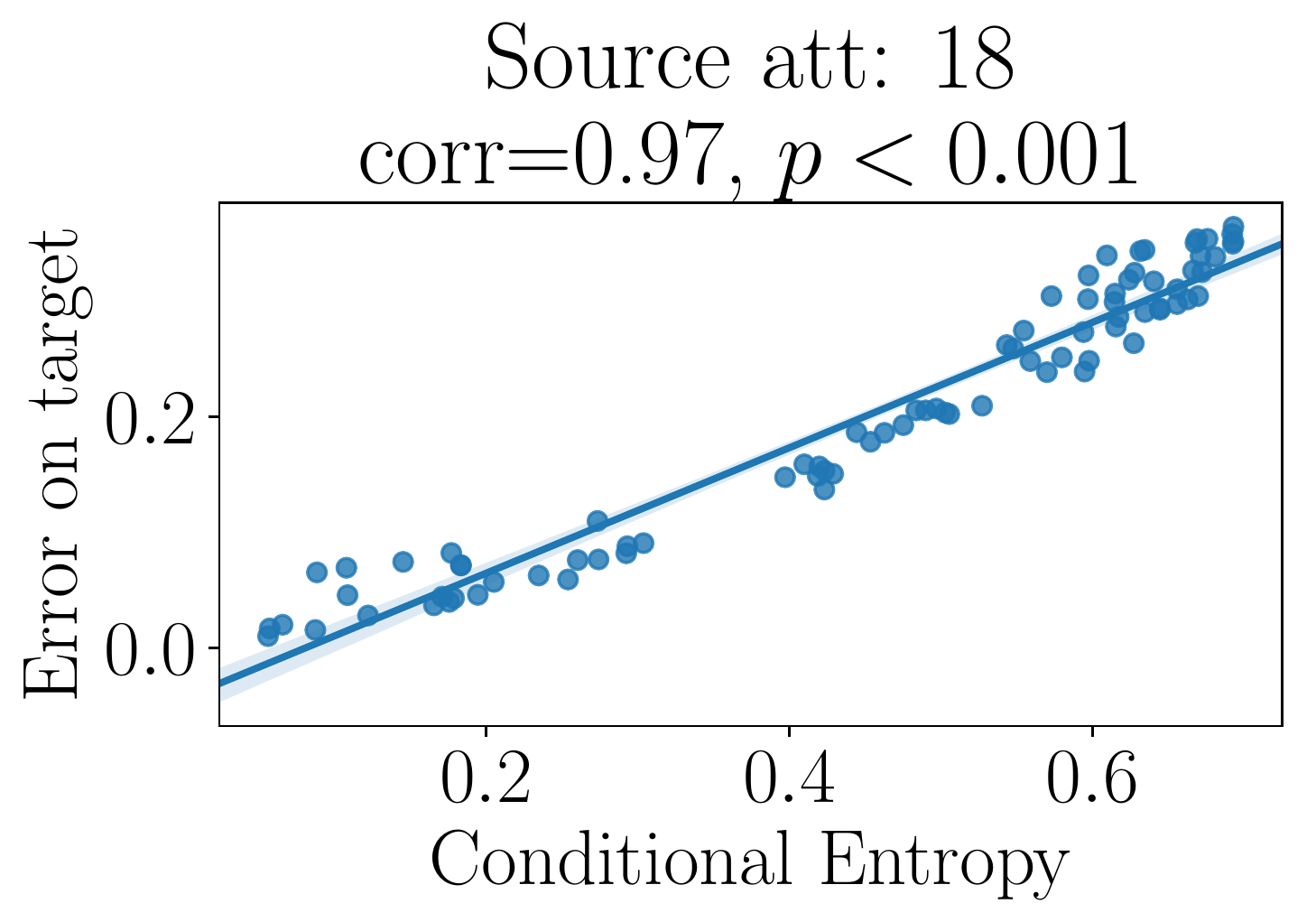}&

\includegraphics[clip, trim=0mm 0mm 0mm 13mm, width=0.19\textwidth]{figures/AWA2_att18.pdf}\\[-2pt]

(15) Small & (16) Bulbous & (17) Lean & (18) Flippers & (19) Hands\\[6pt]
\includegraphics[clip, trim=0mm 0mm 0mm 13mm, width=0.19\textwidth]{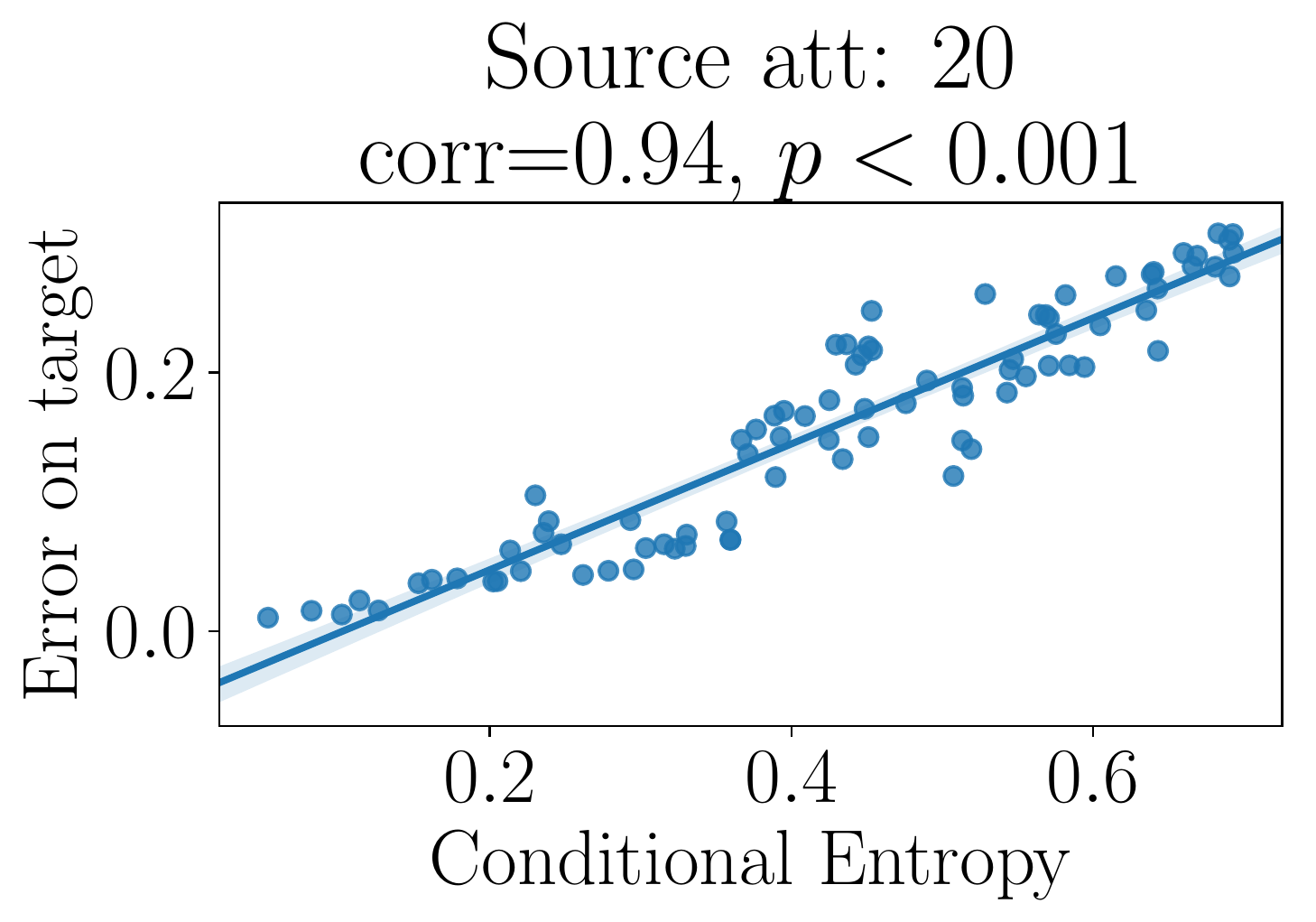}&

\includegraphics[clip, trim=0mm 0mm 0mm 13mm, width=0.19\textwidth]{figures/AWA2_att21.pdf}&

\includegraphics[clip, trim=0mm 0mm 0mm 13mm, width=0.19\textwidth]{figures/AWA2_att22.pdf}&

\includegraphics[clip, trim=0mm 0mm 0mm 13mm, width=0.19\textwidth]{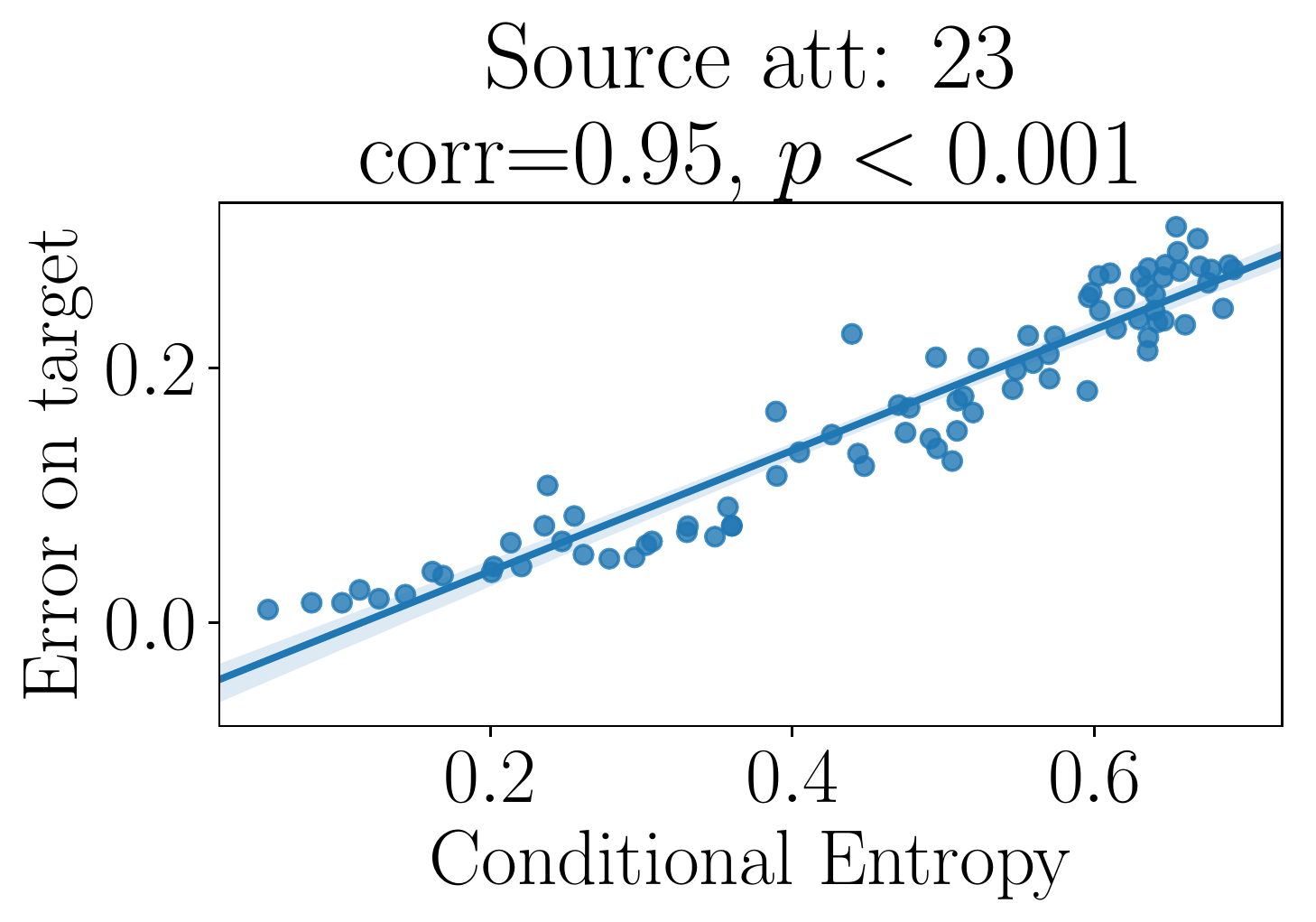}&

\includegraphics[clip, trim=0mm 0mm 0mm 13mm, width=0.19\textwidth]{figures/AWA2_att23.pdf}\\[-2pt]

(20) Hooves & (21) Pads & (22) Paws & (23) Longleg & (24) Longneck\\[6pt]
\includegraphics[clip, trim=0mm 0mm 0mm 13mm, width=0.19\textwidth]{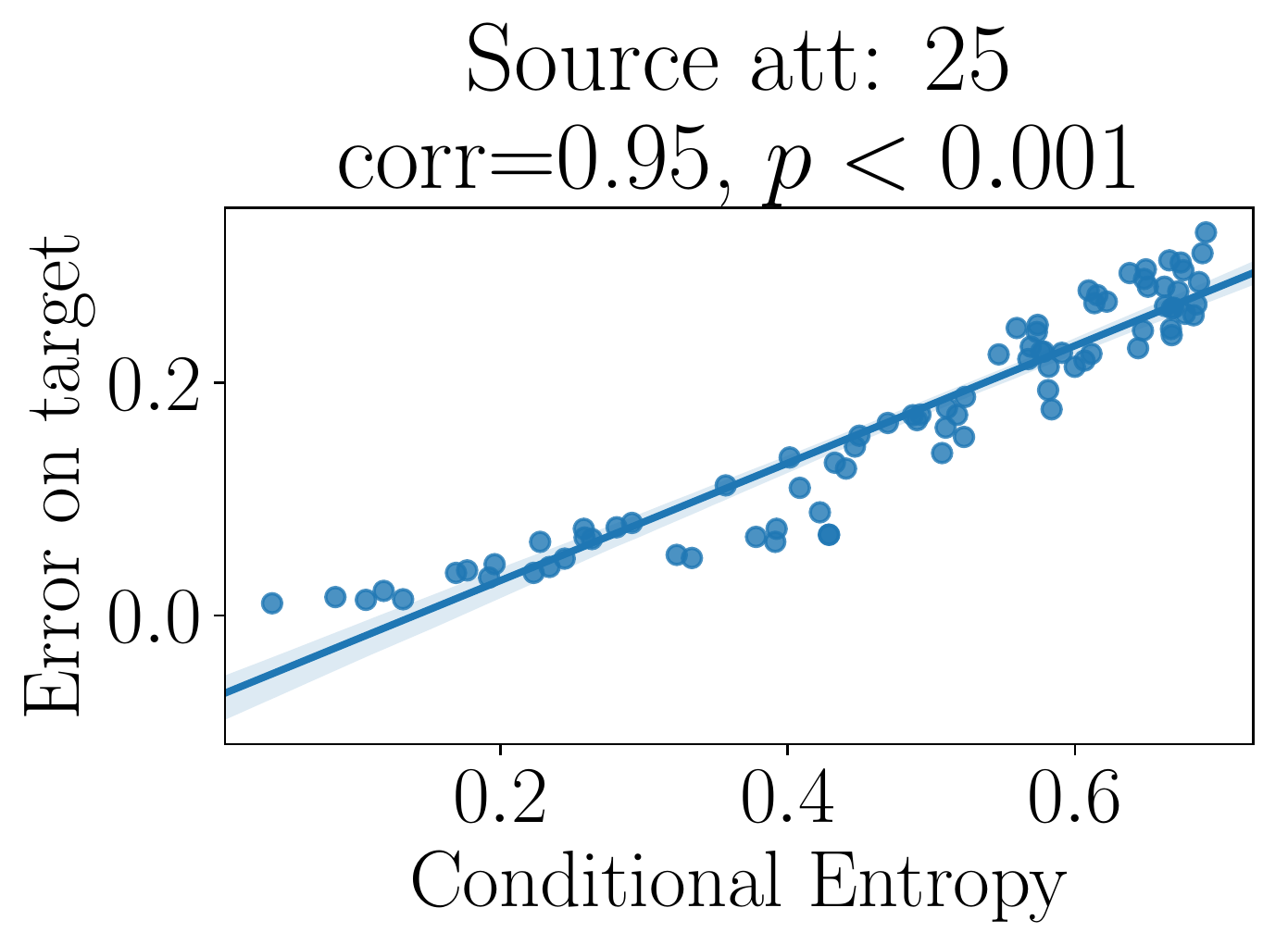}&

\includegraphics[clip, trim=0mm 0mm 0mm 13mm, width=0.19\textwidth]{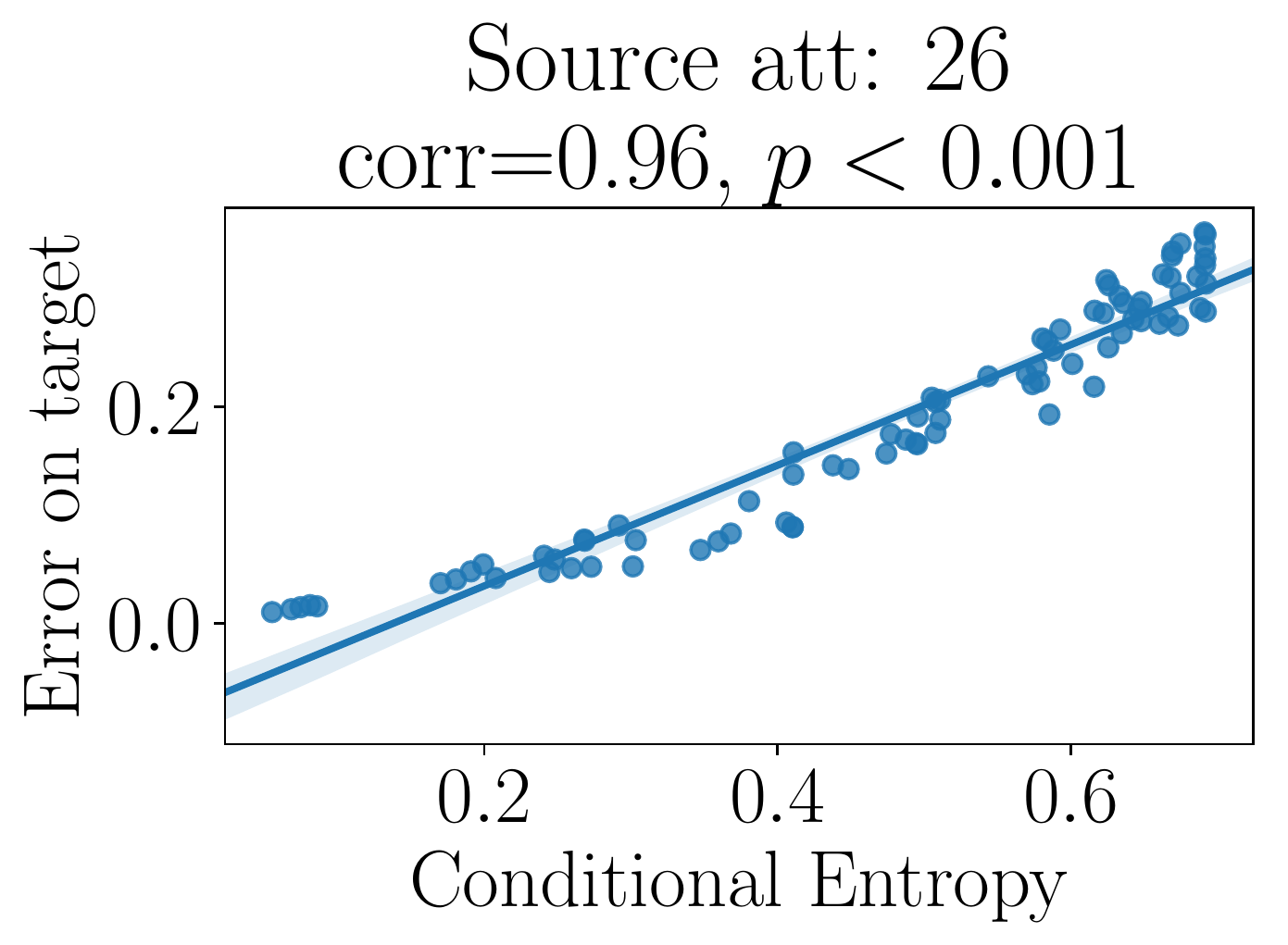}&

\includegraphics[clip, trim=0mm 0mm 0mm 13mm, width=0.19\textwidth]{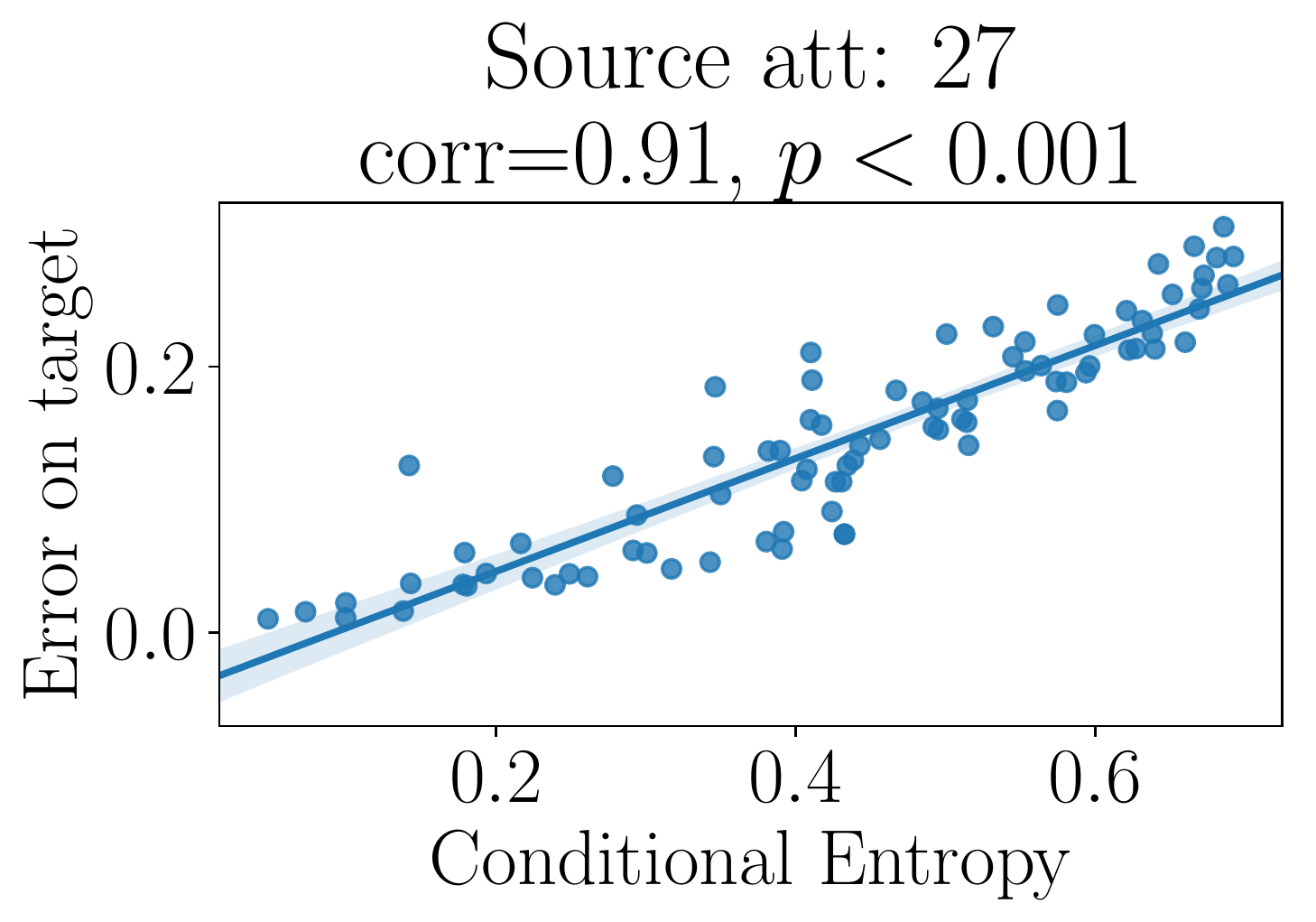}&

\includegraphics[clip, trim=0mm 0mm 0mm 13mm, width=0.19\textwidth]{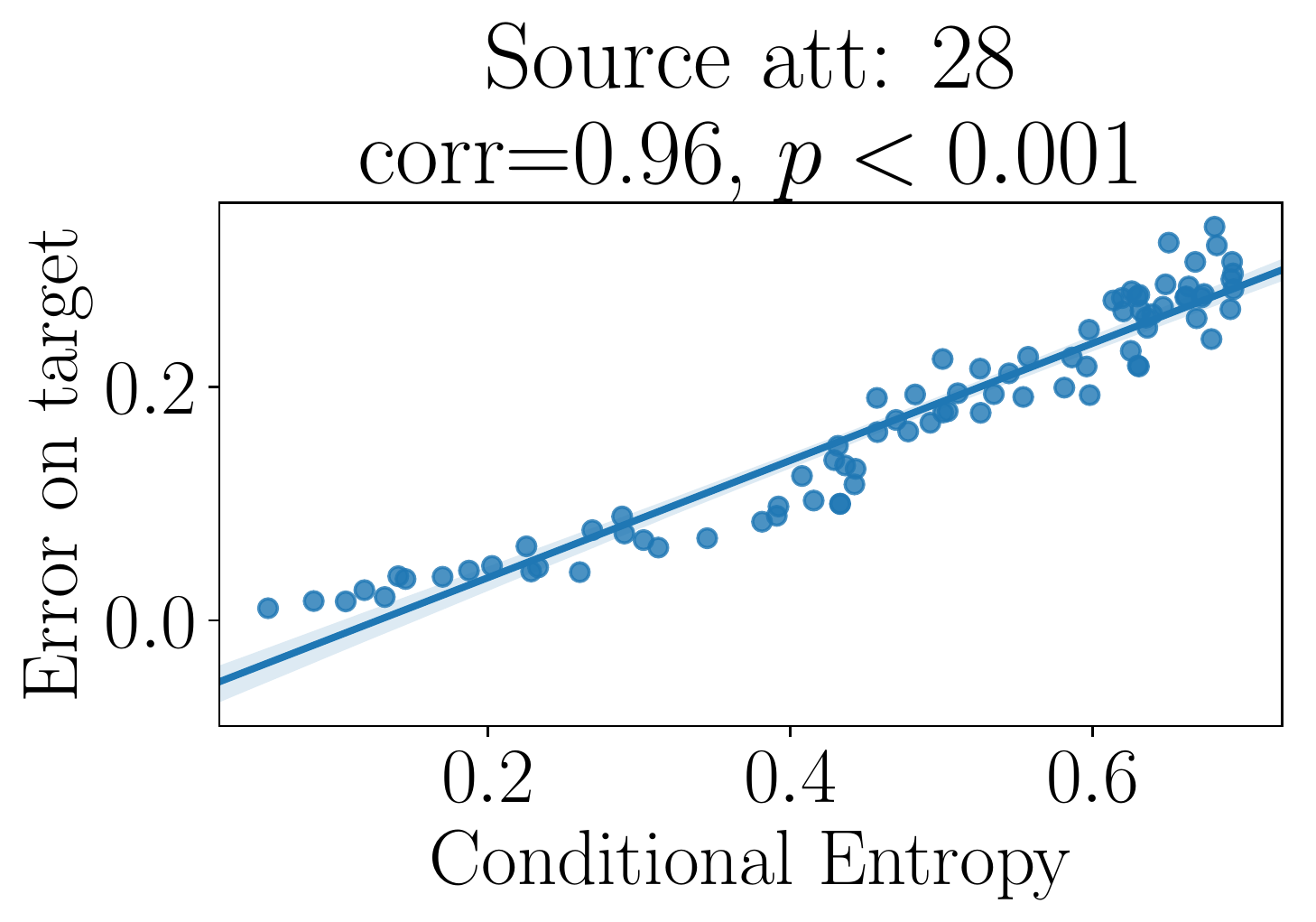}&

\includegraphics[clip, trim=0mm 0mm 0mm 13mm, width=0.19\textwidth]{figures/AWA2_att28.pdf}\\[-2pt]

(25) Tail & (26) Chewteeth & (27) Meatteeth & (28) Buckteeth & (29) Strainteeth\\[6pt]
\includegraphics[clip, trim=0mm 0mm 0mm 13mm, width=0.19\textwidth]{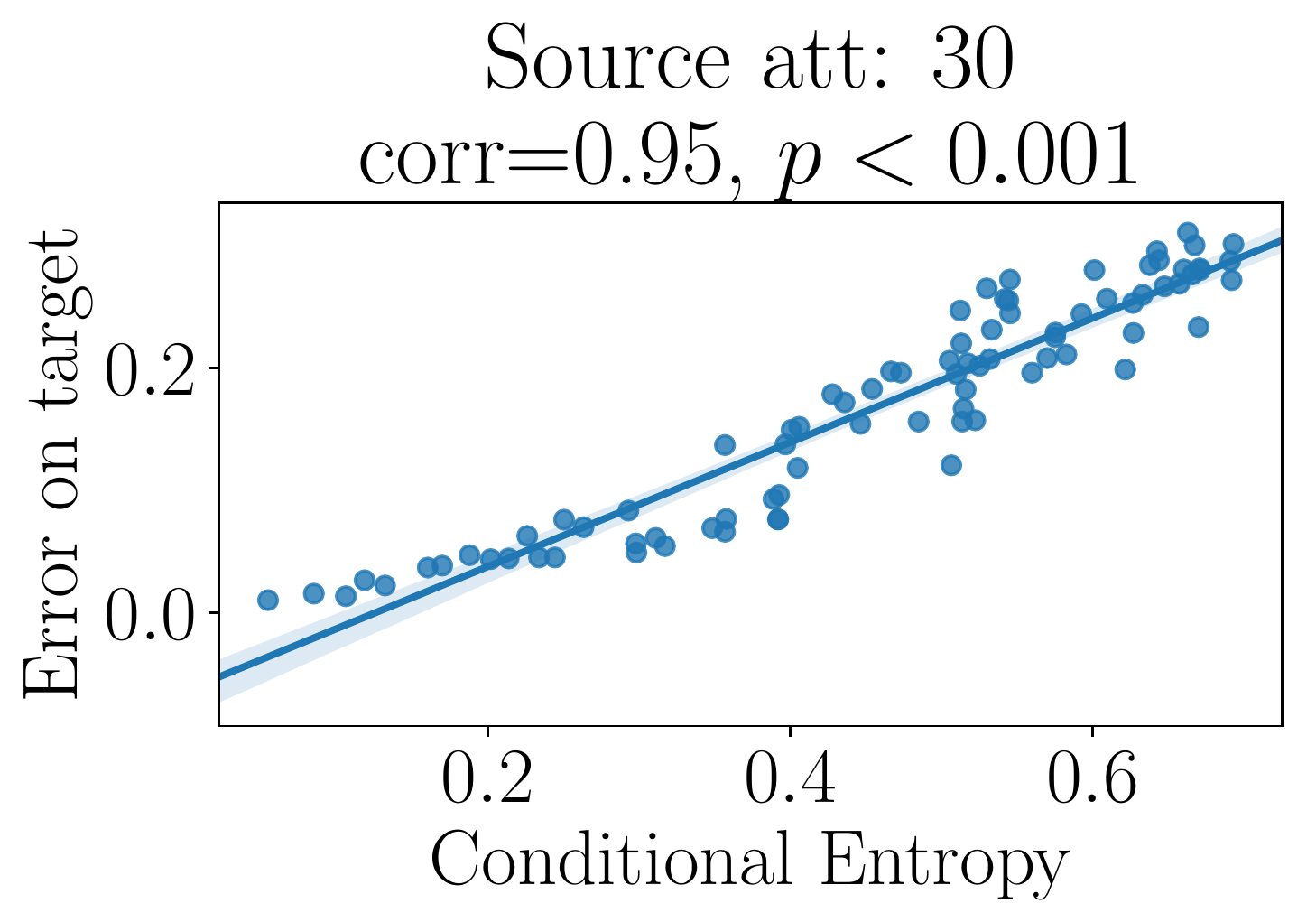}&

\includegraphics[clip, trim=0mm 0mm 0mm 13mm, width=0.19\textwidth]{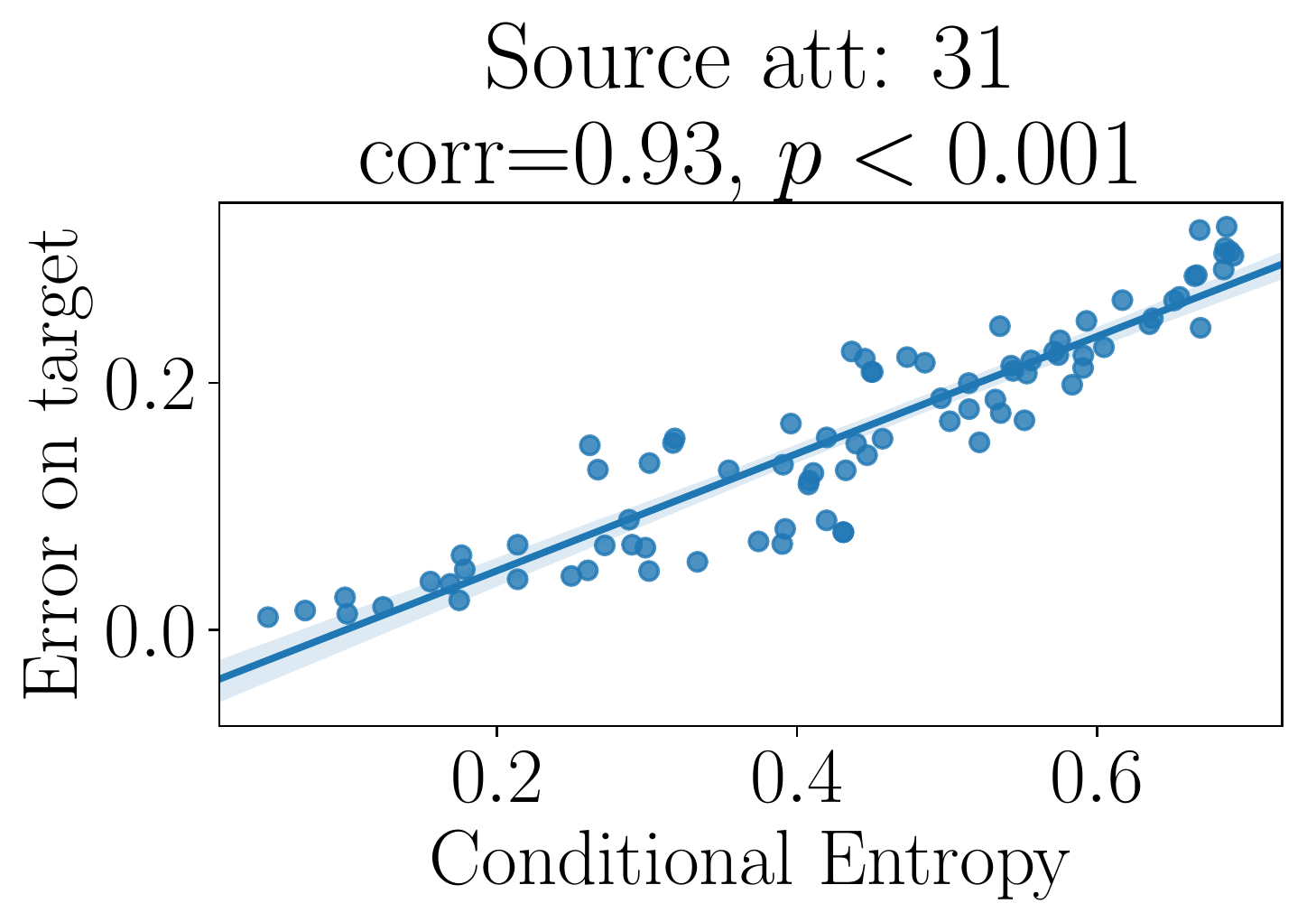}&

\includegraphics[clip, trim=0mm 0mm 0mm 13mm, width=0.19\textwidth]{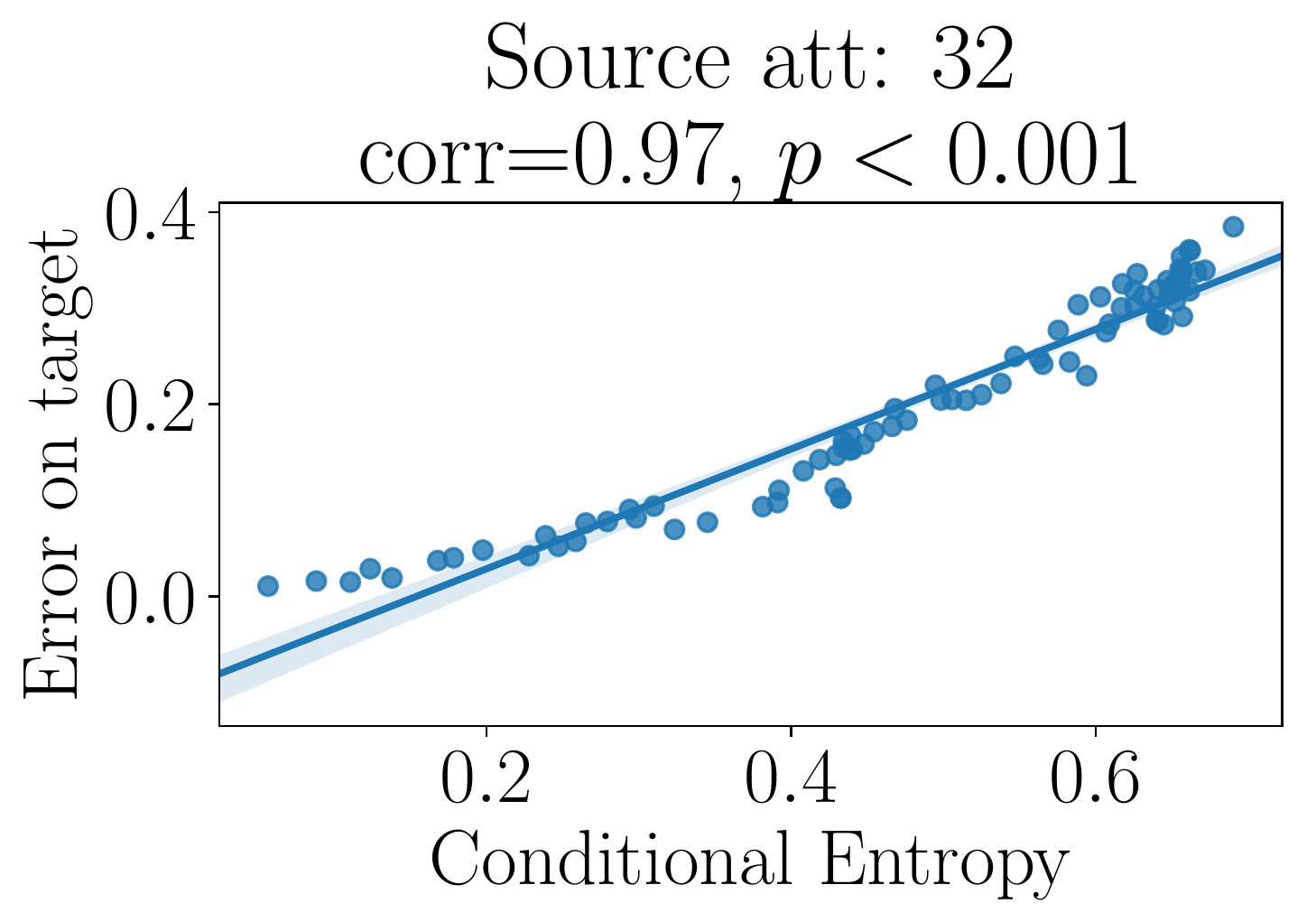}&

\includegraphics[clip, trim=0mm 0mm 0mm 13mm, width=0.19\textwidth]{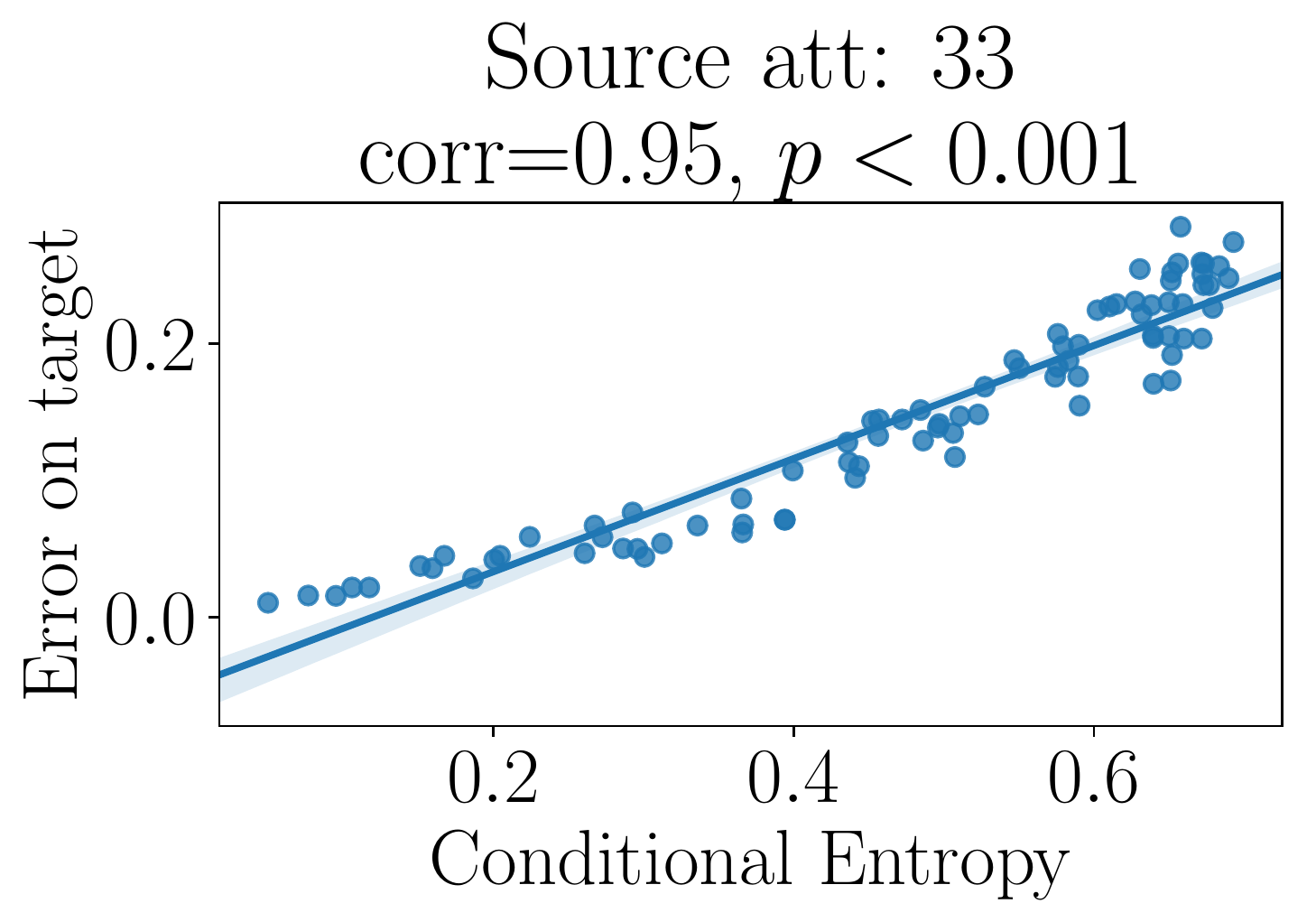}&

\includegraphics[clip, trim=0mm 0mm 0mm 13mm, width=0.19\textwidth]{figures/AWA2_att33.pdf}\\[-2pt]

(30) Horns & (31) Claws & (32) Tusks & (33) Smelly & (34) Flys\\[6pt]

    \end{tabular}
\caption{{\bf Attribute prediction; CE vs. test errors on AwA2 (Extended from Fig.~\ref{fig:face_att_trans}(e-h) in the paper; part 1).} The source attribute, $T^Z$, in each plot is named in the plot title. Points represent different target tasks $T^Y$. Corr is the Pearson correlation coefficient between the two variables and $p$ is the statistical significance of the correlation. In all cases, the correlation is statistically significant.}
\label{fig:trans_awa1}
\end{figure*}

\begin{figure*}[ht]
\centering
\footnotesize
\begin{tabular}{c@{~}c@{~}c@{~}c@{~}c}
\includegraphics[clip, trim=0mm 0mm 0mm 13mm, width=0.19\textwidth]{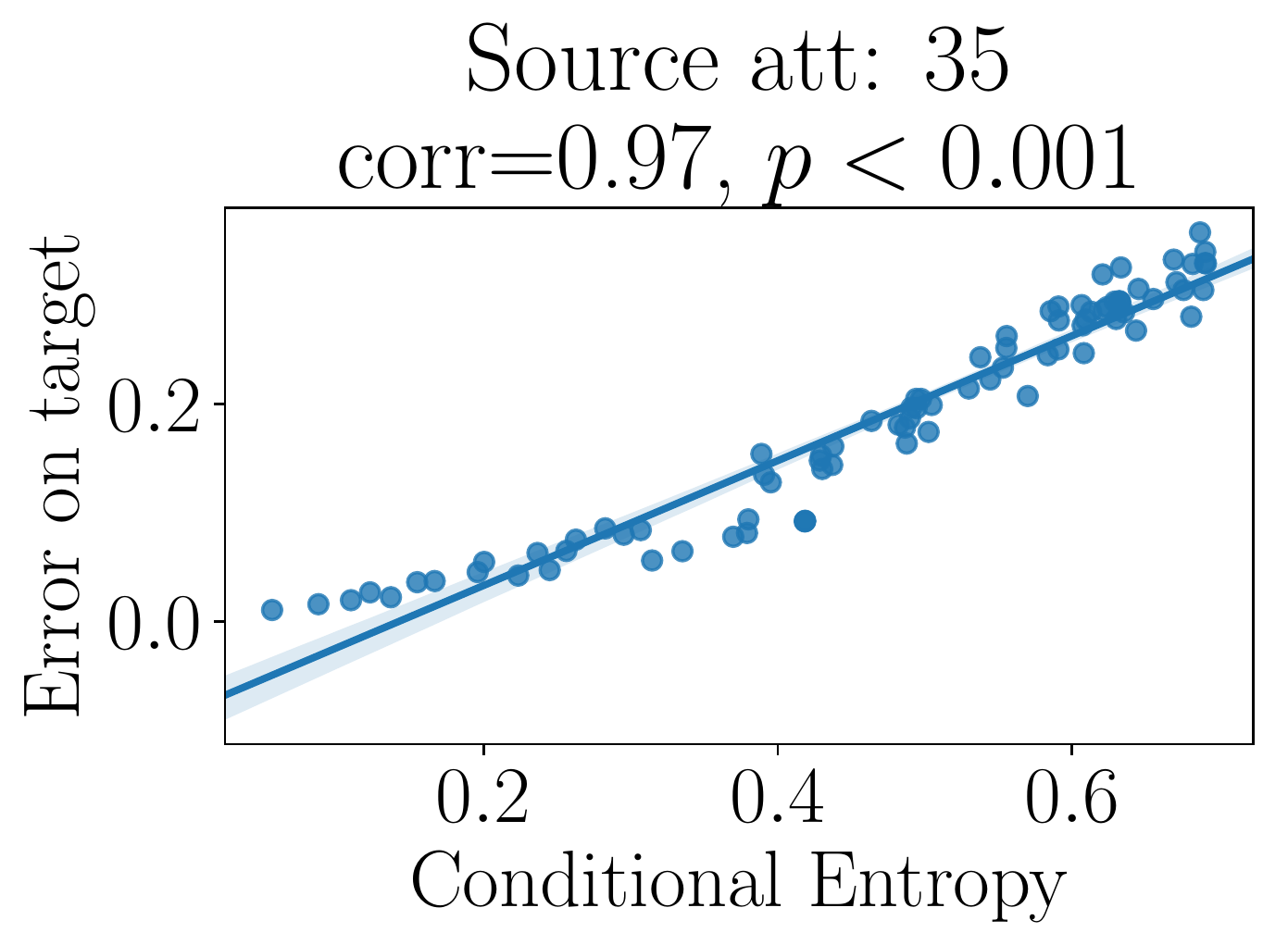}&

\includegraphics[clip, trim=0mm 0mm 0mm 13mm, width=0.19\textwidth]{figures/AWA2_att36.pdf}&

\includegraphics[clip, trim=0mm 0mm 0mm 13mm, width=0.19\textwidth]{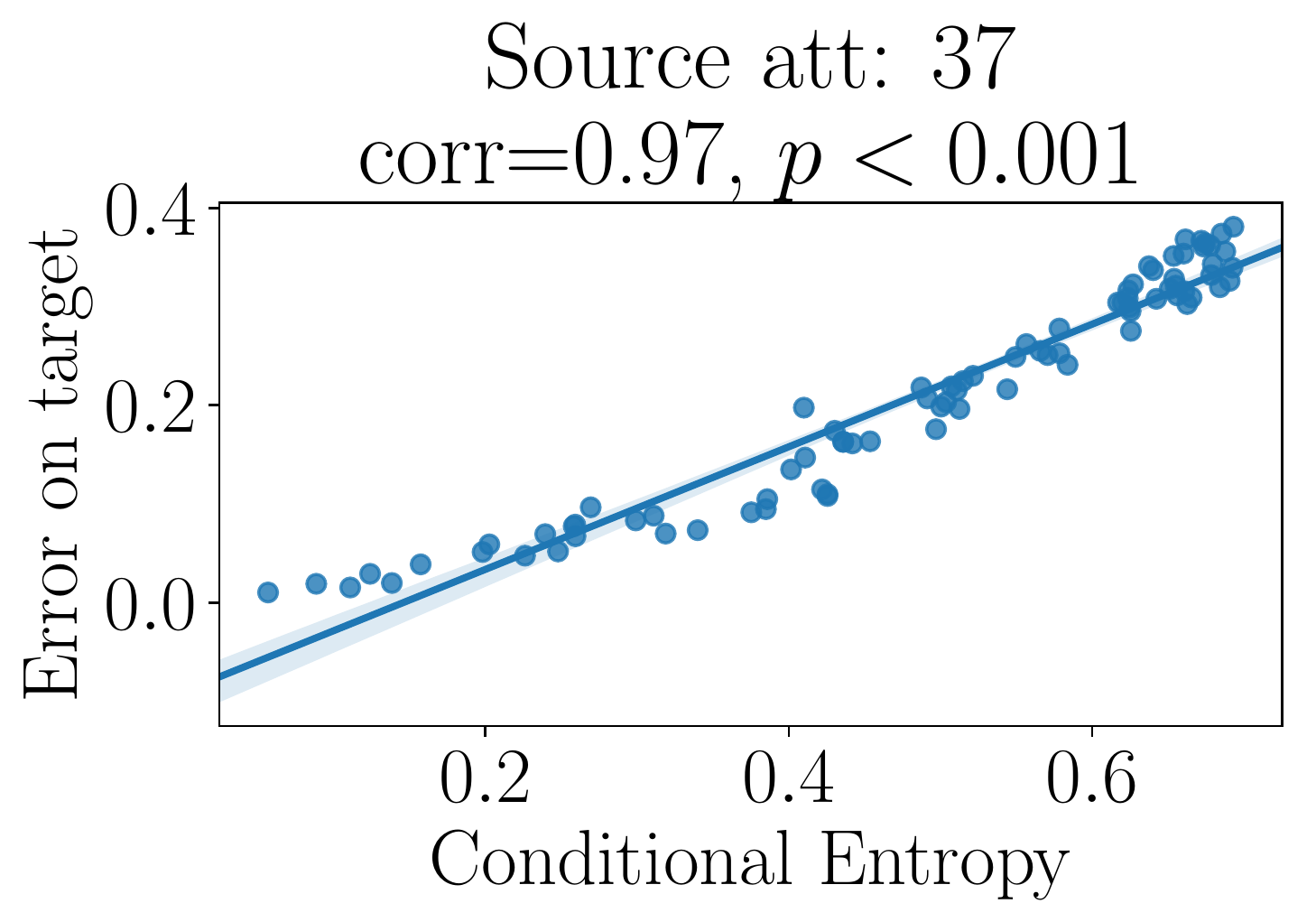}&

\includegraphics[clip, trim=0mm 0mm 0mm 13mm, width=0.19\textwidth]{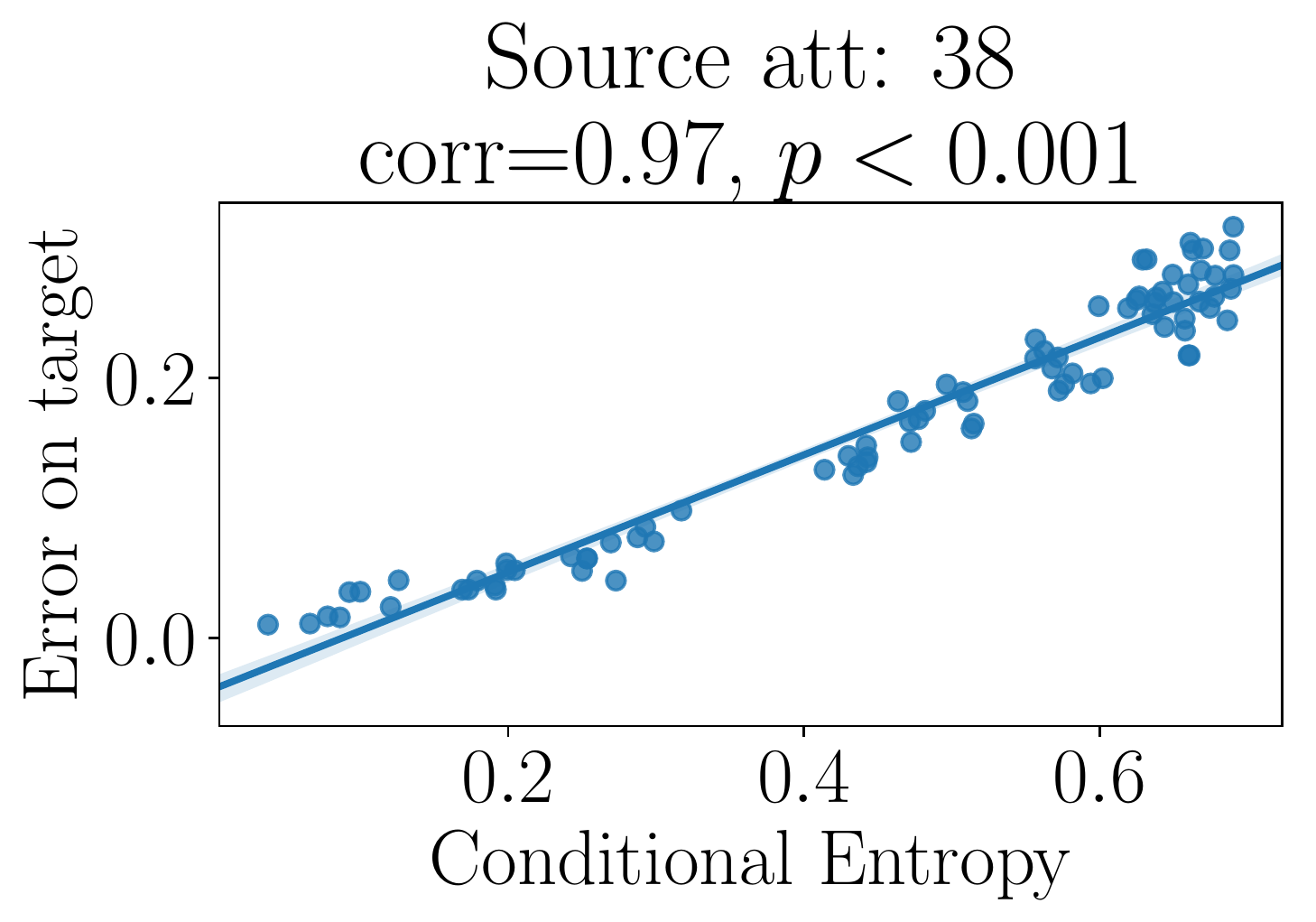}&

\includegraphics[clip, trim=0mm 0mm 0mm 13mm, width=0.19\textwidth]{figures/AWA2_att38.pdf}\\[-2pt]

(35) Hops & (36) Swims & (37) Tunnels & (38) Walks & (39) Fast\\

\includegraphics[clip, trim=0mm 0mm 0mm 13mm, width=0.19\textwidth]{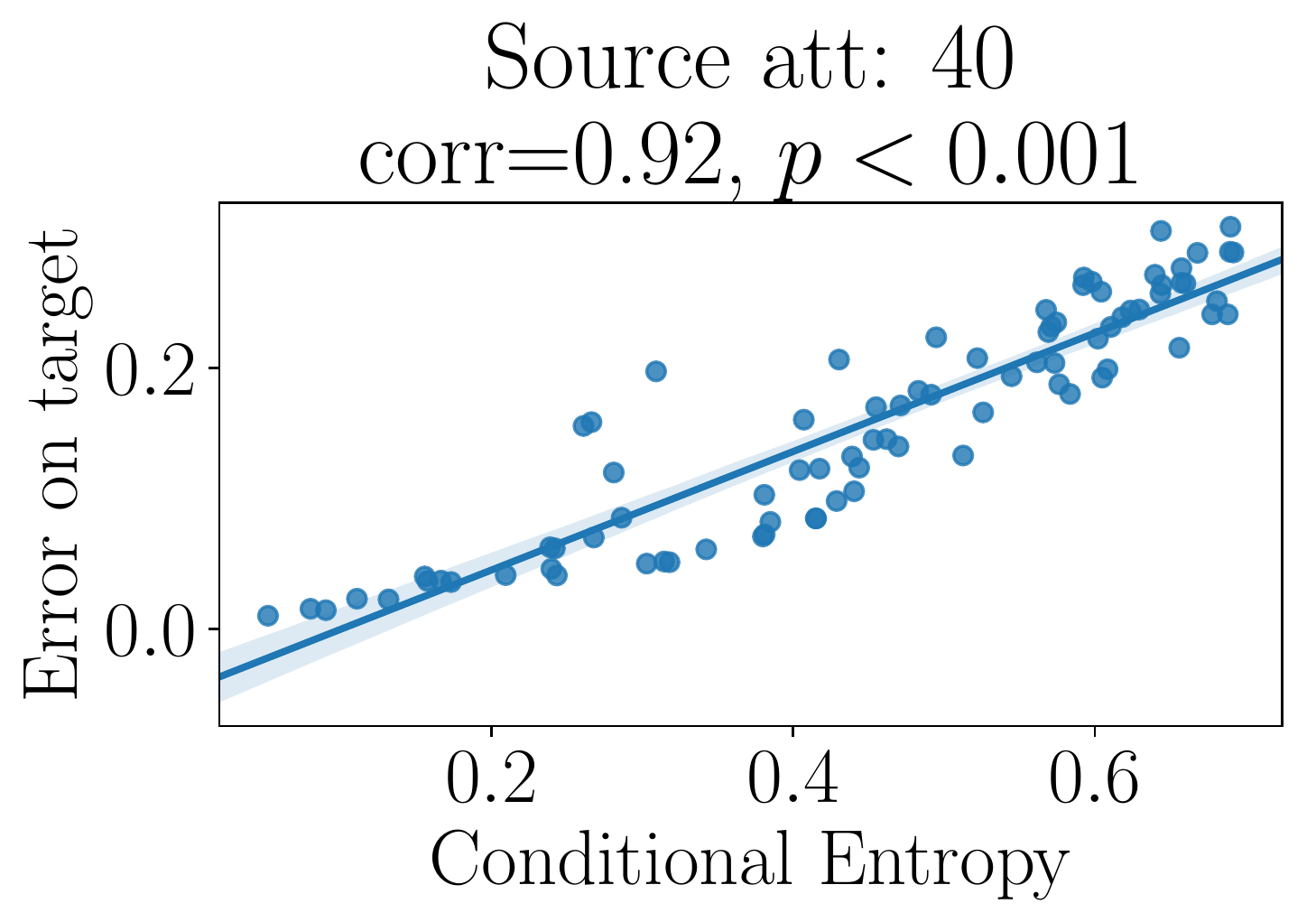}&

\includegraphics[clip, trim=0mm 0mm 0mm 13mm, width=0.19\textwidth]{figures/AWA2_att41.pdf}&

\includegraphics[clip, trim=0mm 0mm 0mm 13mm, width=0.19\textwidth]{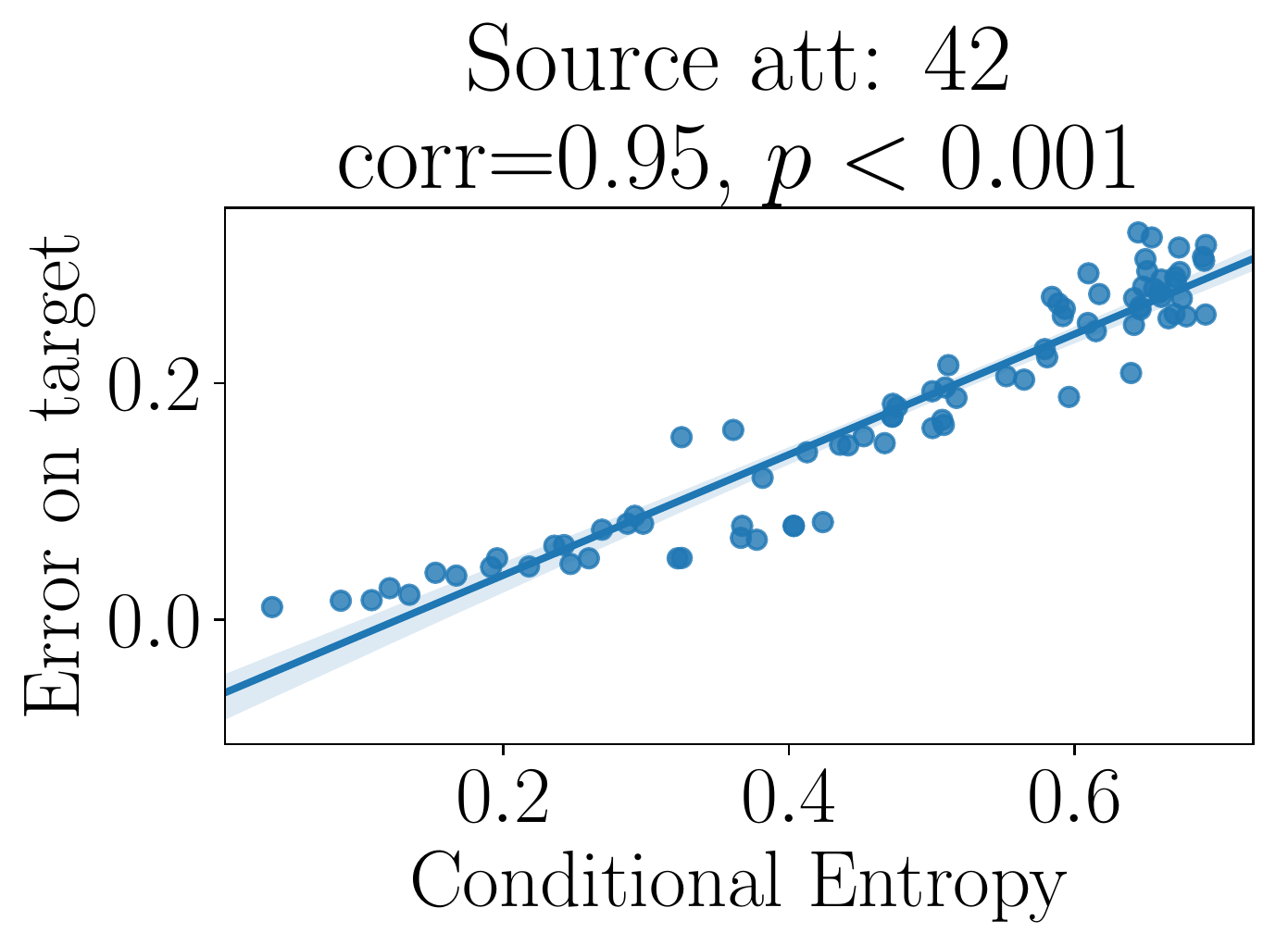}&

\includegraphics[clip, trim=0mm 0mm 0mm 13mm, width=0.19\textwidth]{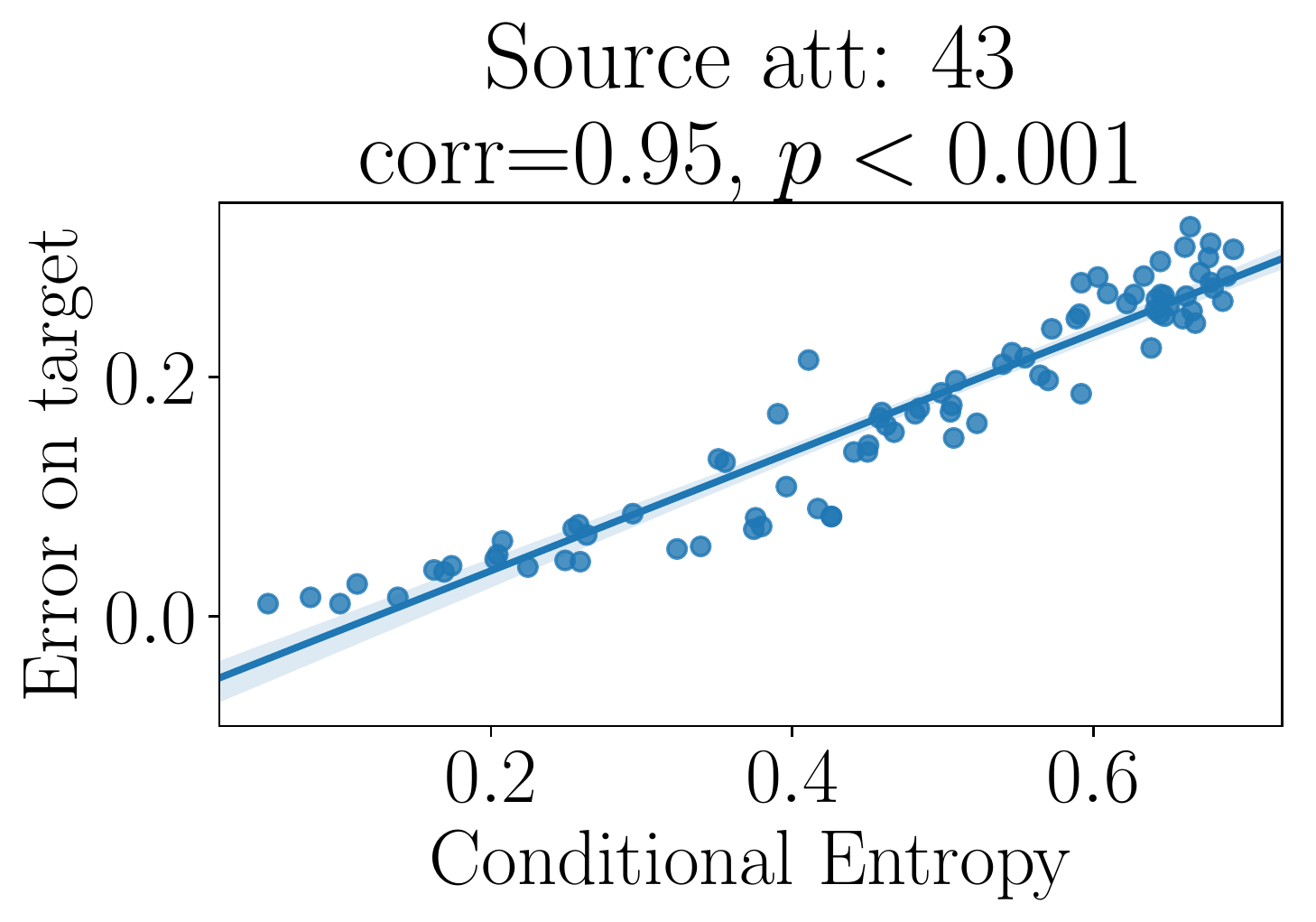}&

\includegraphics[clip, trim=0mm 0mm 0mm 13mm, width=0.19\textwidth]{figures/AWA2_att43.pdf}\\[-2pt]

(40) Slow & (41) Strong & (42) Weak & (43) Muscle & (44) Bipedal\\[6pt]
\includegraphics[clip, trim=0mm 0mm 0mm 13mm, width=0.19\textwidth]{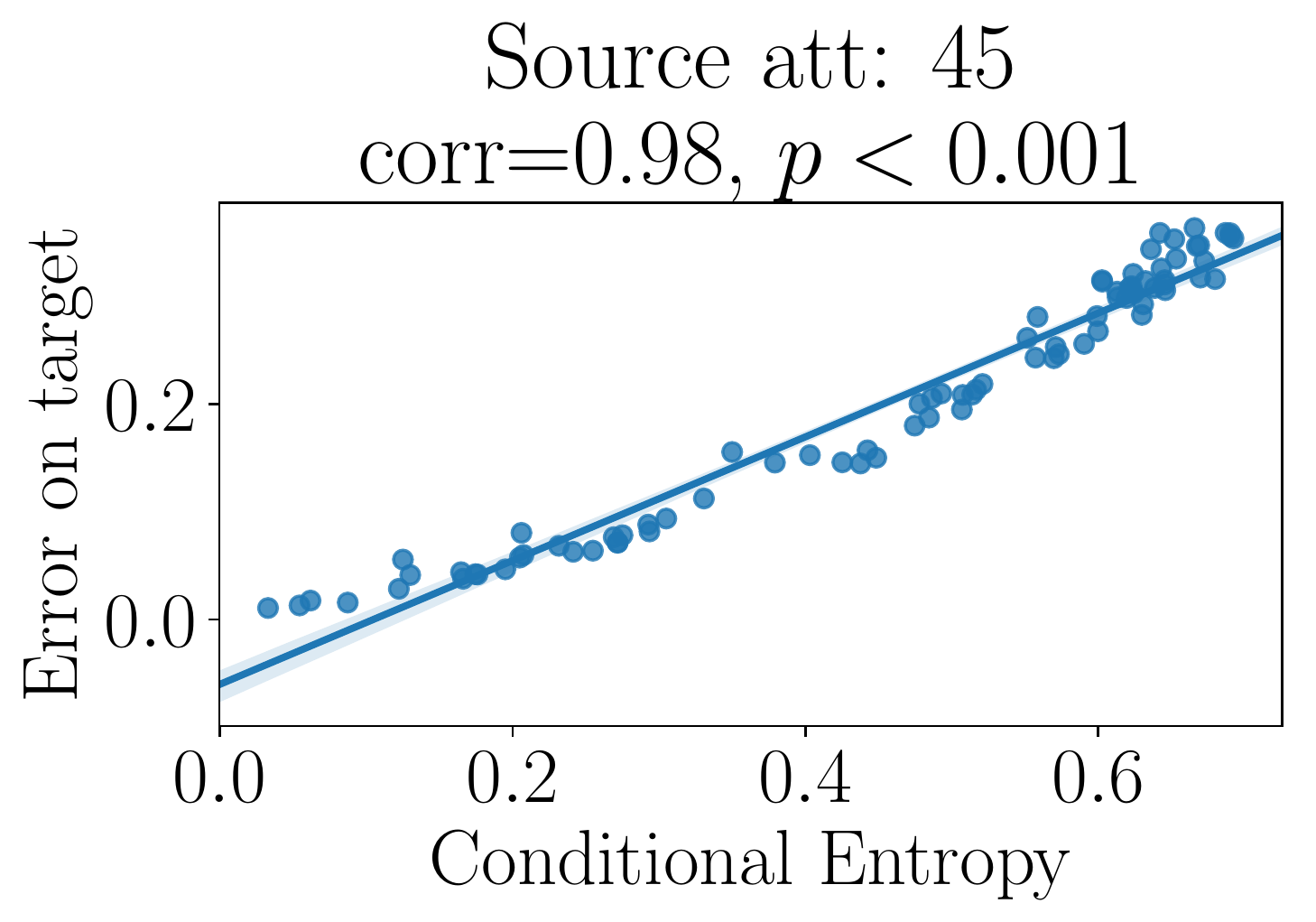}&

\includegraphics[clip, trim=0mm 0mm 0mm 13mm, width=0.19\textwidth]{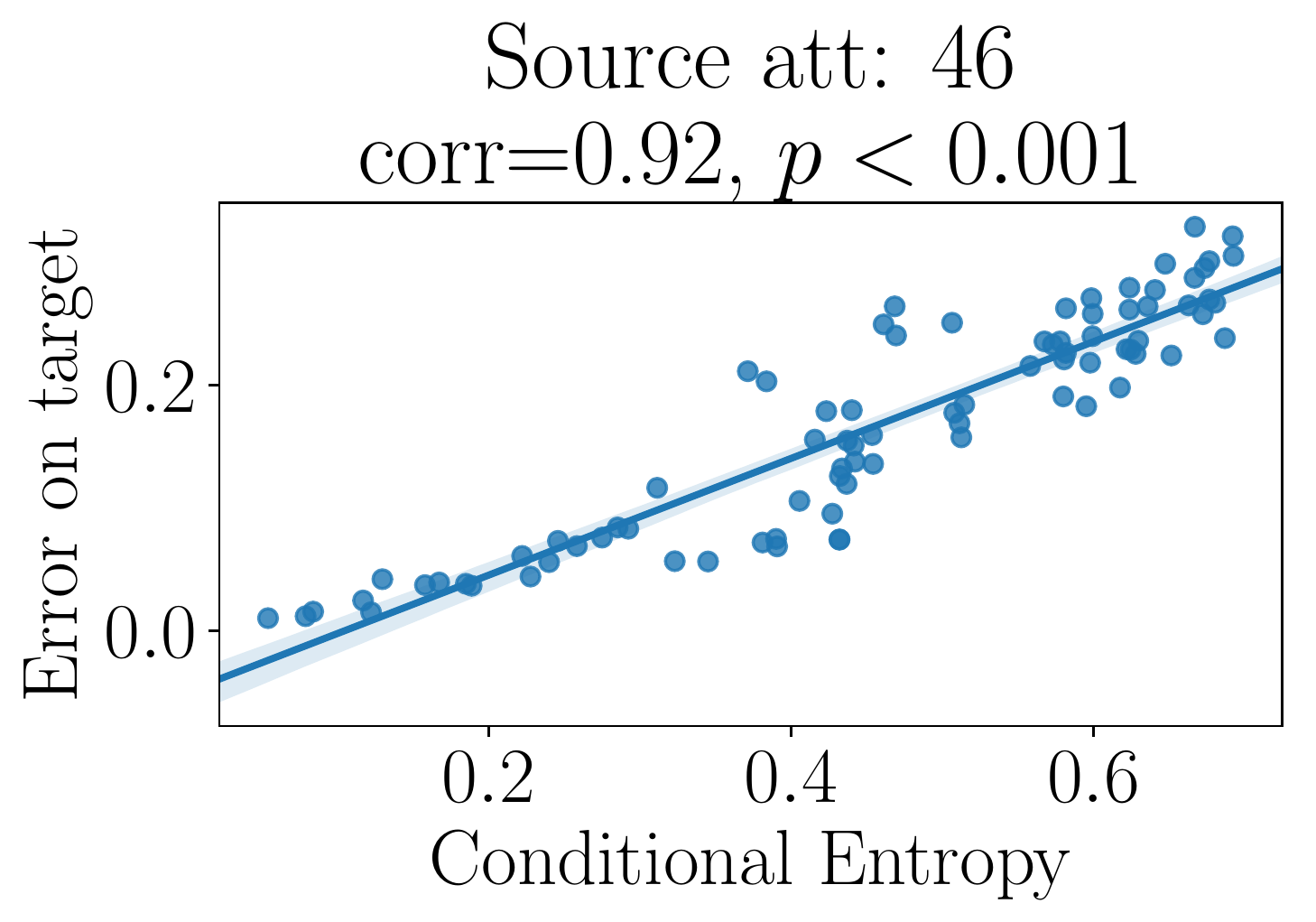}&

\includegraphics[clip, trim=0mm 0mm 0mm 13mm, width=0.19\textwidth]{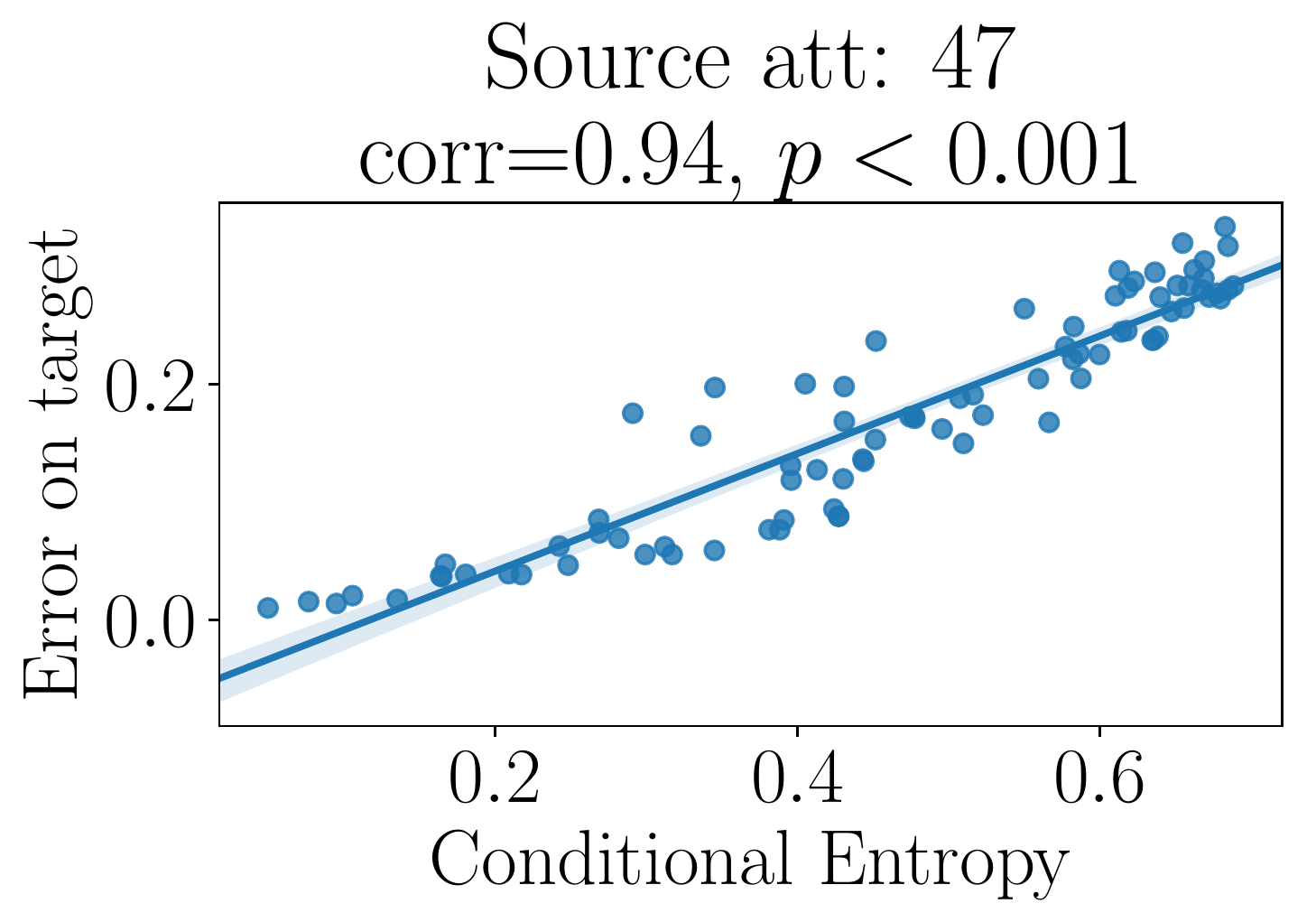}&

\includegraphics[clip, trim=0mm 0mm 0mm 13mm, width=0.19\textwidth]{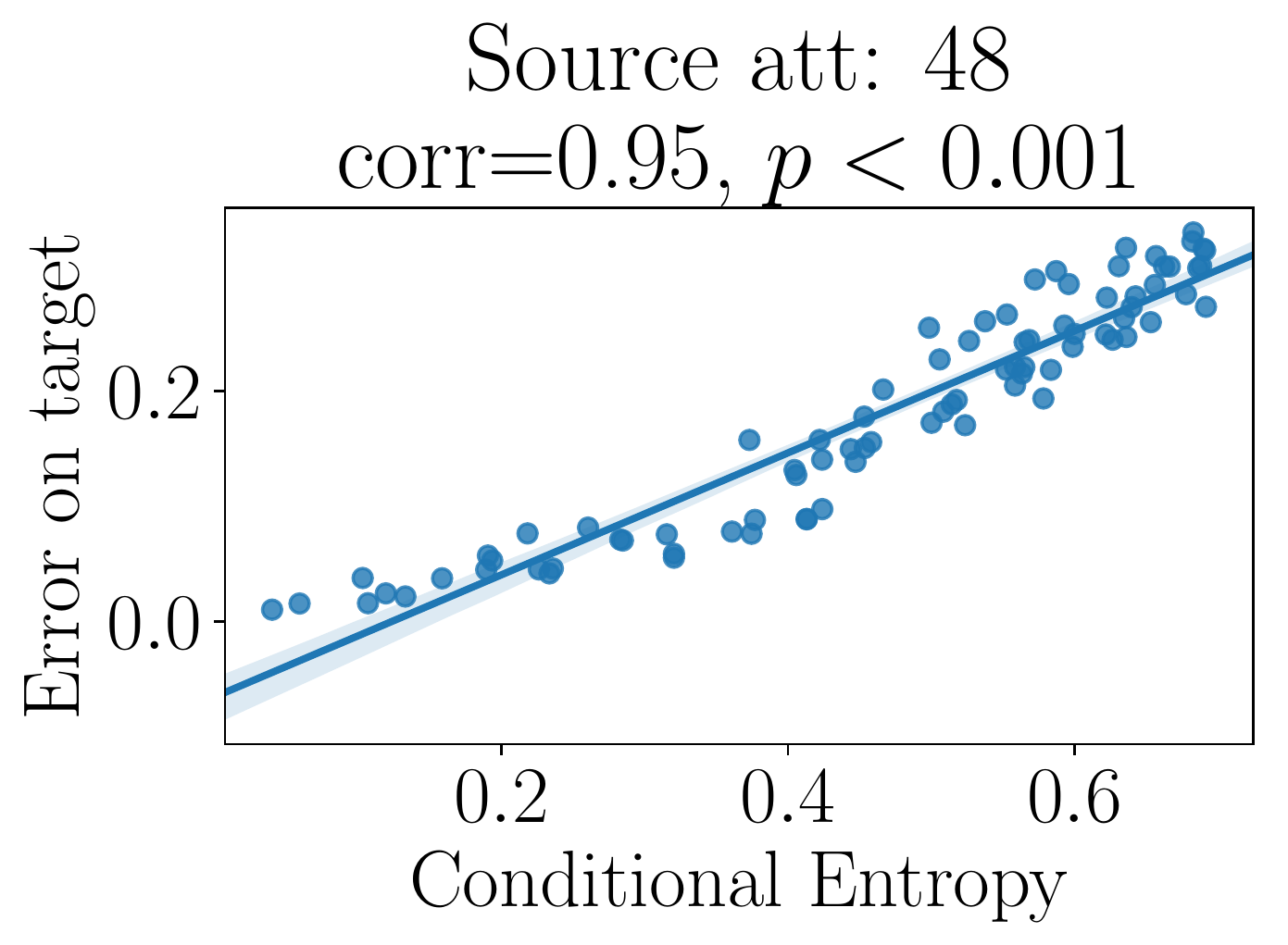}&

\includegraphics[clip, trim=0mm 0mm 0mm 13mm, width=0.19\textwidth]{figures/AWA2_att48.pdf}\\[-2pt]

(45) Quadrapedal & (46) Active & (47) Inactive & (48) Nocturnal & (49) Hibernate\\[6pt]
\includegraphics[clip, trim=0mm 0mm 0mm 13mm, width=0.19\textwidth]{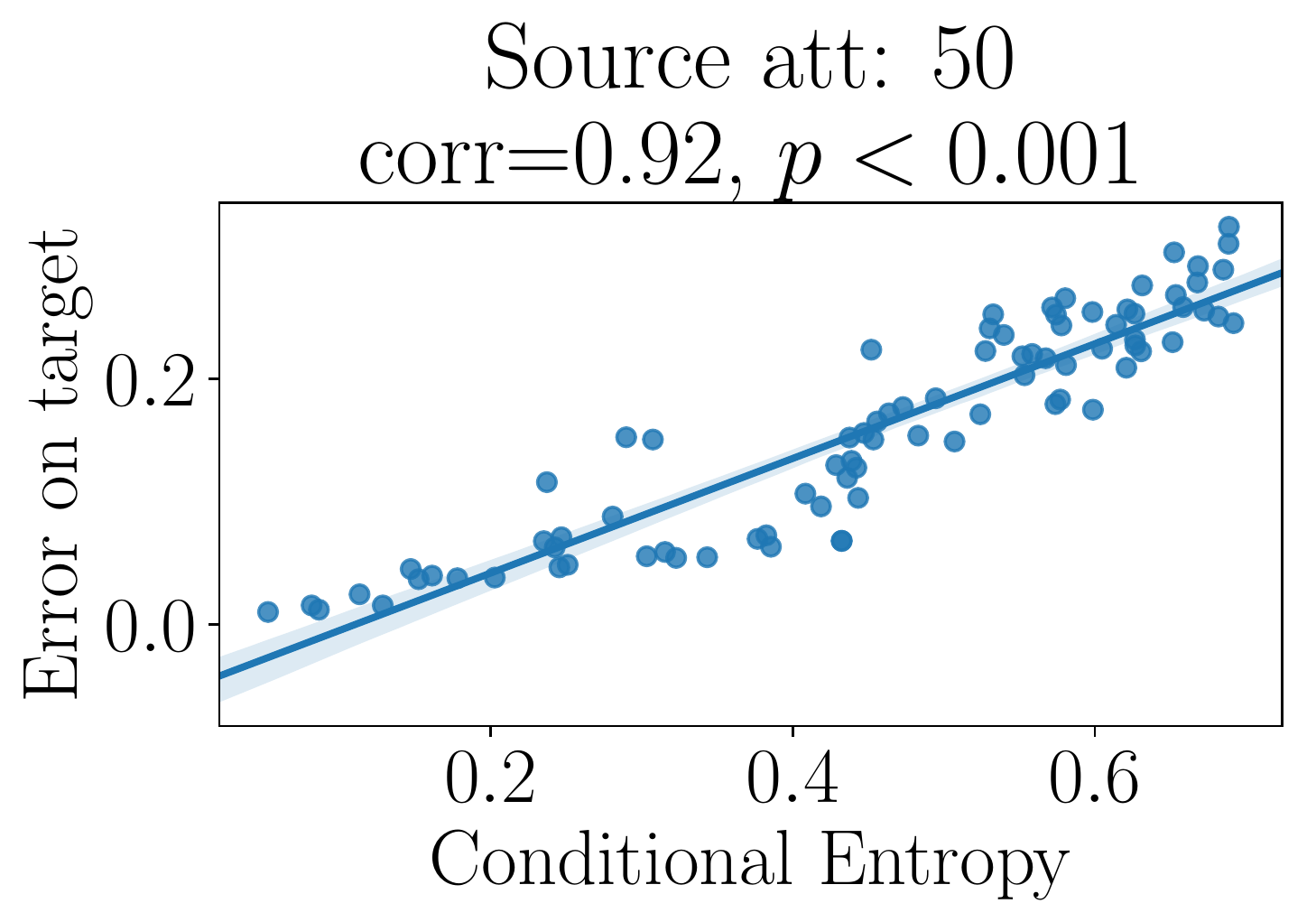}&

\includegraphics[clip, trim=0mm 0mm 0mm 13mm, width=0.19\textwidth]{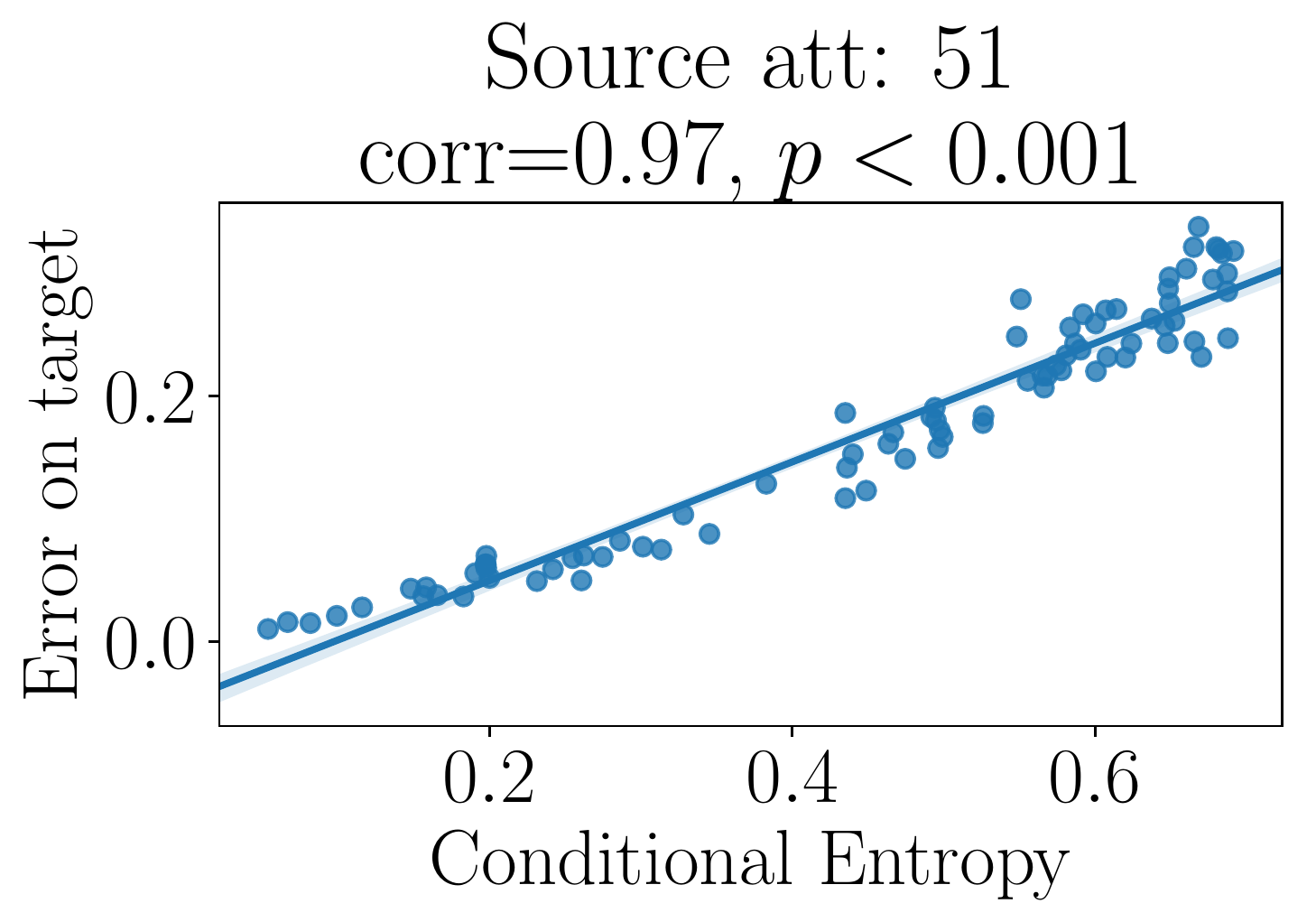}&

\includegraphics[clip, trim=0mm 0mm 0mm 13mm, width=0.19\textwidth]{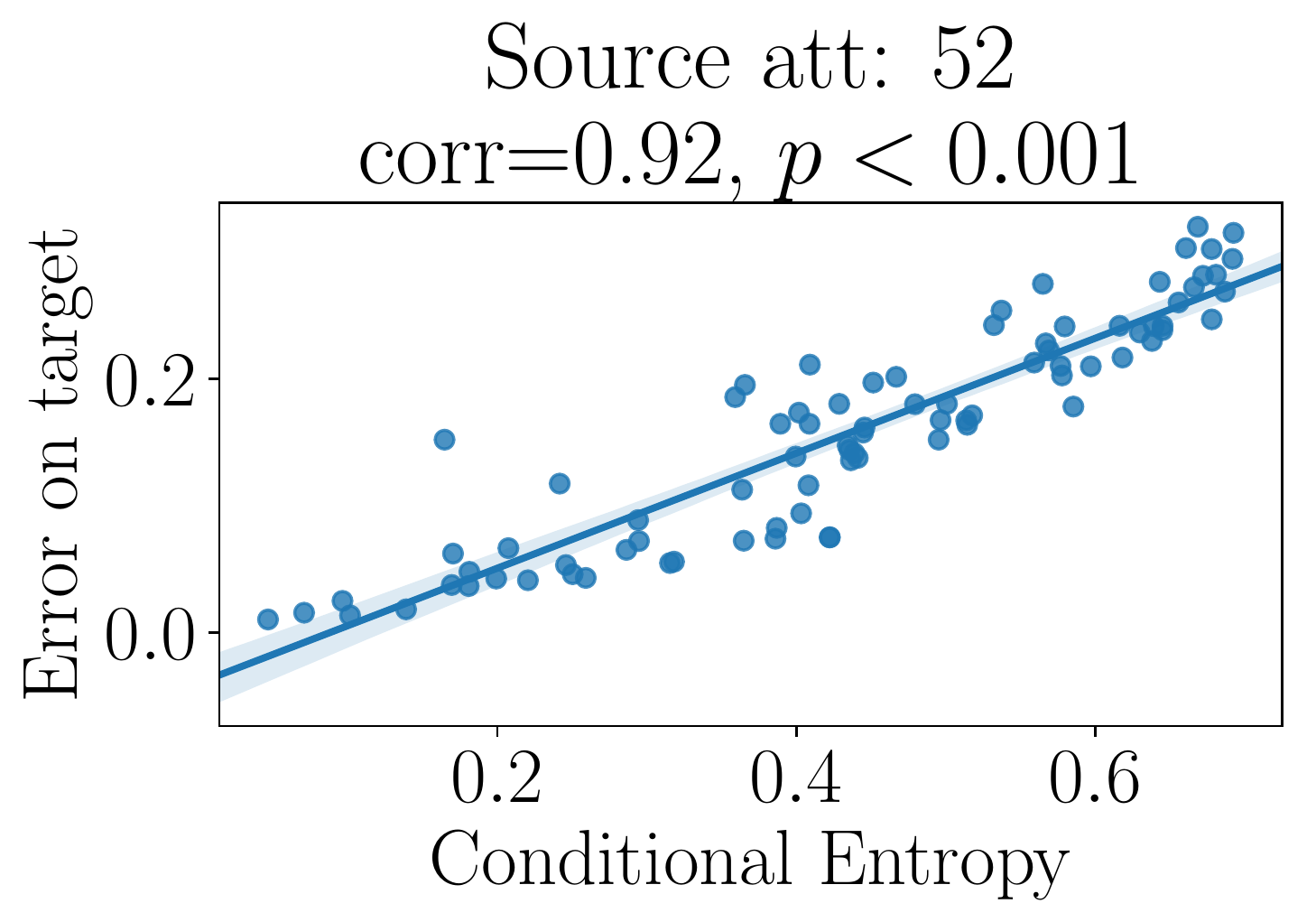}&

\includegraphics[clip, trim=0mm 0mm 0mm 13mm, width=0.19\textwidth]{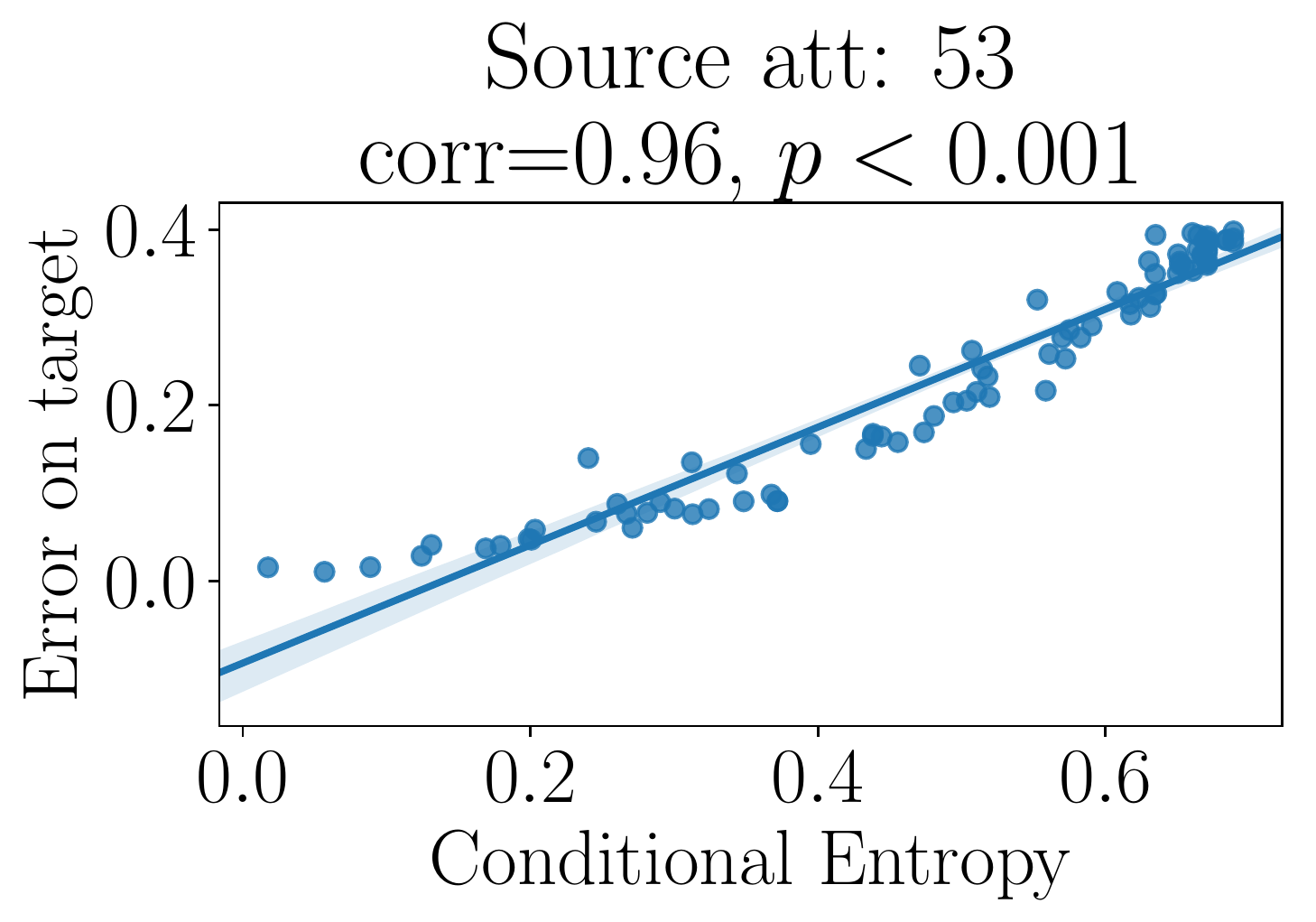}&

\includegraphics[clip, trim=0mm 0mm 0mm 13mm, width=0.19\textwidth]{figures/AWA2_att53.pdf}\\[-2pt]

(50) Agility & (51) Fish & (52) Meat & (53) Plankton & (54) Vegetation\\[6pt]
\includegraphics[clip, trim=0mm 0mm 0mm 13mm, width=0.19\textwidth]{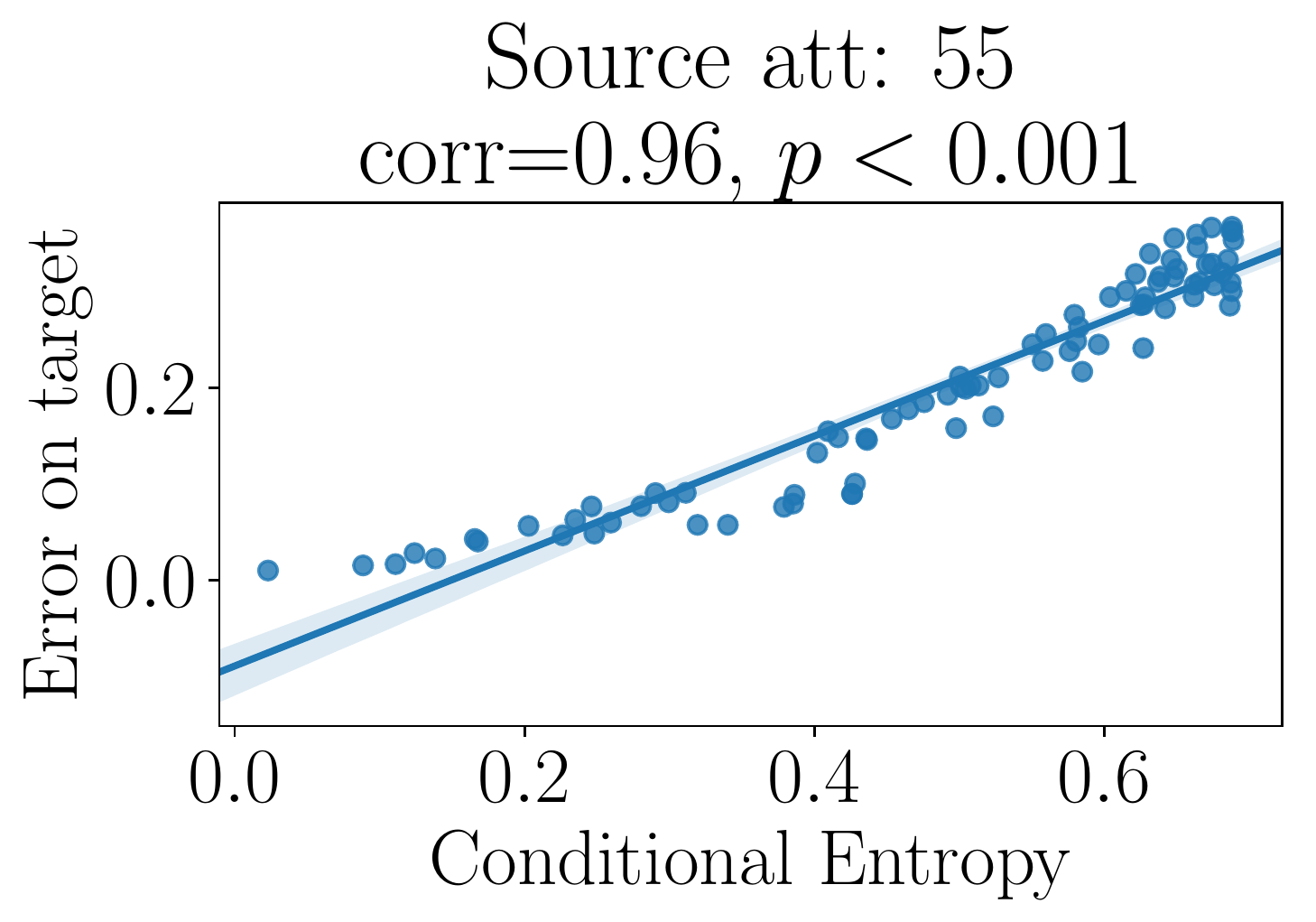}&

\includegraphics[clip, trim=0mm 0mm 0mm 13mm, width=0.19\textwidth]{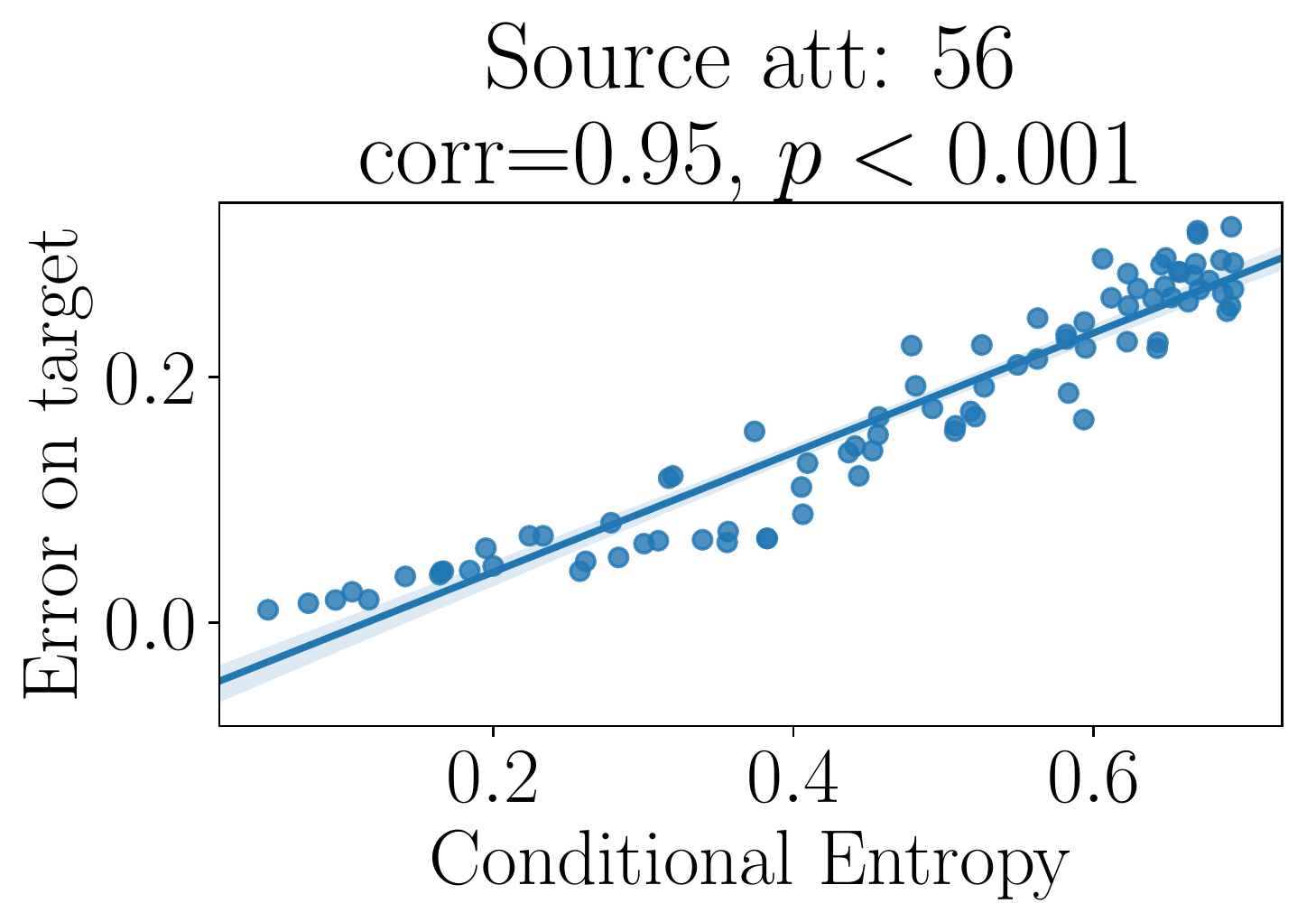}&

\includegraphics[clip, trim=0mm 0mm 0mm 13mm, width=0.19\textwidth]{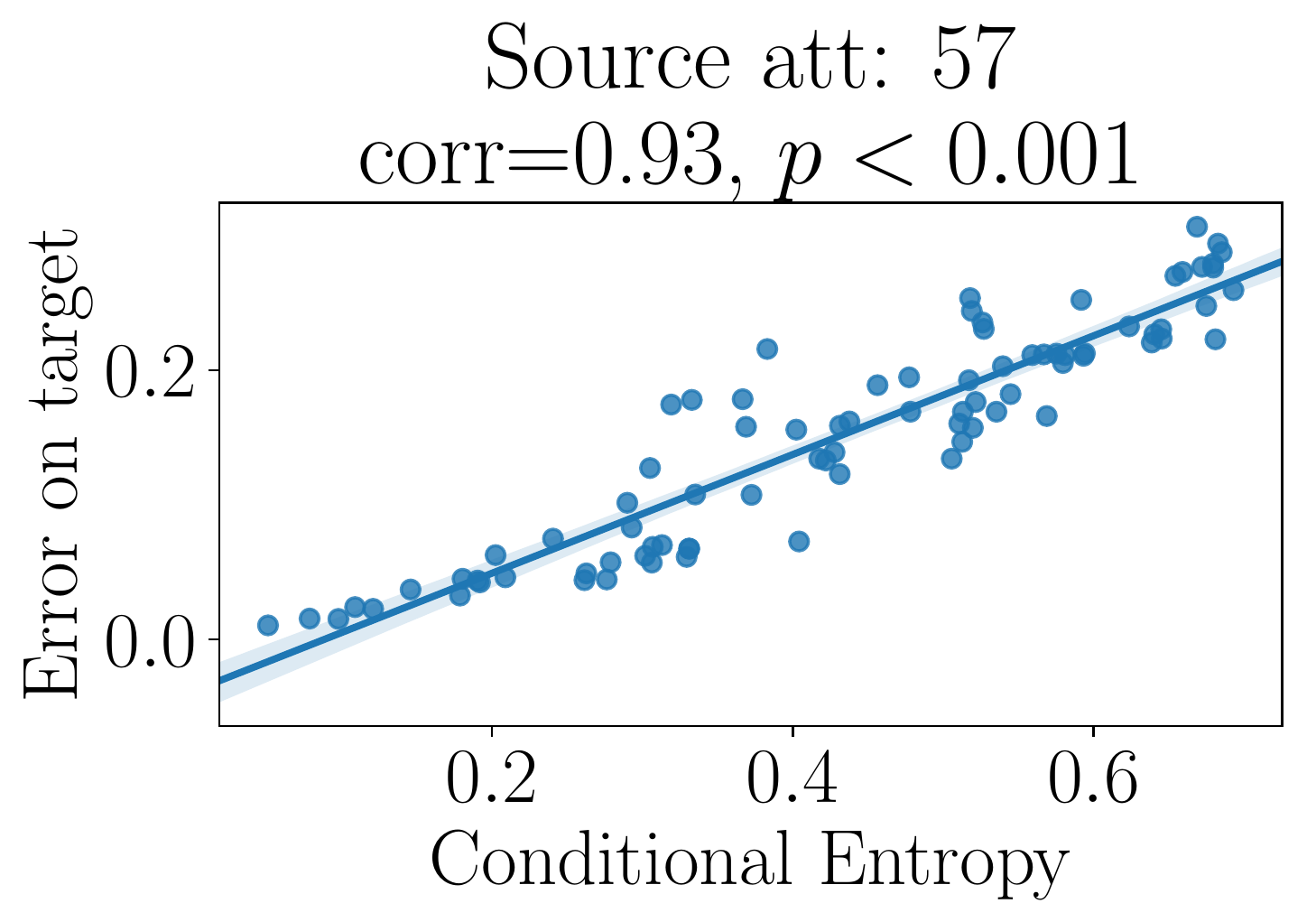}&

\includegraphics[clip, trim=0mm 0mm 0mm 13mm, width=0.19\textwidth]{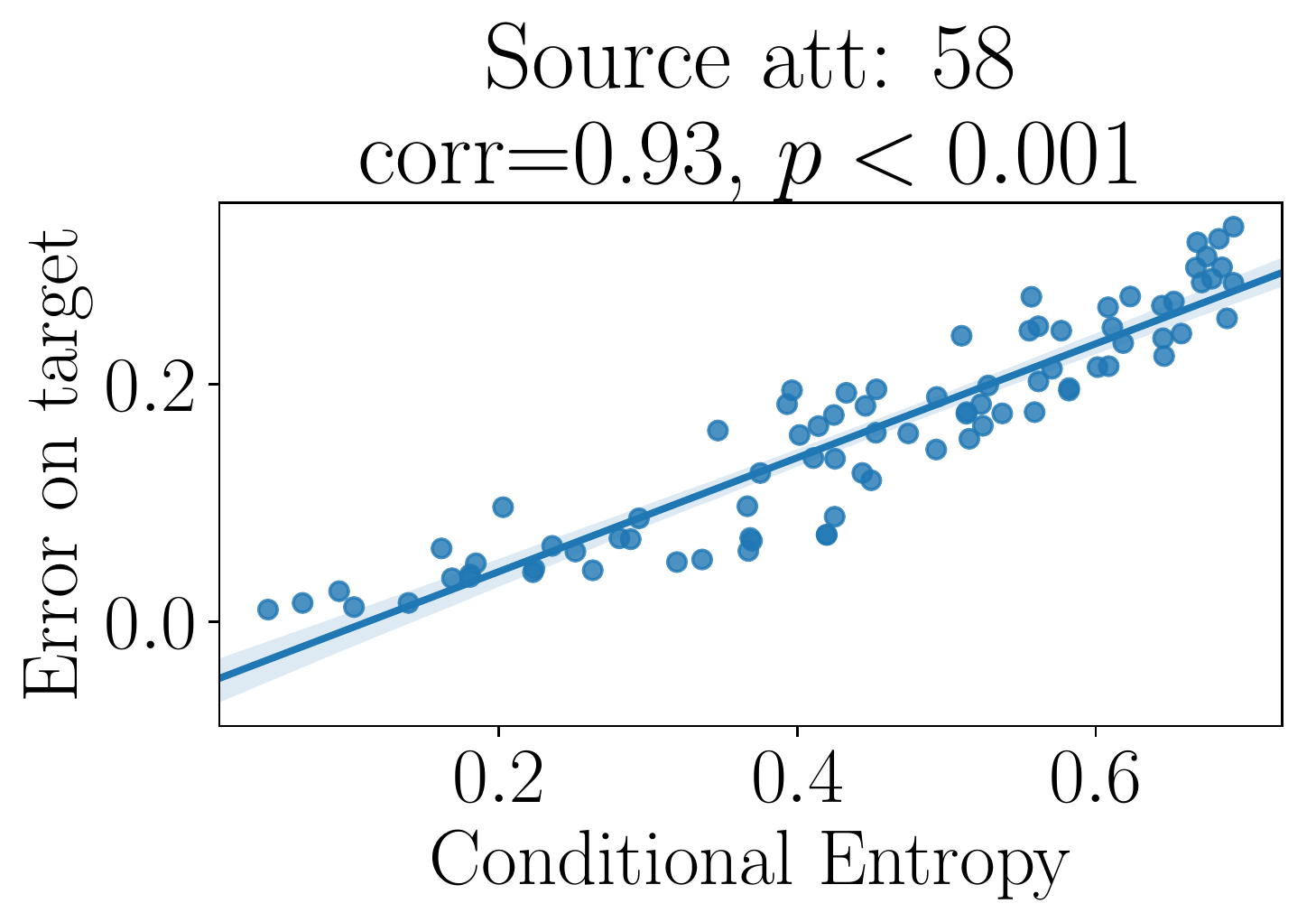}&

\includegraphics[clip, trim=0mm 0mm 0mm 13mm, width=0.19\textwidth]{figures/AWA2_att58.pdf}\\[-2pt]

(55) Insects & (56) Forager & (57) Grazer & (58) Hunter & (59) Scavenger\\[6pt]
\includegraphics[clip, trim=0mm 0mm 0mm 13mm, width=0.19\textwidth]{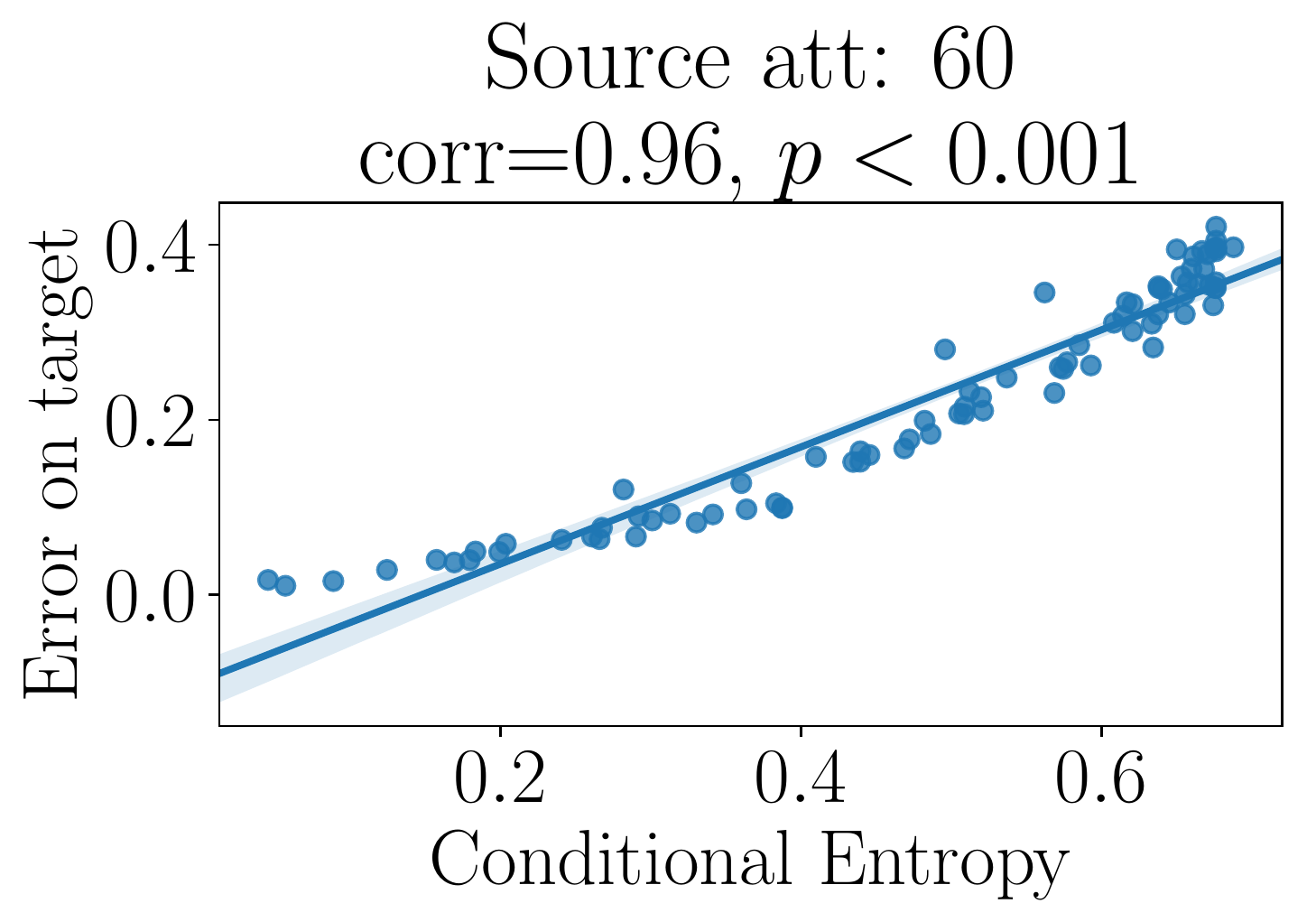}&

\includegraphics[clip, trim=0mm 0mm 0mm 13mm, width=0.19\textwidth]{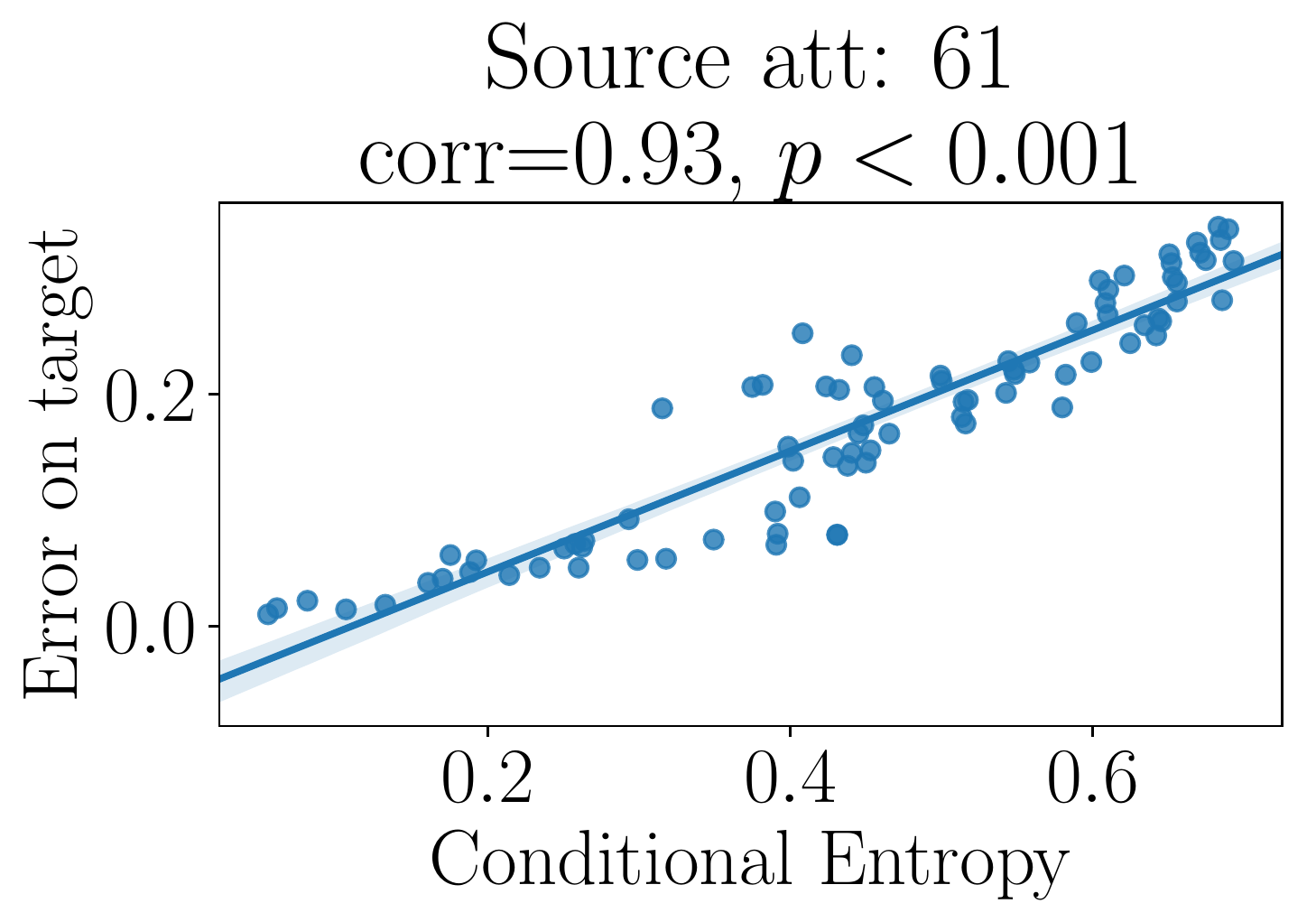}&

\includegraphics[clip, trim=0mm 0mm 0mm 13mm, width=0.19\textwidth]{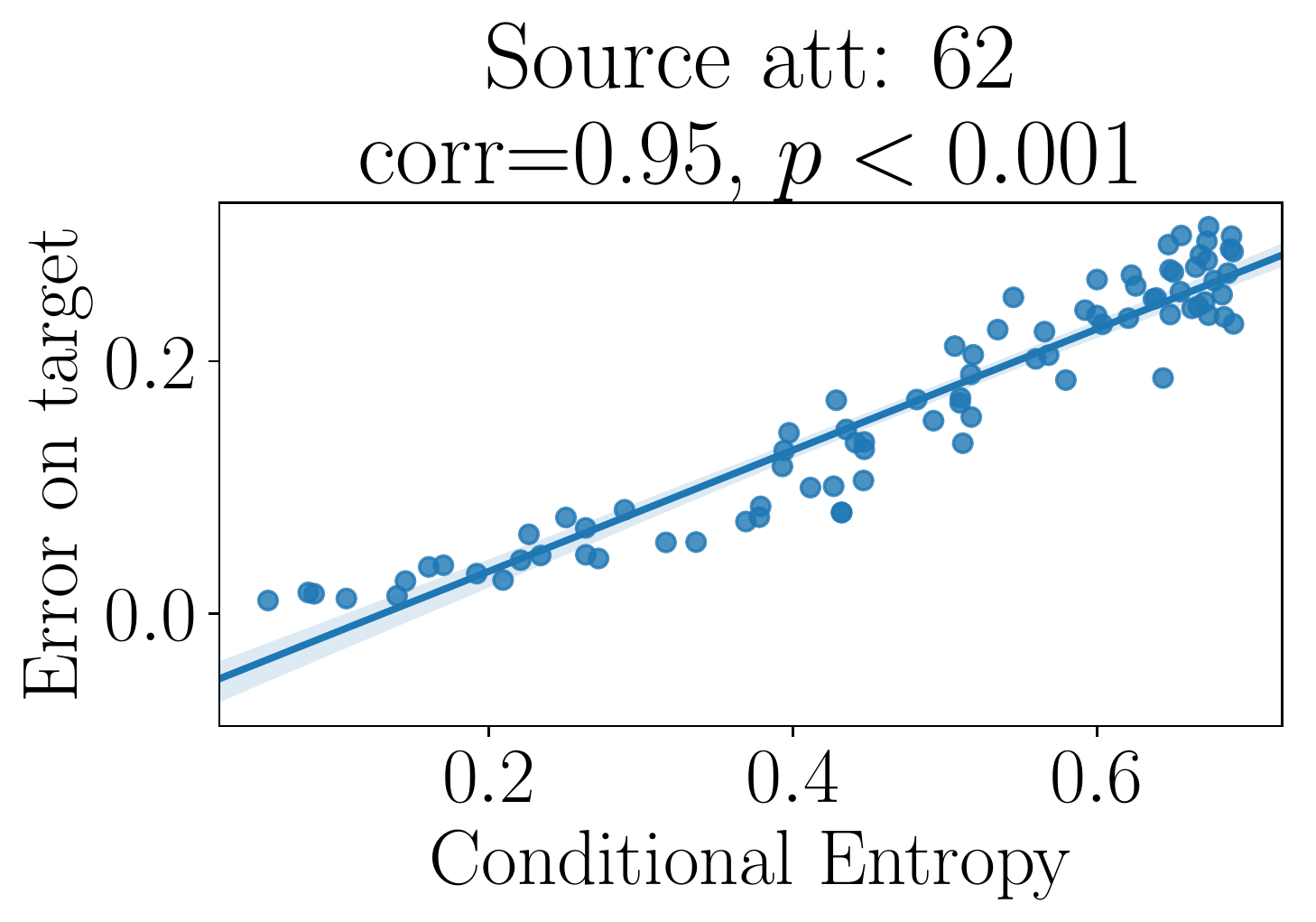}&

\includegraphics[clip, trim=0mm 0mm 0mm 13mm, width=0.19\textwidth]{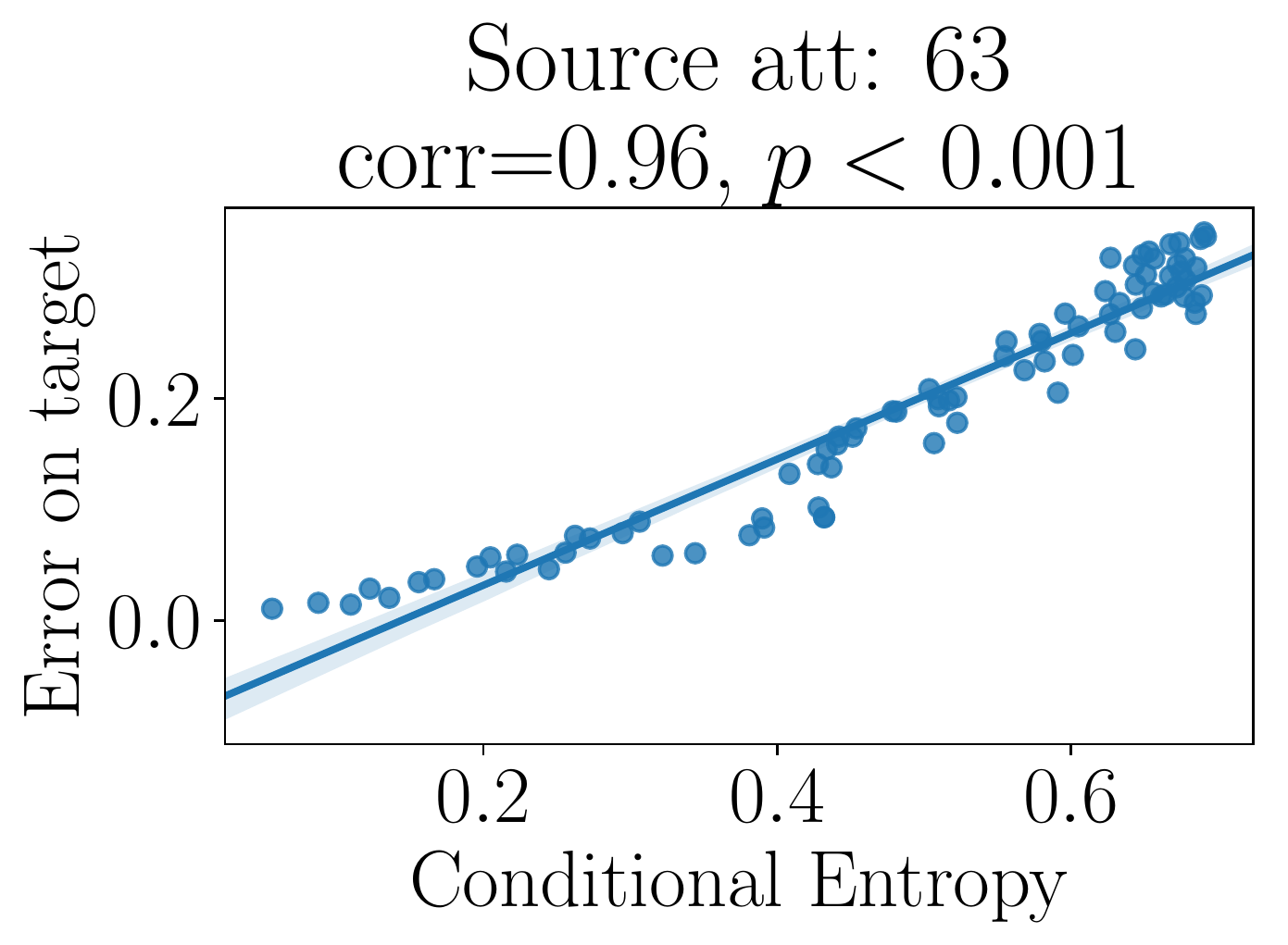}&

\includegraphics[clip, trim=0mm 0mm 0mm 13mm, width=0.19\textwidth]{figures/AWA2_att63.pdf}\\[-2pt]

(60) Skimmer & (61) Stalker & (62) Newworld & (63) Oldworld & (64) Arctic\\[6pt]
\includegraphics[clip, trim=0mm 0mm 0mm 13mm, width=0.19\textwidth]{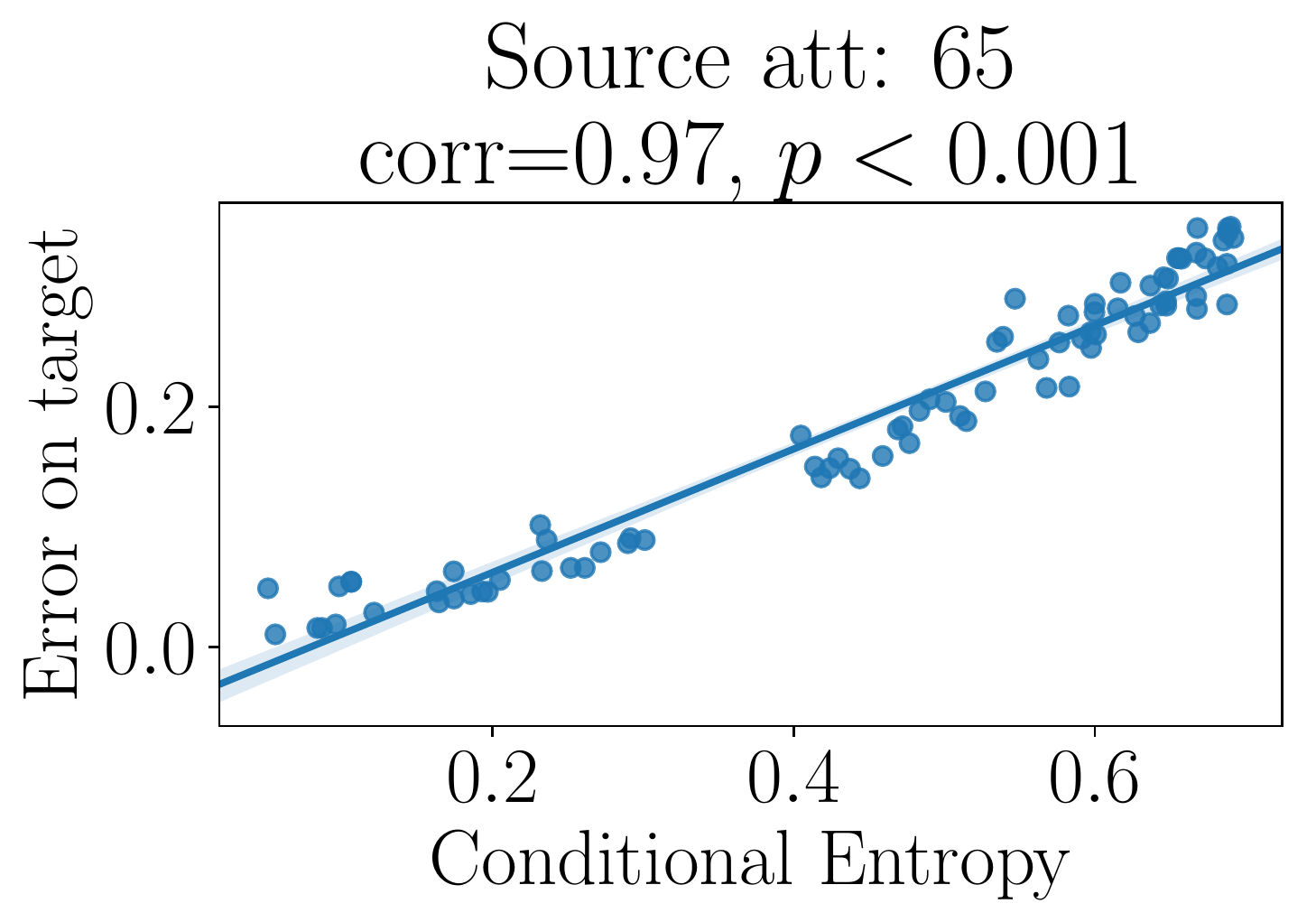}&

\includegraphics[clip, trim=0mm 0mm 0mm 13mm, width=0.19\textwidth]{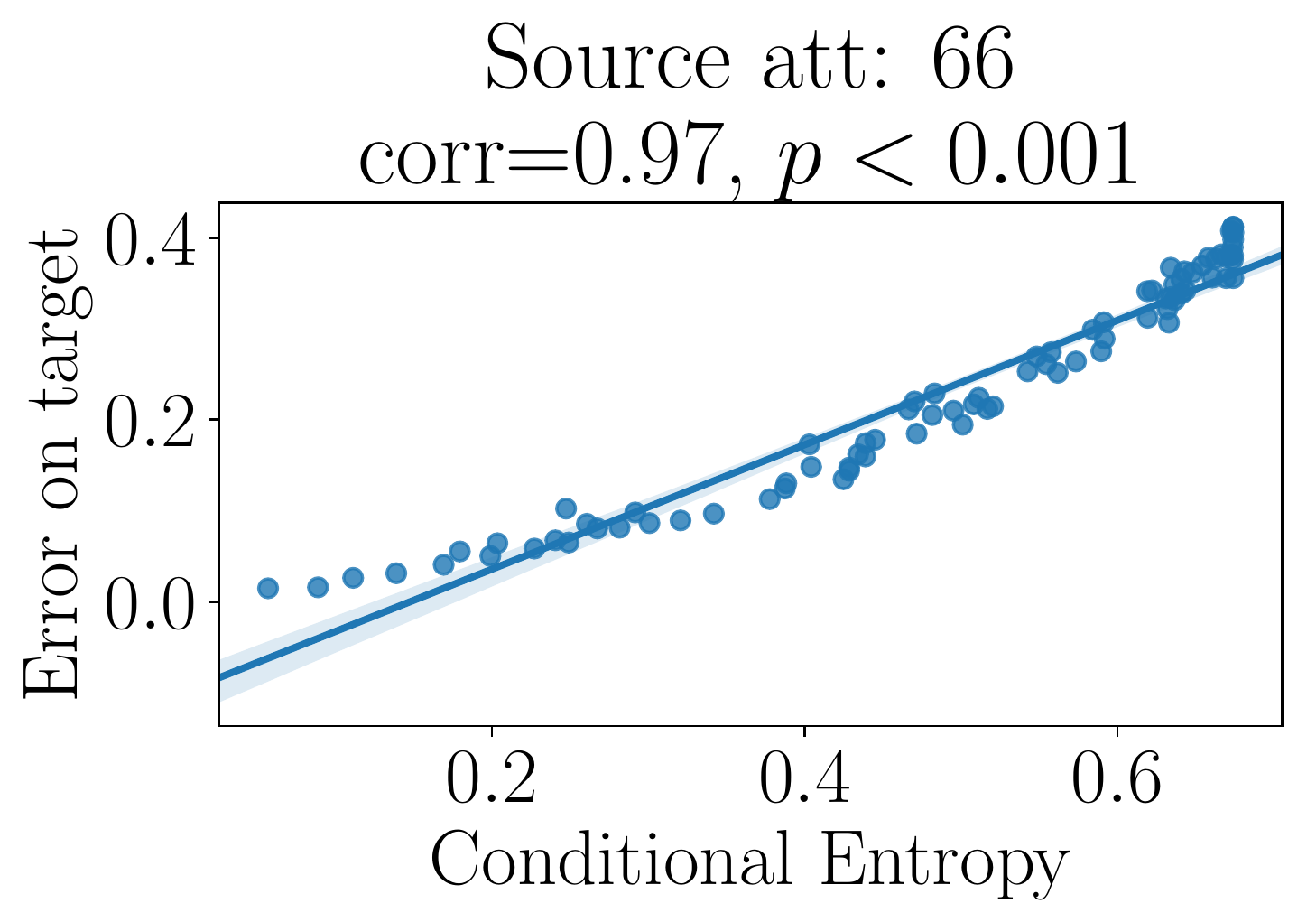}&

\includegraphics[clip, trim=0mm 0mm 0mm 13mm, width=0.19\textwidth]{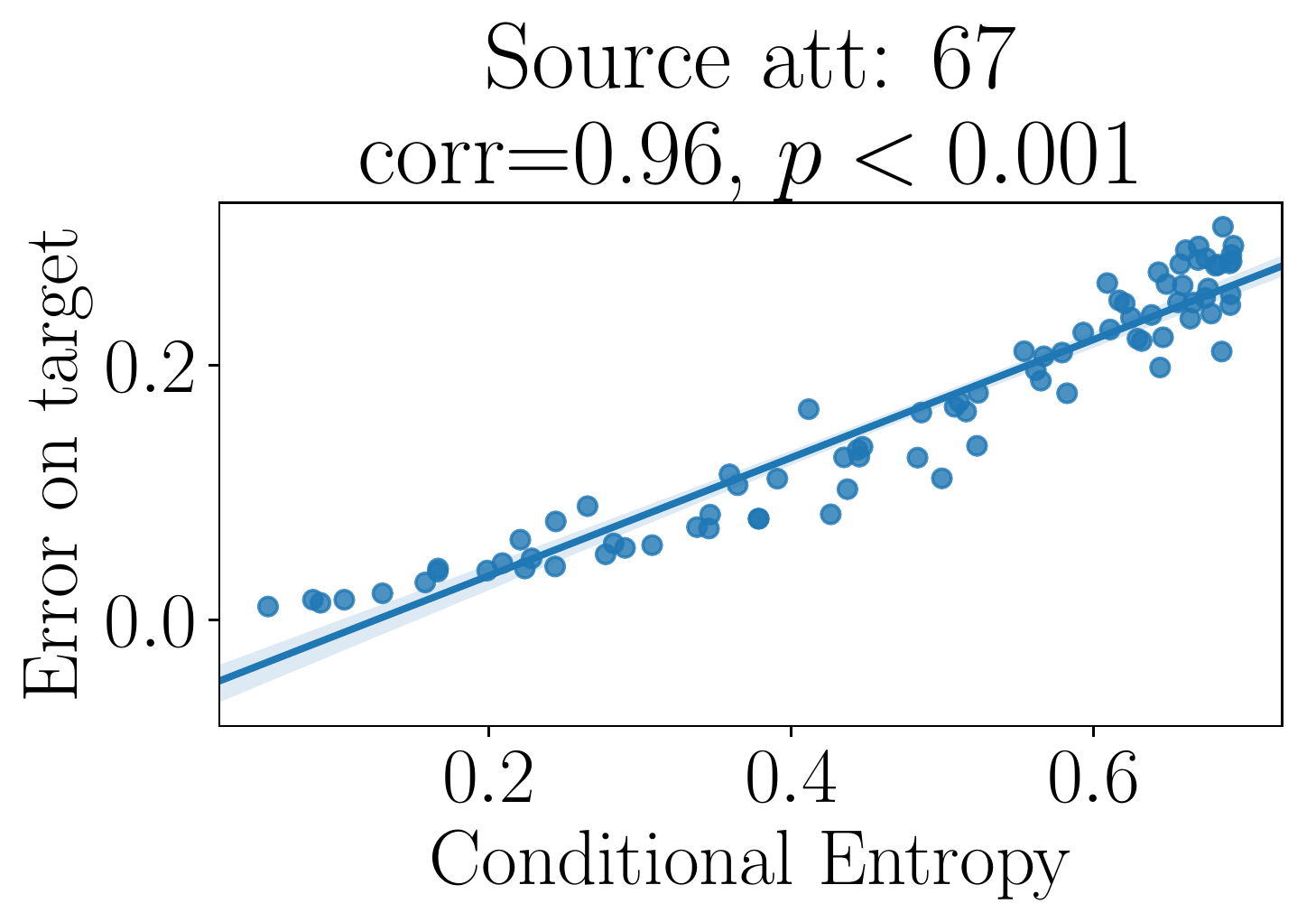}&

\includegraphics[clip, trim=0mm 0mm 0mm 13mm, width=0.19\textwidth]{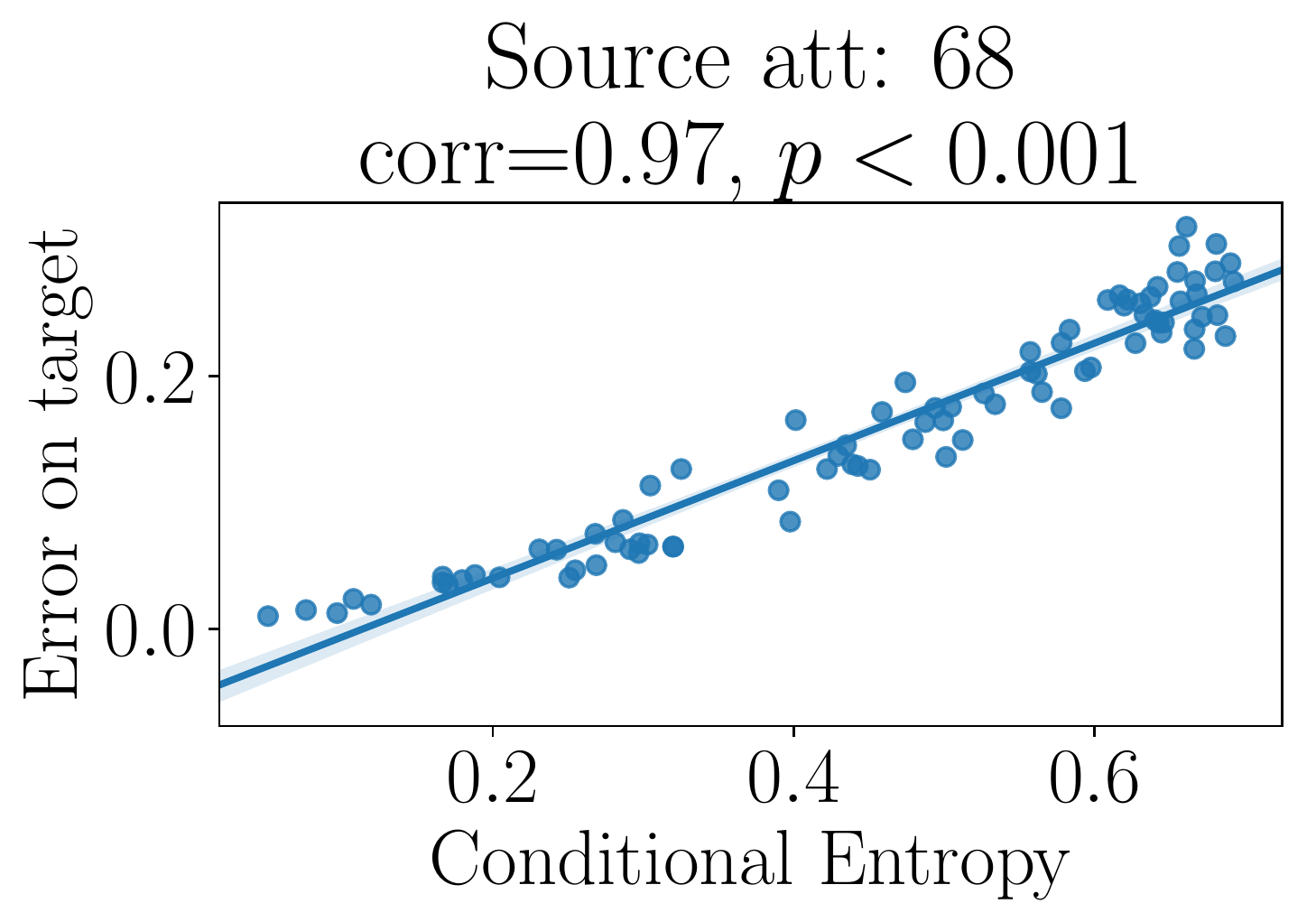}&

\includegraphics[clip, trim=0mm 0mm 0mm 13mm, width=0.19\textwidth]{figures/AWA2_att68.pdf}\\[-2pt]

(65) Coastal & (66) Desert & (67) Bush & (68) Plains & (69) Forest\\[6pt]
\end{tabular}
\caption{{\bf Attribute prediction; CE vs. test errors on AwA2 (Extended from Fig.~\ref{fig:face_att_trans}(e-h) in the paper; part 2).} The source attribute, $T^Z$, in each plot is named in the plot title. Points represent different target tasks $T^Y$. Corr is the Pearson correlation coefficient between the two variables and $p$ is the statistical significance of the correlation. In all cases, the correlation is statistically significant.}
\label{fig:trans_awa2}
\end{figure*}

\begin{figure*}[ht]
\centering
\footnotesize
\begin{tabular}{c@{~}c@{~}c@{~}c@{~}c}
\includegraphics[clip, trim=0mm 0mm 0mm 13mm, width=0.19\textwidth]{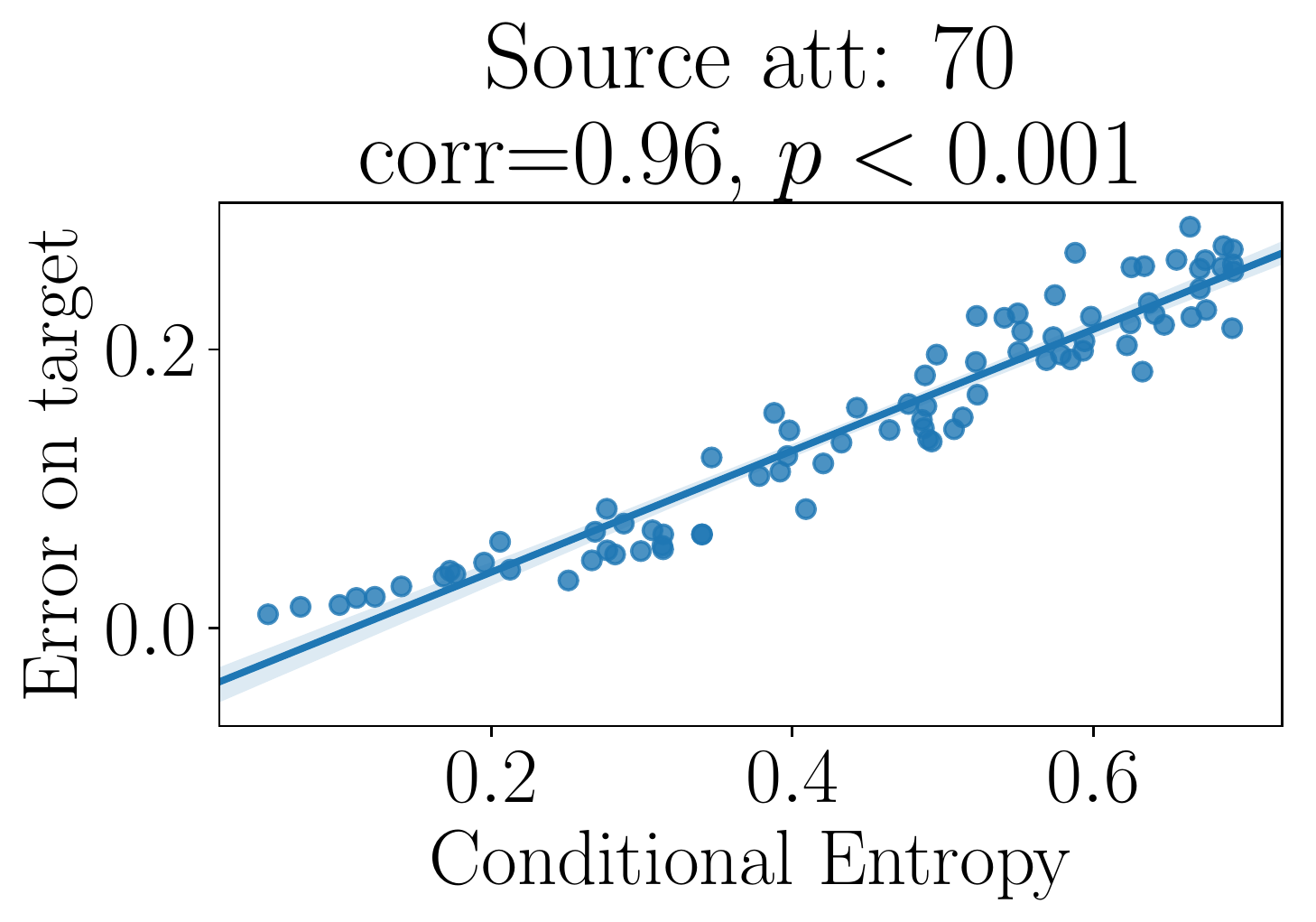}&

\includegraphics[clip, trim=0mm 0mm 0mm 13mm, width=0.19\textwidth]{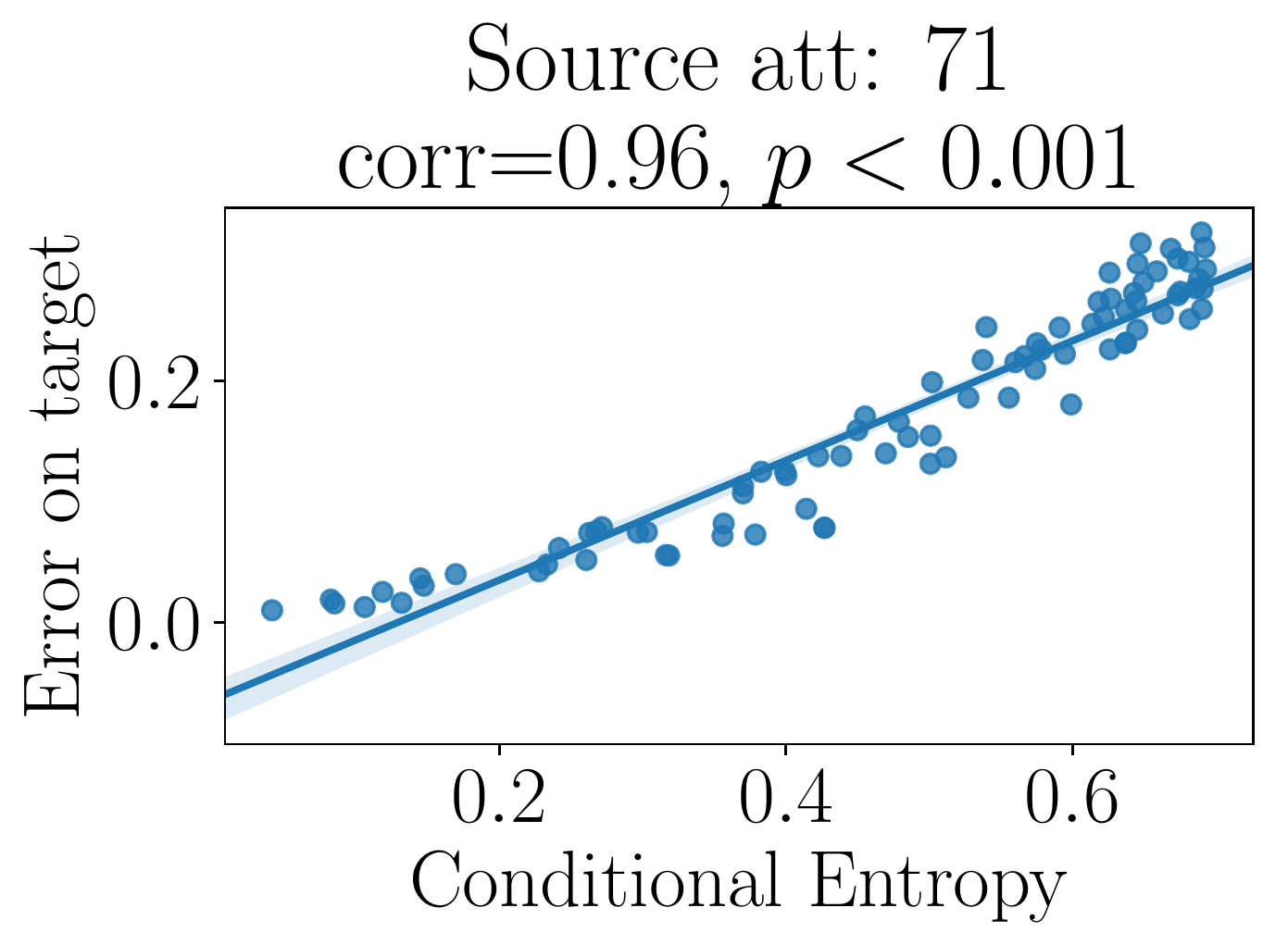}&

\includegraphics[clip, trim=0mm 0mm 0mm 13mm, width=0.19\textwidth]{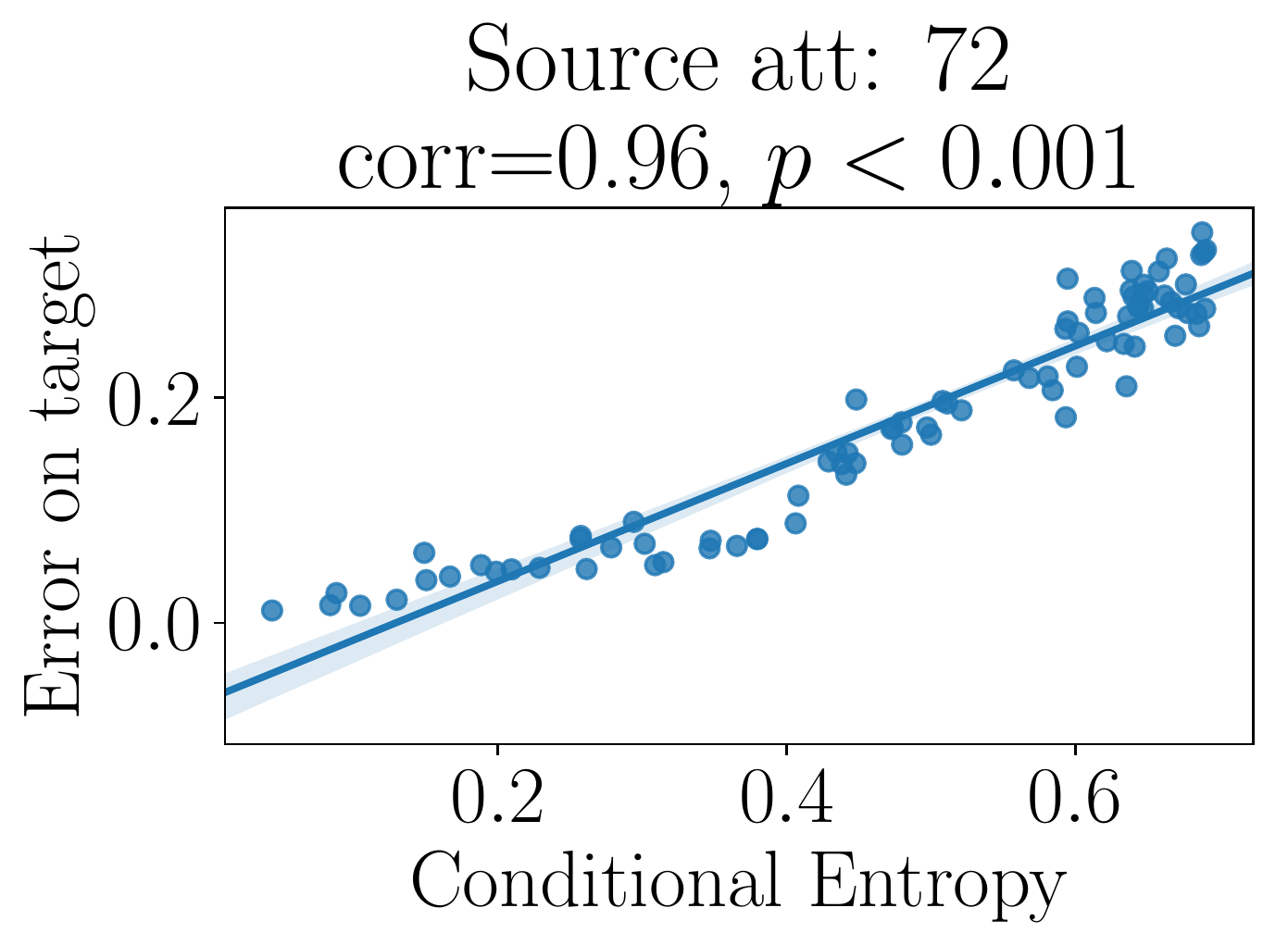}&

\includegraphics[clip, trim=0mm 0mm 0mm 13mm, width=0.19\textwidth]{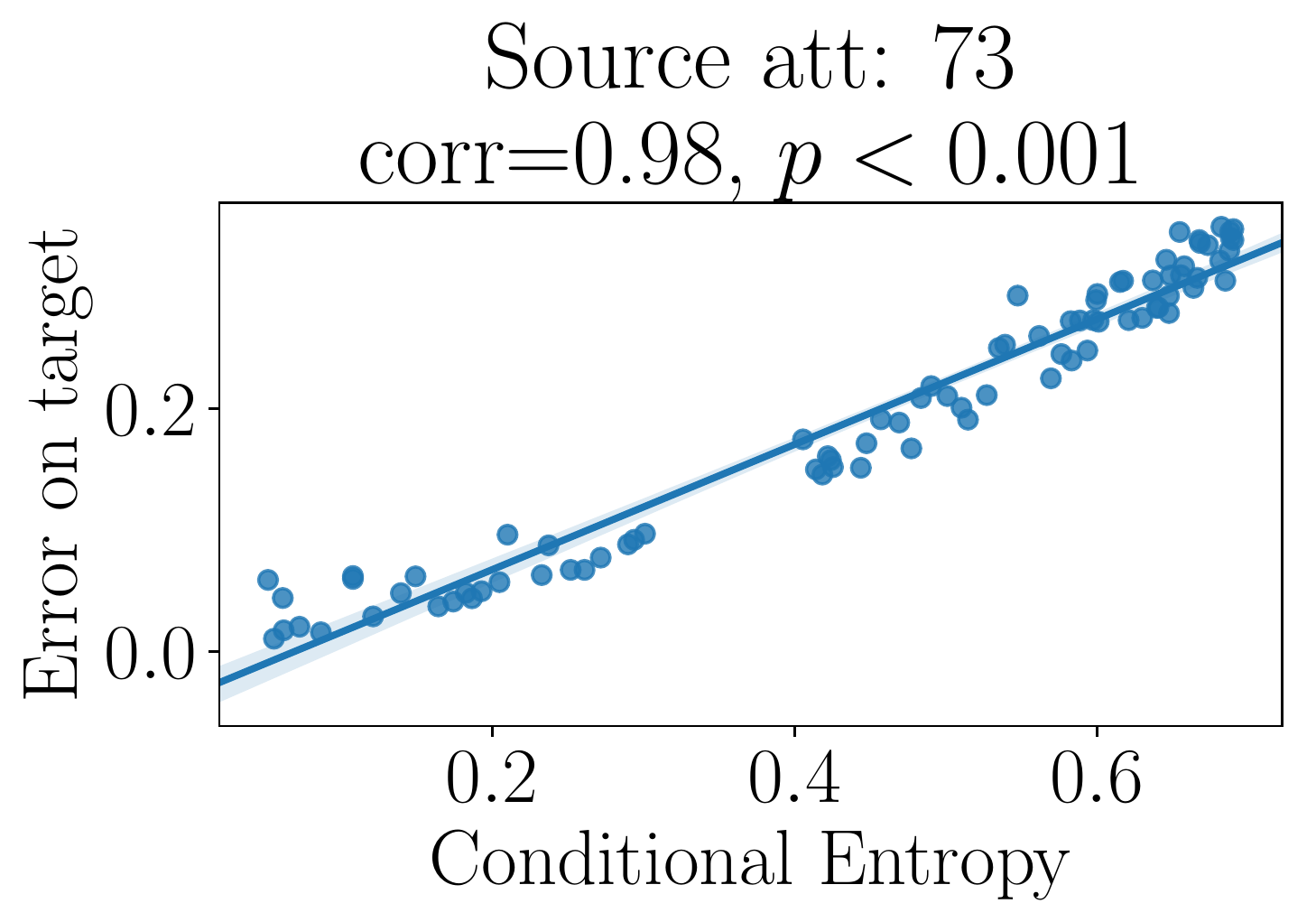}&

\includegraphics[clip, trim=0mm 0mm 0mm 13mm, width=0.19\textwidth]{figures/AWA2_att73.pdf}\\[-2pt]

(70) Fields & (71) Jungle & (72) Mountains & (73) Ocean & (74) Ground\\[6pt]

\includegraphics[clip, trim=0mm 0mm 0mm 13mm, width=0.19\textwidth]{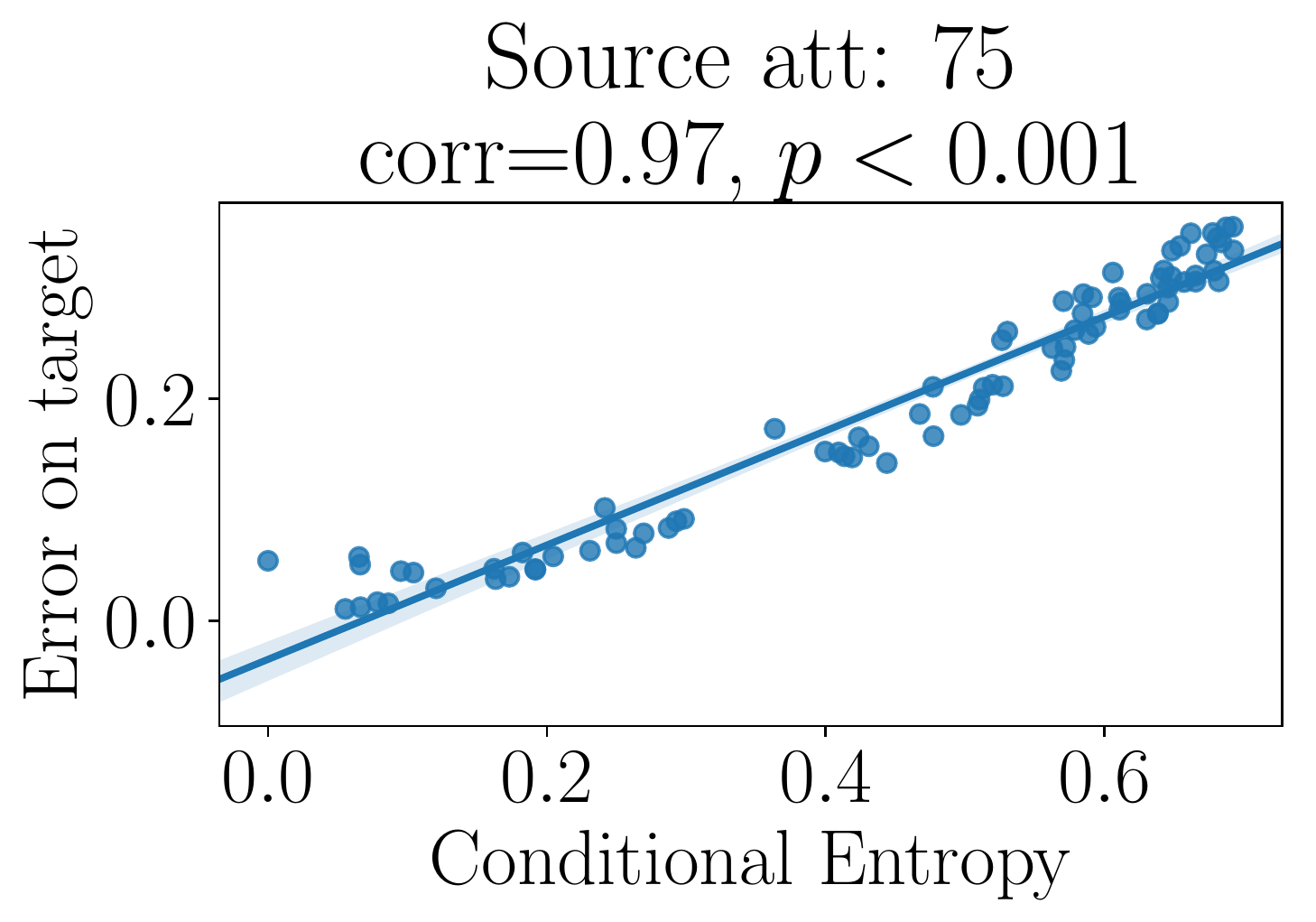}&

\includegraphics[clip, trim=0mm 0mm 0mm 13mm, width=0.19\textwidth]{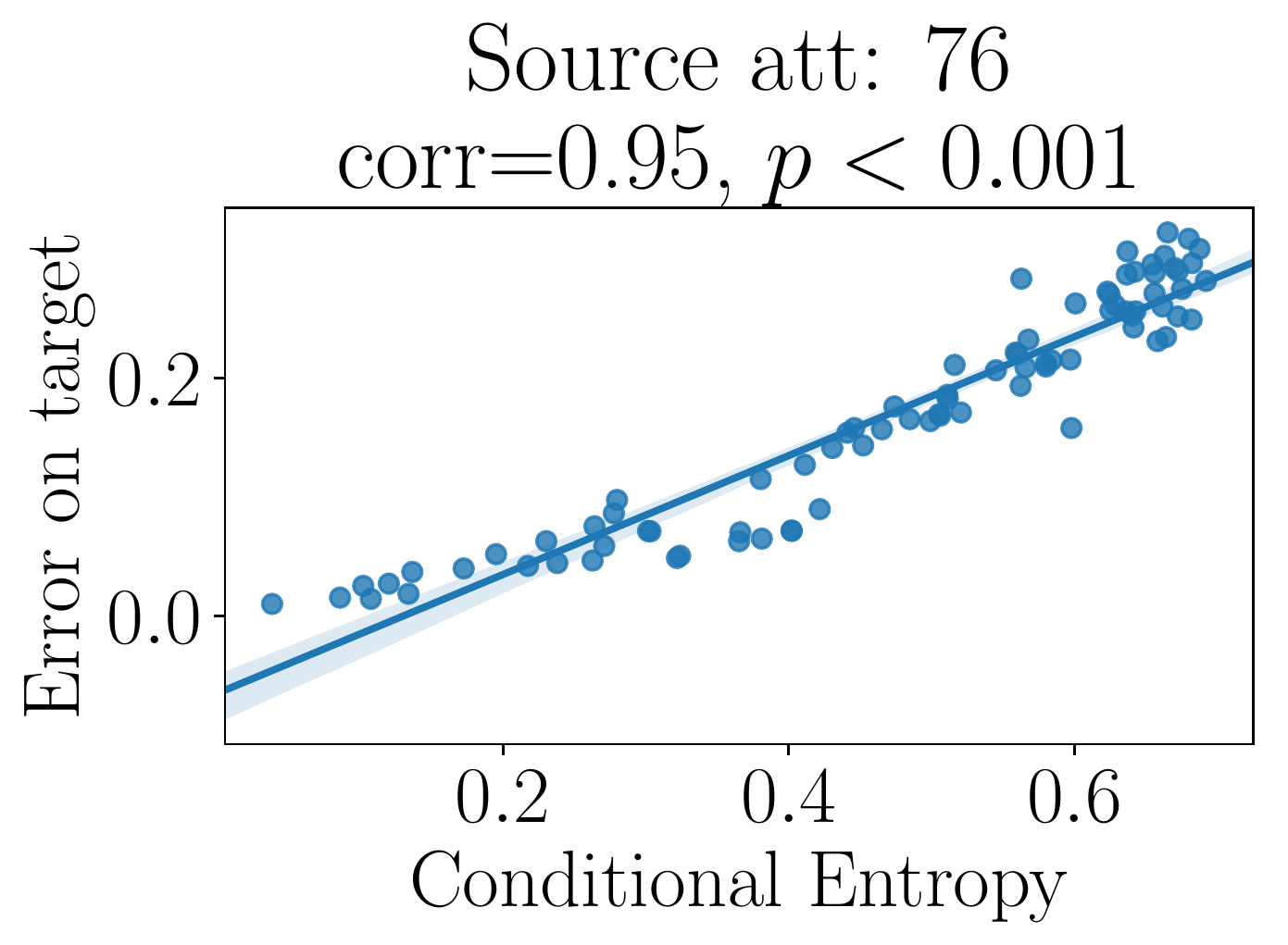}&

\includegraphics[clip, trim=0mm 0mm 0mm 13mm, width=0.19\textwidth]{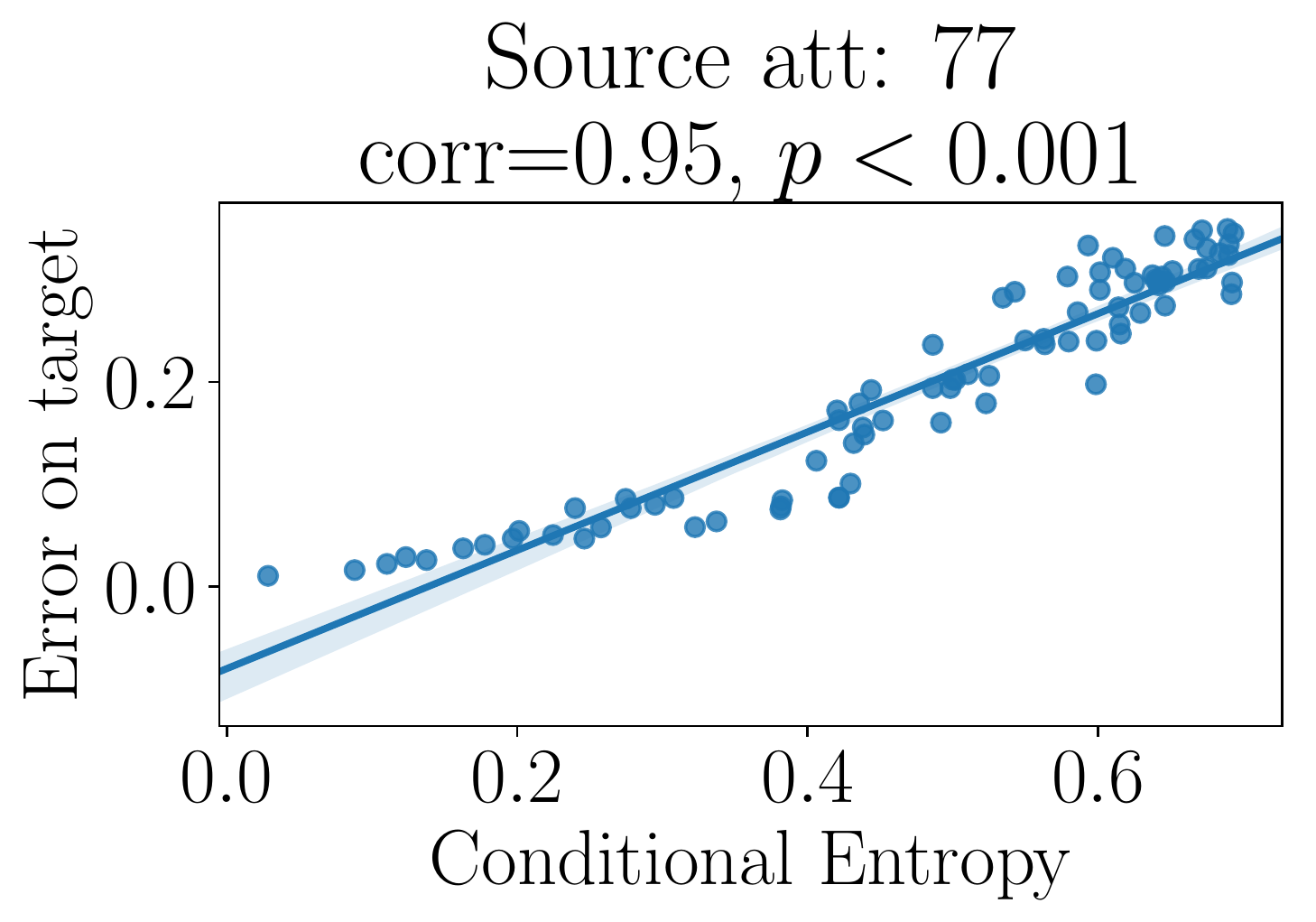}&

\includegraphics[clip, trim=0mm 0mm 0mm 13mm, width=0.19\textwidth]{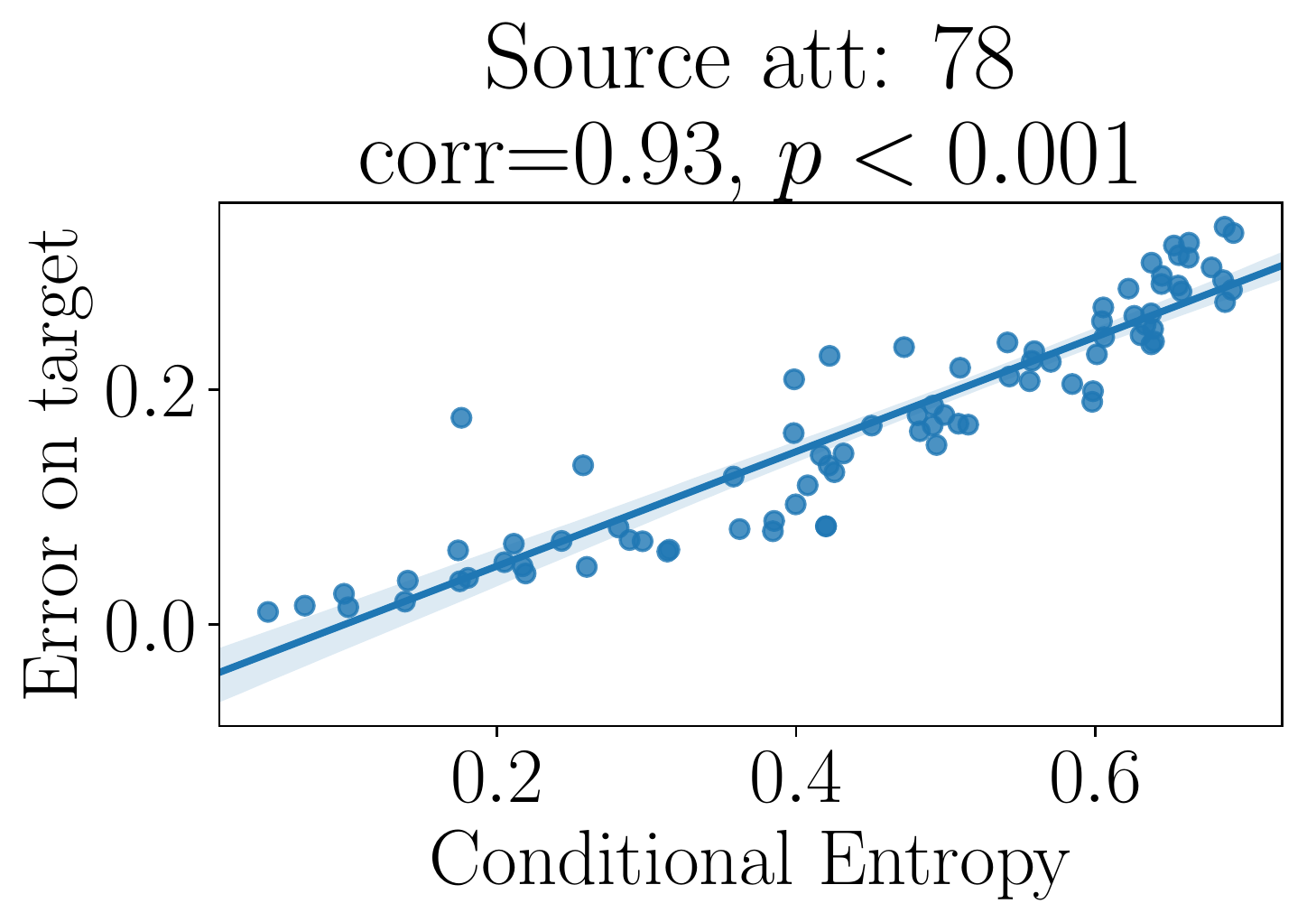}&

\includegraphics[clip, trim=0mm 0mm 0mm 13mm, width=0.19\textwidth]{figures/AWA2_att78.pdf}\\[-2pt]

(75) Water & (76) Tree & (77) Cave & (78) Fierce & (79) Timid\\[6pt]
\includegraphics[clip, trim=0mm 0mm 0mm 13mm, width=0.19\textwidth]{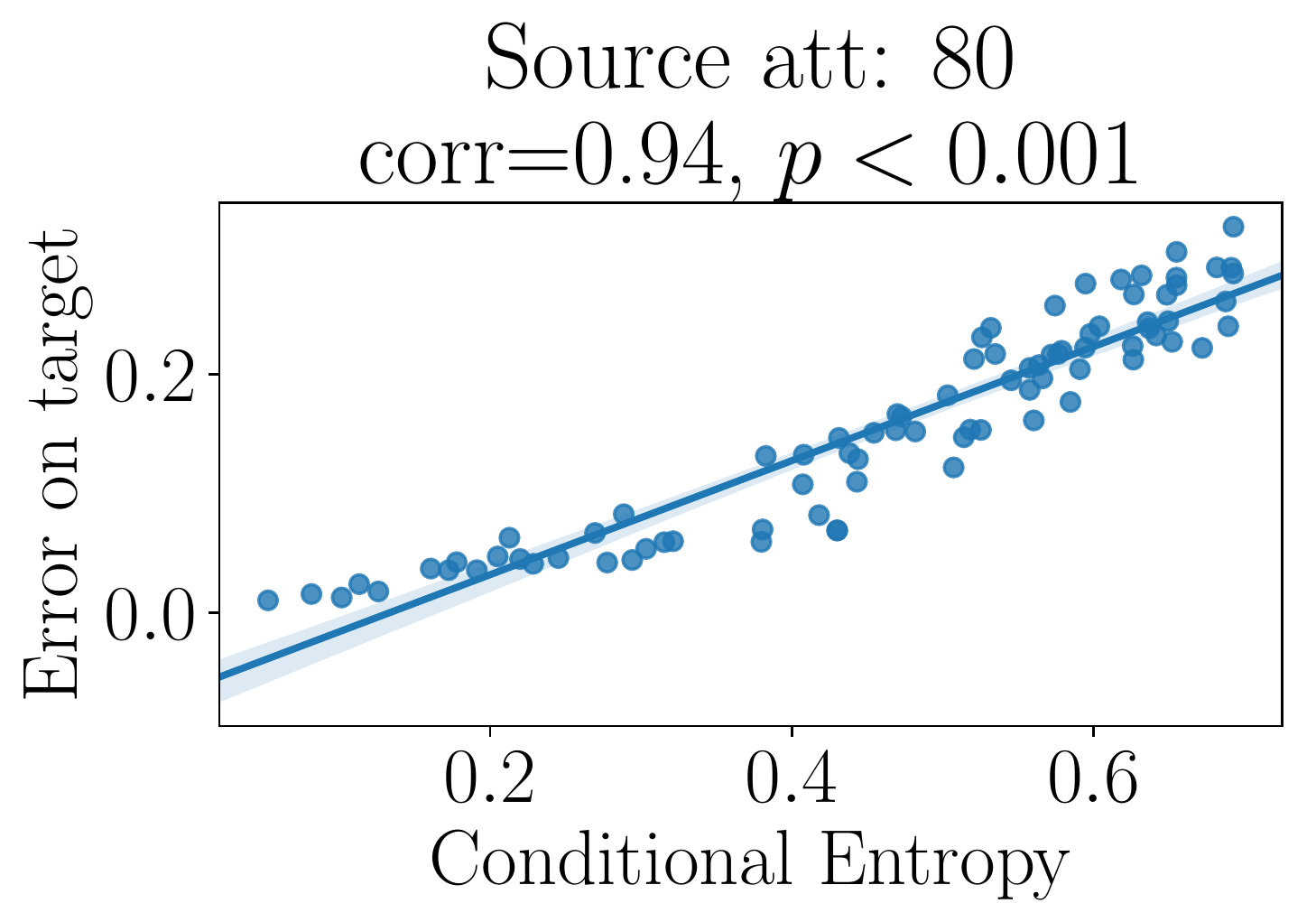}&

\includegraphics[clip, trim=0mm 0mm 0mm 13mm, width=0.19\textwidth]{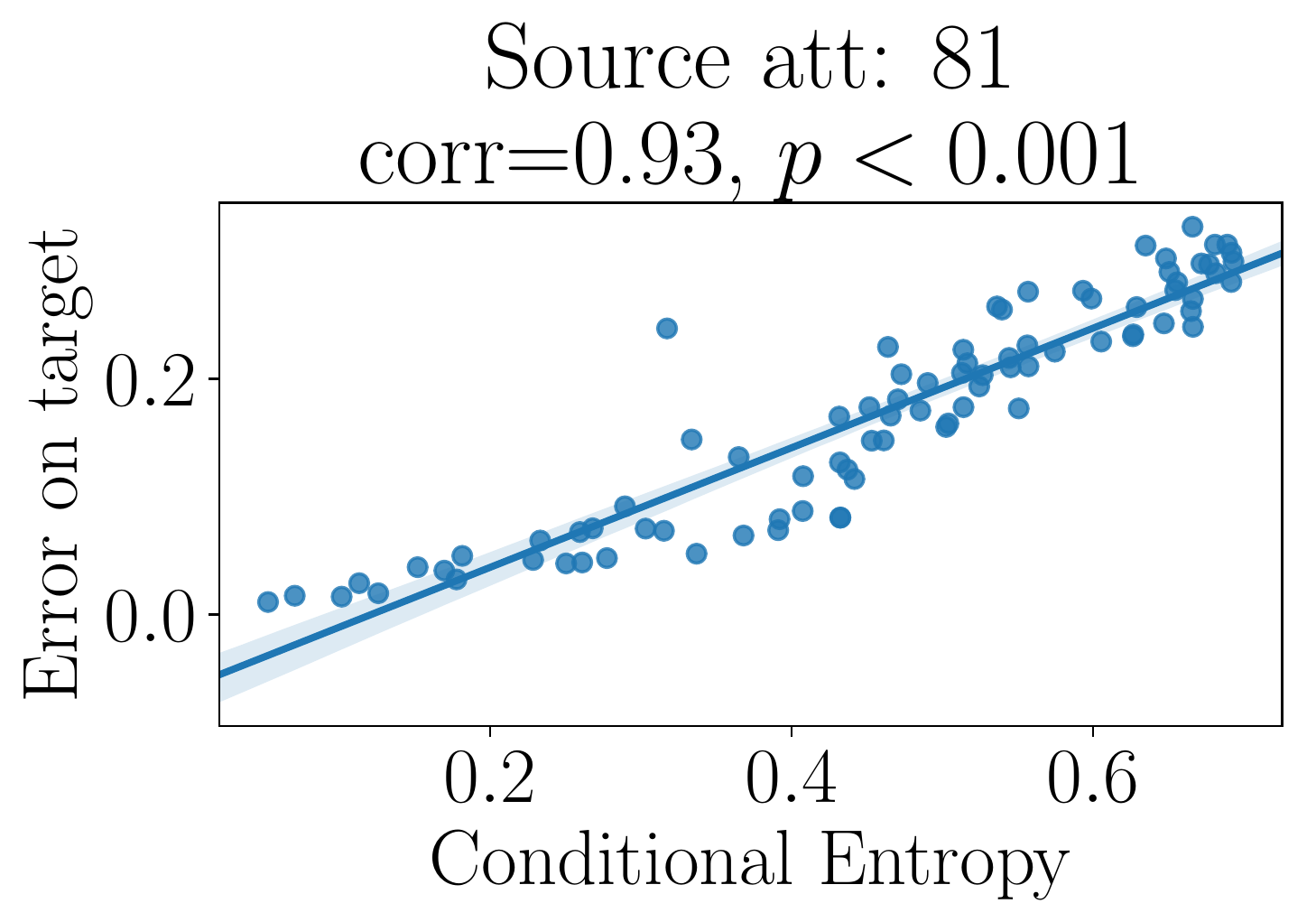}&

\includegraphics[clip, trim=0mm 0mm 0mm 13mm, width=0.19\textwidth]{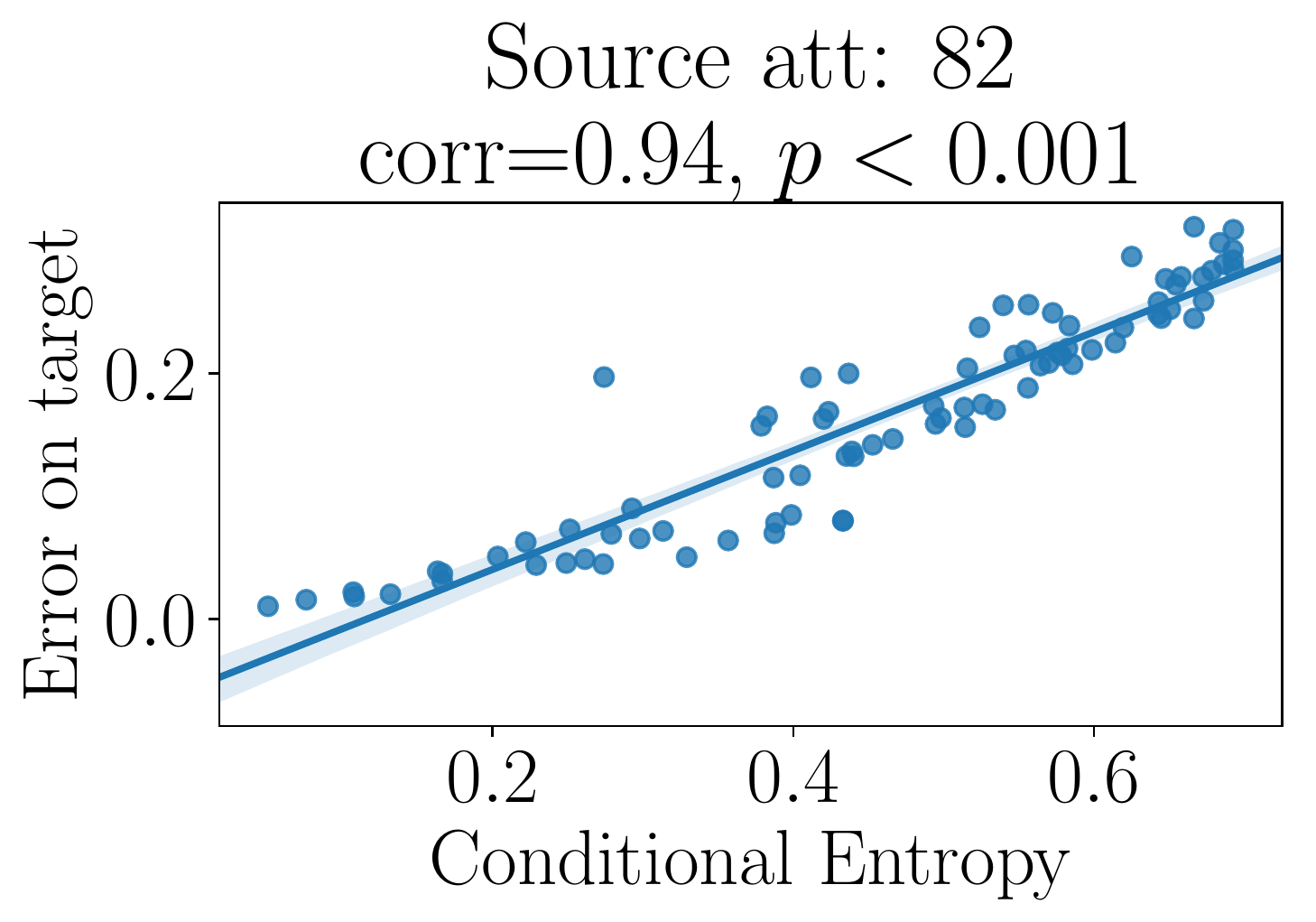}&

\includegraphics[clip, trim=0mm 0mm 0mm 13mm, width=0.19\textwidth]{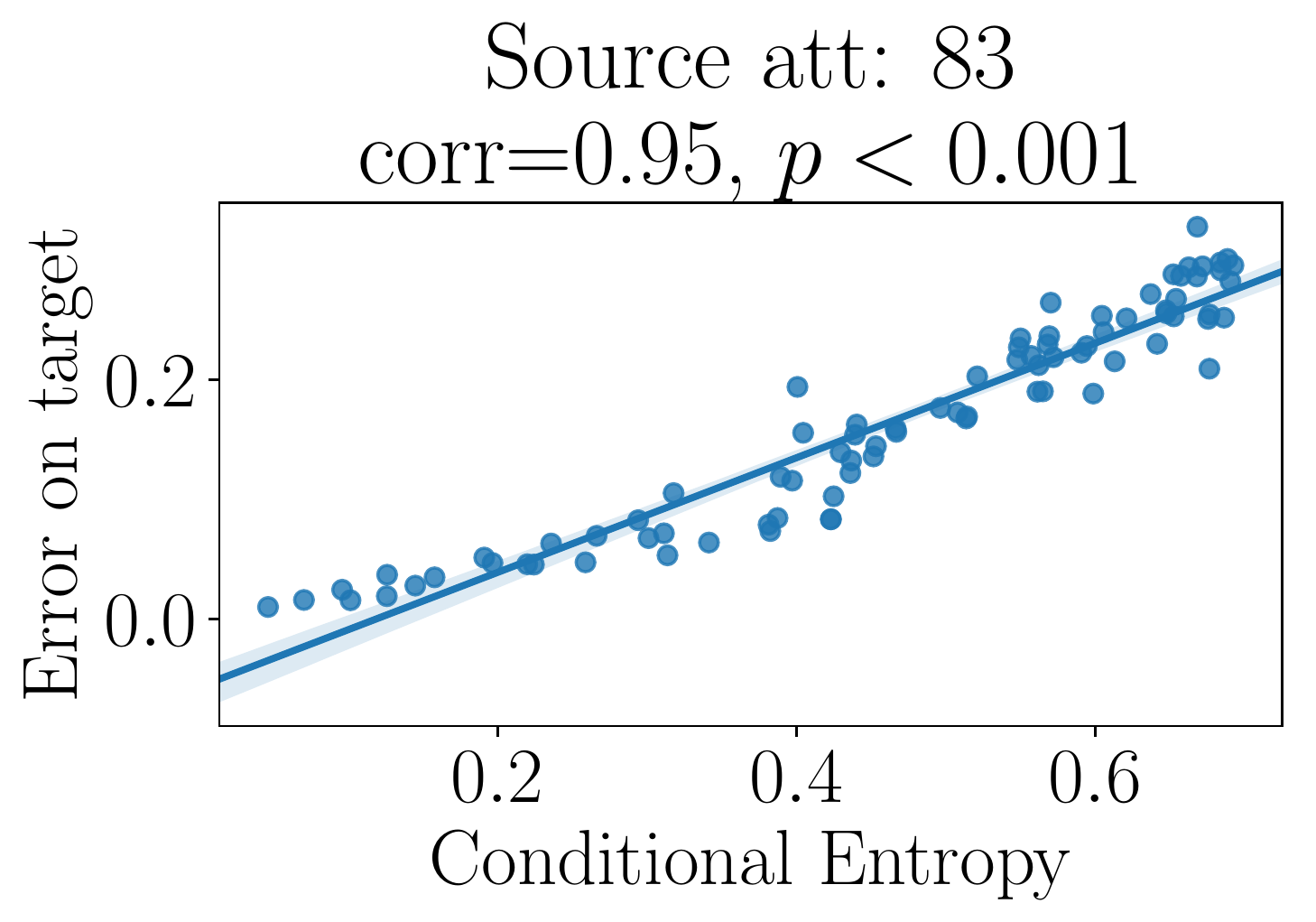}&

\includegraphics[clip, trim=0mm 0mm 0mm 13mm, width=0.19\textwidth]{figures/AWA2_att83.pdf}\\[-2pt]

(80) Smart & (81) Group & (82) Solitary & (83) Nestspot & (84) Domestic\\[6pt]
    \end{tabular}
\caption{{\bf Attribute prediction; CE vs. test errors on AwA2 (Extended from Fig.~\ref{fig:face_att_trans}(e-h) in the paper; part 3).} The source attribute, $T^Z$, in each plot is named in the plot title. Points represent different target tasks $T^Y$. Corr is the Pearson correlation coefficient between the two variables and $p$ is the statistical significance of the correlation. In all cases, the correlation is statistically significant.}
\label{fig:trans_awa3}
\end{figure*}

\begin{itemize}
    \item \textbf{Attribute prediction on CelebA~\cite{liu2015faceattributes}}: see Fig.~\ref{fig:trans_celebA}.
    \item \textbf{CelebA: Transferability from identity to attributes}: see Table~\ref{tab:id2attribute_full}.
    \item \textbf{Attribute prediction on AwA2~\cite{xian2018zero}}: see Fig.~\ref{fig:trans_awa1},~\ref{fig:trans_awa2}, and~\ref{fig:trans_awa3}.
\end{itemize}

\begin{table*}[ht]
    \centering
    \resizebox{1.0\linewidth}{!}{
    \begin{tabular}{c@{~~}lcccccccccc@{~}l}
    \toprule
&Attribute:&Male&Bald&Gray Hair & Mustache & Double Chin & Chubby &Sideburns& Goatee & Young & Wear Hat &\multirow{12}{*}{$\rightarrow$}\\
1&LNets$+$ANet 2015~ \cite{liu2015faceattributes}&0.980&0.980&0.970&0.950&0.920&0.910&0.960&0.950&0.870&0.990&\\
2&Walk and Learn 2016~\cite{wang2016walk}&0.960&0.920&0.950&0.900&0.930&0.890&0.920&0.920&0.860&0.960&\\
3&MOON 2016~\cite{rudd2016moon}&0.981&0.988&0.981&0.968&0.963&0.954&0.976&0.970&0.881&0.990&\\
4&LMLE 2016~\cite{huang2016learning}&0.990&0.900&0.910&0.730&0.740&0.790&0.880&0.950&0.870&0.990&\\
5&CR-I 2017~\cite{dong2017class}&0.960&0.970&0.950&0.940&0.890&0.870&0.920&0.960&0.840&0.980&\\
6&MCNN-AUX 2017~\cite{hand2017attributes}&0.982&0.989&0.982&0.969&0.963&0.957&0.978&0.972&0.885&0.990&\\
7&DMTL 2018~\cite{han2018heterogeneous}&0.980&0.990&0.960&0.970&0.990&0.970&0.980&0.980&0.900&0.990&\\
8&Face-SSD 2019~\cite{jang2019registration}&0.973&0.986&0.976&0.960&0.960&0.951&0.966&0.963&0.876&0.985&\\

\rowcolor{lightblue}9&Conditional Entropy$\uparrow$&0.017&0.026&0.052&0.062&0.083&0.087&0.088&0.089&0.095&0.107&\\
10&Dedicated Res18&0.985&0.990&0.980&0.968&0.959&0.951&0.976&0.974&0.879&0.991&\\
11&FromID SVM&0.992&0.991&0.981&0.968&0.963&0.957&0.976&0.973&0.899&0.988&\\ \hline

&Attribute:& Eye glasses & Pale Skin & Wear Necktie & Blurry & No Beard  & Receding Hairline & 5 clock Shadow & Rosy Cheeks & Blond Hair & Big Lips &\multirow{12}{*}{$\rightarrow$}\\
1&&0.990&0.910&0.930&0.840&0.950&0.890&0.910&0.900&0.950&0.680&\\
2&&0.970&0.850&0.840&0.910&0.900&0.840&0.840&0.960&0.920&0.780&\\
3&&0.995&0.970&0.966&0.957&0.956&0.936&0.940&0.948&0.959&0.715&\\
4&&0.980&0.800&0.900&0.590&0.960&0.760&0.820&0.780&0.990&0.600&\\
5&&0.960&0.920&0.880&0.850&0.940&0.870&0.900&0.880&0.950&0.680&\\
6&&0.996&0.970&0.965&0.962&0.960&0.938&0.945&0.952&0.960&0.715&\\
7&&0.990&0.970&0.970&0.960&0.970&0.940&0.950&0.960&0.910&0.880&\\
8&&0.992&0.957&0.956&0.950&0.949&0.931&0.929&0.943&0.936&0.778&\\

\rowcolor{lightblue}9&&0.109&0.122&0.131&0.139&0.141&0.141&0.145&0.152&0.16&0.161&\\
10&&0.997&0.970&0.963&0.963&0.961&0.936&0.942&0.950&0.961&0.715&\\
11&&0.996&0.958&0.941&0.956&0.958&0.933&0.937&0.939&0.949&0.710&\\ \hline

&Attribute:& Bushy Eyebrows & Wear Lipstick & Big Nose &Bangs& Narrow Eyes & Wear Necklace & Heavy Makeup & Black Hair & Wear Earrings & Arched Eyebrows &\multirow{12}{*}{$\rightarrow$}\\
1&&0.900&0.930&0.780&0.950&0.810&0.710&0.900&0.880&0.820&0.790&\\
2&&0.930&0.920&0.910&0.960&0.790&0.770&0.960&0.840&0.910&0.870&\\
3&&0.926&0.939&0.840&0.958&0.865&0.870&0.910&0.894&0.896&0.823&\\
4&&0.820&0.990&0.800&0.980&0.590&0.590&0.980&0.920&0.830&0.790&\\
5&&0.840&0.940&0.800&0.950&0.720&0.740&0.840&0.900&0.830&0.800&\\
6&&0.928&0.941&0.845&0.960&0.872&0.866&0.915&0.898&0.904&0.834&\\
7&&0.850&0.930&0.920&0.960&0.900&0.890&0.920&0.850&0.910&0.860&\\
8&&0.896&0.926&0.823&0.952&0.890&0.878&0.907&0.879&0.869&0.820&\\ 

\rowcolor{lightblue}9&&0.192&0.202&0.232&0.236&0.252&0.252&0.27&0.286&0.291&0.306&\\
10&&0.927&0.935&0.828&0.961&0.875&0.859&0.916&0.901&0.896&0.834&\\
11&&0.919&0.940&0.845&0.950&0.863&0.865&0.897&0.869&0.853&0.822&\\ \hline

&Attribute:& Brown Hair & Bags U Eyes & Oval Face & Straight Hair & Pointy Nose & Attractive & Wavy Hair & High Cheeks & Smiling & Mouth Open&Average (all)\\
1&&0.800&0.790&0.660&0.730&0.720&0.810&0.800&0.870&0.920&0.920&0.873\\
2&&0.810&0.870&0.790&0.750&0.770&0.840&0.850&0.950&0.980&0.970&0.887\\
3&&0.894&0.849&0.757&0.823&0.765&0.817&0.825&0.870&0.926&0.935&0.909\\
4&&0.870&0.730&0.680&0.730&0.720&0.880&0.830&0.920&0.990&0.960&0.838\\
5&&0.860&0.800&0.660&0.730&0.730&0.830&0.790&0.890&0.930&0.950&0.866\\
6&&0.892&0.849&0.758&0.836&0.775&0.831&0.839&0.876&0.927&0.937&0.913\\
7&&0.960&0.990&0.780&0.850&0.780&0.850&0.870&0.880&0.940&0.940&0.926\\
8&&0.835&0.825&0.748&0.834&0.749&0.813&0.851&0.868&0.918&0.919&0.903\\ 

\rowcolor{lightblue}9&&0.315&0.324&0.339&0.339&0.341&0.361&0.381&0.476&0.521&0.551&\\
10&&0.886&0.834&0.752&0.836&0.769&0.823&0.842&0.878&0.933&0.943&0.911\\
11&&0.854&0.838&0.733&0.812&0.769&0.820&0.800&0.859&0.909&0.901&0.902\\

    \bottomrule
    \end{tabular}
    }
    \vspace{1mm}
    \caption{{\bf Transferability from face recognition to facial attributes. (Extended from Table~\ref{tab:id2attribute} in the paper)} Results for CelebA attributes, sorted in ascending order of row~9 (decreasing transferability). Classification accuracies are shown for all 40 attributes. Subject specific attributes, e.g., {\em male} and {\em bald}, are more transferable than expression related attributes such as {\em smiling} and {\em mouth open}. These identity specific attributes corresponds to the automatic grouping presented in the original CelebA paper~\cite{liu2015faceattributes}. Unlike them, however, we obtain this grouping without necessitating the training of a deep attribute classification model. Unsurprisingly, transfer results (row 11) are best on these subject specific attributes and worst for less related attributes. Rows 1-8 provide published state of the art results. Despite training only an lSVM for attribute, row 11 results are comparable with more elaborate attribute classification systems. For details, see Sec.~\ref{sec:id2attrib}.}
    \label{tab:id2attribute_full}
\end{table*}

\section{Full hardness results}
\begin{itemize}
    \item \textbf{CelebA~\cite{liu2015faceattributes} attribute prediction hardness}: see Table~\ref{tab:CelebAhardorder}.
    \item \textbf{AwA2~\cite{xian2018zero} attribute prediction hardness}: see Table~\ref{tab:AWA2hardorder}.
    \item \textbf{CUB~\cite{WelinderEtal2010} attribute prediction hardness}: see Table~\ref{tab:CUBhardorder}.
\end{itemize}

\begin{table*}[ht]
    \centering
    \resizebox{1.0\linewidth}{!}{
    \begin{tabular}{c@{~~}lcccccccccc@{~}l}
    \toprule
1& Attribute& Bald& Mustache& Gray Hair& Pale Skin& Double Chin& Wearing Hat& Blurry& Sideburns& Chubby& Goatee &$\rightarrow$ \\
\rowcolor{lightblue}2& Conditional Entropy$\uparrow$&0.107&0.173&0.174&0.177&0.189&0.194&0.201&0.217&0.22&0.235&$\rightarrow$\\ \hline
 &$\rightarrow$&Eyeglasses& Rosy Cheeks& Wearing Necktie& Receding Hairline& 5 oClock Shadow& Narrow Eyes& Wearing Necklace& Bushy Eyebrows& Blond Hair& Bangs&$\rightarrow$\\
\rowcolor{lightblue}&$\rightarrow$&0.241&0.242&0.261&0.278&0.349&0.357&0.373&0.409&0.419&0.425&$\rightarrow$\\ \hline
 &$\rightarrow$&No Beard& Wearing Earrings& Bags Under Eyes& Brown Hair& Straight Hair& Young& Big Nose& Black Hair& Big Lips& Arched Eyebrows&$\rightarrow$\\
\rowcolor{lightblue}&$\rightarrow$&0.448&0.485&0.507&0.508&0.512&0.535&0.545&0.55&0.552&0.58&$\rightarrow$\\ \hline
 &$\rightarrow$&Pointy Nose& Oval Face& Wavy Hair& Heavy Makeup& Male& High Cheekbones& Wearing Lipstick& Smiling& Mouth Slightly Open& Attractive &\\
\rowcolor{lightblue}&$\rightarrow$&0.591&0.597&0.627&0.667&0.679&0.689&0.692&0.693&0.693&0.693&\\ \bottomrule

    \end{tabular}
    }
    \vspace{1mm}
    \caption{{\bf CelebA task hardness.} CelebA facial attributes sorted in ascending order of hardness along with their respective hardness scores. Hardness scores listed above are compared with empirical test errors for each task and shown to be strongly correlated (Fig.~\ref{fig:hardness}(a) in the paper). Note that the {\em male} classification task, appearing here as relatively hard, is the easiest task to transfer from face recognition (Table~\ref{tab:id2attribute}).}
    \label{tab:CelebAhardorder}
   
\end{table*}

\begin{table*}[ht]
    \centering
    \resizebox{1.0\linewidth}{!}{
    \begin{tabular}{c@{~~}lccccccccccccccc@{~}l}
    \toprule
1& Attribute& Flys& Red& Skimmer& Desert& Plankton& Insects& Tunnels& Hands& Tusks& Strainteeth& Cave& Blue& Stripes& Scavenger& Hops &$\rightarrow$ \\
\rowcolor{lightblue}2& Conditional Entropy$\uparrow$&0.057& 0.089& 0.112& 0.125& 0.140& 0.170& 0.181& 0.200& 0.205& 0.229& 0.243& 0.251& 0.263& 0.270& 0.284&$\rightarrow$\\ \hline
 &$\rightarrow$&Oldworld& Orange& Yellow& Quadrapedal& Flippers& Ground& Ocean& Coastal& Arctic& Walks& Swims& Water& Weak& Longneck& Bipedal&$\rightarrow$\\
\rowcolor{lightblue}&$\rightarrow$&0.294& 0.303& 0.315& 0.324& 0.345& 0.381& 0.391& 0.392& 0.408& 0.429& 0.433& 0.433& 0.439& 0.444& 0.444&$\rightarrow$\\ \hline
 &$\rightarrow$&Tree& Chewteeth& Hibernate& Nocturnal& Fast& Furry& Stalker& Newworld& Tail& Horns& Hairless& Jungle& Buckteeth& Spots& Active&$\rightarrow$\\
\rowcolor{lightblue}&$\rightarrow$&0.450& 0.454& 0.477& 0.487& 0.501& 0.507& 0.511& 0.514& 0.515& 0.518& 0.524& 0.526& 0.527& 0.562& 0.570&$\rightarrow$\\ \hline
 &$\rightarrow$&Mountains& Strong& Bush& Pads& Fish& Big& Timid& Hunter& Small& Brown& Longleg& Hooves& Agility& Nestspot& Smart&$\rightarrow$\\
\rowcolor{lightblue}&$\rightarrow$&0.580& 0.582& 0.585& 0.593& 0.599& 0.601& 0.617& 0.624& 0.630& 0.643& 0.644& 0.645& 0.646& 0.648& 0.648&$\rightarrow$\\ \hline
 &$\rightarrow$&Group& Meat& Patches& Fierce& Forest& Claws& Black& Muscle& Meatteeth& Slow& Fields& Vegetation& Domestic& Grazer& Gray&$\rightarrow$\\
\rowcolor{lightblue}&$\rightarrow$&0.649& 0.651& 0.656& 0.661& 0.667& 0.667& 0.669& 0.672& 0.674& 0.675& 0.678& 0.679& 0.682& 0.686& 0.690&$\rightarrow$\\ \hline
 &$\rightarrow$&Paws& Plains& Solitary& Toughskin& Bulbous& White& Forager& Inactive& Smelly& Lean&&&&&&\\
\rowcolor{lightblue}&$\rightarrow$&0.691& 0.692& 0.692& 0.692& 0.693& 0.693& 0.693& 0.693& 0.693& 0.693&&&&&&\\ \bottomrule

    \end{tabular}
    }
    \vspace{1mm}
    \caption{{\bf AWA2 task hardness.} AWA2 attributes sorted in ascending order of hardness along with their respective hardness scores. Hardness scores listed above are compared with empirical test errors for each task and shown to be strongly correlated (Fig.~\ref{fig:hardness}(b) in the paper).}
    \label{tab:AWA2hardorder}
   
\end{table*}

\begin{table*}[ht]
    \centering
    \resizebox{.90\linewidth}{!}{
    \begin{adjustbox}{angle=90}
    \begin{tabular}{c@{~~~~}lcccccccccc@{~}l}
    \toprule
1& Attribute& ec::purple& lc::green& ec::green& ec::olive& blc::green& blc::purple& brc::purple& ec::pink& bec::purple& sh::owl-like &$\rightarrow$ \\
\rowcolor{lightblue}2& Conditional Entropy$\uparrow$&0.011& 0.015& 0.015& 0.016& 0.017& 0.019& 0.022& 0.022& 0.025& 0.025&$\rightarrow$\\ \hline
&$\rightarrow$&lc::olive& unc::purple& fc::pink& upc::pink& untc::purple& cc::pink& wc::pink& bec::pink& tc::purple& lc::iridescent&$\rightarrow$\\
\rowcolor{lightblue}&$\rightarrow$&0.025& 0.027& 0.027& 0.028& 0.030& 0.030& 0.031& 0.032& 0.033& 0.034&$\rightarrow$\\ \hline
&$\rightarrow$&tc::pink& wc::purple& brc::pink& uptc::purple& blc::olive& untc::pink& bkc::pink& lc::purple& cc::purple& pc::pink&$\rightarrow$\\
\rowcolor{lightblue}&$\rightarrow$&0.035& 0.035& 0.035& 0.035& 0.036& 0.036& 0.036& 0.037& 0.037& 0.038&$\rightarrow$\\ \hline
&$\rightarrow$&nc::pink& unc::pink& nc::purple& bkc::purple& blc::pink& fc::purple& upc::purple& ec::blue& pc::purple& uptc::pink&$\rightarrow$\\
\rowcolor{lightblue}&$\rightarrow$&0.038& 0.040& 0.040& 0.041& 0.042& 0.042& 0.042& 0.049& 0.051& 0.053&$\rightarrow$\\ \hline
&$\rightarrow$&tc::green& fc::green& bkc::rufous& uptc::rufous& blc::iridescent& sh::long-legged-like& bec::rufous& untc::rufous& ec::rufous& brc::green&$\rightarrow$\\
\rowcolor{lightblue}&$\rightarrow$&0.054& 0.054& 0.057& 0.057& 0.057& 0.061& 0.064& 0.064& 0.064& 0.064&$\rightarrow$\\ \hline
&$\rightarrow$&wc::rufous& lc::rufous& cc::green& upc::rufous& untc::green& bec::green& tc::olive& brc::rufous& blc::rufous& unc::green&$\rightarrow$\\
\rowcolor{lightblue}&$\rightarrow$&0.066& 0.067& 0.069& 0.070& 0.071& 0.072& 0.072& 0.074& 0.074& 0.074&$\rightarrow$\\ \hline
&$\rightarrow$&tc::rufous& brc::iridescent& nc::green& pc::rufous& unc::rufous& uptc::green& nc::rufous& bec::iridescent& ec::buff& tc::iridescent&$\rightarrow$\\
\rowcolor{lightblue}&$\rightarrow$&0.074& 0.075& 0.076& 0.076& 0.076& 0.077& 0.077& 0.080& 0.081& 0.082&$\rightarrow$\\ \hline
&$\rightarrow$&bls::hooked& fc::rufous& lc::blue& ec::orange& untc::iridescent& unc::iridescent& uptc::iridescent& untc::orange& bkc::orange& cc::rufous&$\rightarrow$\\
\rowcolor{lightblue}&$\rightarrow$&0.082& 0.082& 0.082& 0.084& 0.084& 0.089& 0.089& 0.089& 0.089& 0.090&$\rightarrow$\\ \hline
&$\rightarrow$&fc::iridescent& ec::yellow& sh::chicken-like-marsh& nc::orange& cc::iridescent& si::very-large& tc::orange& untc::red& pc::iridescent& uptc::orange&$\rightarrow$\\
\rowcolor{lightblue}&$\rightarrow$&0.092& 0.092& 0.093& 0.093& 0.094& 0.094& 0.095& 0.095& 0.095& 0.097&$\rightarrow$\\ \hline
&$\rightarrow$&bkc::green& cc::orange& ec::grey& pc::green& wc::green& lc::yellow& wc::iridescent& brc::olive& upc::green& nc::iridescent&$\rightarrow$\\
\rowcolor{lightblue}&$\rightarrow$&0.099& 0.099& 0.099& 0.099& 0.100& 0.100& 0.101& 0.103& 0.107& 0.107&$\rightarrow$\\ \hline
&$\rightarrow$&bls::specialized& bkc::iridescent& fc::olive& cc::olive& blc::blue& fc::orange& bec::blue& bls::curved& bls::needle& wc::orange&$\rightarrow$\\
\rowcolor{lightblue}&$\rightarrow$&0.108& 0.108& 0.110& 0.112& 0.112& 0.113& 0.113& 0.113& 0.113& 0.114&$\rightarrow$\\ \hline
&$\rightarrow$&bec::olive& lc::pink& wc::red& sh::upwl& unc::olive& uptc::red& bkc::red& ec::red& upc::orange& upc::iridescent&$\rightarrow$\\
\rowcolor{lightblue}&$\rightarrow$&0.115& 0.117& 0.117& 0.117& 0.118& 0.118& 0.119& 0.120& 0.123& 0.126&$\rightarrow$\\ \hline
&$\rightarrow$&upc::red& unc::blue& sh::hawk-like& brc::orange& nc::olive& brc::blue& bec::orange& sh::upland-ground-like& hp::spotted& tc::blue&$\rightarrow$\\
\rowcolor{lightblue}&$\rightarrow$&0.129& 0.130& 0.133& 0.134& 0.134& 0.137& 0.139& 0.139& 0.141& 0.141&$\rightarrow$\\ \hline
&$\rightarrow$&unc::orange& untc::olive& ec::brown& pc::orange& si::large& lc::red& uptc::olive& hp::unique-pattern& bec::red& ec::white&$\rightarrow$\\
\rowcolor{lightblue}&$\rightarrow$&0.144& 0.145& 0.147& 0.153& 0.155& 0.156& 0.156& 0.156& 0.161& 0.162&$\rightarrow$\\ \hline
&$\rightarrow$&pc::red& nc::red& hp::crested& lc::white& blc::red& sh::swallow-like& tc::red& brc::red& bls::spatulate& unc::red&$\rightarrow$\\
\rowcolor{lightblue}&$\rightarrow$&0.162& 0.162& 0.165& 0.167& 0.167& 0.170& 0.172& 0.173& 0.174& 0.175&$\rightarrow$\\ \hline
&$\rightarrow$&fc::red& bkc::olive& pc::olive& wc::olive& hp::masked& untc::blue& sh::hummingbird-like& ts::forked-tail& sh::sandpiper-like& cc::red&$\rightarrow$\\
\rowcolor{lightblue}&$\rightarrow$&0.175& 0.180& 0.181& 0.182& 0.183& 0.184& 0.184& 0.189& 0.190& 0.193&$\rightarrow$\\ \hline
&$\rightarrow$&upc::olive& fc::blue& wc::blue& uptc::blue& nc::blue& blc::white& blc::yellow& bkc::blue& cc::blue& upc::blue&$\rightarrow$\\
\rowcolor{lightblue}&$\rightarrow$&0.193& 0.205& 0.208& 0.209& 0.212& 0.217& 0.219& 0.220& 0.222& 0.226&$\rightarrow$\\ \hline
&$\rightarrow$&tp::spotted& sh::pigeon-like& bll::longer-than-head& uptc::yellow& sh::duck-like& pc::blue& bep::spotted& bls::hooked-seabird& brp::spotted& sh::tree-clinging-like&$\rightarrow$\\
\rowcolor{lightblue}&$\rightarrow$&0.229& 0.231& 0.231& 0.231& 0.231& 0.235& 0.240& 0.240& 0.248& 0.250&$\rightarrow$\\ \hline
&$\rightarrow$&sh::gull-like& hp::striped& blc::orange& cc::yellow& untc::yellow& bkp::spotted& blc::brown& wp::spotted& nc::yellow& ws::long-wings&$\rightarrow$\\
\rowcolor{lightblue}&$\rightarrow$&0.254& 0.260& 0.262& 0.271& 0.278& 0.284& 0.293& 0.294& 0.294& 0.294&$\rightarrow$\\ \hline
&$\rightarrow$&tc::brown& bkc::yellow& fc::yellow& ws::broad-wings& lc::brown& hp::malar& wc::yellow& bec::brown& lc::orange& ts::fan-shaped-tail&$\rightarrow$\\
\rowcolor{lightblue}&$\rightarrow$&0.298& 0.301& 0.307& 0.313& 0.318& 0.318& 0.318& 0.323& 0.328& 0.330&$\rightarrow$\\ \hline
&$\rightarrow$&ts::squared-tail& upc::yellow& unc::brown& bep::striped& tc::yellow& ec::black& hp::capped& hp::eyeline& hp::eyebrow& cc::white&$\rightarrow$\\
\rowcolor{lightblue}&$\rightarrow$&0.336& 0.346& 0.356& 0.360& 0.363& 0.368& 0.371& 0.374& 0.375& 0.380&$\rightarrow$\\ \hline
&$\rightarrow$&brc::brown& ws::tapered-wings& brp::striped& fc::buff& cc::buff& bls::dagger& ts::rounded-tail& fc::white& pc::yellow& brc::yellow&$\rightarrow$\\
\rowcolor{lightblue}&$\rightarrow$&0.381& 0.382& 0.383& 0.384& 0.386& 0.402& 0.403& 0.412& 0.416& 0.421&$\rightarrow$\\ \hline
&$\rightarrow$&tc::buff& blc::buff& uptc::buff& bec::yellow& untc::buff& bec::black& hp::eyering& bep::multi-colored& unc::yellow& tc::grey&$\rightarrow$\\
\rowcolor{lightblue}&$\rightarrow$&0.425& 0.431& 0.437& 0.440& 0.443& 0.444& 0.446& 0.450& 0.450& 0.451&$\rightarrow$\\ \hline
&$\rightarrow$&nc::buff& tp::striped& uptc::white& fc::brown& bkc::buff& lc::buff& nc::brown& bec::grey& upc::buff& pc::buff&$\rightarrow$\\
\rowcolor{lightblue}&$\rightarrow$&0.456& 0.466& 0.476& 0.483& 0.483& 0.483& 0.484& 0.485& 0.490& 0.490&$\rightarrow$\\ \hline
&$\rightarrow$&brc::buff& si::medium& bkc::white& bec::buff& unc::black& brp::multi-colored& wc::buff& cc::brown& brc::grey& unc::buff&$\rightarrow$\\
\rowcolor{lightblue}&$\rightarrow$&0.492& 0.493& 0.494& 0.494& 0.496& 0.496& 0.498& 0.501& 0.503& 0.510&$\rightarrow$\\ \hline
&$\rightarrow$&unc::grey& bkp::striped& fc::grey& untc::brown& brc::black& uptc::brown& ts::pointed-tail& cc::grey& tc::black& si::very-small&$\rightarrow$\\
\rowcolor{lightblue}&$\rightarrow$&0.513& 0.514& 0.518& 0.520& 0.528& 0.530& 0.531& 0.534& 0.544& 0.549&$\rightarrow$\\ \hline
&$\rightarrow$&nc::white& untc::white& bkp::multi-colored& pc::brown& nc::grey& bkc::brown& upc::white& untc::grey& wp::striped& hp::plain&$\rightarrow$\\
\rowcolor{lightblue}&$\rightarrow$&0.549& 0.554& 0.555& 0.564& 0.572& 0.573& 0.582& 0.584& 0.587& 0.588&$\rightarrow$\\ \hline
&$\rightarrow$&bls::cone& uptc::grey& blc::grey& wc::white& upc::brown& pc::white& nc::black& pc::grey& wc::brown& tp::multi-colored&$\rightarrow$\\
\rowcolor{lightblue}&$\rightarrow$&0.590& 0.591& 0.594& 0.596& 0.599& 0.602& 0.604& 0.605& 0.606& 0.606&$\rightarrow$\\ \hline
&$\rightarrow$&lc::black& bkc::grey& ws::pointed-wings& lc::grey& wp::solid& wp::multi-colored& wc::grey& upc::grey& fc::black& bep::solid&$\rightarrow$\\
\rowcolor{lightblue}&$\rightarrow$&0.610& 0.616& 0.619& 0.622& 0.626& 0.627& 0.628& 0.633& 0.643& 0.646&$\rightarrow$\\ \hline
&$\rightarrow$&cc::black& pc::black& tc::white& bkc::black& bll::same-as-head& ts::notched-tail& uptc::black& brc::white& bls::all-purpose& brp::solid&$\rightarrow$\\
\rowcolor{lightblue}&$\rightarrow$&0.647& 0.651& 0.654& 0.655& 0.659& 0.666& 0.667& 0.669& 0.669& 0.670&$\rightarrow$\\ \hline
&$\rightarrow$&bll::shorter-than-head& bec::white& ws::rounded-wings& unc::white& untc::black& upc::black& blc::black& bkp::solid& tp::solid& wc::black&$\rightarrow$\\
\rowcolor{lightblue}&$\rightarrow$&0.681& 0.681& 0.682& 0.684& 0.685& 0.688& 0.688& 0.691& 0.691& 0.691&$\rightarrow$\\ \hline
&$\rightarrow$&sh::perching-like& si::small&&&&&&&&&\\
\rowcolor{lightblue}&$\rightarrow$&0.693& 0.693&&&&&&&&&\\ \bottomrule
    \end{tabular}
    \end{adjustbox}
    }
    \vspace{1mm}
    \caption{{\bf CUB task hardness.} CUB attributes sorted in ascending order of hardness along with their respective hardness scores. Hardness scores listed above are compared with empirical test errors for each task and shown to be strongly correlated (Fig.~\ref{fig:hardness}(c) in the paper). Attribute names are abbreviated due to space concerns. Full names are provided in Table~\ref{tab:CUBAbb}.}
    \label{tab:CUBhardorder}
\end{table*}

\begin{table*}[ht]
    \centering
    \begin{tabular}{l@{~~}l|l@{~~}l|l@{~~}l|l@{~~}l}
bls &has bill shape &uptc & has upper tail color &untc & has under tail color &tp & has tail pattern \\
wc & has wing color &hp &has head pattern &nc & has nape color &bep &has belly pattern \\
upc & has upperparts color &brc & has breast color &bec & has belly color &pc & has primary color \\
unc & has underparts color &tc & has throat color &ws &has wing shape &lc & has leg color \\
brp &has breast pattern &ec & has eye color &si & has size &blc & has bill color \\
bkc & has back color &bll & has bill length &sh & has shape &cc &has crown color \\
ts & has tail shape &fc &has forehead color &bkp & has back pattern &wp & has wing pattern \\
    \end{tabular}
    \vspace{2mm}
    \caption{{\bf CUB attribute name abbreviations.} Abbreviations used in Table~\ref{tab:CUBhardorder} for the attributes in the CUB dataset~\cite{WelinderEtal2010}.}
    \label{tab:CUBAbb}
\end{table*}

\end{document}